# No Language Left Behind: Scaling Human-Centered Machine Translation

NLLB Team, Marta R. Costa-jussà*, James Cross*, Onur Çelebi*, Maha Elbayad*, Kenneth Heafield*, Kevin Heffernan*, Elahe Kalbassi*, Janice Lam*, Daniel Licht*, Jean Maillard*, Anna Sun*, Skyler Wang*,§ , Guillaume Wenzek*, Al Youngblood*
Bapi Akula, Loic Barrault, Gabriel Mejia Gonzalez, Prangthip Hansanti, John Hoffman, Semarley Jarrett, Kaushik Ram Sadagopan, Dirk Rowe, Shannon Spruit, Chau Tran
Pierre Andrews†, Necip Fazil Ayan†, Shruti Bhosale†, Sergey Edunov†, Angela Fan†,‡, Cynthia Gao†, Vedanuj Goswami†, Francisco Guzmán†, Philipp Koehn†,¶, Alexandre Mourachko†, Christophe Ropers†, Safiyyah Saleem†, Holger Schwenk†, Jeff Wang†

Meta AI, §UC Berkeley, ¶Johns Hopkins University

## Abstract

Driven by the goal of eradicating language barriers on a global scale, machine translation has solidified itself as a key focus of artificial intelligence research today. However, such efforts have coalesced around a small subset of languages, leaving behind the vast majority of mostly low-resource languages. What does it take to break the 200 language barrier while ensuring safe, high quality results, all while keeping ethical considerations in mind? In *No Language Left Behind*, we took on this challenge by first contextualizing the need for low-resource language translation support through exploratory interviews with native speakers. Then, we created datasets and models aimed at narrowing the performance gap between low and high-resource languages. More specifically, we developed a conditional compute model based on Sparsely Gated Mixture of Experts that is trained on data obtained with novel and effective data mining techniques tailored for low-resource languages. We propose multiple architectural and training improvements to counteract overfitting while training on thousands of tasks. Critically, we evaluated the performance of over 40,000 different translation directions using a human-translated benchmark, Flores-200, and combined human evaluation with a novel toxicity benchmark covering all languages in Flores-200 to assess translation safety. Our model achieves an improvement of 44% BLEU relative to the previous state-of-the-art, laying important groundwork towards realizing a universal translation system. Finally, we open source all contributions described in this work, accessible at `https://github.com/facebookresearch/fairseq/tree/nllb`.

*. Equal contribution, alphabetical order
†. Research and engineering leadership, equal contribution, alphabetical order
‡. Corresponding Author. Email: angelafan@fb.com.

## Contents

## 1. Introduction

In Jack Vance (1977)'s sci-fi novel *The Eyes of the Overworld*, its protagonist, Cugel, encounters a wizard who compels him into a task. To assist him, the wizard grants Cugel a magical device: *In order to facilitate your speech, I endow you with this instrument which relates all possible vocables to every conceivable system of meaning.*

Fast-forward half a century later, we now know that Cugel's magical device is really Machine Translation. Conceived as computational systems that translate texts from one language to another, machine translation has been around since the 1940s, but its recent migration from statistical (Brown et al., 1993; Koehn, 2009; Lopez, 2008) to neural systems has pushed the technology to new frontiers (Bahdanau et al., 2015; Cho et al., 2014; Kalchbrenner and Blunsom, 2013; Wu et al., 2016). This shift has not only advanced translation quality at breakneck speed, but it has also furthered the expansion of machine translation into new applications. Today, machine translation impacts how people all over the world communicate, work, travel, learn, access information, and more (Khoong and Rodriguez, 2022; Koehn and Germann, 2014; Lee, 2020).

While machine translation continues to grow, the fruits it bears are unevenly distributed (Fan et al., 2020). In fact, the vast majority of improvements made in machine translation in the last decades have been for high-resource languages, or languages that have large quantities of training data available digitally. For instance, those who communicate in English, French, German or Russian—languages which have long enjoyed institutional investments and data availability—stand to gain substantially more from the maturation of machine translation than those who speak Catalan, Assamese, Ligurian, or Kinyarwanda.

Many languages of this latter group attract less attention and resources, even though most languages spoken globally today are *Low-Resource* languages (Joshi et al., 2020). Many of these languages escape researchers' gaze for a confluence of reasons, including constraints conjured up by past investments (or lack thereof), research norms, organizational priorities, and Western-centrism to name a few. Without an effort to course correct, much of the internet could continue to be inaccessible to speakers of these languages. Research indicates that while only 25.9 percent of internet users speak English, 63.7 percent of all websites are in English (the next on the list is Russian at 6.8 percent; Richter, 2022). For many low-resource language communities, *The Polyglot Internet* (Zuckerman, 2008), an instrumental medium that could propel education access and social mobility, remains out of reach because the web has long prioritized content tailored to high-resource language speakers.

Expanding machine translation to more low-resource languages is further curtailed by technical challenges (Haddow et al., 2022). Compared to their high-resource counterparts, training data for low-resource languages are expensive and logistically challenging to procure (Kuwanto et al., 2021; Nekoto et al., 2020; Orife et al., 2020). Without sufficient training data, standard techniques may not stand the test of emerging demands. These hurdles have become ever more pronounced as the popularity of data-hungry techniques such as large-scale pre-training and model scaling have become mainstream (Conneau and Lample, 2019; Conneau et al., 2020; Kenton and Toutanova, 2019; Radford et al., 2019).

To overcome these barriers, much existing work on low-resource translation has focused on leveraging *multilingual* systems, or models capable of handling multiple languages. These models have the advantage of crosslingual transfer (Nguyen and Chiang, 2017; Zoph et al., 2016), allowing related languages to learn from one another (Arivazhagan et al., 2019; Fan et al., 2020; Zhang et al., 2020). While multilingual models have demonstrated promising performance improvement compared to bilingual models (Tran et al., 2021), enabling the representation of hundreds of languages while retaining strong translation quality remains an open area of research. Another strategy aimed at mitigating the low-resource challenge is to acquire more language data. Some of these attempts have focused on collecting human translations, while others have leveraged large-scale data mining and monolingual data pipelines to consolidate data found across the web (Bañón et al., 2020; Karakanta et al., 2018; Ramesh et al., 2022; Schwenk et al., 2021b). The latter techniques are often plagued by noise and biases, making it difficult to validate the quality of the created datasets (Kreutzer et al., 2022). Finally, developing translation models for low-resource languages requires the existence of high-quality, human-translated evaluation benchmarks. Datasets such as Flores-101 (Goyal et al., 2022) work towards this, but coverage is capped at 100 languages.

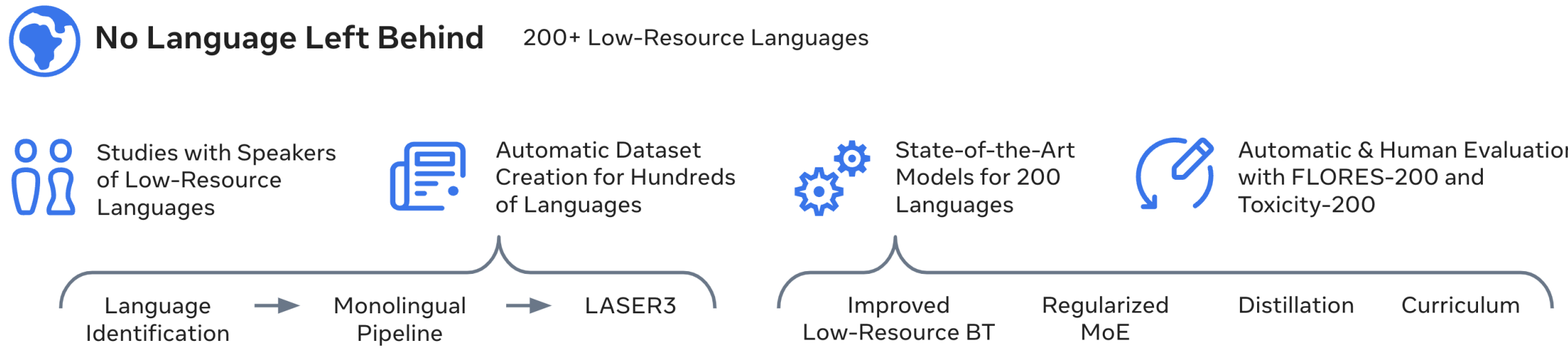

Figure 1: **No Language Left Behind:** Our low-resource translation effort focuses on four cornerstones. (1) We strive to understand the low-resource translation problem from the perspective of native speakers. (2) We study how to automatically create training data to move low-resource languages towards high-resource. (3) We utilize this data to create state-of-the-art translation models. (4) We evaluate every language we aim to translate.

In this article, we ask: *What does it take to double the language coverage of most existing translation models while ensuring high quality and safe translations?* More concretely, how do we use a human-centric approach (Robertson et al., 2021) to create fluent, meaning-preserving translations for over 200 languages, many of which belong to a class of low-resource languages that remain underserved by existing translation technologies? And how can we do so while minimizing potential harm from catastrophic and toxic translations hallucinated by neural MT models — infrequent occurrences that nevertheless have an out-sized adverse impact on the human user?

We take on this challenge in the *No Language Left Behind* (NLLB) effort. We begin by creating Flores-200, a many-to-many multilingual dataset that allows us to measure translation quality through any of the 40,602 total translation directions. We developed a distillation-based sentence encoding technique, LASER3 (Heffernan et al., 2022), that helped us mine web data to create parallel datasets for low-resource languages. Using both mined data and a set of human-translated *seed data*, we trained multilingual Mixtures-of-Experts models with state of the art performance. Despite doubling the number of languages, our final model performs 40% better than the previous state of the art on Flores-101. To detect and prevent potentially harmful translations that are hallucinated by the translation models, we created a dataset of toxic words for all 200 languages by combining automatic and human evaluations. We proposed and conducted human evaluations on many languages our models cover, in addition to the common automatic metrics, to gain qualitative insight into the impact of the translation. Finally, beyond creating these models, we also reflect on the creation process, analyzing the risks and benefits of our research from a societal standpoint. We open source all the benchmarks, data, scripts, and models described in this effort to support further research.[1] In addition, we focus on the practical applicability of our work for low-resource speaking communities. We deploy our techniques to provide translation support to Wikipedia editors, enabling them to create new articles more efficiently for languages that are not supported by other translation systems.

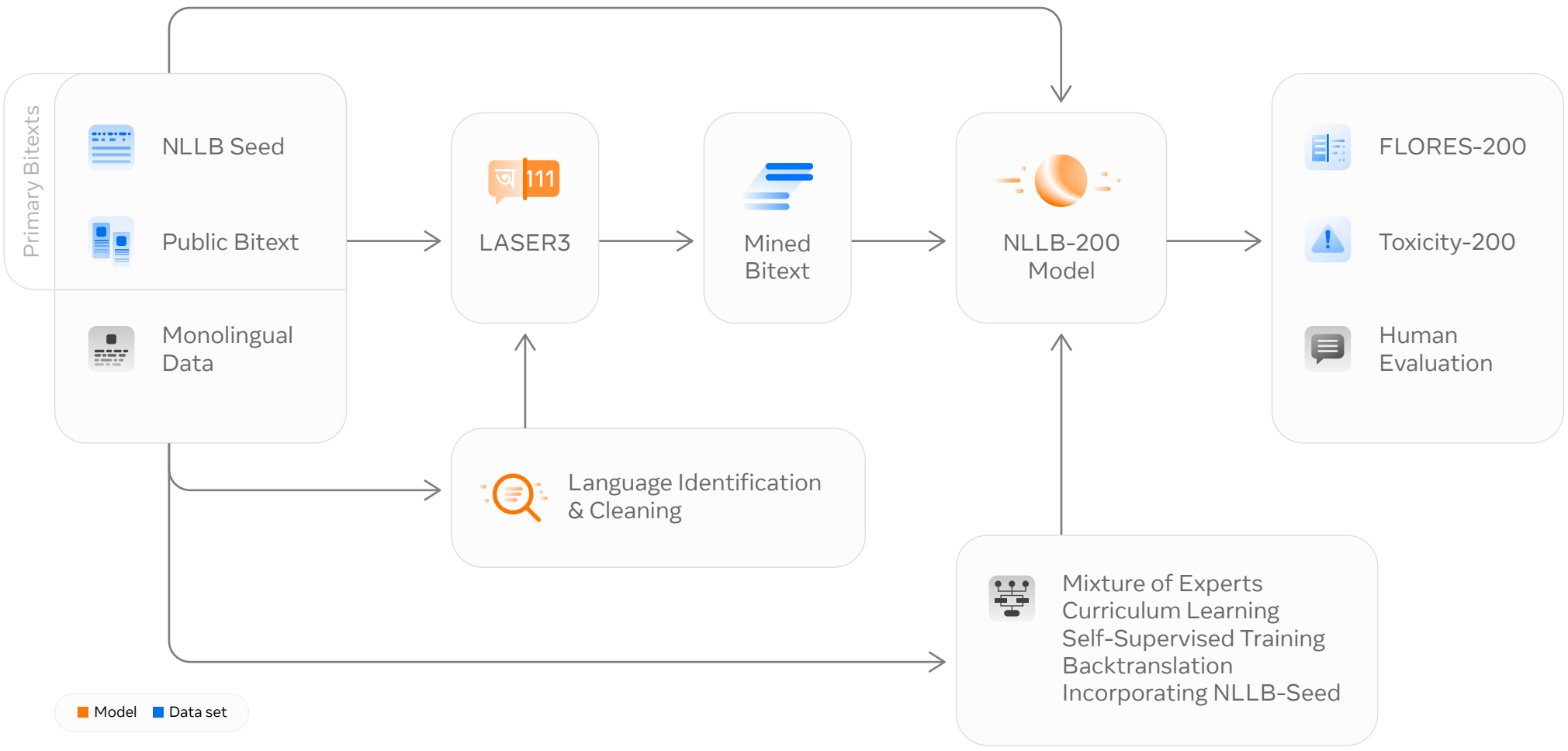

Figure 2: **How the Pieces Fit Together, a Bird's-Eye View:** We depict the technical components of No Language Left Behind and how they fit together. We display the interaction between data, how data is utilized in the models we develop (orange), and how models are evaluated. Datasets shown in blue are novel datasets created in No Language Left Behind.

The rest of the article is structured as follows, with Figure 2 as an overview: Section 2 describes the open challenges in low-resource translation and analyzes the widespread use of translation systems. Section 3 presents the languages we focus on and how we arrived at this set of languages. Section 4 summarizes the creation process of Flores-200 and NLLB-Seed + NLLB-MD, our translation seed datasets, with quality analysis. Section 5 overviews the creation of monolingual and mined bilingual data, which enables the creation of models for hundreds of languages. Section 6 details various modeling techniques developed to improve the performance of low-resource languages. Section 7 traces the automatic and human evaluation of our translations, including the detection of catastrophic and toxic translations. We integrate the aforementioned datasets and techniques into NLLB-200, a

1. All are available here: `https://github.com/facebookresearch/fairseq/tree/nllb`

model that currently supports 202 languages, and analyze its quality and performance in Section 8. We conclude in Section 9, where we reflect on the social impact of our research and lay out future possibilities and challenges. It is our hope that our contribution would guide future researchers who, like us, are eager to see Cugel's magical device — machine translation covering all languages — transform from a conceptual chimera into a reality.

To make our work available to the community, we open source the following:

- **Human-Translated Datasets**

  Flores-200: Evaluation dataset in 204 languages

  NLLB-Seed: Seed training data in 39 languages

  NLLB-MD: Seed data in different domains in 6 languages to assess generalization

  Toxicity-200: wordlists to detect toxicity in 200 languages

- **Tools to Create Large Scale Bitext Datasets**

  Language Identification for more than 200 languages

  LASER3: sentence encoders for identifying aligned bitext for 148 languages

  `stopes`: a data mining library that can be used to process and clean monolingual data, then create aligned bitext

  Training data recreation: Scripts that recreate our training data

- **Translation Models covering 202 languages**

  NLLB-200: A 54.5B Sparsely Gated Mixture-of-Experts model

  3.3B and 1.3B Dense Transformer models

  1.3B and 600M Dense transformer models distilled from NLLB-200

  Training and generation scripts to reproduce our models

## 2. Human-Centered Low-Resource Language Translation

To situate our goal of providing high-quality translation for hundreds of languages, we first explore the importance of this research to those who matter the most to us: low-resource language communities. Inspired by *Value Sensitive Design* (Friedman and Hendry, 2019; Van Der Hoven and Manders-Huits, 2020), we attribute community-level interests and values as the cornerstone of our research. Adopting this framework propels us to start with people and prioritize how they interact with technology, with direct emphasis on ethical and social considerations (Mukhija et al., 2021). To understand how low-resource language speakers perceive machine translation, we conducted an interview study with 44 low-resource language speakers. As stakeholders likely to be impacted by No Language Left Behind (NLLB), their contributions helped us envision the promises many believe machine translation could deliver to their communities. Punctuating their careful optimism were concrete suggestions on ways to maximize social gains while minimizing risks. Moreover, many interviewees painted illustrative pictures of the cultural and political environments their languages live in, the ways in which language and social experiences intertwine, and how NLLB could potentially shake up the cultural status quo.

## 2.1 Exploratory Interview Study Research Design

We designed a semi-structured interview protocol aimed at exploring the needs and concerns of low-resource language speakers vis-à-vis machine translation. Although low-resource languages could be deemed low-resource for a variety of reasons, including being under-researched, digitized, or taught (Cieri et al., 2016; Magueresse et al., 2020), for the purpose of the study, we define low-resource as languages which had less than 1 million sentences of publicly available example translations at the time of the study. The interviews captured a broad array of attitudes and understandings, including the usage and application of low-resource languages, perceived value of translation technology, and how translation systems ought to be developed.

Overall, our recruitment effort led us to 44 native speakers of low-resource languages from diverse backgrounds, with ages ranging from 23 to 58. Covering a total of 36 languages, the distribution is as follows: 5 languages are spoken predominantly in North America, 8 in South America, 4 in Europe, 12 in Africa, and 7 in Asia. Although our sample has breadth in terms of race, education, and location, the majority of our participants are immigrants living in the U.S. and Europe, and about a third of them ($n = 17$) identify as tech workers. All interviews were conducted remotely via video conferencing software. On average, the interviews lasted 1.5 hours. Two-third of the interviews were recorded and transcribed. For unrecorded interviews, two researchers took extensive notes throughout. Bringing all 44 interviews together, responses were then systematically coded to allow major themes and ideas to emerge.

We acknowledge that sampling low-resource language speakers from diasporic contexts comes with its limitations. For one, as immigrants, their perspectives may not consummately capture the sentiments of their communities back home. That said, some scholars have argued that in technologically underdeveloped nations, where many low-resource language communities reside, people tend to view technology more optimistically and aspirationally than those who live in places with higher levels of technological development (Kapania et al., 2022; Kozyreva et al., 2021; Sambasivan, 2021; Sambasivan et al., 2021). Thus, being exposed to critical technological discourses (especially in recent times) could in fact make many of our interviewees more cognizant of the risks behind technological advancement, affording them a more balanced outlook. Moreover, immigration scholars remind us that global movement today is a transnational process, where those in receiving societies maintain cultural ties with those who remain in sending societies via a variety of communicative and media platforms (Baldassar et al., 2016; Levitt and Jaworsky, 2007; Levitt and Lamba-Nieves, 2011). Because we found strong evidence of such processes in our interviews, we trust that our participants are in a unique position to speak both critically and knowledgeably about the sociological underpinnings of their languages.

Over-sampling tech workers may introduce another form of selection bias. More specifically, research suggests that tech workers, given their insider status, are likely to espouse techno-optimism — a positive outlook with respect to technological development (McLennan, 2016). While such an effect cannot be downplayed, tech workers' personal affinity with technological practices could in fact imbue in them a critical reflexivity we were eager to tap into. As projected, while many participants speculated on the benefits of our research, they were equally keen on underscoring the potential risks such an intervention might impose on their very own language communities. These nuanced perspectives were vital in shaping our research processes and procedures.

### 2.1.1 Why should we prioritize low-resource languages?

Language is not only a way for people to communicate with one another, but it also conveys culture, history, and self-identity (Demichelis and Weibull, 2008; Hall, 2013). As a binding agent, language fosters community by extending the tradition and heritage of a common people. Even though many of our low-resource language interviewees are also fluent English speakers, almost all of them maintain that their native tongue remains a foundational part of their identity. Drawing parallels between themselves and their networks back home, more than half of the participants of our study lament that without sustained efforts at prioritizing the usage and application of their native languages, many of them would face endangerment in time to come.

**Decline of Native Language and Culture.** The fear that the low-resource languages might be undergoing a state of decline reverberated throughout the interviews. Such assertions typically attributed the decline to two causes: cultural and economical. Cultural theory suggests that as more and more aspects of our lives become digitally-mediated, prolonged exposures to content found on the web and social media platforms (e.g., YouTube, Facebook) leads to the prioritization of high-resource languages. By extension, this phenomenon spotlights Western epistemology and ideas over other ways of knowing (Nurullah, 2008). As few interviewees pointed out, the cultural dominance of the West applies intense pressure onto more localized media productions. As low-resource language speakers gravitate towards books, movies, and social media content tailored to high-resource language audiences, interest in content produced in their native tongue could be crowded out. Without sustained audiences, cultural products in low-resource languages risk displacement.

Another camp attributes low-resource languages' decline to the sways of the global political economy. For many low-resource language speakers who come from developing nations, a high-resource language like English is seen as both a vehicle for global competitiveness and social mobility. Prioritizing the *lingua franca* of the global economy means directing more resources at English education and tethering local communities to the needs of the knowledge economy—much of it driven by the demands of the West. Viewed through a zero-sum lens, many interviewees believe that the promotion of English might spell an increasing peripheralization of native languages in public life. Under such pressures, the status of many low-resource languages risks continued relegation.

Noting these trends, many low-resource language speakers remind us that machine translation could be a critical tool in promoting language and cultural preservation. As an Igbo speaker urged, improving machine translation for his native language would allow more people to produce cultural knowledge in that language, later adding that websites like Wikipedia could be a vital platform that enable others to learn about his culture's history and practices. Echoing such sentiments, another interviewee points to the importance of such bi-directional learning, noting that having the ability to translate means people who do not speak their language could read and understand Wikipedia articles about their culture, which further motivates other writers to write more. Thus, bi-directional learning not only illuminates the intricate relationship between machine translation and culture preservation, it provides an opportunity to disrupt the entrenching nature of Western-centric knowledge dissemination. The centrality of Wikipedia in these stories tells us that supporting one of the world's most frequented knowledge-sharing portal could deeply amplify the impact of our effort.

**Coverage and Quality of Existing Automatic Translation.** When asked about translation coverage, most low-resource language speakers express comfort in the fact that their respective languages are supported by existing systems. A few interviewees said that being included by commercially available services makes them feel seen and raises the visibility of their languages. However, such sentiments are not uniformly shared. For a select group of low-resource language speakers, whose languages contain multiple scripts and variants, full coverage remains lacking.

For instance, a Moroccan Arabic speaker said that fully supporting the Arabic language requires us to take the various extant Arabic languoids[2] into account so that we do not end up favoring one form over another. This concern similarly applies to languages with dual or multiple scripts (i.e., Banjar, Kanuri etc.). By excluding certain languoids or scripts and propping more well-resourced variants as the "default" option (Sunstein and Thaler, 2003), we not only jeopardize accurate cultural representation, but also exacerbate the unequal field that already plagues language distribution and usage across different parts of the world.

On the other hand, quality concerns resonated across the board with our participants. Reflecting on a sizable quality gap pitching high-resource language translation against low-resource language translation (Joshi et al., 2019), many interviewees cite poor and unreliable results as the key reason for irregular or discontinued use. For instance, a Bhojpuri speaker says that translating a sentence in their language using a commercially available system and then editing it takes more time than doing so manually. Another interviewee asserted that it is not perfection that she wants, but rather a technology that is reasonably usable for translation to and from their language. A few interviewees even mentioned that the lack of care given to their languages in some translation platforms have led to occasional toxic or crude translations, further eroding their confidence in these systems. These perspectives remind us that even though language inclusion is an important first step, striving for safe and high quality translation is still what matters most at the end of the day.

**Who stands to gain?** Discussions around the value of machine translation among low-resource language speakers evince the deep socioeconomic gaps that divide one community from another, impacting the perceived utility of the technology. While machine translation primarily helps those from more advantaged backgrounds learn new languages or travel more effectively, its presence in financially impoverished communities could be instrumental for social mobility or even economic survival. For instance, a Tigrinya speaker notes that in Ethiopia, where less than 20 percent of the country has internet access, actual access to what the web offers is even more restricted due to the lack of quality translation. They later stressed that language can be an entrenched barrier to education and employment. Many low-resource language speakers from Africa echo these sentiments, reminding us of the consequences of chronic marginalization and its impact on people (Alupo et al., 2021), and the wide spectrum of gains machine translation could deliver to different populations.

2. For a discussion of the notion of languoid, see Good and Hendryx-Parker (2006).

Zooming into individual communities themselves, we see similar forms of divide. For instance, most interviewees agree that those with technological know-how would benefit more from machine translation than those who do not. One interview hints that younger individuals in their communities are more well-suited to exploit the utility of machine translation than their older counterparts. Citing the recent COVID-19 pandemic as an example, she noted that in places where science-backed information was sparse due to the lack of trust-worthy formal institutions, seniors of these communities were dependent on their more tech-savvy network and family members to acquire timely, translated health information derived from international organizations. In the same vein, those with higher levels of technology know-how would also be better able to repel misinformation, fake news, or online scams that could arise from the expansion of translation technologies into low-resource languages.

Taken collectively, it is important to note that low-resource language communities are not a monolithic group; they each navigate unique sociopolitical and cultural contexts. In speaking to their constituents, we learn that realizing quality translation, while important for several reasons, remains one solution to a massive puzzle that is fair language representation and equitable knowledge access. That said, by offering up one solution, we hope to galvanize other actors into action. As one low-resource language speaker opined, incorporating more low-resource languages in machine translation helps de-prioritized languages gain digital visibility on a global scale, which could compel local institutions to take native languages more seriously and invest more resources into preserving or teaching them. This perspective underscores both the symbolic and material benefits machine translation could bring. The positive encouragements from low-resource language speakers throughout the course of the study remind us that by taking a human-centric approach and focusing on languages that have historically been left behind, we can help communities maintain a connection to their native languages—a quintessential part of many people's culture and identity.

### 2.2 No Language Left Behind: Guiding Principles

Combining insights drawn from interviews with low-resource language speakers and good practices distilled from literature on responsible AI (Arrieta et al., 2020; Bender et al., 2021; Blodgett et al., 2022; Paullada et al., 2021; Sambasivan and Holbrook, 2018), we introduce four key guiding principles underlying our research:

1. **Prioritize the needs of underserved communities**. As aforementioned, we put the needs of low-resource language communities at the front and center of our effort. Recognizing that machine translation is a value-laden technological artifact that has historically de-prioritized certain populations, we use this effort to redistribute power and resources to underserved communities. By elevating the needs of low-resource language communities, we hope that our contribution is part of a collective effort that propels digital representation into a more equitable era.

2. **Sharing through open-sourcing**. Low-resource language speakers across the board remind us that transparency ought to be a key emphasis when developing NLLB. With the dual intent to foster transparency and avoid a duplication of effort, we decided early on that we were going to open source NLLB. This way, the research community at large could directly benefit from our contribution. Creating NLLB with open-sourcing in mind also motivates us to be intentional and deliberative in our approach throughout the developmental process. We hope that the impact of our work could be amplified as other scientists and practitioners build on this effort to advance the field of machine translation as a whole.

3. **Being interdisciplinary in our approach**. As cogently put by a low-resource language speaker, machine translation is not just a coding problem, for at its very core, it is a human matter. To avoid the 'alignment problem' (Christian, 2020) and allow our system to perform in a way that is both value-sensitive and socially responsible, our research effort is taken on by an interdisciplinary team with scholars from a wide array of humanities (i.e., Philosophy, Ethics), social scientific (i.e., Sociology, Linguistics), and technical (i.e., Computer Science, Statistics) backgrounds. Bolstering the diversity of our team not only expands our methodological and analytic toolkit, it also affords us a chance to leverage different skills to tackle disparate aspects of the challenge.

4. **Being reflexive in our efforts**. Finally, reflexivity motivates us to critically examine our own judgments, practices, and belief systems throughout NLLB's creation process to ensure that we mitigate biases commonly found in the development of artificial intelligence systems. Concretely, we offer up detailed documentation of how we arrived at various decisions below to allow different stakeholders to comb through our intentions and motivations. We acknowledge that with ambitious efforts like these, trade-offs have to be made and perfection remains elusive. As such, it is our hope that our current effort would invite critical examinations of existing practices, which would then allow us to make more informed decisions in future iterations of NLLB.

Now that we have described our motivation and values, we move on to the next part of the story—overcoming the technical challenges involved in realizing machine translation for 200 languages, from language identification to training data, models, and evaluation. As is the case with any cutting edge interventions, big problems require novel adaptions. Below, we describe the journey we took to materialize the technical dimensions of NLLB, detailing ethical and social considerations along the way. First, let's meet our language candidates.

## 3. Languages

Broadly accessible machine translation systems support around 130 languages; our goal is to bring this number up to 200. In deciding what languages to offer, we first parsed through the 101 languages covered in Flores-101, a dataset for translation evaluation covering predominantly low-resource languages. From there, we generated a preliminary list of over 250 possible language candidates, eventually trimming it down to around 210 for final expansion from 101 to 200+ languages.

The creation process of the preliminary list is as follows. First, we considered all languages with a Wikipedia presence. As noted in the section above, Wikipedia is a key site of knowledge dissemination for many speaking low-resource languages, making it a pertinent place to start. Currently, Wikipedia supports over 300 languages, extending mindfully its content beyond English (Johnson and Lescak, 2022), and new languages can be added

| Code | Language | Script | Family | Subgrouping | 🌐 | Res. | Specification |
|---|---|---|---|---|---|---|---|
| ace_Arab[NEW] | **Acehnese** | Arabic | Austronesian | Malayo-Polynesian | ✗ | Low | North Acehnese |
| ace_Latn[NEW] | **Acehnese** | Latin | Austronesian | Malayo-Polynesian | ✗ | Low | North Acehnese |
| acm_Arab[NEW] | **Mesopotamian Arabic** | Arabic | Afro-Asiatic | Semitic | ✗ | Low | Baghdadi |
| acq_Arab[NEW] | **Ta'izzi-Adeni Arabic** | Arabic | Afro-Asiatic | Semitic | ✗ | Low | |
| aeb_Arab[NEW] | **Tunisian Arabic** | Arabic | Afro-Asiatic | Semitic | ✗ | Low | Derja |
| afr_Latn | **Afrikaans** | Latin | Indo-European | Germanic | 🌐 | High | |
| ajp_Arab[NEW] | **South Levantine Arabic** | Arabic | Afro-Asiatic | Semitic | ✗ | Low | Ammani |
| aka_Latn[NEW] | **Akan** | Latin | Atlantic-Congo | Kwa Volta-Congo | ✗ | Low | Asante |
| amh_Ethi | **Amharic** | Ge'ez | Afro-Asiatic | Semitic | 🌐 | Low | Addis Ababa |
| apc_Arab[NEW] | **North Levantine Arabic** | Arabic | Afro-Asiatic | Semitic | ✗ | Low | |
| arb_Arab | **Modern Standard Arabic** | Arabic | Afro-Asiatic | Semitic | 🌐 | High | |
| arb_Latn[NEW] | **Modern Standard Arabic** | Latin | Afro-Asiatic | Semitic | ✗ | Low | |
| ars_Arab[NEW] | **Najdi Arabic** | Arabic | Afro-Asiatic | Semitic | ✗ | Low | |
| ary_Arab[NEW] | **Moroccan Arabic** | Arabic | Afro-Asiatic | Semitic | ✗ | Low | |
| arz_Arab[NEW] | **Egyptian Arabic** | Arabic | Afro-Asiatic | Semitic | ✗ | Low | |
| asm_Beng | **Assamese** | Bengali | Indo-European | Indo-Aryan | 🌐 | Low | Eastern |
| ast_Latn | **Asturian** | Latin | Indo-European | Italic | ✗ | Low | Central |
| awa_Deva[NEW] | **Awadhi** | Devanagari | Indo-European | Indo-Aryan | ✗ | Low | Ayodhya |
| ayr_Latn[NEW] | **Central Aymara** | Latin | Aymaran | Central Southern Aymara | 🌐 | Low | Aymara La Paz jilata |
| azb_Arab[NEW] | **South Azerbaijani** | Arabic | Turkic | Common Turkic | ✗ | Low | Tabrizi |
| azj_Latn | **North Azerbaijani** | Latin | Turkic | Common Turkic | 🌐 | Low | Shirvan |
| bak_Cyrl[NEW] | **Bashkir** | Cyrillic | Turkic | Common Turkic | 🌐 | Low | Literary |
| bam_Latn[NEW] | **Bambara** | Latin | Mande | Western Mande | 🌐 | Low | |
| ban_Latn[NEW] | **Balinese** | Latin | Austronesian | Malayo-Polynesian | ✗ | Low | |
| bel_Cyrl | **Belarusian** | Cyrillic | Indo-European | Balto-Slavic | 🌐 | Low | Central |
| bem_Latn[NEW] | **Bemba** | Latin | Atlantic-Congo | Benue-Congo | ✗ | Low | Central |
| ben_Beng | **Bengali** | Bengali | Indo-European | Indo-Aryan | 🌐 | High | Rarhi |
| bho_Deva[NEW] | **Bhojpuri** | Devanagari | Indo-European | Indo-Aryan | 🌐 | Low | |
| bjn_Arab[NEW] | **Banjar** | Arabic | Austronesian | Malayo-Polynesian | ✗ | Low | Banjar Kuala |
| bjn_Latn[NEW] | **Banjar** | Latin | Austronesian | Malayo-Polynesian | ✗ | Low | Banjar Kuala |
| bod_Tibt[NEW] | **Standard Tibetan** | Tibetan | Sino-Tibetan | Bodic | 🌐 | Low | Lhasa |
| bos_Latn | **Bosnian** | Latin | Indo-European | Balto-Slavic | 🌐 | High | |
| bug_Latn[NEW] | **Buginese** | Latin | Austronesian | Malayo-Polynesian | ✗ | Low | Bone |
| bul_Cyrl | **Bulgarian** | Cyrillic | Indo-European | Balto-Slavic | 🌐 | High | |
| cat_Latn | **Catalan** | Latin | Indo-European | Italic | 🌐 | High | |
| ceb_Latn | **Cebuano** | Latin | Austronesian | Malayo-Polynesian | 🌐 | Low | |
| ces_Latn | **Czech** | Latin | Indo-European | Balto-Slavic | 🌐 | High | |
| cjk_Latn[NEW] | **Chokwe** | Latin | Atlantic-Congo | Benue-Congo | ✗ | Low | |
| ckb_Arab | **Central Kurdish** | Arabic | Indo-European | Iranian | 🌐 | Low | |
| crh_Latn[NEW] | **Crimean Tatar** | Latin | Turkic | Common Turkic | ✗ | Low | |
| cym_Latn | **Welsh** | Latin | Indo-European | Celtic | 🌐 | Low | Y Wyndodeg |
| dan_Latn | **Danish** | Latin | Indo-European | Germanic | 🌐 | High | |
| deu_Latn | **German** | Latin | Indo-European | Germanic | 🌐 | High | |
| dik_Latn[NEW] | **Southwestern Dinka** | Latin | Nilotic | Western Nilotic | ✗ | Low | Rek |
| dyu_Latn[NEW] | **Dyula** | Latin | Mande | Western Mande | ✗ | Low | |
| dzo_Tibt[NEW] | **Dzongkha** | Tibetan | Sino-Tibetan | Bodic | ✗ | Low | |
| ell_Grek | **Greek** | Greek | Indo-European | Graeco-Phrygian | 🌐 | High | |
| eng_Latn | **English** | Latin | Indo-European | Germanic | 🌐 | High | |
| epo_Latn[NEW] | **Esperanto** | Latin | Constructed | Esperantic | 🌐 | Low | |
| est_Latn | **Estonian** | Latin | Uralic | Finnic | 🌐 | High | |
| eus_Latn[NEW] | **Basque** | Latin | Basque | – | 🌐 | High | |
| ewe_Latn[NEW] | **Ewe** | Latin | Atlantic-Congo | Kwa Volta-Congo | 🌐 | Low | Aŋlo |
| fao_Latn[NEW] | **Faroese** | Latin | Indo-European | Germanic | 🌐 | Low | |
| fij_Latn[NEW] | **Fijian** | Latin | Austronesian | Malayo-Polynesian | 🌐 | Low | Bau |
| fin_Latn | **Finnish** | Latin | Uralic | Finnic | 🌐 | High | |
| fon_Latn[NEW] | **Fon** | Latin | Atlantic-Congo | Kwa Volta-Congo | ✗ | Low | |
| fra_Latn | **French** | Latin | Indo-European | Italic | 🌐 | High | |
| fur_Latn[NEW] | **Friulian** | Latin | Indo-European | Italic | ✗ | Low | Central |
| fuv_Latn | **Nigerian Fulfulde** | Latin | Atlantic-Congo | North-Central Atlantic | ✗ | Low | Sokoto |
| gla_Latn[NEW] | **Scottish Gaelic** | Latin | Indo-European | Celtic | ✗ | Low | Northern Hebrides |

| Code | Language | Script | Family | Subgrouping | 🌐 | Res. | Specification |
|---|---|---|---|---|---|---|---|
| gle_Latn | **Irish** | Latin | Indo-European | Celtic | 🌐 | Low | |
| glg_Latn | **Galician** | Latin | Indo-European | Italic | 🌐 | Low | |
| grn_Latn[NEW] | **Guarani** | Latin | Tupian | Maweti-Guarani | 🌐 | Low | |
| guj_Gujr | **Gujarati** | Gujarati | Indo-European | Indo-Aryan | 🌐 | Low | Amdavadi/Surti |
| hat_Latn[NEW] | **Haitian Creole** | Latin | Indo-European | Italic | 🌐 | Low | |
| hau_Latn | **Hausa** | Latin | Afro-Asiatic | Chadic | 🌐 | Low | |
| heb_Hebr | **Hebrew** | Hebrew | Afro-Asiatic | Semitic | 🌐 | High | |
| hin_Deva | **Hindi** | Devanagari | Indo-European | Indo-Aryan | 🌐 | High | |
| hne_Deva[NEW] | **Chhattisgarhi** | Devanagari | Indo-European | Indo-Aryan | ✗ | Low | |
| hrv_Latn | **Croatian** | Latin | Indo-European | Balto-Slavic | 🌐 | High | |
| hun_Latn | **Hungarian** | Latin | Uralic | – | 🌐 | High | |
| hye_Armn | **Armenian** | Armenian | Indo-European | Armenic | 🌐 | Low | Yerevan |
| ibo_Latn | **Igbo** | Latin | Atlantic-Congo | Benue-Congo | 🌐 | Low | Central |
| ilo_Latn[NEW] | **Ilocano** | Latin | Austronesian | Malayo-Polynesian | 🌐 | Low | |
| ind_Latn | **Indonesian** | Latin | Austronesian | Malayo-Polynesian | 🌐 | High | |
| isl_Latn | **Icelandic** | Latin | Indo-European | Germanic | 🌐 | High | |
| ita_Latn | **Italian** | Latin | Indo-European | Italic | 🌐 | High | |
| jav_Latn | **Javanese** | Latin | Austronesian | Malayo-Polynesian | 🌐 | Low | |
| jpn_Jpan | **Japanese** | Japanese | Japonic | Japanesic | 🌐 | High | |
| kab_Latn[NEW] | **Kabyle** | Latin | Afro-Asiatic | Berber | ✗ | Low | North Eastern |
| kac_Latn[NEW] | **Jingpho** | Latin | Sino-Tibetan | Brahmaputran | ✗ | Low | |
| kam_Latn | **Kamba** | Latin | Atlantic-Congo | Benue-Congo | ✗ | Low | Machakos |
| kan_Knda | **Kannada** | Kannada | Dravidian | South Dravidian | 🌐 | Low | Central |
| kas_Arab[NEW] | **Kashmiri** | Arabic | Indo-European | Indo-Aryan | ✗ | Low | Kishtwari |
| kas_Deva[NEW] | **Kashmiri** | Devanagari | Indo-European | Indo-Aryan | ✗ | Low | Kishtwari |
| kat_Geor | **Georgian** | Georgian | Kartvelian | Georgian-Zan | 🌐 | Low | Kartlian |
| knc_Arab[NEW] | **Central Kanuri** | Arabic | Saharan | Western Saharan | ✗ | Low | Yerwa |
| knc_Latn[NEW] | **Central Kanuri** | Latin | Saharan | Western Saharan | ✗ | Low | Yerwa |
| kaz_Cyrl | **Kazakh** | Cyrillic | Turkic | Common Turkic | 🌐 | High | |
| kbp_Latn[NEW] | **Kabiyè** | Latin | Atlantic-Congo | North Volta-Congo | ✗ | Low | Kɛ̀wɛ |
| kea_Latn[NEW] | **Kabuverdianu** | Latin | Indo-European | Italic | ✗ | Low | Sotavento |
| khm_Khmr | **Khmer** | Khmer | Austroasiatic | Khmeric | 🌐 | Low | Central |
| kik_Latn[NEW] | **Kikuyu** | Latin | Atlantic-Congo | Benue-Congo | ✗ | Low | Southern |
| kin_Latn[NEW] | **Kinyarwanda** | Latin | Atlantic-Congo | Benue-Congo | 🌐 | Low | |
| kir_Cyrl | **Kyrgyz** | Cyrillic | Turkic | Common Turkic | 🌐 | Low | Northern |
| kmb_Latn[NEW] | **Kimbundu** | Latin | Atlantic-Congo | Benue-Congo | ✗ | Low | |
| kmr_Latn[NEW] | **Northern Kurdish** | Latin | Indo-European | Iranian | 🌐 | Low | |
| kon_Latn[NEW] | **Kikongo** | Latin | Atlantic-Congo | Benue-Congo | ✗ | Low | |
| kor_Hang | **Korean** | Hangul | Koreanic | Korean | 🌐 | High | |
| lao_Laoo | **Lao** | Lao | Tai-Kadai | Kam-Tai | 🌐 | Low | Vientiane |
| lij_Latn[NEW] | **Ligurian** | Latin | Indo-European | Italic | ✗ | Low | Zeneise |
| lim_Latn[NEW] | **Limburgish** | Latin | Indo-European | Germanic | ✗ | Low | Maastrichtian |
| lin_Latn | **Lingala** | Latin | Atlantic-Congo | Benue-Congo | 🌐 | Low | |
| lit_Latn | **Lithuanian** | Latin | Indo-European | Balto-Slavic | 🌐 | High | |
| lmo_Latn[NEW] | **Lombard** | Latin | Indo-European | Italic | ✗ | Low | Western |
| ltg_Latn[NEW] | **Latgalian** | Latin | Indo-European | Balto-Slavic | ✗ | Low | Central |
| ltz_Latn | **Luxembourgish** | Latin | Indo-European | Germanic | 🌐 | Low | |
| lua_Latn[NEW] | **Luba-Kasai** | Latin | Atlantic-Congo | Benue-Congo | ✗ | Low | |
| lug_Latn | **Ganda** | Latin | Atlantic-Congo | Benue-Congo | 🌐 | Low | |
| luo_Latn | **Luo** | Latin | Nilotic | Western Nilotic | ✗ | Low | |
| lus_Latn[NEW] | **Mizo** | Latin | Sino-Tibetan | Kuki-Chin-Naga | 🌐 | Low | Aizawl |
| lvs_Latn | **Standard Latvian** | Latin | Indo-European | Balto-Slavic | 🌐 | High | |
| mag_Deva[NEW] | **Magahi** | Devanagari | Indo-European | Indo-Aryan | ✗ | Low | Gaya |
| mai_Deva[NEW] | **Maithili** | Devanagari | Indo-European | Indo-Aryan | 🌐 | Low | |
| mal_Mlym | **Malayalam** | Malayalam | Dravidian | South Dravidian | 🌐 | Low | |
| mar_Deva | **Marathi** | Devanagari | Indo-European | Indo-Aryan | 🌐 | Low | Varhadi |
| min_Arab[NEW] | **Minangkabau** | Arabic | Austronesian | Malayo-Polynesian | ✗ | Low | Agam-Tanah Datar |
| min_Latn[NEW] | **Minangkabau** | Latin | Austronesian | Malayo-Polynesian | ✗ | Low | Agam-Tanah Datar |
| mkd_Cyrl | **Macedonian** | Cyrillic | Indo-European | Balto-Slavic | 🌐 | High | |
| plt_Latn[NEW] | **Plateau Malagasy** | Latin | Austronesian | Malayo-Polynesian | 🌐 | Low | Merina |
| mlt_Latn | **Maltese** | Latin | Afro-Asiatic | Semitic | 🌐 | High | |

| Code | Language | Script | Family | Subgrouping | 🌐 | Res. | Specification |
|---|---|---|---|---|---|---|---|
| mni_Beng[NEW] | **Meitei** | Bengali | Sino-Tibetan | Kuki-Chin-Naga | ✗ | Low | |
| khk_Cyrl | **Halh Mongolian** | Cyrillic | Mongolic-Khitan | Mongolic | 🌐 | Low | |
| mos_Latn[NEW] | **Mossi** | Latin | Atlantic-Congo | North Volta-Congo | ✗ | Low | Ouagadougou |
| mri_Latn | **Maori** | Latin | Austronesian | Malayo-Polynesian | 🌐 | Low | Waikato-Ngapuhi |
| mya_Mymr | **Burmese** | Myanmar | Sino-Tibetan | Burmo-Qiangic | 🌐 | Low | Mandalay-Yangon |
| nld_Latn | **Dutch** | Latin | Indo-European | Germanic | 🌐 | High | |
| nno_Latn[NEW] | **Norwegian Nynorsk** | Latin | Indo-European | Germanic | ✗ | Low | |
| nob_Latn | **Norwegian Bokmål** | Latin | Indo-European | Germanic | 🌐 | Low | |
| npi_Deva | **Nepali** | Devanagari | Indo-European | Indo-Aryan | 🌐 | Low | Eastern |
| nso_Latn | **Northern Sotho** | Latin | Atlantic-Congo | Benue-Congo | 🌐 | Low | |
| nus_Latn[NEW] | **Nuer** | Latin | Nilotic | Western Nilotic | ✗ | Low | |
| nya_Latn | **Nyanja** | Latin | Atlantic-Congo | Benue-Congo | 🌐 | Low | |
| oci_Latn | **Occitan** | Latin | Indo-European | Italic | ✗ | Low | |
| gaz_Latn[NEW] | **West Central Oromo** | Latin | Afro-Asiatic | Cushitic | 🌐 | Low | |
| ory_Orya | **Odia** | Oriya | Indo-European | Indo-Aryan | 🌐 | Low | Baleswari (Northern) |
| pag_Latn[NEW] | **Pangasinan** | Latin | Austronesian | Malayo-Polynesian | ✗ | Low | |
| pan_Guru | **Eastern Panjabi** | Gurmukhi | Indo-European | Indo-Aryan | 🌐 | Low | Majhi |
| pap_Latn[NEW] | **Papiamento** | Latin | Indo-European | Italic | ✗ | Low | Römer-Maduro-Jonis |
| pes_Arab | **Western Persian** | Arabic | Indo-European | Iranian | 🌐 | High | |
| pol_Latn | **Polish** | Latin | Indo-European | Balto-Slavic | 🌐 | High | |
| por_Latn | **Portuguese** | Latin | Indo-European | Italic | 🌐 | High | Brazil |
| prs_Arab[NEW] | **Dari** | Arabic | Indo-European | Iranian | 🌐 | Low | Kabuli |
| pbt_Arab | **Southern Pashto** | Arabic | Indo-European | Iranian | 🌐 | Low | Literary |
| quy_Latn[NEW] | **Ayacucho Quechua** | Latin | Quechuan | Chinchay | 🌐 | Low | Southern Quechua |
| ron_Latn | **Romanian** | Latin | Indo-European | Italic | 🌐 | High | |
| run_Latn[NEW] | **Rundi** | Latin | Atlantic-Congo | Benue-Congo | ✗ | Low | |
| rus_Cyrl | **Russian** | Cyrillic | Indo-European | Balto-Slavic | 🌐 | High | |
| sag_Latn[NEW] | **Sango** | Latin | Atlantic-Congo | North Volta-Congo | ✗ | Low | |
| san_Deva[NEW] | **Sanskrit** | Devanagari | Indo-European | Indo-Aryan | 🌐 | Low | |
| sat_Olck[NEW] | **Santali** | Ol Chiki | Austroasiatic | Mundaic | ✗ | Low | |
| scn_Latn[NEW] | **Sicilian** | Latin | Indo-European | Italic | ✗ | Low | Literary Sicilian |
| shn_Mymr[NEW] | **Shan** | Myanmar | Tai-Kadai | Kam-Tai | ✗ | Low | |
| sin_Sinh[NEW] | **Sinhala** | Sinhala | Indo-European | Indo-Aryan | 🌐 | Low | |
| slk_Latn | **Slovak** | Latin | Indo-European | Balto-Slavic | 🌐 | High | |
| slv_Latn[NEW] | **Slovenian** | Latin | Indo-European | Balto-Slavic | 🌐 | High | |
| smo_Latn[NEW] | **Samoan** | Latin | Austronesian | Malayo-Polynesian | 🌐 | Low | |
| sna_Latn | **Shona** | Latin | Atlantic-Congo | Benue-Congo | 🌐 | Low | |
| snd_Arab | **Sindhi** | Arabic | Indo-European | Indo-Aryan | 🌐 | Low | Vicholi |
| som_Latn | **Somali** | Latin | Afro-Asiatic | Cushitic | 🌐 | Low | Nsom |
| sot_Latn[NEW] | **Southern Sotho** | Latin | Atlantic-Congo | Benue-Congo | 🌐 | High | |
| spa_Latn | **Spanish** | Latin | Indo-European | Italic | 🌐 | High | Latin American |
| als_Latn[NEW] | **Tosk Albanian** | Latin | Indo-European | Albanian | 🌐 | High | |
| srd_Latn[NEW] | **Sardinian** | Latin | Indo-European | Italic | ✗ | Low | Logudorese and Campidanese |
| srp_Cyrl | **Serbian** | Cyrillic | Indo-European | Balto-Slavic | 🌐 | Low | |
| ssw_Latn[NEW] | **Swati** | Latin | Atlantic-Congo | Benue-Congo | ✗ | Low | |
| sun_Latn[NEW] | **Sundanese** | Latin | Austronesian | Malayo-Polynesian | 🌐 | Low | |
| swe_Latn | **Swedish** | Latin | Indo-European | Germanic | 🌐 | High | |
| swh_Latn | **Swahili** | Latin | Atlantic-Congo | Benue-Congo | 🌐 | High | Kiunguja |
| szl_Latn[NEW] | **Silesian** | Latin | Indo-European | Balto-Slavic | ✗ | Low | |
| tam_Taml | **Tamil** | Tamil | Dravidian | South Dravidian | 🌐 | Low | Chennai |
| tat_Cyrl[NEW] | **Tatar** | Cyrillic | Turkic | Common Turkic | 🌐 | Low | Central and Middle |
| tel_Telu | **Telugu** | Telugu | Dravidian | South Dravidian | 🌐 | Low | Coastal |
| tgk_Cyrl | **Tajik** | Cyrillic | Indo-European | Iranian | 🌐 | Low | |
| tgl_Latn | **Tagalog** | Latin | Austronesian | Malayo-Polynesian | 🌐 | High | |
| tha_Thai | **Thai** | Thai | Tai-Kadai | Kam-Tai | 🌐 | High | |
| tir_Ethi[NEW] | **Tigrinya** | Ge'ez | Afro-Asiatic | Semitic | 🌐 | Low | |
| taq_Latn[NEW] | **Tamasheq** | Latin | Afro-Asiatic | Berber | ✗ | Low | Kal Ansar |
| taq_Tfng[NEW] | **Tamasheq** | Tifinagh | Afro-Asiatic | Berber | ✗ | Low | Kal Ansar |
| tpi_Latn[NEW] | **Tok Pisin** | Latin | Indo-European | Germanic | ✗ | Low | |
| tsn_Latn[NEW] | **Tswana** | Latin | Atlantic-Congo | Benue-Congo | ✗ | High | Sehurutshe |
| tso_Latn[NEW] | **Tsonga** | Latin | Atlantic-Congo | Benue-Congo | 🌐 | Low | |

| Code | Language | Script | Family | Subgrouping | 🌐 | Res. | Specification |
|---|---|---|---|---|---|---|---|
| tuk_Latn[NEW] | **Turkmen** | Latin | Turkic | Common Turkic | 🌐 | Low | Teke |
| tum_Latn[NEW] | **Tumbuka** | Latin | Atlantic-Congo | Benue-Congo | ✗ | Low | Rumphi |
| tur_Latn | **Turkish** | Latin | Turkic | Common Turkic | 🌐 | High | |
| twi_Latn[NEW] | **Twi** | Latin | Atlantic-Congo | Kwa Volta-Congo | 🌐 | Low | Akuapem |
| tzm_Tfng[NEW] | **Central Atlas Tamazight** | Tifinagh | Afro-Asiatic | Berber | ✗ | Low | |
| uig_Arab[NEW] | **Uyghur** | Arabic | Turkic | Common Turkic | 🌐 | Low | |
| ukr_Cyrl | **Ukrainian** | Cyrillic | Indo-European | Balto-Slavic | 🌐 | High | |
| umb_Latn | **Umbundu** | Latin | Atlantic-Congo | Benue-Congo | ✗ | Low | |
| urd_Arab | **Urdu** | Arabic | Indo-European | Indo-Aryan | 🌐 | Low | Lashkari |
| uzn_Latn | **Northern Uzbek** | Latin | Turkic | Common Turkic | 🌐 | High | |
| vec_Latn[NEW] | **Venetian** | Latin | Indo-European | Italic | ✗ | Low | Venice |
| vie_Latn | **Vietnamese** | Latin | Austroasiatic | Vietic | 🌐 | High | |
| war_Latn[NEW] | **Waray** | Latin | Austronesian | Malayo-Polynesian | ✗ | Low | Tacloban |
| wol_Latn | **Wolof** | Latin | Atlantic-Congo | North-Central Atlantic | ✗ | Low | Dakkar |
| xho_Latn | **Xhosa** | Latin | Atlantic-Congo | Benue-Congo | 🌐 | High | Ngqika |
| ydd_Hebr[NEW] | **Eastern Yiddish** | Hebrew | Indo-European | Germanic | 🌐 | Low | Hasidic |
| yor_Latn | **Yoruba** | Latin | Atlantic-Congo | Benue-Congo | 🌐 | Low | Ọyọ and Ibadan |
| yue_Hant[NEW] | **Yue Chinese** | Han (Traditional) | Sino-Tibetan | Sinitic | 🌐 | Low | |
| zho_Hans | **Chinese** | Han (Simplified) | Sino-Tibetan | Sinitic | 🌐 | High | |
| zho_Hant | **Chinese** | Han (Traditional) | Sino-Tibetan | Sinitic | 🌐 | High | |
| zsm_Latn | **Standard Malay** | Latin | Austronesian | Malayo-Polynesian | 🌐 | High | |
| zul_Latn | **Zulu** | Latin | Atlantic-Congo | Benue-Congo | 🌐 | High | |

Table 1: **204 Languages of No Language Left Behind:** We display the language *Code*, language name, *Script*, and language *Family*. The symbol 🌐 indicates machine translation support by Google and/or Microsoft, whereas ✗ indicates support by neither. *Res.* indicates if we classify the language as high or low-resource. *Specification* contains, if available, additional information on the language variant collected in Flores-200. The superscript[NEW] indicates new languages added to Flores-200 compared to Flores-101.

as part of a community request process.[3] Next, we solicited lists of languages spoken in various regions by native speakers, focusing particularly on African languages—a category of languages that have historically been underrepresented in translation efforts (Nekoto et al., 2020). We then examined language coverage in multiple existing datasets in the natural language processing community, paying focused attention on training datasets without accompanying evaluation datasets. Finally, we considered the adoption and usage of each language by looking at the approximate number of native speakers and other community-level variables relevant to our work.

Next, for each of the language candidates, we partnered with linguists from various specialized language service providers to understand if each of these languages has a standardized written form. We did this because having a reliable, high-quality evaluation dataset is critical to accelerated experimental progress. However, prioritizing languages with fairly standardized written forms has notable downsides (see Appendix A). For one, many languages have natural variations and are being written in different standards or scripts in different regions. For instance, languages such as Fulah include several distinct varieties and languages such as Kashmiri and Central Kanuri contain multiple scripts in common use. Systematically documenting these dimensions helped us assess how we could best support multiple variants of different languages (such as languages with multiple writing systems or natural variation).

3. https://meta.wikimedia.org/wiki/Language_proposal_policy

In tandem with these considerations, deciding which languages to include in the final list ultimately came down to assessing the potential impact we might have on the respective low-resource language communities. For instance, we exclude languages with extremely low number of native speakers. Without a concerted plan to thoroughly understand the needs of these communities and potential risks we could cause, we do not feel comfortable including their languages in our effort. Keeping in line with our guiding principles, many of the languages that made the final cut have a presence on Wikipedia and are from historically underrepresented regions. Last but not least, it is worth noting that in this work, we exclude many languages that do not have written standards or are predominantly oral. It is our hope that future research could direct more attention at languages with different modalities.

**Language Information.** In accordance with the #BenderRule (Bender, 2019), we summarize information about each of our 204 supported languages in Table 1.

**Code.** We represent each language with a BCP 47 tag sequence using a three-letter ISO 639-3 code as the base subtag, which we complement with ISO 15924 script subtags, as we collected resources for several languages in more than one script.

**Language.** There may be multiple ways to refer to the same language; due to formatting limitations, only one of the versions is displayed. The language names have been cross-referenced with major linguistic information platforms such as Ethnologue (Lewis, 2009) and Glottolog (Hammarström et al., 2022).

**Script.** The English name of the script is provided. As some languages are written in more than one script, we work towards supporting this natural variation.

**Family and Subgrouping.** We provide Language family information for each language based on the Glottolog database (Hammarström et al., 2022).

**Web Support.** We examine if each language is supported by Google Translate[4] and/or Microsoft Translate.[5] The symbol 🌐 indicates that either or both platforms supports the language. The symbol ✗ indicates that neither platform supports the language.[6]

**Resource-Level (Res).** We categorize a language as *low-resource* if there are fewer than 1M publicly available, de-duplicated bitext samples with any other language within our set of 200 languages. Note this goes beyond counting English-centric training data, as many languages may have available datasets in languages spoken more prominently in their region. For example, many countries in Africa are Francophone. This results in 150 languages classified as low-resource.

**Specification.** This column contains, if available, additional information regarding the specific language variety or region represented.

The language information provided in Table 1 reflects the resources gathered through the Flores-200 collection efforts, which are described in the next section.

4. `https://translate.google.com/`
5. `http://www.bing.com/translator`
6. Information was accessed on June 15, 2022

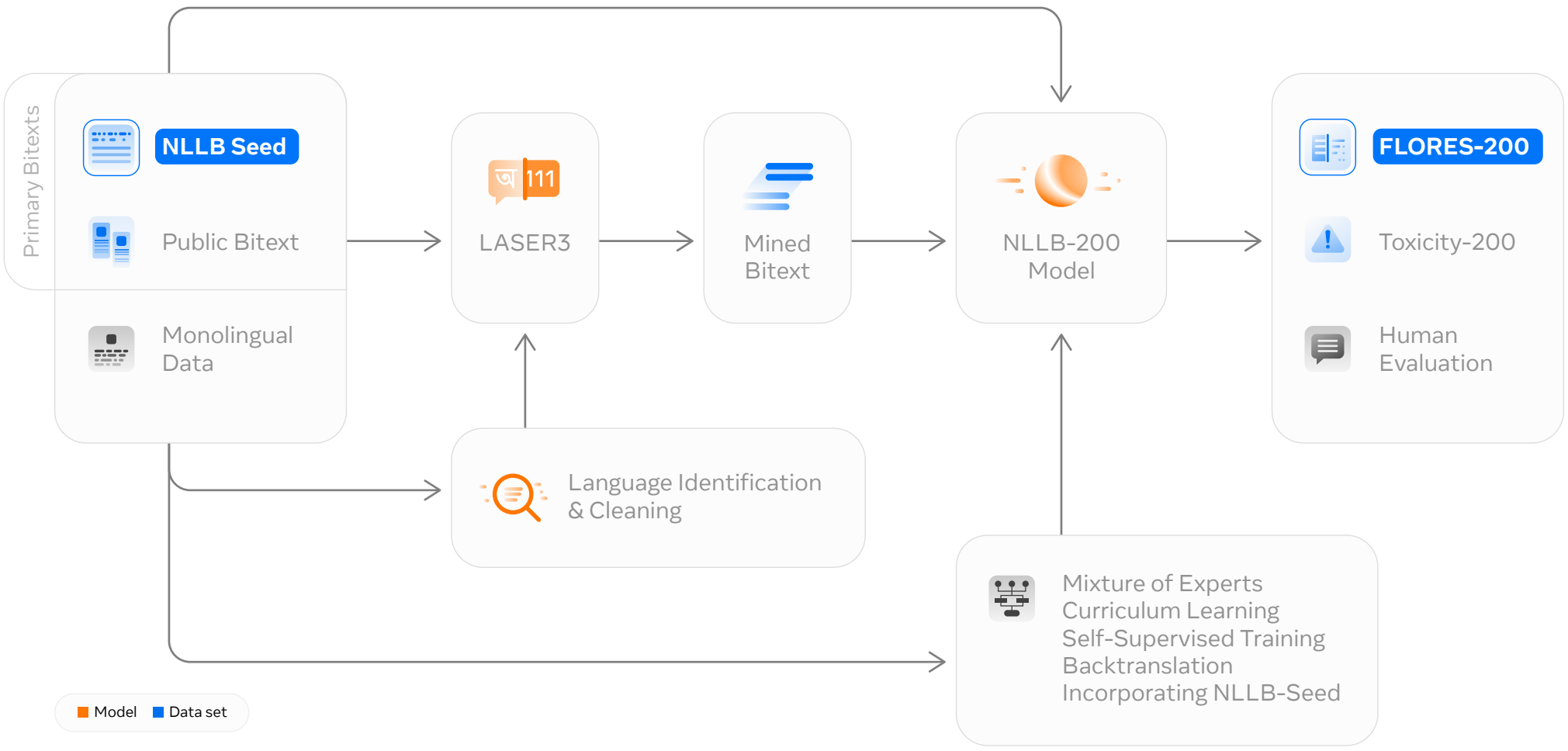

Figure 3: **Human-Translated Dataset Contributions of No Language Left Behind**: As highlighted, these datasets enable model training and evaluation.

## 4. Creating Professionally Translated Datasets: FLORES-200 and NLLB-Seed

Low-resource translation faces several challenges, first and foremost that of data availability. In this section, we describe three components devised to overcome this problem, shown in Figure 3. First, we describe the creation of Flores-200, a high quality, many-to-many benchmark dataset that doubles the language coverage of a previous effort known as Flores-101. Then, we trace the development process of professionally-translated *seed bitext* data in 39 low-resource languages, giving us the ability to train any models that require parallel data. Finally, we describe NLLB-MD, a dataset in multiple different domains to evaluate generalizable translation capability. These resources enable the evaluation and creation of models for languages that previously had marginal support.

### 4.1 FLORES-200

A major area of focus in machine translation research has been on the development of high-quality evaluation datasets, or benchmarks that can be reliably used to assess progress in the field. The ability to evaluate allows us to compare different approaches and understand what requires further research and development. The creation of benchmark datasets at the yearly Workshop on Machine Translation (Akhbardeh et al., 2021) led to rapid progress on translation directions such as English to German and English to French. We are also seeing recent work on creating low-resource translation datasets as illustrated by the SALT (Akera et al., 2022; Babirye et al., 2022) and the AmericasNLI (Ebrahimi et al., 2022) datasets. Beyond the field of translation, evaluation benchmarks such as SQuAD (Rajpurkar et al., 2016), GLUE (Wang et al., 2018), and even the Penn Treebank language modeling benchmark (Mikolov and Zweig, 2012) propelled significant research advances.

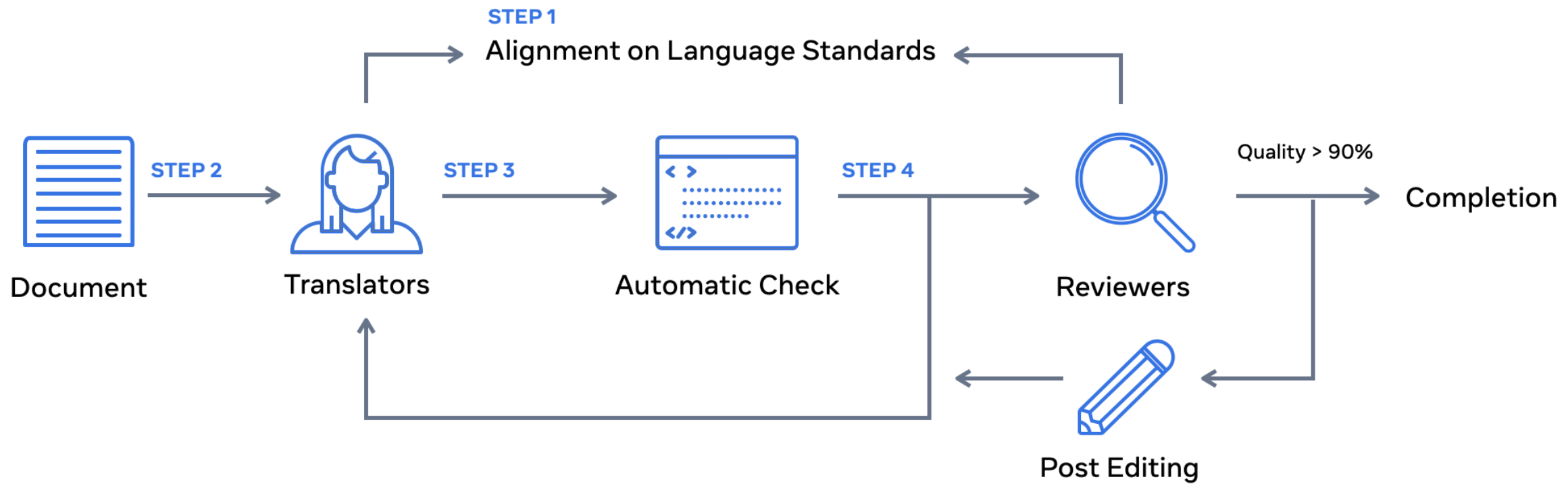

Figure 4: **FLORES-200 Translation Workflow:** We created a complex, multi-step process to ensure quality. First, professional translators and reviewers aligned on language standards. Next, translators translated the full set of FLORES-200 sentences, followed by automated checks. Subsequently, the group of independent reviewers reviewed the quality, and based on their assessment, we sent some translations out for post-editing. If the quality assessment indicated that the quality is above 90 percent, the language is considered ready for inclusion in FLORES-200.

The creation of FLORES-200 seeks to double the existing language coverage of FLORES-101. This raises significant challenges due to the even more low-resource nature of the languages we have introduced in this effort. More specifically, these languages may require ever increasingly specialized professional translators, have less standardization, and the verifying process to ensure translation quality becomes more complex. Below, following a brief summary of the characteristics of FLORES-101, we describe in detail how we overcome these new challenges in the creation of FLORES-200, paying particular attention to the adapted protocol and quality assurance mechanisms. Then, we present an analysis on the overall quality of our evaluation benchmark.

#### 4.1.1 Benchmark Creation for Low-Resource Languages

**Preliminaries.** As a significant extension of FLORES-101, FLORES-200 consists of 3001 sentences sampled from English-language Wikimedia projects for **204** total languages. Approximately one third of sentences are collected from each of these sources: Wikinews, Wikijunior, and Wikivoyage. The content is professionally translated into 200+ languages to create FLORES-200. As we translate the same set of content into all languages, FLORES-200 is a *many-to-many multilingual* benchmark. We refer the reader to Goyal et al. (2022) for greater detail.

**Finding Professional Translators and Translation Reviewers.** FLORES-200 is created with professional human translators who translate the FLORES source dataset into the target languages and a separate group of independent translation reviewers who perform quality assessments of the human translations and provide translation feedback to the translators. Both translators and reviewers undergo vetting processes, handled by language service providers (LSPs). Translators are required to be native speakers and educated in the target language and have a high level fluency in English. Translators are required to have at least two to three years of translation experience in the relevant language pair if they have an academic degree in translation or linguistics and three to five years of translation experience if they do not have any relevant academic qualification. Translators also undergo a translation test every 18 months to assess their translation quality. Further, Flores-200 reviewers are also required to be native speakers of the target language. Reviewers typically have a translation degree, at least five years of experience working as a translator, translation review experience, and where possible are accredited by a relevant translation board.

We note that these are stringent standards, and extensions of Flores-200 to even more low-resource languages in the future may be difficult. Already for many languages, finding a reviewer that meets the criteria above is very challenging. In these cases, we modified the qualification process to accept applications from reviewers with more general language degrees such as Linguistics or African Language Studies, or no degree provided they have had extensive commercial translation experience (e.g. >10 years). To cover even more low-resource languages in the future, we believe that there are several ways to work with experienced and skilled translators while maintaining high quality standards. For instance, one of such solutions is to translate from non-English source languages. We pilot this process and describe it in greater detail in Section 4.1.2.

**Flores-200 Translation Workflow.** The Flores-200 data creation workflow incorporates the original Flores-101 processes along with a few new initial phases as shared in detail below.

- **Alignment Phase:** We have introduced an initial alignment phase to the workflow for the translators and reviewers before translating Flores-200. There are several steps incorporated in alignment between the translation and quality assurance agencies – aligning on resourcing and target regions, linguistic topics between the translators and reviewers per language through a new alignment template, and query logs between the linguists on both sides. The alignment template helped linguists identify approaches on the language script, standardization, spelling, borrowed terms, neologisms, informative content style, and resources for glossaries, and sample content in the target language. This has been especially helpful for languages with less established standards for translation.

- **Translation Phase:** Translation then begins with an initial translation phase, where the same 200 sentences are translated by all participating translators for each language. The initial translation data contains an even split across the three sources — Wikinews, Wikijunior, and Wikivoyage, with the segments corresponding to the same articles for context and continuity. The initial translations are then sent to the QA LSP team for review. The main focus of the initial translation and QA steps is to understand and align on the translation approach between the translators and reviewers. The report contains sentence-level feedback (identified error category, error severity level and comments where possible) and high-level feedback on locale expectations, use of specified script, use of borrowings and/or neologisms, named entities, and overall style and register.

- **Iteration:** Translation LSP teams may respond to the initial QA reports with arbitration. Adjustments are then made to all alignment materials where needed and the translation approach is updated and re-aligned on. The full translation of all 3000 sentences then begins (see Goyal et al. (2022) for details).

- **Completion:** When full translation is completed, the QA LSP team performs a final QA review and assesses a 20% sample data. Optional arbitration, rework and QA spot checks may follow if the final quality score of the translation dataset is below 90%.

#### 4.1.2 Benchmark Creation for Non-English Directions

The standard Flores-200 workflow focuses on translation only from English. While this standardizes the process for all languages, it has clear limitations. For example, there are many qualified translators who may not speak English, but are able to translate between several non-English languages. Further, several languages may be easier to translate from a non-English source. Instead, we focus on adaptation and transliteration and design customized QA workflows to support this.

**Translation of Arabic Languoids.** We apply this workflow to create datasets for various variants of Arabic, expanding our language coverage beyond Modern Standard Arabic to regional variants such as Moroccan Arabic. To create Flores-200 for Arabic variants, LSP teams analyzed the linguistic characteristics of each Arabic languoid and how much they differed from Modern Standard Arabic on various linguistic aspects such as vocabulary differences, grammatical and structural differences, influence from other regional languages and informative content style. Based on these analyses, Arabic languoids were either translated directly from English or adapted from the Modern Standard Arabic dataset with the English source provided as context.[7] For each languoid that implemented adaptation, LSP teams also created a termlist consisting of terms from Modern Standard Arabic and an equivalent term in the target Arabic languoid to ensure consistent adaptation.

Two tiers of quality assessment were created for adaptation from Modern Standard Arabic. One tier encompassed a partial QA review where the reviewer assessed a 10% sample data and reviewed the termlist. This process was applied to languoids that were assessed to have mainly vocabulary differences, some structural differences and some influence from other regional languages. Another tier required the reviewer to only assess the termlist as the languoids mainly differed from Modern Standard Arabic minimally and on vocabulary usage. The 90% quality threshold is applied as usual.

**Script Transliteration.** There were four languages (`ace_Arab`, `bjn_Arab`, `min_Arab`, `taq_Tfng`) that were transliterated from their Latin script counterparts. The translation LSP performs transliteration into the appropriate scripts. The QA LSP reviews a 20% sample of the transliterated text with the English source and Latin script data provided for context. In the QA report, transliteration errors are flagged only by severity level; there are no error categories for transliteration errors. Two or more errors found in one segment would be flagged with a *major* severity level. Anything fewer would be flagged as *minor*. The quality threshold for transliteration is 95%.

7. `acm_Arab`, `acq_Arab`, `aeb_Arab`, and `ars_Arab` were adapted.

| Overview Statistics | |
|---|---|
| # of sentences | 3001 |
| Avg # of words/sentence | 21 |
| # of articles | 842 |
| **Split** | **# of sentences** |
| dev | 997 |
| devtest | 1012 |
| test | 992 |
| # of Languages requiring Re-translation | 10 |
| Avg # of Re-translations | 1 |
| Max # of Re-translations | 2 |
| Avg # of Days to Translate 1 language | 42 |
| Avg # of Days to align | 28 |
| Avg # of Days for 1 language | 119 |
| Shortest Turnaround (days) for 1 language | 70 |
| Longest Turnaround (days) for 1 language | 287 |

Table 2: **FLORES at a Glance. (left)** FLORES is divided into three evaluation splits, totaling 3001 sentences. **(right)** Summary of Quality Control based on the statistics of 73 languages that implemented the new FLORES-200 workflow.

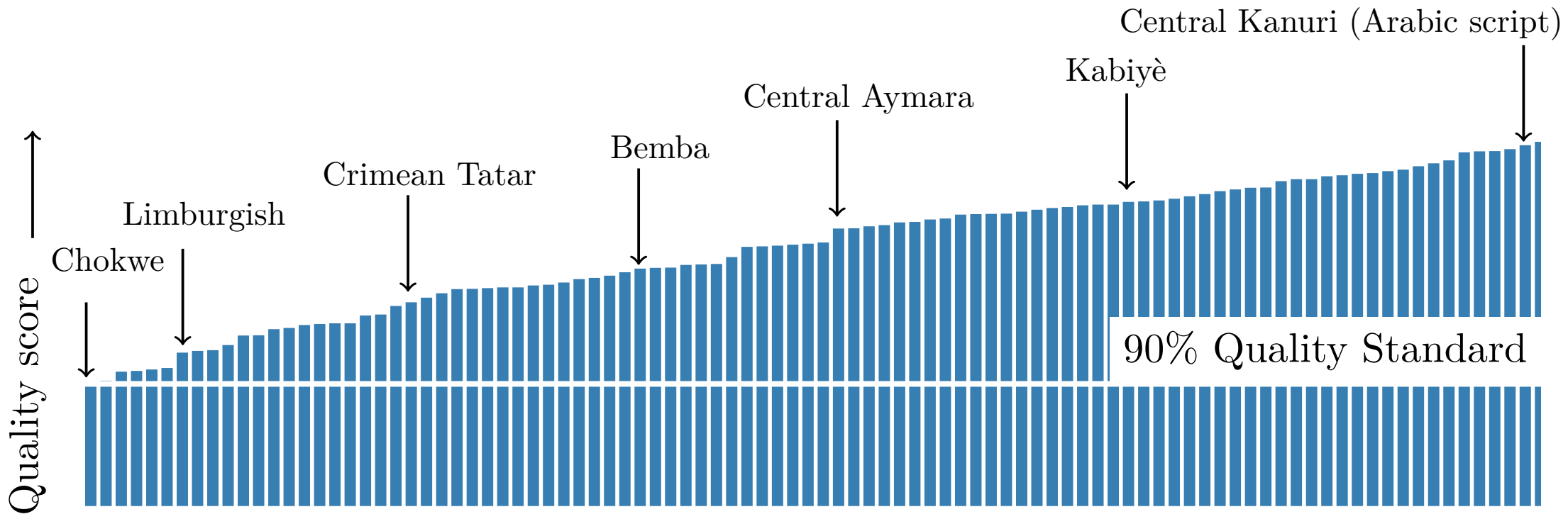

Figure 5: **Quality of FLORES-200**: We depict the quality assurance score for the languages in FLORES-200. The minimum acceptable standard is 90 percent.

#### 4.1.3 FLORES-200 AT A GLANCE

**Overview.** FLORES-200 consists of translations from 842 distinct web articles, totaling 3001 sentences. These sentences are divided into three splits: dev, devtest, and test. We release the full text of the dev and devtest splits, and keep the test set hidden through an evaluation server. On average, sentences are approximately 21 words long. We summarize this information in Table 2 (left).

**Quality.** To consider a language ready for inclusion in FLORES-200 requires a final human quality assurance evaluation. We display the quality scores of our languages in Figure 5 with several example languages labeled. Mistranslation and unnatural translation errors were still the most common errors found while assessing the quality of the human translations. These were mainly due to influences from other non-English languages that may be prominently used in the target communities, leading to excessive borrowings of vocabulary and grammar, literal translations due to infrequent usage of the target language in a formal, informative content style and the lower levels of standardization. There has also been an increasing trend in spelling inconsistencies in the human translations due to lower levels of standardization leading to inconsistent or even subjective or preferential approaches.

**Challenges in Creating Datasets for Very Low-Resource Languages.** Overall, compared to FLORES-101, our new translation workflow substantially streamlines the translation effort. For example, the number of languages requiring re-translation (see Table 2, right) is only 10, down from 45 in FLORES-101. However, despite these improvements, we continued to experience similar challenges as in FLORES-101 — but at even greater scale due to the increasing low-resource nature of the languages. For example, low-resource languages are not as often worked with in the localization or translation fields. As a result, there are lower levels of industry-wide standardization, leading to a more challenging path to navigate (Skadiņš et al., 2014a). This led to longer turnaround times, and often required finding new translators and reviewers several times. These challenges were especially felt during some of the more difficult languages such as Sicilian and Buginese, which have taken significantly longer periods of time to complete (287 days).

### 4.2 NLLB Seed Dataset

Machine learning is notoriously data-hungry, leading to many areas of research aimed at reducing the amount of required supervision. Recent advances in zero-shot learning (Chen et al., 2021; Gu et al., 2019; Johnson et al., 2017; Zhang et al., 2020) and self-supervised learning (Bapna et al., 2022; Liu et al., 2020; Ma et al., 2021), for instance, seek to reduce this reliance. However, generation tasks such as translation likely are unable to reach the desired quality levels without some starter data. For instance, it is challenging to produce a good translation without seeing a minimum number of sentences in a new language. Similarly, it may be difficult to classify which language a sentence is in without seeing reliable examples of text in different languages. To this end, we create NLLB-SEED, a set of professionally-translated sentences in the Wikipedia domain. NLLB-SEED consists of around six thousand sentences in 39 languages.[8]

Such a dataset has numerous potential uses. Critically, NLLB-SEED contains data that is definitely in the specified language, as it is fully professionally translated by humans. NLLB-SEED's target-side data in various languages can be utilized for language identification models that classify which language an arbitrary piece of input text is in. The dataset can also be used for its aligned bitext, for example to train translation models. Another option is to utilize NLLB-SEED to do domain finetuning, such as adapting general-purpose translation models to the Wikipedia domain.

**Source Sentence Selection.** Data for NLLB-SEED was sampled from Wikimedia's *List of articles every Wikipedia should have*,[9] a collection of 10,000 Wikidata IDs corresponding to notable topics in different fields of knowledge and human activity. These are split into 11 categories such as *People*, *History*, *Philosophy and Religion*, *Geography*. We uniformly sampled a subset of IDs from which we would draw data, and mapped these to the corresponding English Wikipedia articles. From each of these articles we then sampled the data that would be sent to translators. Instead of extracting individual sentences, which would have left translators with little context to work with, we chose to sample triplets of

8. Note that we focus on 39 for NLLB-SEED as these were the languages where there did not exist publicly available high-quality bitext for training in large quantities.
9. `https://meta.wikimedia.org/wiki/List_of_articles_every_Wikipedia_should_have/Expanded`

contiguous sentences, ensuring no more than one triplet per article was used (similar to Flores-200).

We note that like Flores-200, NLLB-Seed's source data is English-centric and sampled from English Wikipedia.[10] This has an important effect: the content reflects what Wikipedia editors find is relevant for English Wikipedia, and likely does not cover diverse content from different cultures. Further, the target text in NLLB-Seed is ultimately translated by humans, and thus potentially contains effects of translationese (often defined as awkward, unnatural, or overly literal translations) (Volansky et al., 2015).

**Translation Workflow.** Script, specification, spelling and translation approaches were first established and aligned on from Flores-200. Translators referenced these linguistic alignments while working on Seed Data Translations. The datasets were translated directly from English for 39 languages while two Arabic script languages (Acehnese and Banjar) and Tamasheq in Tifinagh script were transliterated from their respective Latin script datasets that were first translated from English.[11] Following the translation or transliteration phase was a linguistic quality assessment phase in which the completed datasets were checked against the linguistic alignments from FLORES along with automatic quality control checks. The datasets were then finalized and completed.

We note that NLLB-Seed has a key distinction compared to evaluation benchmarks such as Flores-200. Critically, NLLB-Seed is meant to be used for *training* rather than *model evaluation*. Due to this difference, NLLB-Seed does not go through the human quality assurance process present in Flores-200.

### 4.3 NLLB Multi-Domain Dataset

Avoiding overfitting and achieving strong out-of-domain performance remains a major challenge in neural machine translation (Koehn and Knowles, 2017). While both Flores-200 and NLLB-Seed cover a large number of topics, we want to ensure that models perform well on text coming from different domains. Additionally, since potential users might be interested in tuning general translation models for specific applications, we want to investigate how effectively our system can be fine-tuned on a dataset covering a new domain. More specifically, we want to answer the following two questions: **(1)** How well do models generalize to non-Wikimedia domains? **(2)** Does fine-tuning on high quality in-domain parallel text lead to good performance? In order to investigate these questions, we create the NLLB-MD parallel dataset, covering six directions and made up of 3,000 professionally-translated sentences in each of four different domains.

**Language Selection.** NLLB-MD covers the following six languages: Central Aymara (`ayr_Latn`), Bhojpuri (`bho_Deva`), Dyula (`dyu_Latn`), Friulian (`fur_Latn`), Russian (`rus_Cyrl`) and Wolof (`wol_Latn`). Along with five low-resource languages, we also chose to include one high-resource language to enable comparisons with other models and datasets. We chose low-resource languages related to other high-resource ones (e.g., `fur_Latn` is related to `ita_Latn`), so as to enable future studies investigating language transfer.

10. Note: There is no overlap between the sentences in Flores-200 and NLLB-Seed
11. We had a specific process for Ligurian: half the data for Ligurian were first translated from English to Italian, then translated from Italian to Ligurian while the other half was translated directly from English. As we are lucky to have Ligurian native speaker, we developed this process to improve quality.

**Domain Selection.** We collected 3,000 English sentences in each of four different domains, and sent them to professional translators to be translated into each of NLLB-MD's six target languages. The translation workflow used is analogous to the one followed for NLLB-Seed. The domains included are:

- **News**: We translate the English side of the WMT21 English-German development set, containing a sample of newspapers from 2020 (Akhbardeh et al., 2021).
- **Scripted formal speech**: We translate text extracted from a series of scripted English-language talks covering a variety of topics.
- **Unscripted informal speech**: We extract 3,000 utterances from the multi-session chat dataset of Xu et al. (2022), which contains on average 23 words per turn.
- **Health**: We translated one World Health Organisation report (Donaldson and Rutter, 2017) and combined it with sentences translated from the English portion of the TAUS Corona Crisis Report.[12]

### 4.4 Conclusion

To summarize, Flores-200, which enables reliable evaluation of over 200 languages, is critical for ensuring the quality of the results our systems generate. NLLB-Seed plays an important role for training both sentence encoders (see Section 5) and translation models (see Section 6.5). Finally, we utilize NLLB-MD to measure the generalizability of our translation models across multiple domains (see Section 8.3). Now that we have described the creation of three human-translated datasets and their uses, we visit how we acquired training data for our effort in the subsequent section.

## 5. Automatically Creating Translation Training Data for Hundreds of Languages

The current techniques used for training translation models are difficult to extend to low-resource settings — that is, when data for a language is limited in both aligned textual data (*bitext*, or pairs of translated sentences) and single language data (*monolingual*, or data in one language only). In fact, many low-resource languages are supported only through small targeted bitext datasets such as the Christian Bible (McCarthy et al., 2020), which are extremely limited in domain diversity. In this section, we detail how we built a large scale dataset that covers hundreds of languages and discuss the challenges we faced with noisy data at web-scale.

For context, publicly available bitext data is often scarce (Gowda et al., 2021). Our approach centers around extending existing datasets by collecting non-aligned monolingual data and using large-scale data mining (Schwenk et al., 2021b) to identify sentences that have a high probability of being translations of each other in different languages. To enable this for hundreds of languages, we first develop language identification systems (LID, Section 5.1) that label which language a given piece of text is written in. Subsequently, we curate available monolingual data, apply sentence splitting and LID along with various filtering

12. `https://md.taus.net/corona`

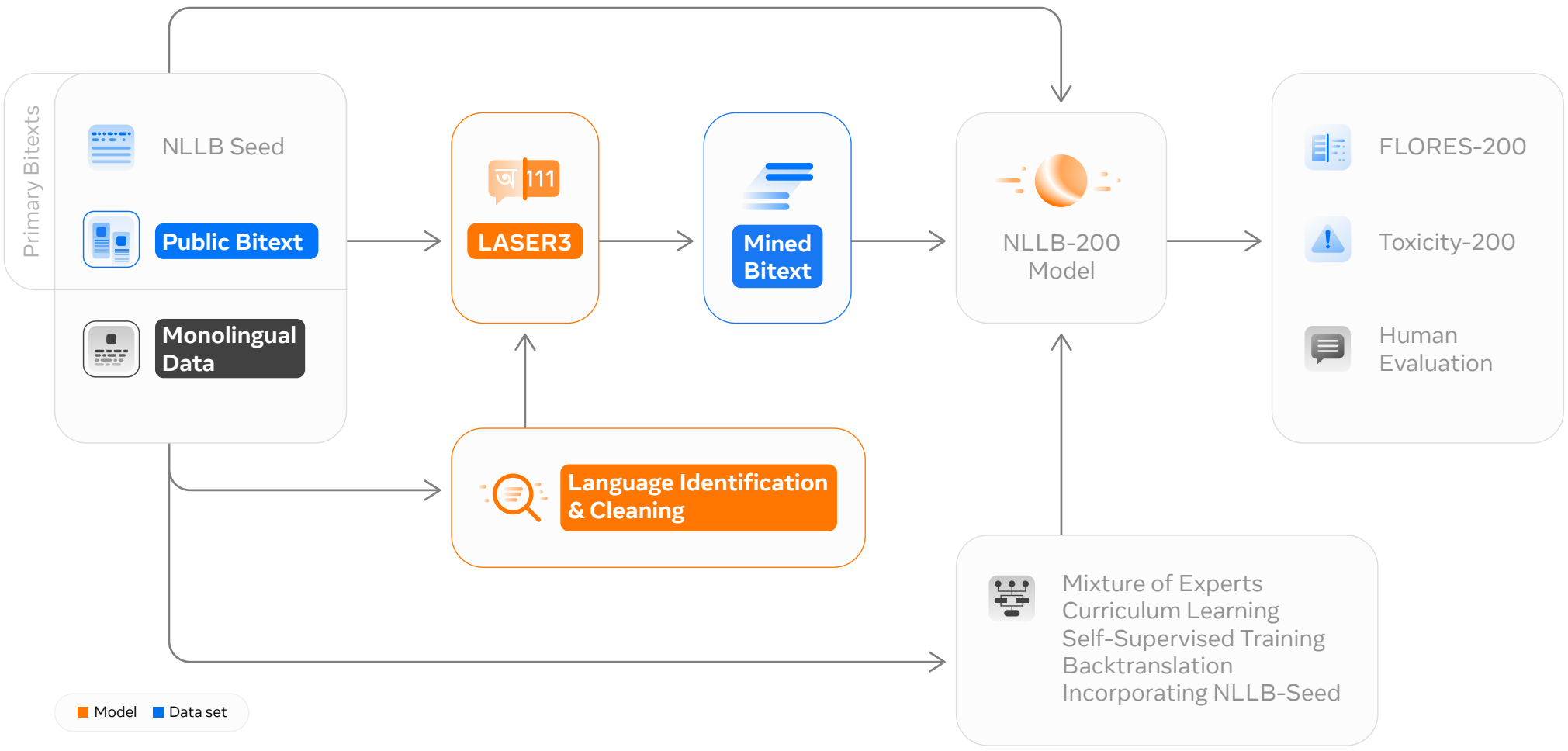

Figure 6: **Automatic Dataset Creation Contributions of No Language Left Behind**: As highlighted, we create language identification and a monolingual data cleaning process, then describe the training of LASER3 to produce large-scale mined bitext for hundreds of languages.

mechanisms (Section 5.2), and then move ahead with mining aligned pairs (Section 5.3). An overview of this process is presented in Figure 7.

## 5.1 Language Identification

Language identification (LID) is the task of predicting the primary language for a span of texts. It is widely used in commercial applications (such as the *detect language* feature embedded in some web browsers) and is of particular importance in natural language processing research. The rise of large-scale pretraining, particularly the increasing focus on multilingual models, is strongly dependent on the existence and identification of monolingual data at scale. Advances in cross-lingual representation learning (Conneau and Lample, 2019; Wang et al., 2020b) such as large-scale bitext mining (Bañón et al., 2020; Ramesh et al., 2022; Schwenk et al., 2021b), unsupervised machine translation (Conneau et al., 2020; Ren et al., 2019; Yang et al., 2018) and back-translation at scale (Edunov et al., 2018) require large quantities of clean monolingual data. These disparate approaches, including our focus on large-scale data mining of aligned sentences, involve taking large quantities of input text often drawn from web corpora such as CommonCrawl[13] and labeling them with corresponding languages.

There are a few well-known challenges associated with large-scale and accurate language identification using web data (Caswell et al., 2020): **(1)** Domain mismatch could occur due to the scarcity of text reliably labeled by language. For example, the Christian Bible has

13. `https://commoncrawl.org/`

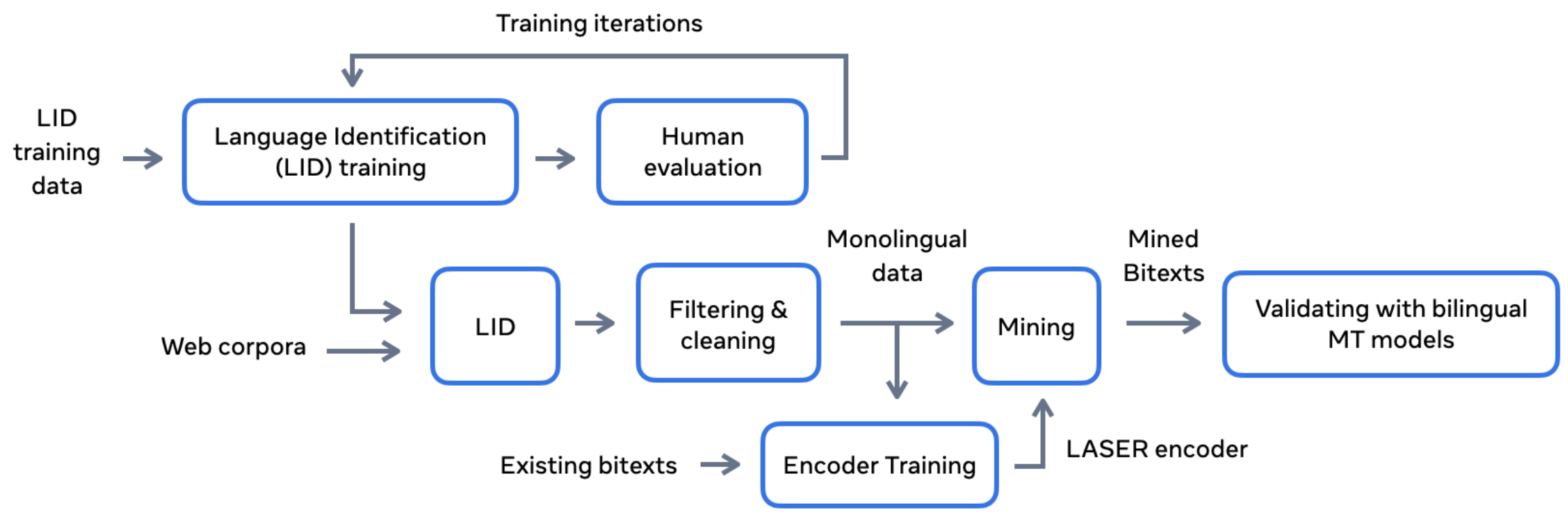

Figure 7: **Overview of our Bitext Mining Pipeline.** Language identification is applied on web corpora to extract monolingual sentences. Aligned pairs are later identified with LASER3.

been translated into a wide array of languages. However, an LID system trained on this corpus would not reliably classify sentences from non-Biblical domains. Properly extending training data is not trivial: while the web contains data in thousands of languages (Prasad et al., 2018; Scannell, 2007), most of it is unlabeled. Filling in this gap is Wikipedia, which is frequently used for training language identification (Thoma, 2018) on a broader scale beyond the Christian Bible (although such relatively clean formal text is not representative of the web at large); **(2)** Severe class imbalance could exist because many of the low-resource languages of interest to us have low presence on the web. For classifiers to work, they must have an extremely low false positive rate. Otherwise, low-resource languages are prone to misidentification; **(3)** Efficiency to run over large web collections remains low. Even though classification is massively parallelizable, running it on all texts makes speed critical. In this section, we describe our approach to language identification and how we strike a necessary balance between predictive performance and scalability.

#### 5.1.1 Related Work

There is extensive literature dedicated to the development of LID systems. Jauhiainen et al. (2019) give a recent and comprehensive overview of the features and algorithms used in the literature. While LID could be seen as a solved problem in some domains (McNamee, 2005), it remains an open challenge for web data (Abadji et al., 2022; Caswell et al., 2020; Zampieri et al., 2015b). Specifically, issues coalesce around **(1)** scaling successful approaches to more languages (Jauhiainen et al., 2017); **(2)** incidents where there is significant domain mismatch (Widdows and Brew, 2021) in the cases of short tweets or multiple languages (Duvenhage, 2019); and **(3)** distinguishing similar languages (Goutte et al., 2016).

**Scaling LID to More Languages.** Devoted attention to advance LID techniques have led to a noticeable increase in both language coverage and accuracy over time. CLD3[14] and `fasttext` (Grave et al., 2018) are two readily available models offering high detection performance for 107 and 187 languages respectively. By using numerous public datasets, Dunn

14. `https://github.com/google/cld3`

(2020) and Brown (2014) report even higher coverage, supporting 464 and 1366 languages respectively. That said, developments around low-resource languages face slow advancement due to the emphasis on religious texts and constraints brought about by software localization. Caswell et al. (2020) scale up to 1,629 languages using wordlists and self-supervision to bootstrap training data found on the web. These approaches using found data suffer from domain imbalance: because the available text domains vary by language, the classifier conflates domain with language. In contrast, we curate Flores-200 to use as development set, so that our LID system performance is tuned over a uniform domain mix. One could of course use the Christian Bible as a uniform domain. However, we believe Flores-200 is closer to web content.

**Domain Mismatch.** Because the web covers a very broad set of domains and reliably labeled text is scarce, there is almost always a domain mismatch between training data and the web text being classified. Widdows and Brew (2021) describe a new feature based on the rank of words within frequency tables that enhances robustness of LID systems to domain mismatches. They train their classifier on Wikipedia and report results on a Twitter test set, unfortunately covering only 22 languages. Short text is tackled in Duvenhage (2019) for South African Languages with a stacked classifier. Neural network-based strategies are also derived in Ansari et al. (2021); Shekhar et al. (2020) to handle text written in a mix of English and Indian languages (code mixing). Caswell et al. (2020) thoroughly analyze and classify failure modes of language identification on web corpora. They suggest using a series of filters along with a new unsupervised learning approach to drastically improve precision at limited cost on recall. These filters are costly to devise and tune for all languages however. Some of them were successfully put into practice in Abadji et al. (2022) to release a cleaner version of the OSCAR dataset. Our approach combines a data-driven `fasttext` (Grave et al., 2018) model trained on Flores-200 with a small set of handwritten rules to address human feedback on classification errors.

**Handling Similar Languages.** Distinguishing between similar languages has been an active research topic, for instance, with the shared task on Discriminating between Similar Languages within the VarDial workshop (Goutte et al., 2016). Several common machine learning algorithms along with standard neural networks are compared in Haas and Derczynski (2021) for Nordic languages. Duvenhage (2019); Goutte et al. (2014); Zampieri et al. (2015a) explore various hierarchical approaches that first predict the language group of input text, then apply a more specialized classifier to distinguish between languages within that group. In this work, we collaborate in close partnership with linguists to understand which languages can be easily confused and analyze the model performance while employing a flat classification strategy.

#### 5.1.2 Models

We utilize `fasttext` to train language identification models (Bojanowski et al., 2017; Joulin et al., 2017). `fasttext` is widely used for text classification tasks due to its simplicity and speed, while achieving good quality. We embed character-level n-grams from the input text, then leverage a multi-class linear classifier on top. The lightweight nature of `fasttext` enables our LID models to handle web-scale data. Additionally, a linear model has the benefit of being easily explainable, allowing us to trace any classification error back to its root cause. This is instrumental in addressing common pitfalls that arise when detecting language on web corpora (Caswell et al., 2020).

**Classifier Design.** We experimented with two different designs. **(1)** A combination of multiple binary classifiers where the final decision is obtained by selecting the language having the highest score after a threshold is applied. We apply threshold optimization so that when the confidence of a classifier is low, the corresponding language is not considered for the final decision. If none of the classifiers surpass its threshold, the sentence is filtered out. **(2)** A multiclass classifier using softmax over all possible languages. In this case, the threshold optimization is done after the softmax.

Our experiments motivated us to focus on the second approach, which offers several advantages. First, changing the threshold for one language does not impact the performance of the other , while this is not true in the first setting. Second, we found that this approach generalizes better to out of domain data which is our primary use case (Wikipedia → Web data). Finally, a single classifier has the added benefit of being computationally simpler, thus streamlining the language identification process.

**Training Data and Handling Massive Class Imbalance.** We use publicly available datasets to train our LID system, partially covering our of interest. We supplement these with NLLB-Seed (see Section 4.2) for any missing language. However, the amount of data available for each language is far from uniform, and massive class imbalance in the raw training data exists (Caswell et al., 2020; Dunn, 2020). For example, English alone represents 10.1% of our training data, while Minangkabau (Latin script) represents only 0.06%. Following Arivazhagan et al. (2019), we experimented with multiple settings of temperature upsampling for under represented , where sentences from a language $l$ representing $p_l$ percent of the dataset are sampled proportionally to $p_l^{\frac{1}{T}}$. Optimal performance was obtained at $\frac{1}{T} = 0.3$.

**Training Parameters.** Our best model was trained with softmax loss over two epochs with a learning rate of 0.8 and embeddings with 256 dimensions. We discarded words with less than a thousand occurrences after upsampling and picked a minimum and maximum character n-gram length of two and five respectively, which were assigned a slot in buckets of size 1,000,000. All hyperparameters were tuned on Flores-200 `dev`.

#### 5.1.3 Improving LID with Linguistic Analysis

Language identification is a challenging task where numerous failure modes exist, often exacerbated by the gap between the clean data that LID models are trained on and the noisy data that LID models are applied to. LID models that are trained in a supervised manner on fluently written sentences may have difficulty identifying grammatically incorrect and incomplete strings extracted from the web. Furthermore, models can easily learn spurious correlations that are not meaningful for the task itself. In light of these challenges, we collaborated closely with a team of linguists throughout different stages of LID development to identify proper areas of focus, mitigate issues, and explore solutions.

**LID Inspection Interface.** We leveraged the linearity of `fasttext` to build an easy-to-use interface for linguists to peek into its inner workings. The tool enabled linguists to analyze model errors and discern model patterns. As illustrated in Figure 8, we visualize how much each n-gram contributed to the final prediction. In one of the applications, the tool led linguists to notice the similarity in phonotactics between Standard Malay and Indonesian, which are one of the most frequently confused language pairs, and to find out through linguistic research that in spite of obvious differences, a certain degree of mutual intelligibility exists between the two.

| Label | Label Score | Text |
|---|---|---|
| eng | 0.681 | (French: Déclaration des droits de l'homme et du citoyen de 1789), set by France's National Constituent Assembly in 1789, is a human civil rights document |
| fra | 0.16 | (French: Déclaration des droits de l'homme et du citoyen de 1789), set by France's National Constituent Assembly in 1789, is a human civil rights document |

Figure 8: **LID Inspection Interface,** used on an example sentence from the English Wikipedia containing a short passage in French. The top 2 labels with highest probability are displayed, along with their score. N-grams that contributed the most (either positively or negatively) to the predictions are highlighted (in green and red respectively).

**Filtering Training Data.** To mitigate the learning of spurious correlations due to noisy training samples while modeling hundreds of languages, we worked in collaboration with linguists to develop several filters, illustrated in Table 3 and described below. All are subsequently applied on our raw training dataset.

- **Character Distribution Filtering**: The public datasets we used for training were mostly built from webpages. Through investigation by linguists, numerous occurrences of mislabeled sentences were found, likely caused by short passages in a different language within a page, such as Indonesian sites that display a collection of Javanese poems. We also noticed random creative use of unexpected scripts, typically used for decoration or emphasis as pointed out in Caswell et al. (2020). Table 3 gives a few examples. To address this problem, we searched for distribution shifts in characters, either by computing character histograms or by looking at the language's expected script unicode range.

  **Character Histograms**: We computed the character distributions of each language on our development set and defined an arbitrary *accepted character set* for each of them by considering all characters falling within the first $95^{\text{th}}$ percentile. We consequently filtered out any sentence from our training set that was composed of less than 80% of such accepted characters.

  **Script Detection**: For languages whose script spans thousands of characters, the character histogram method mentioned above was not as effective since the character distribution trends were less prominent. As an alternative, linguists provided Unicode ranges to define accepted character sets. Any sentence containing less than 50% of characters from that set was eventually discarded. For example, the sentences shown in Table 3 for Japanese and Chinese do not contain the right `Jpan` and `Hans` scripts.

| Filter | Label | Filtered Sentence |
|---|---|---|
| **Histogram** | urd_Arab | ┣████══─𝓜☺️♓𝓜 ɑ Đ ɑ█ ےیاشاپےضترم : |
| | dan_Latn | అనంతపూర్ డిస్ట్రిక్ , urdu: عاضد دابآ لدآ ) er et distrikt i den |
| **Script** | jpn_Jpan | 4.0, CUDA 対応。消費電力は 40W。Quadro FX 380 コア 450MHz |
| | zho_Hant | 容存档于 2009 年 2 月 10 日）． Satellite map 維基衛星 |
| **English** | tur_Latn | A module is said to be semisimple if it is the sum of simple submodules. |
| | nld_Latn | Line drawing and design: From the book Brazil and the Brazilians, 1857 |

Table 3: **Examples of Sentences Filtered** from our LID training dataset.

- **English-specific Filtering**: Linguists also pointed out that many mislabeled training samples were actually plain English sentences. This can be explained by the massive prevalence of English on the web, even on pages primarily written in other languages. We built a simple, dedicated binary `fasttext` classifier to filter these samples out of our training dataset.

#### 5.1.4 Results

This section presents a comparison of our approach to existing publicly available models on both Flores-200 and annotated noisy web data, followed by an error analysis.

**Evaluation on the Flores-200 Benchmark.** We analyze the performance of our LID models on the Flores-200 dataset from Section 4 and compare to other open-source models. We utilize Flores-200 for evaluation as the target-side text is human-verified as being in the right language. Utilizing standard public datasets for evaluation is less reliable given they often contain untrustworthy language labels and are quite noisy (Kreutzer et al., 2022).

We compare our LID model with three publicly available models: `CLD3`, [15] `LangId`[16] and `LangDetect`.[17] Table 4 reports performance of our final LID model on the set of various language intersections covered by all four models. Micro F1 scores and False Positive Rates across all languages found in Flores-200 are displayed in Table 5. Given the different scopes of languages supported, we report on 3 cascading intersections with Flores-200: **(1)** the 51 languages also supported by `LangId`, `LangDetect` and `CLD3`, **(2)** the 78 languages also supported by `LangId` and `CLD3` and **(3)** the 95 languages also supported by `CLD3`. We report metrics of all models across all intersections to reflect the impact of false positives on unseen languages.

Our model is capable of handling the 200 languages of Flores-200 (compared to the 107 languages supported by `CLD3`) while achieving significantly higher performance than all three of `LangId`, `LangDetect` and `CLD3`. Furthermore, the gain in F1 score is accompanied by a noticeable improvement in False Positive Rate, suggesting a much stronger fit for extracting low-resource languages from web corpora (Caswell et al., 2020).

**Human Evaluation on Noisy Web Data.** Despite strong results on Flores-200, we expect a sizable gap in performance when applying our LID model to our target web data.

15. `https://github.com/google/cld3`
16. `https://github.com/saffsd/langid.py`
17. `https://pypi.org/project/langdetect/`

| | # Supported Languages | Flores-200 ∩ CLD3 ∩ LangId ∩ LangDetect 51 Labels | | Flores-200 ∩ CLD3 ∩ LangId 78 Labels | | Flores-200 ∩ CLD3 95 Labels | |
|---|---|---|---|---|---|---|---|
| | | **F1** | **FPR** | **F1** | **FPR** | **F1** | **FPR** |
| `LangDetect` | 55 | 97.3 | 0.0526 | 64.4 | 0.4503 | 53.1 | 0.4881 |
| `LangId` | 97 | 98.6 | 0.0200 | 92.0 | 0.0874 | 75.8 | 0.2196 |
| `CLD3` | 107 | 98.2 | 0.0225 | 97.7 | 0.0238 | 97.0 | 0.0283 |
| Ours | **218** | **99.4** | **0.0084** | **98.8** | **0.0133** | **98.5** | **0.0134** |

Table 4: **Comparison of Open-Source Language Identification Models with various intersections of labels.** F1 is the micro F1 score and FPR is the micro False Positive Rate.

| | **Micro F1** | **Macro F1** | **Macro Precision** | **Macro Recall** | **Macro FPR** | **Micro FPR** |
|---|---|---|---|---|---|---|
| Low-Resource | 95.63 | 95.9 | 97.6 | 95.4 | 0.01213 | 0.0235 |
| All Flores-200 | 95.85 | 95.5 | 94.0 | 95.7 | 0.02110 | 0.0210 |

Table 5: **Performance of our LID system on FLORES-200**. Arabic languoids and Akan/Twi have been merged after linguistic analysis.

Indeed, various sources of noise such as language mixing, creative use of various scripts, and leetspeak are widespread online. Extracting sentences from internet pages is also prone to unexpected artifacts introduced after parsing. There is no readily available evaluation set from the web domain on which to properly assess and tune performance, let alone iterate on design choices when modeling. This motivated us to audit the performance of our system with human annotators.

To this end, we select 74 low-resource languages on which our preliminary LID model yield low F1 scores. After a first run of language identification on web data, we randomly selected several thousand sentences across various languages for which prediction scores fell between 50% and 90%. That hard threshold was chosen upon manual inspection, noticing that many classification errors were found within that range. Human annotators were were tasked with inspecting our random sentences and assessing whether each was indeed in the predicted language.

Based on these annotations, we built a challenge set for language identification to benchmark our final LID model. As shown on Table 6, we achieve lower performance than on the Flores-200 dataset, hinting at a non-negligible domain mismatch. We also compare performance against `CLD3`. As suggested in Caswell et al. (2020), we report False Positive Rates (FPR) on top of F1 scores, to get a better picture of how well our model would fare on web corpora with extreme class imbalance. Despite the extended language coverage, our approach yields both a higher F1 score at a lower FPR, suggesting a good fit for the downstream pipeline of Section 5.2. We share performance of our model and `CLD3` on all the selected languages in Table 50 in the appendix. It should be noted that since the sentences to annotate were chosen based on a previous model, this challenge set is biased by that underlying intermediate model.

| Language | **Ours F1** | **CLD3 F1** | **Ours FPR** | **CLD3 FPR** |
|---|---|---|---|---|
| Micro | **79.14** | 64.41 | **0.79** | 1.12 |
| Macro | **74.16** | 60.13 | **0.77** | 1.12 |

Table 6: **Comparison of `CLD3` and Our Model on a Challenge Set built from Human Annotations.** Only the average performance of languages supported by both `CLD3` and our model is shown. Full table in Appendix Table 50.

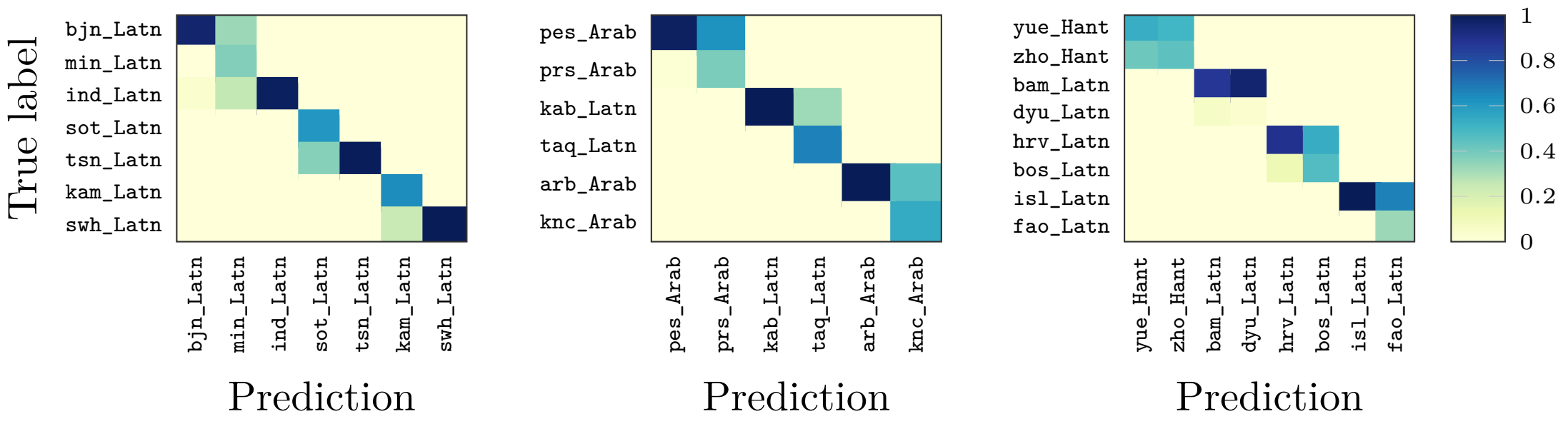

Figure 9: **Confusion Matrix on Flores-200** on the 21 languages with lowest accuracy

| Reference (English) | Asante Twi (aka_Latn) and Akuapem Twi (twi_Latn) |
|---|---|
| No one was inside the apartment | Na obiara nni odan no mu |
| There's a tiny cave near the top that must be passed through | ɔbodan ketewa bi bɛn soro ho a ɛsɛ sɛ wofa mu |
| For a few pennies some children will tell you the story | Sɛ woma mmofra bi sika ketewa bi a, wobɛka anansesɛm yi akyerɛ wo |

Table 7: **Examples of Identical Sentences** in Akan and Twi, two similar languages.

**Analysis on Challenging Language Pairs.** Figure 9 brings to light a small group of confusable language pairs found to be the most difficult for our LID system: Akan/Twi, Dyula/Bambara, Faroese/Icelandic, Western Persian/Dari, and Bosnian/Croatian. We worked closely with linguists to analyze them. Upon inspection, we found that these language pairs correspond to highly similar languages, displaying major vocabulary and grammar overlap. For example, Asante Twi (`aka_Latn`) and Akuapem Twi (`twi_Latn`) are two mutually intelligible languoids of the Akan language continuum, which share common words, phrases, or even identical sentence translations. Examples can be found in Table 7. This suggests that from a linguistic point of view, the LID confusion found in these similar languages is to be expected and is not a symptom of a deeper modeling issue. In practice, this means prediction performance might be underestimated for some languages and calls for collecting and accepting multiple language labels in future work.

**Impact of Sentence Length.** We noticed that predictions tend to be more robust for long sentences. Figure 10 gives an overview of the difference in performance as a function of input length. This is consistent with an observation by Jauhiainen et al. (2017) and could be further investigated. A potential mitigation strategy would be to tune our models on a more balanced development set with respect to length. Shorter test sentences could be synthetically created from our current development samples. In our current approach, we mitigate this issue by applying length filters in the downstream monolingual pipeline described in Section 5.2.

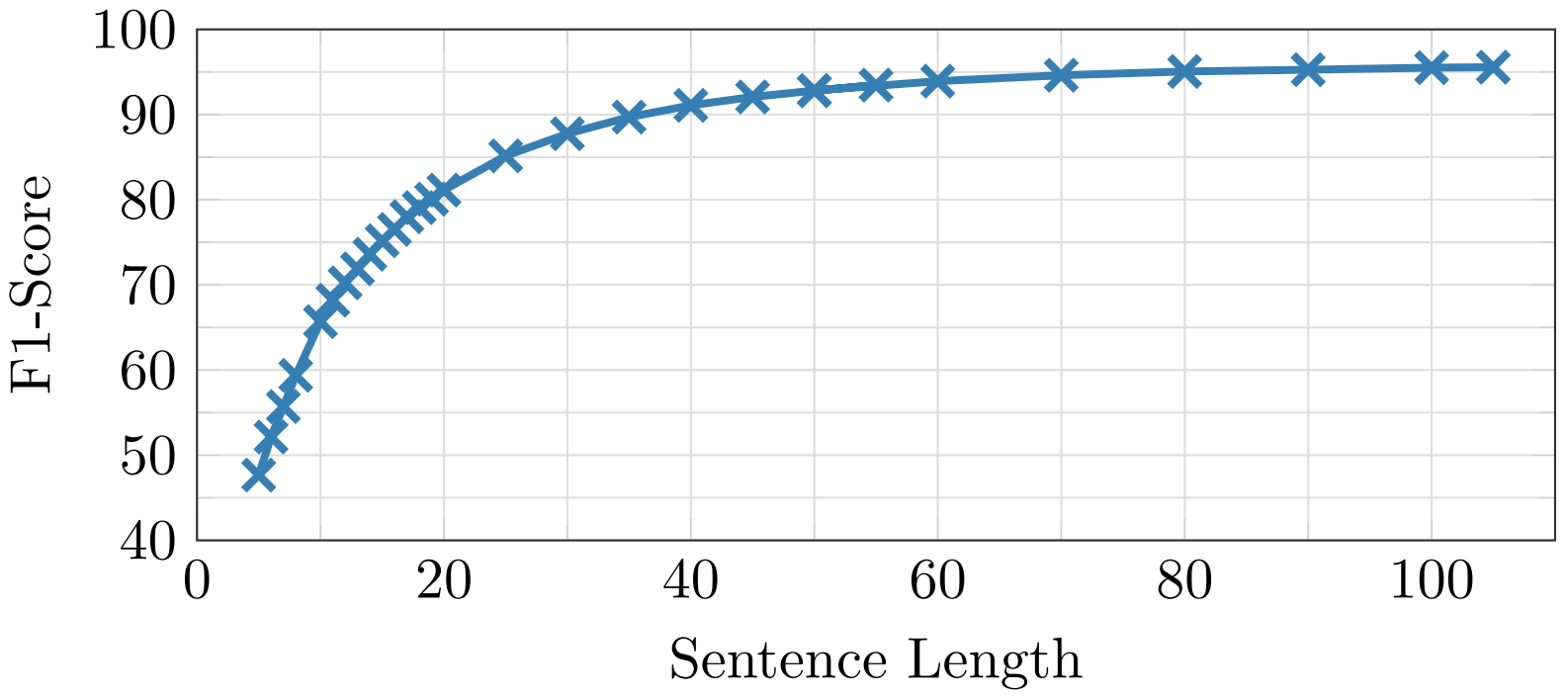

Figure 10: **Effect of Sentence Length on Performance.** We synthetically create test samples of a specific length without cutting words, except for languages with continuous scripts.

## 5.2 Gathering and Cleaning Monolingual Data at Scale

Monolingual data is a valuable resource which can be used for a variety of downstream tasks such as bitext mining, backtranslation, and language model training. Additionally, data quality can have a large impact on the performance of such tasks. In order to maximize the potential benefits of leveraging these data sources, we aim to produce high quality and clean monolingual data. As discussed earlier in this section, such data can be scarce, particularly in the low-resource setting. We therefore decided to extend the work done in CCNet, CCMatrix (Schwenk et al., 2021b; Wenzek et al., 2020), and others like OSCAR (Ortiz Suárez et al., 2019). In this section, we describe our end-to-end process for both curating and cleaning monolingual data.

### 5.2.1 Description of our Monolingual Pipeline

**Data Sources.** We begin with web data as our starting point, provided by CommonCrawl (CC)[18] and ParaCrawl (Bañón et al., 2020). This data has been preprocessed to remove all markup and (approximately) normalize encoding to UTF-8. HTML stripping converts block tags to newlines while inline tags are removed. The resulting lines can contain many sentences or simply a short snippet of text; we refer to them as "paragraphs".

**Applying Language Identification.** To convert the raw web text in paragraph form to sentences, we apply language identification in a hierarchical fashion. First, we apply LID to each web paragraph. Subsequently, we use the predicted language to choose a sentence splitter for the language.[19]

The raw paragraphs sometimes contain a mix of different languages or might include code switching. To avoid having a mix of languages, once we have split the documents in sentences, we re-run LID to identify the language of each sentence. If the sentence-level LID does not match the paragraph-level LID, we discard the sentence to be sure we keep

18. In `wet` format, `https://commoncrawl.org/`
19. We use a mix of custom splitting rules, `indicnlp` (Kunchukuttan, 2020), `https://github.com/mediacloud/sentence-splitter`, `https://github.com/Esukhia/botok`, `khmer-nltk` (Hoang, 2020) and LaoNLP `https://github.com/wannaphong/laonlp`

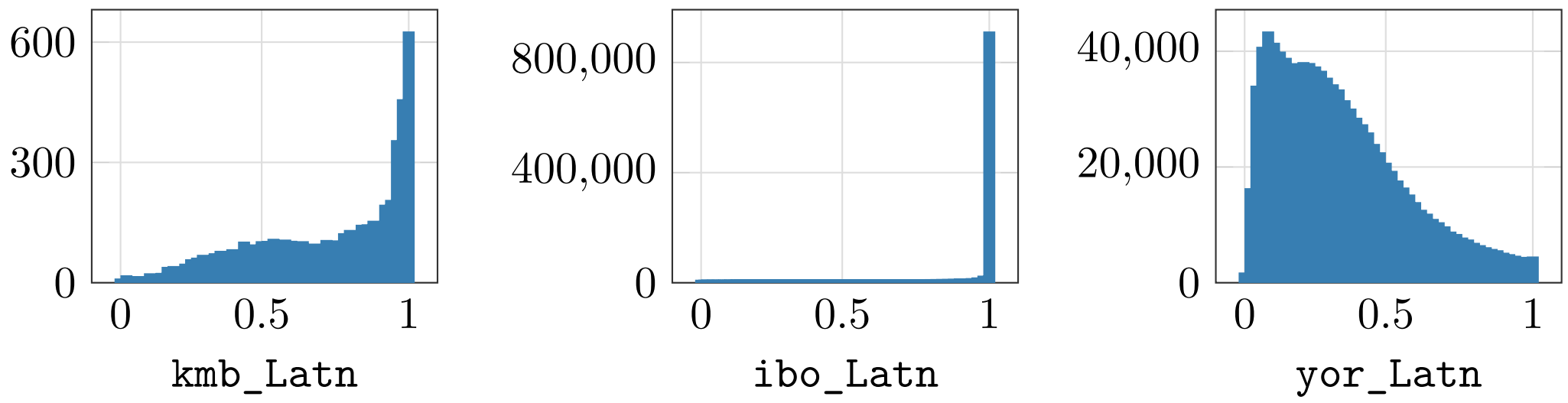

Figure 11: **LID Score Distribution Patterns on ParaCrawl**, illustrated with Kimbundu, Igbo and Yoruba.

high-confidence sentences in the targeted language. We also discard sentences if they do not use the expected script for the target language.

Note that many sentences are extremely noisy. In particular, they often contain long URLs or lists of hashtags. These confuse the LID and script identification process as they are in Latin script and not always in the same language as the original sentence. To identify the actual language of the sentence, we truncate the URLs and hashtags before running the language identification.

Given the domain mismatch (see Section 5.1), our development set could not be utilized to tune the detection thresholds of our `fasttext` classifier to a desired performance level. Instead, we relied on the distribution of model scores on the monolingual data of the ParaCrawl dataset (Bañón et al., 2020) across all predicted languages. We chose that dataset based on the assumption that its language distribution would realistically match that of CommonCrawl, despite the inevitable bias induced by the LID model used in the creation of ParaCrawl itself. The vast majority of languages fell into one of three score distribution patterns, as illustrated in Figure 11 with Kimbundu, Igbo and Yoruba.

1. Left-skewed distribution, where scores rarely go below the 0.5 mark

2. Extremely left-skewed, where almost all scores landed above 0.9

3. Right-skewed, where scores very rarely go beyond the 0.5 mark

This analysis motivated our choice of 0.5 as a default threshold in the first two cases, except for high-resource languages, where we could afford a more stringent value of 0.9 and still collect enough monolingual data in our downstream pipeline depicted in Section 5.2. In the last case, we picked values corresponding to the peak of each distribution (ranging from 0.2 to 0.4), in the hope to collect a sizable amount of data from our pipeline.

**Heuristics for Data Cleaning.** We subsequently apply a few heuristics to remove sentences that do not match reasonable quality criteria: minimum and maximum length, space/punctuation/number/emoji ratios, and maximum number of repeated characters. For example, if a sentence contains over 20% punctuation, it likely is not a well-formed sentence. As our model encoders (see Section 5.3) were not trained on substantial content with emojis, we prefer to strip emojis from all text to avoid losing sentences that could match if they did not have the special characters. Some sentences were dominated by hashtags and URLs (e.g.

| Filter | Example | Reason |
|---|---|---|
| Low LID threshold | Internet Plus € 58,50 | `eng_Latn` at 0.19 LID score |
| LID mismatch | Best véto ever! | doc. LID French, sent. LID Czech |
| Numbers | Vol.180 Sep. (2011) | exceeded numbers ratio |
| Punctuation | . * sApEvAte cHe... » (Previous page) | exceeded punctuation ratio |
| Emoji | 💪💪💪 #gymgirl | exceeded emoji ratio |

Table 8: **Examples of Filtered Sentences** in our monolingual pipeline.

tweet-like sentences). In these cases, we remove these parts of the sentences and apply the heuristics on the truncated sentence instead.

We note that these ratios should differ between languages, and are not universally applicable. In certain languages, concepts may take many more words to convey, meaning that setting length-related thresholds is problematic. Similarly, other languages may utilize more punctuation or have shorter words. Thus, we do not set extremely stringent filters, and we examine the amount of text filtered across all 200 languages for each of the filters. Table 8 provides some typical examples of filtered sentences.

**Deduplication.** The margin-based criterion of our mining approach requires unique sentences (see Section 5.3), but the sentence splitting and cleaning process might generate a lot of duplicate content, so we run a global deduplication process over all sentences of the same language.

**Language Model Filtering.** As we are interested in keeping high quality sentences in our datasets to later train our final multilingual translation models, when possible, we also run a Language Model (LM) filtering. In practice, it is difficult to train high-quality language models for low-resource languages, so we focus on applying language model filters on a few high-resource languages only. Because we do most of the mining where one side of the pair is English, we believe that if we have high-quality content in the English corpus, the mining alignment process will also output high-quality content on the other side of the pair. For English, we use the `KenLM` (Heafield, 2011) model from CCNet (Wenzek et al., 2020).

**Computational Challenges.** We processed around 37.7 Petabytes of data through the whole pipeline. This was a challenge for data management and disk usage. In particular, we had to make hard decisions when filtering high-resource languages to artificially keep only around 30% of data from the most voluminous languages. This threshold was identified as the limit under which the LM score would identify sentences of low quality. Processing was distributed over many machines and several months to be able to get to the final monolingual dataset. We realize that processing such volume of data is not always possible and are open-sourcing much of our results and code to make it easier for everyone to benefit from our effort.

#### 5.2.2 Monolingual Data at a Glance

We started with 107.9 billion paragraphs from the web (97.4% high-resource, 2.6% low-resource) and discarded 2.8 billion sentences (85.3% high, 14.7% low) because of LID/Script mismatch or low LID score (see Section 5.1) to produce 43.7 billion monolingual sentences (90.8% high, 9.2% low) to feed to the mining. 21.5 billion sentences are in English. See Table 9 for details on data volume. Note that we have such a big drop in number of sentences because we drastically filtered high-resource languages to keep the top 30% of sentences based on LM score.

| | Total | Min | | Max | | Median | | Mean |
|---|---|---|---|---|---|---|---|---|
| **Low-Resource** | | | | | | | | |
| Raw Data (Para.) | 2.4B | 27.1K | tzm_Tfng | 465.8M | nob_Latn | 3.3M | mai_Deva | 17.9M |
| LID/Script Mismatch (Sent.) | 0.3B | 0.2K | tzm_Tfng | 47.8M | nob_Latn | 0.3M | fao_Latn | 2.2M |
| Clean Sentences | 3.6B | 1.3K | tuk_Latn | 330.3M | glg_Latn | 4.4M | tso_Latn | 26.8M |
| **High-Resource** | | | | | | | | |
| Raw Data (Para.) | 105.4B | 4247.8K | tsn_Latn | 73.2B | eng_Latn | 83.1M | eus_Latn | 3401.2M |
| LID/Script Mismatch (Sent.) | 2.5B | 124.4K | ben_Beng | 1.6B | eng_Latn | 5.0M | als_Latn | 81.2M |
| Clean Sentences | 40.1B | 5153.8K | tsn_Latn | 21.5B | eng_Latn | 234.3M | als_Latn | 1294.5M |

Table 9: **Key Statistics** for the processing of low- and high-resource monolingual data.

### 5.3 Mining Bitexts for Low-Resource Languages

Machine translation, like many machine learning applications, is heavily data-driven. Previous works have clearly established that translation quality generally increases with the amount of available high-quality training data (Koehn and Knowles, 2017). Existing parallel corpora for low-resource languages are often opportunistically drawn from known collections of multilingual content, such as the Christian Bible or publications of multinational organizations. These are often limited in quantity and domain. In this section, we describe how we automatically create translation training datasets for low-resource languages through *bitext mining*. We mainly focus on bitexts paired with English, but we are also interested in mining through other language pairs as it was shown that these can improve the overall performance of a multilingual translation system (Fan et al., 2020).

**What is Bitext Mining?** The underlying idea of our bitext mining approach is to first learn a multilingual sentence embedding space and to use a similarity measure in that space to decide whether two sentences are parallel or not. This comparison can be done for all possible pairs in two collections of monolingual texts, termed *global mining* in Schwenk et al. (2021a). Another common alternative approach is referred to as *hierarchical or local mining* and comprises of first performing a selection of potential document pairs, and then limiting the mining to sentences within each document pair. The European ParaCrawl project is a typical example for this approach (Bañón et al., 2020). Earlier work along these lines handled both web sites with parallel text (Resnik and Smith, 2003) and comparable data (Fung and Cheung, 2004). In this work we follow the approach from WikiMatrix (Schwenk et al., 2021a) and CCMatrix (Schwenk et al., 2021b), which both used global mining.

As our mining approach requires a multilingual embedding space, we faced several challenges when scaling this representation to the 200 languages of the No Language Left Behind effort. For example, how do we make sure that all languages are well-learned? And how should we account for large imbalances of available training data? Training a massively multilingual sentence encoder from scratch each time a new set of languages is added would be computationally very expensive. Furthermore, such an approach has the drawback that the learned embedding spaces from each new model are not (naturally) mutually compatible. This can make mining intractable as for each new encoder, the entirety of available monolingual data needs to be re-embedded, and for English alone, this means tens of billions of sentences and large compute resources. In order to overcome these issues, one approach is to train smaller mutually compatible sentence encoders using the teacher-student distillation technique proposed by Reimers and Gurevych (2020). Several extensions of this underlying idea were proposed by Heffernan et al. (2022) that we adopt (see Section 5.3.2).

### 5.3.1 Related work

**Mining Methodology.** In order to find aligned texts, early approaches focused on information beyond the text itself. One notable example is the STRAND algorithm which looked for articles with a similar document structure to find translated web pages (Resnik, 1999; Resnik and Smith, 2003). Work subsequently used text in the pages as the basis for alignment using techniques such as crosslingual document retrieval (Munteanu and Marcu, 2005; Utiyama and Isahara, 2003), bag of words or language models (Buck and Koehn, 2016), Jaccard similarity (Azpeitia et al., 2017, 2018; Etchegoyhen and Azpeitia, 2016), or translation (Abdul-Rauf and Schwenk, 2009; Bouamor and Sajjad, 2018). More recent approaches have begun to leverage advancements in representation learning by encoding texts into an embedding space, and then using a distance-based method to determine if a pair of texts in different languages have a similar meaning. Works such as España-Bonet et al. (2017); Guo et al. (2018); Hassan et al. (2018); Yang et al. (2019) used bilingual embeddings, but this has the limitation of not being able to directly mine across many languages. In order to address this, learning a massively multilingual embedding space allows for any pair of languages to be encoded and mined (Artetxe and Schwenk, 2019a,b; Feng et al., 2020; Kvapilíková et al., 2020; Schwenk, 2018). When attempting to mine at scale, a few approaches have been applied to large quantities of language pairs. One such example is the ParaCrawl project[20] which mined data for all official EU languages. A similar approach is that of the ccAligned project (El-Kishky et al., 2020). More recently, Schwenk et al. (2021b) presented CCMatrix which successfully mined billions of sentences from the web using the LASER multilingual embedding space (Artetxe and Schwenk, 2019b). The Samanantar project focused on providing a large mined corpus for eleven Indian languages (Ramesh et al., 2022).

**Multilingual Sentence Representation Learning.** There are a wide range of works covering the learning of multilingual representations such as mBERT (Devlin et al., 2019), XLM (Conneau and Lample, 2019), and XLM-R (Conneau et al., 2020). However, when applying these approaches to obtain representations at the sentence level, they can often suffer from the lack of an explicit sentence-based criterion during training, resulting in poor performance on tasks such as bitext retrieval (Hu et al., 2020). In order to overcome this issue, methods have been explored with an explicit sentence objective such as SentenceBERT (SBERT) (Reimers and Gurevych, 2019). SBERT is based on a siamese network, subsequently fine-tuned on NLI data (Bowman et al., 2015), and produces sentence embeddings by classifying pairs of sentences. Similar to this siamese network, LaBSE also uses a dual-

20. `https://paracrawl.eu/`

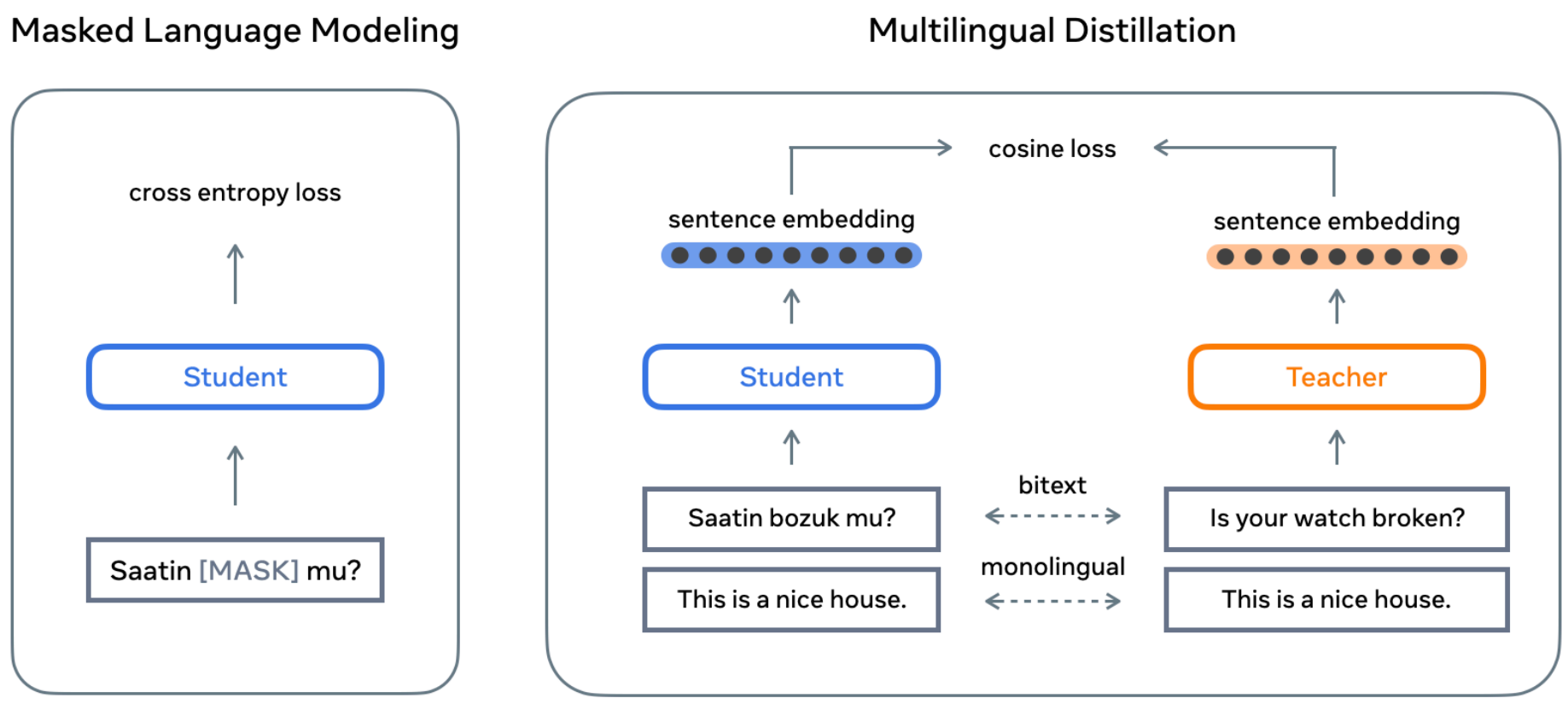

Figure 12: **Architecture of the LASER3 Teacher-Student Approach.** We refer the reader to Heffernan et al. (2022) for more details.

encoder approach with BERT. However, pre-training for the BERT encoders is done using both a masked language modelling (MLM) and translation language modelling (TLM) objective (Conneau et al., 2020). Sentence embeddings are then produced by passing bilingual translation pairs through the dual-encoder setup and then applying an additive margin softmax loss (Yang et al., 2019). Another popular multilingual embedding model is LASER (Artetxe and Schwenk, 2019b). Unlike the previous BERT-based approaches, there is no pre-training, and sentence embeddings are produced by using an encoder-decoder architecture, and then max-pooling over the encoder outputs.

When attempting to learn a multilingual embedding space, one of the limitations of many existing approaches is that each time a model is to be expanded to include a new set of languages, the entire model must be retrained from scratch, which is very costly. In order to address this limitation, Wang et al. (2020b) provide a technique which is able to extend mBERT to low-resource languages by first increasing the size of the existing vocabulary, and then continuing self-supervised training using low-resource monolingual data. Another approach introduced by Reimers and Gurevych (2020) uses a teacher-student approach. In this distillation setup, English-paired bitexts are used to both learn the English embedding space of a monolingual teacher (SBERT), while also using the non-English side to learn a new language. Heffernan et al. (2022) further built upon this approach by experimenting with different architectures for teacher and student (e.g., BiLSTM and Transformer), and also applying such distilled sentence representations to the task of bitext mining.

#### 5.3.2 Student-Teacher Mining Approach

The overall approach focuses on starting with a massively multilingual sentence encoder *teacher* model and adapting it to several different low-resource *student* models. This enables us to add low-resource languages without needing to compete with high-resource languages for capacity, to avoid retraining the full model from scratch, and to maintain compatibility of the multilingual embedding spaces for subsequent mining. Figure 12 summarizes the overall architecture of the teacher-student approach. The teacher, LASER2, is an improved version of the open-source LASER encoder.[21] The original training procedure (Artetxe and Schwenk, 2019b) was changed as follows: the use of SentencePiece tokenization and upsampling of low-resource languages. The architecture of the 5-layer BiLSTM encoder and the max pooling method to obtain sentence embeddings were left unchanged. Training was performed on the same 93 languages with public resources obtained from OPUS (Tiedemann, 2012). The reader is referred to Artetxe and Schwenk (2019b) for details on the original LASER training procedure. Training of the students follows the approach described in greater detail in Heffernan et al. (2022):

- students are specialized for one language or several similar languages;
- students are randomly initialized since we want to handle low-resource language for which we don't have a pre-trained LM;
- students may have a dedicated SentencePiece vocabulary different from the teacher, to better accommodate scripts and tokens in the student languages (see Section 5.3.3)
- students learn to minimize the cosine loss with the teacher, since we also use cosine distance for bitext mining (see Figure 12);
- students can have an MLM loss to leverage student language monolingual data (see Figure 12 and Section 5.3.3).

**Training Parameters.** Our student encoders used a 12-layer transformer, hidden size of 1024, with 4 attention heads, totalling around 250M parameters. All students were trained with available bitexts in their respective language, complemented by two million sentences of English/English and English/Spanish. The motivation is to *anchor* the students to the English embedding space, make it more robust by including English/Spanish bitexts from CCMatrix, and jointly learn new languages. This technique is particularly useful when only limited amounts of bitexts are available to train the students. Teacher-student training was performed on 16 GPUs, ADAM optimizer, a learning rate of 0.0005, and a batch size of 10,000. We trained student encoders for 148 languages and named these models LASER3.

**Proxy Metric for New Encoders.** Mined bitexts will be subsequently utilized to improve translation quality for 200 languages. Consequently, our primary metric is neural machine translation (NMT) quality. However, mining and NMT training are computationally expensive, and it is intractable to systematically perform this evaluation for many different sentence encoder variants. As an evaluation proxy, we use a mining-based multilingual similarity search error rate, referred to here as `xsim`. As opposed to cosine accuracy which aligns embeddings based on the highest cosine score, `xsim` aligns source and target embeddings based on the highest margin score, which was shown to be beneficial in mining (Artetxe and Schwenk, 2019a). The margin-based score is defined as:

$$\text{score}(x, y) = \text{margin}\left( cos(x, y), \sum_{z \in NN_k(x)} \frac{cos(x, z)}{2k} + \sum_{v \in NN_k(y)} \frac{cos(y, v)}{2k} \right) \tag{1}$$

21. `https://github.com/facebookresearch/LASER`

where $x$ and $y$ are the source and target sentences, and $NN_k(x)$ denotes the $k$ nearest neighbors of $x$ in the other language. We set $k$ to 4. All `xsim` results are calculated on Flores-200 devtest, using the *ratio* margin where $\text{margin}(a, b) = \frac{a}{b}$. Additionally, all scores are calculated into English (i.e. xxx $\rightarrow$ eng). English is encoded by the teacher and the other language by the LASER3 student. To facilitate further research using `xsim`, we also open-source this evaluation method to the community.[22]

**End-to-end Encoder Evaluation.** Once we have identified the best sentence encoder for each language using the `xsim` scores, we perform mining, add the mined data to the existing bitexts, and train a bilingual NMT system. Initial experiments indicated that a threshold on the margin of 1.06 seems to be the best compromise between precision and recall for most of the languages. For these NMT baselines, we do not apply additional filtering on the bitexts and leave this to the training procedure of our massively multilingual NMT system (see Section 6.4). Thus, the translation performance is compared to training the same NMT system on existing bitexts only. This enables us to asses the improvements brought *only* by mined bitext. We further limit our experimentation to the translation from foreign languages into English. This helps to compare resources and assess the translation quality among all languages. The English corpus we mine in has 21.5 billion unique sentences.

We do not attempt to optimize the architecture and parameters of the bilingual NMT systems to the characteristics of each language pair, namely the size of available bitexts, but use the same architecture for all. Therefore, the reported results should not be interpreted as the best possible ones given the available resources – they are mainly provided to validate the mined bitexts. We use a 12 layer encoder and decoder and train for 100 epochs. The SentencePiece (SPM) vocabulary has 7,000 tokens. We look for best performance on the Flores-200 development set, and report detokenized BLEU on Flores-200 devtest.

#### 5.3.3 Language-Specific Encoder Training

The original LASER encoder, as used in the CCMatrix project (Schwenk et al., 2021b) performs very well on several high-resource languages like Arabic, Chinese, Czech, German, Japanese, Polish or Indonesian. We used the CCMatrix bitexts directly for these languages. We then trained new sentence encoders and performed mining for the 148 remaining ones. In the following, we discuss several representative individual languages and families. To perform mining, we apply the same algorithms and optimizations as proposed in Schwenk et al. (2021b).

**Improving LASER.** To highlight the improvements upon LASER using our teacher-student approach, Figure 13 shows the `xsim` error rates for some of LASER's supported languages, which were re-trained using our LASER3 student models. On average the `xsim` error is brought down from 61 to 0.91. In particular, some languages such as Burmese (`mya_Mymr`) and Irish (`gle_Latn`) saw the biggest reductions in `xsim`, with decreases of 93.3 $\rightarrow$ 0.89 and 92.5 $\rightarrow$ 0.79, respectively.

**European Minority Languages.** There is a large variety of languages which are spoken locally in various regions of Europe. We consider here 10 languages. For most of them, we were not able to find meaningful amounts of public bitexts and heavily utilize NLLB-Seed

22. `https://github.com/facebookresearch/LASER/`

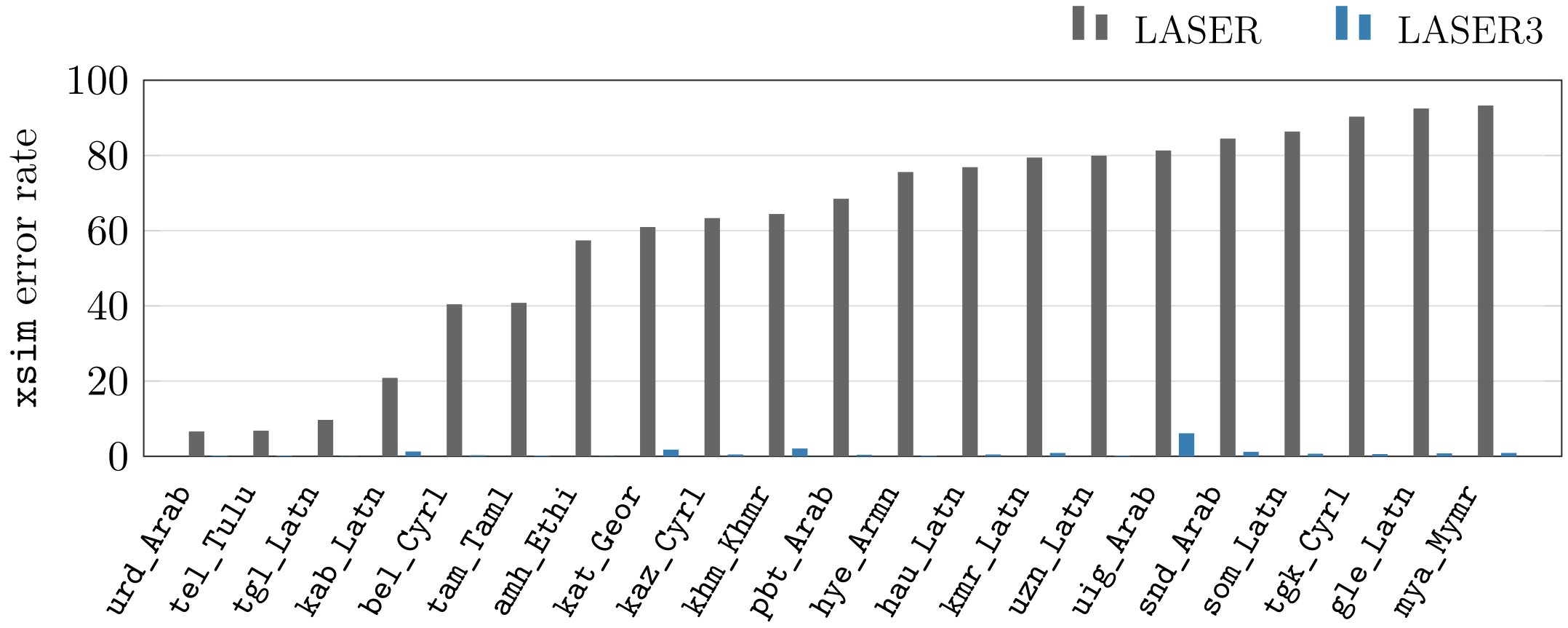

Figure 13: **Comparison of `xsim` Error Rates** from LASER3 and the original LASER encoder. LASER3 has significant improvements (lower is better).

| **Lang.** | `fao` | `fur` | `lij` | `lim` | `lmo` | `ltz` | `srd` | `szl` | `vec` | `ydd` |
|---|---|---|---|---|---|---|---|---|---|---|
| **Script** | Latn | Latn | Latn | Latn | Latn | Latn | Latn | Latn | Latn | Hebr |
| **Addtl. Lang** | deu | ita | ita | nld | ita | deu | ita | pol | ita | deu |
| **Bitexts** [k] | 6.6 | 6.3 | 2.2 | 5.4 | 1.3 | 9.8 | 1.4 | 6.4 | 1.2 | 6.2 |
| **BLEU** | 0 | 0 | 0 | 0 | 0 | 0 | 0 | 0 | 0 | 0 |
| `xsim` [%] | 2.57 | 0.1 | 0.2 | 16.1 | 1.09 | 0.59 | 0.1 | 0.69 | 2.77 | 0.1 |
| **Monolingual** | 1.2M | 737k | 106k | 15M | 61M | 123M | 515k | 2.5M | 12M | 12M |
| **Mined** | 1.6M | 532k | 631k | 2.0M | 4.1M | 5.5M | 723k | 1.0M | 2.5M | 3.3M |
| **BLEU** | 10.6 | 23.5 | 13.4 | 5.5 | 20.7 | 37.0 | 20.9 | 18.9 | 17.8 | 30.1 |

Table 10: **Statistics and Mining Performance** for European minority languages. BLEU scores are foreign into English.

(see Section 4.2). Therefore, we chose to train individual sentence encoders specific to each of these languages and complemented the training data with 2M bitexts in a similar higher-resource European languages. Typical examples are Sicilian paired with Italian or Silesian paired with Polish. Detailed information for all 10 languages are given in Table 10.

We observe that pairing the minority language with another similar major language yields encoders with very low `xsim` error rates for most of the languages, and we were able to mine large amounts of bitexts yielding good NMT performance. Three languages reach a BLEU score superior to 20 (`fur_Latn`, `lmo_Latn` and `srd_Latn`), and two languages superior to 30 (`ltz_Latn` and `ydd_Hebr`).

**Creole Languages.** We applied a similar strategy for Creole languages. Linguistically speaking, Creole languages are a possible outcome of a contact situation between languages over a fairly brief period of time (Lent et al., 2022). When training the sentence encoders, we paired each Creole language with a "similar" high-resource language (see Table 11). The sentence encoders have a very low `xsim` error rate for all languages except Sango (`sag_Latn`) for which we were not able to identify a similar language with sufficient resources.

| **Lang.** | hat | kea | pap | sag | tpi |
|---|---|---|---|---|---|
| **Script** | Latn | Latn | Latn | Latn | Latn |
| **Addtl. Lang** | fra | por | spa por | lin | eng |
| **Bitexts** | 334 | 6 | 5 | 282 | 458 |
| **BLEU** | 20.2 | 0 | 0 | 4.8 | 14.7 |
| xsim [%] | 1.19 | 1.19 | 0.1 | 8.6 | 0.2 |
| **Monolingual** | 14M | 227k | 28M | 645k | 1.7M |
| **Mined** | 8.0M | 656k | 7.3M | 1.9M | 1.2M |
| **BLEU** | 29.2 | 4.9 | 40.9 | 5.3 | 16.1 |

Table 11: **Statistics and Mining Performance** for Creole languages.

| **Lang.** | kab | taq | taq | tzm |
|---|---|---|---|---|
| **Script** | Latn | Latn | Tfng | Tfng |
| **bitexts** | 72k | 10.2k | 4k | 6.2k |
| **BLEU** | 1.2 | 0 | 0 | 0 |
| xsim [%] | 0.99 | 24.11 | 35.57 | 3.66 |
| **Monolingual** | 3.4M | 23k | 5k | 59k |
| **Mined** | 3.1M | 240k | - | 111k |
| **BLEU** | 6.2 | 1.2 | - | 3.8 |

Table 12: **Statistics and Mining Performance** for Berber languages.

Let us compare two extreme settings: Papiamento (pap_Latn) and Kabuverdianu (kea_Latn). For the former, we were able to crawl 28M sentences of monolingual data, while we have less than 300k for the latter. Both have less than ten thousand sentences of existing bitexts which is largely insufficient to train an NMT system. We were able to mine more than 7M bitexts for Papiamento which yielded an impressive BLEU score of 40.9, while we only achieve BLEU 4.9 for Kabuverdianu. This highlights that the amount of available monolingual data is crucial to make bitext mining successful.

**Berber Languages.** We considered several languages of the Berber family, namely Kabyle (kab_Latn), Tamashek (taq) and Central Atlas Tamazight (tzm). We consider Tamazight written in the Tifinagh script and Tamashek both in Latin and Tifinagh script. All are very low-resource languages with barely 10,000 sentences of available bitext, and 72,000 for Kabyle. It was also very challenging to collect monolingual data for these Berber languages. These statistics are given in Table 12.

We were able to mine bitexts for Kabyle and reach a modest BLEU score of 6.2. xsim error rates for Tamashek are above 20% and insufficient amount of monolingual data make it impossible to mine bitext of good quality for Tamashek. For Tamazight, a very small amount of bitexts could be mined. Tamashek and Tamazight are typical examples of very low-resource languages for which it seems to be very hard to collect written material to support training of machine translation system.

**Malayo-Polynesian Languages.** Let us consider a larger language family: the Malayo-Polynesian family. We discuss here 13 languages from this family, and use a single encoder for ten of them. The languages Fijian, Maori, and Samoan are handled by a separate encoder. The result overview is given in Table 13.

We observe very low xsim error rates for most of the languages, although several languages have less than hundred thousand sentences of bitexts. Training all languages together in one specific encoder for this family seems to be very beneficial for these very low-resource languages. In addition, we were able to collect several million sentences of monolingual text for most of the languages. This gives us optimal conditions for mining and we achieve substantial improvements in the BLEU score compared to training a bilingual NMT system on the public bitexts only. We have two languages which improved by more than twenty

| Lang. | bitexts | BLEU | xsim % | Monol. | Mined | BLEU |
|---|---|---|---|---|---|---|
| ace_Latn | 39.2k | 0 | 2.37 | 2.2M | 1.4M | 10.3 |
| bug_Latn | 21.8k | 0 | 1.58 | 0.7M | 717k | 4.2 |
| ceb_Latn | 1.1M | 34.4 | 0.1 | 23.6M | 8.1M | 39.0 |
| ind_Latn | 11M | - | 0.1 | - | - | - |
| jav_Latn | 86k | 11.1 | 0.1 | 27.2M | 8.5M | 31.2 |
| pag_Latn | 327k | 15.6 | 0.69 | 3.9M | 1.9M | 18.5 |
| sun_Latn | 32.3k | 1.5 | 0.59 | 8.2M | 6.1M | 28.5 |
| tgl_Latn | 1.3M | 40.2 | 0.1 | 89M | 33M | 43.8 |
| war_Latn | 331k | 26.5 | 0.2 | 26.9M | 4.9M | 36.5 |
| zsm_Latn | 2.3M | 34.4 | 0.0 | 640M | 40.5M | 41.4 |
| fij_Latn | 667k | 15.0 | 0.3 | 1.5M | 2.2M | 14.6 |
| mri_Latn | 45k | 5.6 | 1.38 | 3.8M | 3.2M | 20.5 |
| smo_Latn | 419k | 20.0 | 0.2 | 3.5M | 3.7M | 25.6 |

Table 13: **Statistics and Mining Performance** for Malayo-Polynesian languages.

BLEU points: Javanese (jav_Latn): 11.1 → 31.2 and Sundanese (sun_Latn): 1.5 → 28.5. Even high-resource languages like Tagalog (tgl_Latn) and Standard Malay (zsm_Latn) see significant improvements of the BLEU score. We did not mine new bitexts for Indonesian since CCMatrix already provides 70 million parallel sentences, but we included it to help learning the sentence encoder for those languages. We also observe very good performance for Maori: the BLEU score improves from 5.6 to 20.5, while mined bitexts did not improve NMT performance for Fijian.

**African Languages.** Among the 200 languages of the NLLB project, 55 are spoken in the African continent, which is more than a quarter of all languages we handle. Except for seven languages — Modern Standard Arabic, Afrikaans, Southern Sotho, Swahili, Tswana, Xhosa and Zulu — all are low-resource languages, i.e. with less than one million publicly available sentence pairs. Twelve of them even have less than hundred thousand sentence pairs, which we named very low-resource languages. In addition, we struggled to curate meaningful amounts of monolingual data. Given these facts, training sentence encoders for African languages and mining high quality bitexts turned out to be a major challenge — even in the broader community. In fact, collecting resources, training NMT systems, and performing evaluations for African languages is the focus of several works (Abbott and Martinus, 2019; Azunre et al., 2021c; Dabre and Sukhoo, 2022; Emezue and Dossou, 2020; Hacheme, 2021; Nekoto et al., 2020; Siminyu et al., 2021). A detailed description and analysis of our effort is reported in Heffernan et al. (2022).

The average BLEU score over 44 languages increased from 11.0 to 14.8 with help of the mined bitexts. We also deployed a new training procedure which combines supervised training, i.e. minimizing the cosine loss between the teacher and student embedding, and unsupervised masked LM training (see left part of Figure 12). This enabled us to benefit from monolingual data during encoder training. This new approach yielded improved encoders for difficult languages such as Wolof (Heffernan et al., 2022).

| Training | SPM | #train | `amh_Ethi` | `tir_Ethi` |
|---|---|---|---|---|
| LASER2 | 50k joint | 220M | 34.9 | 92.9 |
| Semitic | 50k joint | 9M | 0.2 | 1.19 |
| Ge'ez | 8k specific | 0.7M | 0.1 | 0.89 |
| LaBSE | 501k joint | $\approx$ 6000M | 0 | 13.74 |

Table 14: **`xsim` Error Rates on FLORES devtest for Amharic and Tigrinya and different training strategies.** The specified amount of training data excludes 4M sentences of English for our models.

**Handling Specific Scripts.** In massively multilingual systems, a common approach is to utilize the same SentencePiece vocabulary. In our initial experimentation, we re-used the LASER2 teacher 50k vocabulary for all student encoders. This has some advantages as shared vocabulary could ease generalization when the same tokens appear in multiple languages. However, despite upsampling, low-resource languages could be poorly represented in a joint vocabulary. We next explore utilizing specific vocabularies for small subsets of languages. Table 14 summarizes the results for different vocabulary strategies.

Amharic was part of LASER2 but the `xsim` error rate is rather high, and LASER2 generalizes badly to Tigrinya. We first explore training an encoder for three Semitic languages: Amharic, Tigrinya, and Maltese. This yields a significant improvement: `xsim`=0.2% and 1.19% respectively, highlighting the usefulness of teacher-student training and specific encoders for a small set of similar languages. We then trained an encoder for Amharic and Tigrinya only, paired with English as in all our experiments, and a specific 8k SentencePiece vocabulary to better support the Ge'ez script. This brought `xsim` down to 0.1% and 0.89%, respectively, even though we use less training data. Our best model is on par with LaBSE (which includes only Amharic), and significantly outperforms it for Tigrinya.

#### 5.3.4 Mining at a Glance

Overall, we mined 148 bitexts paired with English which totals to 761 million sentence pairs with an alignment score of at least 1.06. As mentioned above, bitexts aligned with English for the remaining languages were taken from CCMatrix (Schwenk et al., 2021b). We also provide 1465 non-English bitext pairs. This corresponds to 302 million sentence pairs. This includes all language pairs among African, Indic and Malayo-Polynesian families, respectively, as well as alignments of all African languages with French.

#### 5.3.5 Limitations of Large-Scale Mining

We note that for some languages, we are only able to create a small amount of bitext through data mining. The limiting factor is predominantly the lack of monolingual data. Many low-resource languages have limited presence on the web, and the data that we curate can be heavily filtered at many stages: language identification, aggressive cleaning of monolingual data, or even poorly aligned bitext. Even with our best efforts, those challenges compound, and they can affect certain languages far more than others. Further, note that even in our mined bitext, we still end up including some content that is already available from other sources — a good example is the Christian Bible. An important final consideration is the web already has machine translated content. For example, many websites may use translation to 'internationalize' their content. A positive is that a large majority of the languages we focus on are not contained in most existing commercial translation services (see Table 1), however as we mine against higher-resource languages, it is likely our mined datasets contain translated content.

#### 5.3.6 Ethical Considerations for Mining Research

To close this section, we would like to reflect on the issue of data ownership. While we performed our due diligence on deployments of all licensed datasets, the ownership of low-resource language data remains an open debate. In our interview study, many low-resource stakeholders express that sharing language access might in fact be a necessary trade-off for technological advancement. Blocking such access meant blocking any future benefits that could positively impact low-resource language communities. However, we stress that access and ownership are two disparate concepts. Even though we deploy many low-resource language datasets, ownership ultimately belongs to the speakers of these languages.[23]

### 5.4 Conclusion

We describe our methodology for automatically creating aligned translation training data for low-resource languages. We face significant challenges, as bitext data to train sentence encoders such as LASER and monolingual data to mine in is extremely scarce. To offset these, we develop **(1)** a high-quality language identification system for over 200 languages that outperforms publicly available LIDs in both Flores-200 and web domains, **(2)** a detailed, documented monolingual dataset curation and cleaning pipeline, and **(3)** a teacher-student based multilingual sentence encoder training methodology that enables transfer to extremely low-resource languages with minimal supervised bitext. These contributions combine to create over 1.1 billion new sentence pairs of training data for 148 languages.

**Democratizing Large-Scale Dataset Creation.** We are releasing all the code that was used to train the LID model, and run the monolingual sentence splitting and filtering. We are also releasing the mining code, under an open source license, with tools to run mining with our open-sourced LASER encoders. We have built a new mining library: `stopes`[24] that can run locally or scale over any cluster of your choice. This library is shipped to use the LASER encoders described in this article, but also any encoder available on Huggingface sentence transformers.[25] It also enables users to run the end-to-end mining pipeline with simple configurable tooling. Our hope is that this will allow more people to create high quality bitext datasets in their language of choice, and extend the research on language mining and translation. The encoders can be found in LASER repository, and the teacher-student code in the nllb branch of the fairseq repository.

---

23. For example, consider the Kaitiakitanga License `https://github.com/TeHikuMedia/Kaitiakitanga-License/blob/tumu/LICENSE.md`
24. `https://github.com/facebookresearch/stopes`
25. `https://huggingface.co/sentence-transformers`

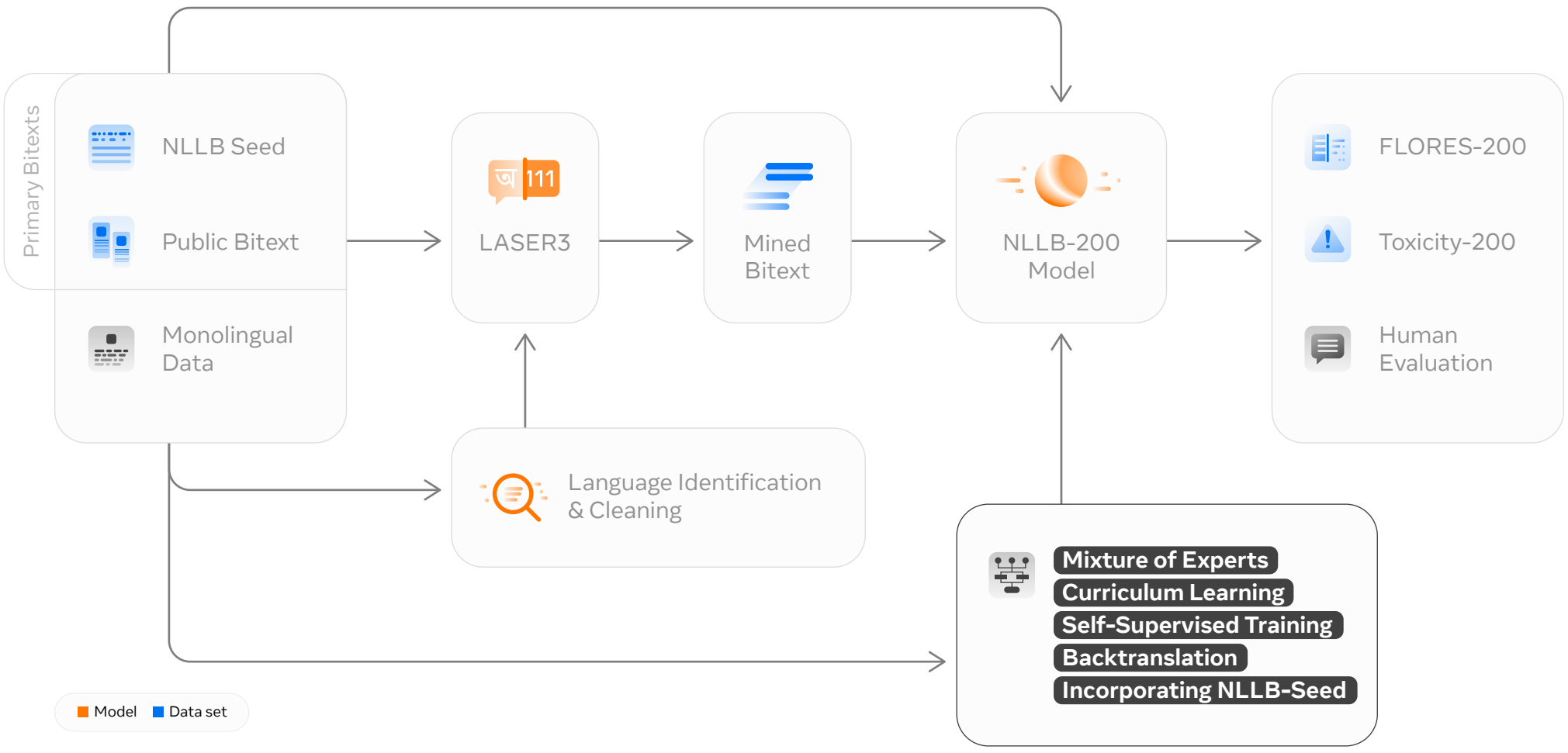

Figure 14: **Modeling Contributions of No Language Left Behind**: As highlighted, we describe several modeling techniques to enable coverage of hundreds of languages in one model. We focus on effectively scaling model capacity while mitigating overfitting, as well as how to improve backtranslation for low-resource languages and incorporate NLLB-Seed.

## 6. Modeling

Existing research in massively multilingual machine translation has been predominantly restricted to about 100 languages, which represent only a fraction of globally written languages (Arivazhagan et al., 2019; Fan et al., 2020; Zhang et al., 2020). Some works have extended beyond (Mueller et al., 2020; Siddhant et al., 2022), but critically lack reliable performance evaluation. Despite much research, the translation quality for languages with low volumes of data is typically poor. Further, adding extremely low-resource languages beyond 100 is very challenging because there is often very little existing good quality available bitext, and even large scale bitext mining (Schwenk et al., 2021b) can struggle to create sufficiently large datasets (see Section 5.3). Data often varies widely in available volume across languages, creating imbalance between different language directions. This complicates massively multilingual training, as some language directions have begun overfitting while others have not yet even converged.

In this section, we investigate how to most effectively scale multilingual machine translation to hundreds of languages. We develop several novel techniques that tackle the major challenges of low-resource translation, such as training models with sufficient capacity to represent 200+ languages while adjusting to variable data capacity per language pair. We present several sets of experiments that help us identify the most performant model architecture and training strategy that can scale to hundreds of languages (see Section 8.1). These involve **(1)** conditional compute models to minimize interference between unrelated language directions, **(2)** studying the best self-supervision strategies on monolingual data, and **(3)** data augmentation or backtranslation for the lowest volume and low accuracy language directions.

## 6.1 Preliminaries

We first describe the multilingual machine translation task setup including tokenization, model architecture, and the ablation dataset we use in detail.

### 6.1.1 Task Setup

We model multilingual neural machine translation as a sequence-to-sequence task, where we condition on an input sequence in the source language with an encoder and generate the output sequence in the expected target language with a decoder (Bahdanau et al., 2015). We train to maximize the probability of the translation in the target language $T$, given the source sentence $S$, the source language $\ell_s$, and the target language $\ell_t$, i.e., $P(T|S, \ell_s, \ell_t)$. In this section, we discuss details of **(1)** tokenization of the text sequences in the source and target languages, **(2)** model architecture with the input and output specifically designed for multilingual machine translation, and **(3)** the multilingual machine translation dataset we use for ablation experimentation of our ideas at a smaller, but representative scale.

**Segmentation with SentencePiece.** To tokenize our text sequences, we train a single SentencePiece (SPM) (Kudo and Richardson, 2018) model for all languages. We sample a total of 100M sentences from primary bitext data. For more details on the primary bitext data, see Section 8.1.2. To ensure low-resource languages are well-represented in the vocabulary, we downsample high-resource and upsample low-resource languages with a sampling temperature of 5 (Arivazhagan et al., 2019). Vocabulary size is a critical hyperparameter in multilingual translation models involving low-resource languages (Oladipo et al., 2022; Rajab, 2022). The vocabulary size of our trained SPM model is 256,000. Such a large vocabulary size ensures adequate representation across the wide spectrum of languages we support. See Section 8.1.1 for more details.

**Model Architecture.** Our sequence-to-sequence multilingual machine translation model is based on the Transformer encoder-decoder architecture (Vaswani et al., 2017). The encoder transforms the source token sequence into a sequence of token embeddings. The decoder attends to the encoder output and autoregressively generates the target sentence token by token. More precisely, the encoder takes the sequence of tokens $W = (w_1, \ldots, w_S)$ and the source language $\ell_s$, and produces a sequence of embeddings $H = (h_1, \ldots, h_S)$, which are then provided to the decoder with the target language $\ell_t$ to produce the target tokens $V = (v_1, \ldots, v_T)$ sequentially, that is,

$$H = \texttt{encoder}(W,\ \ell_s), \quad (2)$$

$$\forall i \in [1, \ldots, T],\ v_{i+1} = \texttt{decoder}(H,\ \ell_t,\ v_1,\ \ldots,\ v_i). \quad (3)$$

Note that we prefix the source sequence with the source language, as opposed to the target language as previously done in several works (Arivazhagan et al., 2019; Johnson et al., 2017). This is primarily because we prioritize optimizing zero-shot performance of our model on any pair of 200 languages at a minor cost to supervised performance. Empirically, we find zero-shot performance to be negatively affected when conditioning the encoder on the target language. When the source is conditioned on only the source language, the encoder generalizes better to pairs of source and target languages not encountered during training (Fan et al., 2020).

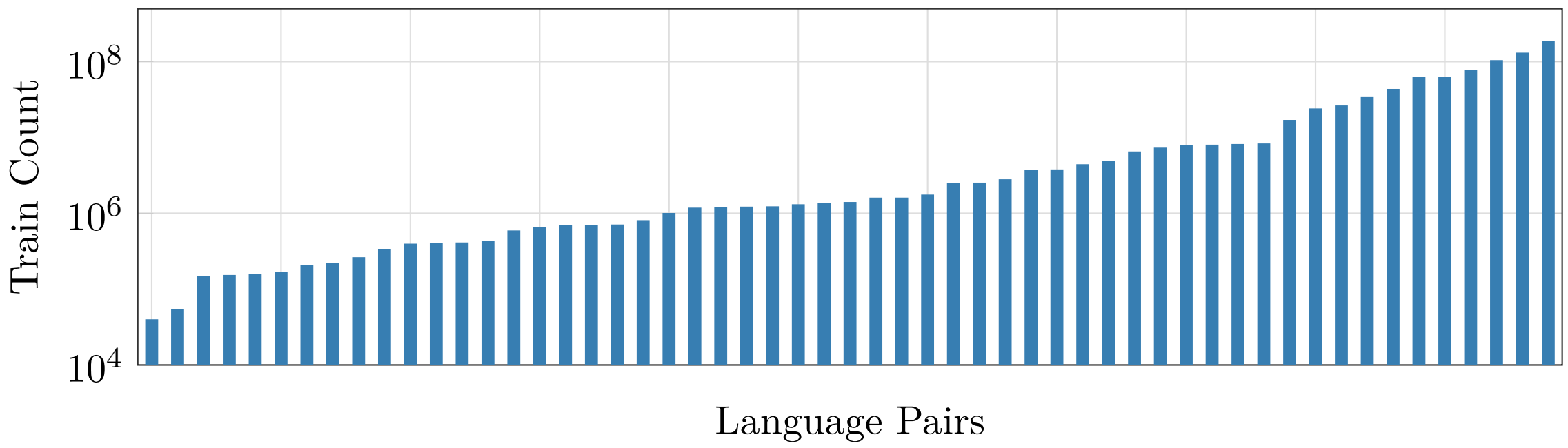

Figure 15: **Ablation Dataset Counts** depicting the amount of training data across all language pairs, ranking from 39,992 to 18.7 million sentence pairs.

Both the encoder and decoder are stacks of Transformer layers. Each Transformer layer takes a sequence of embeddings as input and outputs a sequence of embeddings. In the encoder, Transformer layers are composed of two sub-layers, a self-attention and a feed-forward layer. These are applied sequentially and are both preceded by a LayerNorm (Ba et al., 2016) and followed by a residual connection (He et al., 2015):

$$Z = X + \texttt{self-attention}(\texttt{norm}(X)), \quad (4)$$
$$Y = Z + \texttt{feed-forward}(\texttt{norm}(Z)). \quad (5)$$

We apply LayerNorm at the beginning of each sub-layer (Pre-LN), as opposed to applying LayerNorm after the residual connection at the end of each sub-layer (Post-LN). This is because Pre-LN is more stable in practice compared to Post-LN (Xiong et al., 2020). The self-attention layer is an attention layer that updates each element of the sequence by looking at the other elements, while the feed-forward layer (FFN) passes each element of the sequence independently through a 2-layer MLP. In the decoder, there is an additional third sub-layer, between the self-attention and the feed-forward, which computes attention over the output of the encoder. We refer the reader to Vaswani et al. (2017) for further details.

#### 6.1.2 Ablation Dataset

We construct a multilingual machine translation benchmark such that it is representative of our final benchmark on 200+ languages. We choose a representative sub-sample of 53 out of 202 languages and a total of 110 translation directions (see Table 51 in the appendix). These consist of 45 directions out of English (aggregated as `eng_Latn-xx`), 45 directions into English (aggregated as `xx-eng_Latn`) and 20 non-English directions (aggregated as `xx-yy`). In terms of resource level, there are 40 high-resource and 70 low-resource directions (see Table 1 and Table 51). Out of 70 low-resource directions, 22 are *very low-resource*, i.e., have less than 100K training examples. The dataset is composed of publicly available bitext in all 110 language directions (see Section 8.1.2) and large scale mined data (see Section 5.3) in English-centric directions.[26] There are a total of 864M examples in this benchmark. The highest resource language pair has 186M examples and the lowest resource language pair has 40K examples, thus representing the extreme skew characteristic of the final dataset with 202 languages. Figure 15 shows the data distribution over language pairs sorted by the example count per pair. We call this dataset our *ablation* dataset and use this throughout all experiments in this section.

### 6.2 Conditional Compute for Massively Multilingual Machine Translation

A massively multilingual translation model is trained on several translation directions at once, utilizing the same shared model capacity. This can lead to beneficial crosslingual transfer between related languages at the risk of increasing interference between unrelated languages (Conneau et al., 2020; Fan et al., 2020). Sparsely Gated Mixture of Experts (MoE) models are a type of conditional compute models (Almahairi et al., 2016; Bengio et al., 2013) that activate a subset of model parameters per input, as opposed to *dense* models that activate all model parameters per input. MoE models unlock significant representational capacity while maintaining the same inference and training efficiencies in terms of FLOPs as compared to the core dense architecture. In this section, we study how we can use Sparsely Gated Mixture of Experts models (Du et al., 2021; Hwang et al., 2022; Lepikhin et al., 2020; Lewis et al., 2021; Shazeer et al., 2017; Zoph et al., 2022) to achieve a more optimal trade-off between crosslingual transfer and interference and improve performance for low-resource languages.

**Sparsely Gated Mixture of Experts.** As illustrated in Figure 16, we replace the FFN sublayer in dense models with an MoE sublayer once every $f_{\mathrm{MoE}}$ layers in both the encoder and decoder. The MoE sublayer consists of $E$ feed-forward networks (FFN), denoted with $(\mathrm{FFN}_1, \mathrm{FFN}_2, \ldots, \mathrm{FFN}_E)$, each with input and output projections $W_i^{(e)}$ and $W_o^{(e)}$. A gating network, consisting of a softmax-normalized linear layer with weights $W_g$, is attached to each MoE sublayer to decide how to route tokens to experts. Given an input token $x_t$ the output of the MoE sublayer is evaluated as:

$$\mathrm{FFN}_e(x_t) = W_o^{(e)}\,\mathrm{ReLU}(W_i^{(e)} \cdot x_t), \quad (\forall e \in \{1, \ldots, E\}) \tag{6}$$

$$G_t = \mathrm{softmax}(W_g \cdot x_t), \quad \mathcal{G}_t = \text{Top-k-Gating}(G_t), \tag{7}$$

$$\mathrm{MoE}(x_t) = \sum_{e=1}^{E} \mathcal{G}_{te} \cdot \mathrm{FFN}_e(x_t), \tag{8}$$

with $\mathcal{G}_t \in \mathbb{R}^E$ the routing vector computed by the gating network, i.e., for each expert, $\mathcal{G}_{t,e}$ is the contribution of the $\mathrm{e}^{\mathrm{th}}$ expert ($\mathrm{FFN}_e$) in the MoE output. We follow the Top-k-Gating algorithm of Lepikhin et al. (2020) and dispatch each token to at most $k = 2$ experts. We always choose the top 2 scoring experts per token, and do not add randomization to the choice of the second expert.

26. For this ablation dataset, we only include English-centric data to manage experimental iteration speed. In Section 8, we include thousands of non-English training directions.

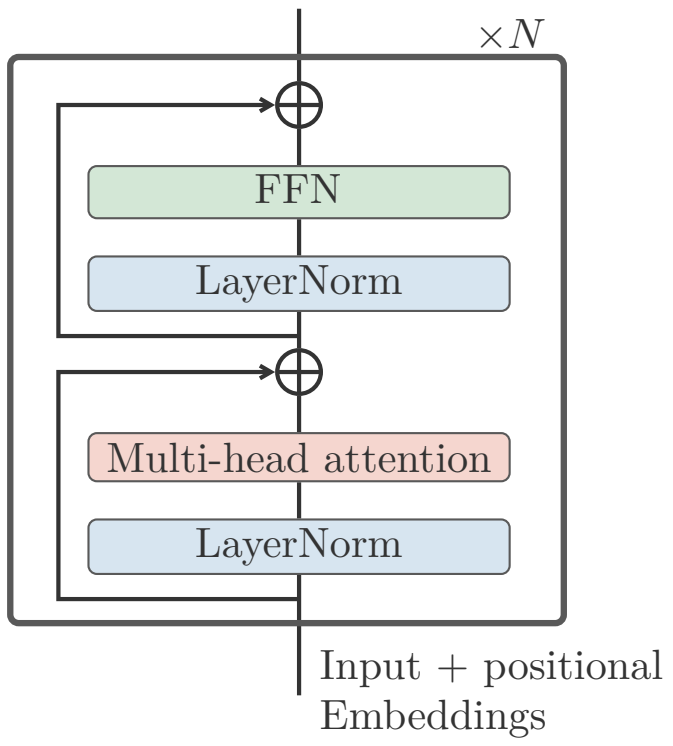

(a) Dense Transformer

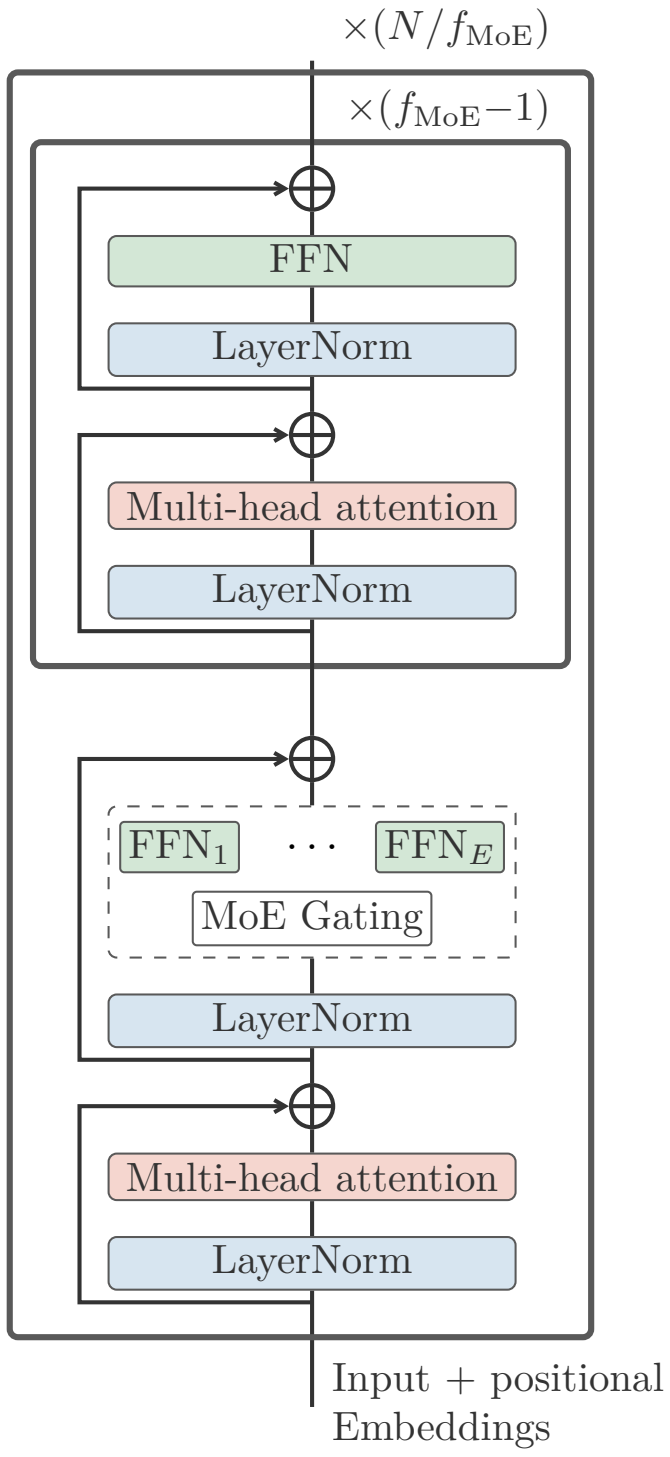

(b) MoE Transformer

Figure 16: **Illustration of a Transformer encoder with MoE layers** inserted at a 1:$f_{\text{MoE}}$ frequency. Each MoE layer has $E$ experts and a gating network responsible for dispatching tokens.

The Transformer encoder-decoder model, supplemented with MoE layers and their respective gating networks, learns to route input tokens to the corresponding top-2 experts by optimizing a linearly weighted combination of label-smoothed cross entropy (Szegedy et al., 2015) and an auxiliary load balancing loss (Shazeer et al., 2017). This additional loss term (LB) pushes the tokens to be uniformly distributed across experts and is evaluated as:

$$LB = E \cdot \sum_{e=1}^{E} f_e p_e, \quad p_e = \frac{1}{T} \cdot \sum_{t=1}^{T} G_{te}, \tag{9}$$

where $f_e$ is the fraction of tokens routed to the $\text{e}^{\text{th}}$ expert, as their first choice, through Top-k-Gating, and $p_e$ is the average routing probability to that expert over the $T$ tokens in the mini-batch. We refer the reader to Lepikhin et al. (2020) for more on the optimization of MoE models.

In the rest of this section, we first detail how we train vanilla Sparsely Gated MoE models for multilingual machine translation on our benchmark and show how they compare to dense models. We then discuss why these vanilla Sparsely Gated MoE models are suboptimal for low-resource language pairs (Section 6.2.1). We propose in Section 6.2.2 a series of architectural changes that significantly improve the performance on low-resource language pairs with MoE models. Finally, we devise and study a simple but effective curriculum learning strategy (Section 6.2.3) as another approach to get improvement on low-resource pairs with these models.

| | eng_Latn-xx | | | | xx-eng_Latn | | | | xx-yy |
|---|---|---|---|---|---|---|---|---|---|
| | all | high | low | v.low | all | high | low | v.low | all |
| Dense 615M | 41.7 | 54.2 | 36.6 | 30.4 | 51.1 | 61.5 | 46.8 | 44.0 | 39.4 |
| MoE-64 (~Dense 615M) | 43.0 | 55.3 | 37.9 | 30.3 | 52.6 | 63.3 | 48.3 | 44.2 | 39.8 |
| Δ | +1.3 | +1.1 | +1.3 | -0.1 | +1.5 | +1.8 | +1.5 | +0.2 | +0.4 |
| Dense 615M - $p_{\text{drop}}$=0.1 | 41.9 | 54.1 | 37.0 | 31.1 | 51.8 | 61.8 | 47.8 | 45.3 | 39.6 |
| MoE-64 (~Dense 615M) - $p_{\text{drop}}$=0.1 | 43.6 | 55.8 | 38.7 | 32.0 | 53.4 | 63.6 | 49.3 | 45.9 | 41.1 |
| Δ | +1.7 | +1.7 | +1.7 | +0.9 | +1.6 | +1.8 | +1.5 | +0.6 | +1.5 |
| Dense 1.3B | 43.3 | 55.4 | 38.4 | 31.6 | 53.5 | 63.6 | 49.4 | 46.5 | 41.3 |
| MoE-64 (~Dense 1.3B) | 43.3 | 55.9 | 38.2 | 29.7 | 52.9 | 63.9 | 48.4 | 43.7 | 39.3 |
| Δ | +0.0 | +0.5 | -0.2 | -2.0 | -0.6 | +0.3 | -0.9 | -2.8 | -2.1 |
| Dense 1.3B - $p_{\text{drop}}$=0.1 | 43.7 | 55.4 | 39.0 | 33.1 | 54.4 | 63.8 | 50.6 | 47.9 | 41.9 |
| MoE-64 (~Dense 1.3B) - $p_{\text{drop}}$=0.3 | 44.3 | 56.0 | 39.5 | 32.5 | 54.4 | 63.9 | 50.6 | 47.7 | 41.9 |
| Δ | +0.6 | +0.6 | +0.5 | -0.6 | +0.0 | +0.1 | +0.0 | -0.2 | +0.0 |

Table 15: **Vanilla Sparsely Gated MoE with and without overall dropout (validation set chrF++).** We report averages in each set of directions: `eng_Latn-xx`, `xx-eng_Latn` and `xx-yy` as *all*. For `eng_Latn-xx` and `xx-eng_Latn` we breakdown the pairs by resource level: high-resource (high), low-resource (low) and very low resource (v.low) — We see that a vanilla MoE model does not outperform the corresponding 1.3B dense model on the ablation benchmark. On adding overall dropout, we see a significant improvement in all directions on MoE models. At smaller computational cost per update (615M), MoE with overall dropout shows larger gains.

#### 6.2.1 Vanilla Sparsely Gated MoE and its drawbacks for Low-Resource Languages

The motivation behind sparsely activating expert subnetworks in an MoE model is to allow different parameters to model different aspects of the input space. We hypothesize that the added expert capacity should help higher resource language pairs that might otherwise be constrained to share the same dense model capacity with many other language pairs. We also hypothesize that with a massive number of translation directions, the added expert capacity would reduce interference, thus benefiting tasks of all resource levels. To verify this claim, and to understand the limits of vanilla MoE models, we compare in the following set of experiments the performance of MoE models to that of their dense counterparts on our ablation dataset.

**Experimental Setup.** We train a baseline dense Transformer encoder-decoder model with 1.3B parameters with model dimension 1024, FFN dimension 8192, 16 attention heads, 24 encoder layers and 24 decoder layers. Next, we train a corresponding Sparsely Gated MoE model by replacing the dense FFN sublayer with an MoE sublayer in every alternate

Transformer layer of the model ($f_{\text{MoE}}$=2). Each MoE sublayer has 64 experts (close to the number of languages in the benchmark, i.e., 53) and routes input tokens to the top-2 expert FFN sublayers in the MoE layer as in Lepikhin et al. (2020). All models are trained for 100k updates with an effective batch size of 1M tokens per update. For dense models, the objective function is label-smoothed cross-entropy ($\epsilon = 0.1$) (Szegedy et al., 2015), and for MoE models, the objective function is a weighted sum of label-smoothed cross-entropy and the load balancing loss (Equation (9)) with weights 1.0 and 0.01, respectively. We optimize with Adam (Kingma and Ba, 2015) using $(\beta_1, \beta_2, \epsilon) = (0.9, 0.98, 10^{-6})$. We linearly increase the learning rate up to 0.004 through 8000 warmup updates, then follow the inverse square root learning rate schedule. For Top-2-Gating, we set the expert capacity to $2 \times T/E$, i.e., we enforce that each expert processes, at most, $2 \times T/E$ tokens, where $T$ is the number of tokens in the mini-batch and $E$ is the number of experts. During generation, we set the capacity to $T$ so that all tokens can be routed to whichever expert they choose. We use the chrF++ metric to compare the model performance (see Section 7.1).

**Results.** In Table 15, we see that the Sparsely Gated MoE model with 64 experts (MoE-64), while computationally similar, shows good improvements when compared to the dense 615M model. We see 1+ chrF++ score improvements on all subsets except for very low resource pairs (v.low) and non-English pairs (`xx-yy`). There are worse trends when scaling up the computational cost per update — for the MoE-64 model (computationally similar to the dense 1.3B model), we see neutral or worse performance compared to the dense 1.3B model.

Adding overall dropout (sweeping over $p_{\text{drop}} \in \{0.1, 0.2, 0.3\}$) significantly improves the performance of MoE-64 in both the 615M and 1.3B variants — For the 615M compute equivalent variant, Moe-64 with $p_{\text{drop}}$=0.1 outperforms dense 615M with dropout by +1.5 to +1.7 chrF++ points across all subsets of pairs. Importantly, when increasing the dropout from 0.0 to 0.1 for MoE 64, we see that the relative decline of -0.1 chrF++, changes into a relative improvement of +0.9 chrF++ for very low resource pairs translating out of English. For the 1.3B compute equivalent variant, we see +0.5 to +0.6 chrF++ improvement in the performance of high resource and low resource language pairs translating out of English, but no gains on translation into English or non-English pairs. This indicates that once we scale the computational cost per update, we see milder improvements on high-resource language pairs as well as low-resource pairs. We hypothesize two potential reasons for this: **(1)** we use a temperature of 1.0 for sampling, i.e., we do not upsample datasets from low-resource pairs. This preserved imbalance drives the 1.3B dense model to allocate capacity proportional to the resource level of each language pair. As a result, high-resource pairs will likely have enough capacity in the 1.3B dense model (given the size and nature of our ablation dataset) and will not benefit as much from the additional capacity of MoE models. **(2)** As we increase computational cost per update, the propensity for low or very low-resource pairs to overfit increases thus causing performance to deteriorate.

To further understand the training regimes of MoE models, we look at their learning curves in Figure 17. We observe in the case of `eng_Latn-kon_Latn`, a very low-resource pair, that the model continues to face significant overfitting when trained for 100k updates. This is unsurprising, as iterating over a small training set with large capacity causes overfitting. Training for more steps is important for high-resource pairs, but we want to avoid negatively affecting low-resource pairs in the process. In the next two sections, we discuss two strategies that address this issue and improve model performance on low-resource pairs.

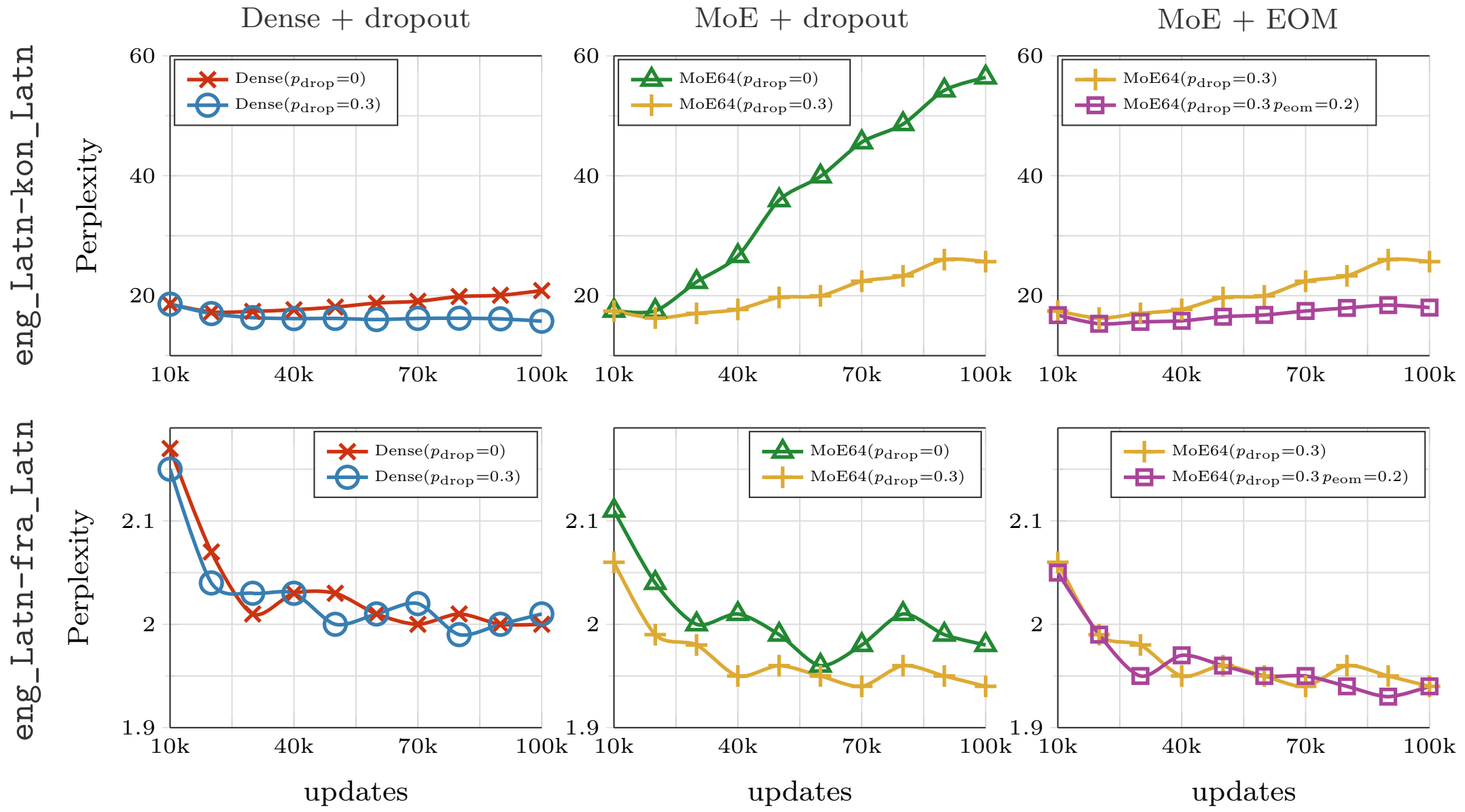

Figure 17: **Validation Perplexities with Various Dropout Strategies** for a low-resource direction (`eng_Latn-kon_Latn` in the top row) and a high-resource direction (`eng_Latn-fra_Latn` in the bottom row). Our proposed MoE Expert Output Masking (EOM) is an effective regularizer compared to overall dropout on `eng_Latn-kon_Latn`.

#### 6.2.2 Regularizing Massively Multilingual Mixtures of Experts

Although overall dropout is sufficient to regularize dense models, MoE models with overall dropout still significantly overfit on low-resource pairs as seen in Figure 17. To remedy this, we design and test different regularization strategies specific to MoE architectures. We describe each of these strategies and report results on our ablation dataset.

**MoE Expert Output Masking (EOM).** MoE models enable specialized expert capacity to be activated based on the input token. However, with increased capacity, the learned token-expert assignment can cause the models to overfit, especially on low-resource translation directions. In this proposed regularization strategy, we mask the *expert output* for a random fraction ($p_{\text{eom}}$) of the input tokens. For input tokens with dropped expert outputs, the first and/or second expert is effectively skipped. As illustrated in Figure 18b, we mask both experts for the first token ($x_1$ in red), we did not mask any of the expert outputs for the second token ($x_2$ in blue), and the last scenario is that of the last token ($x_3$ in green), where only one expert is masked. Note that although this masking will zero out some combination weights $\mathcal{G}_{t,e}$ in Equation (13), it will not affect the weights used in the load balancing loss in Equation (9).

We compare EOM to Gating Dropout (Liu et al., 2022), a strategy for reducing cross-machine communication in MoE layers which also has a regularizing effect. Gating Dropout skips the `All-to-All` communication between GPUs with probability $p_{gd}$, routing tokens to the local experts instead.

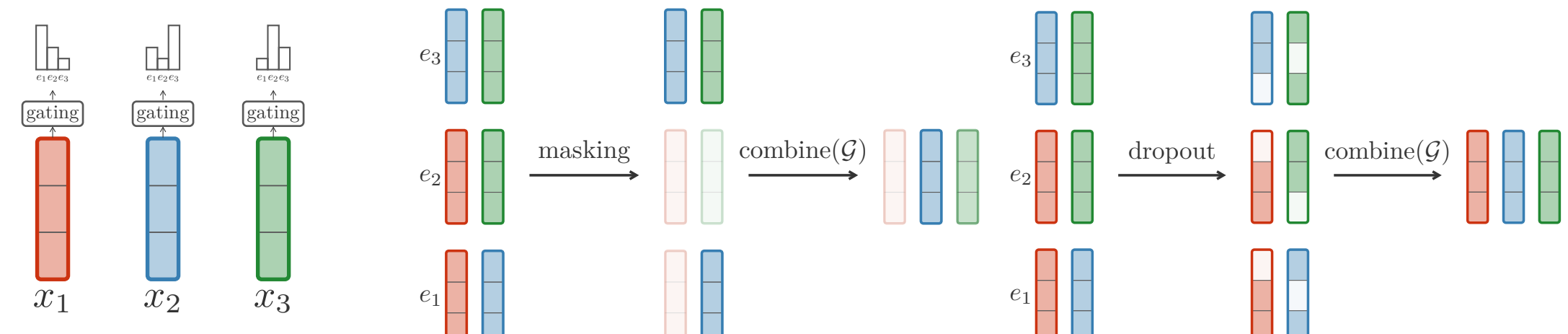

(a) Routing tokens (b) MoE Expert Output Masking (c) Overall dropout

Figure 18: **Illustration of MoE Expert Output Masking in contrast to overall dropout for MoE layers**: a color represents a token, and each token is dispatched to two experts (Top-k-Gating). Faded colors correspond to dropped units or masked outputs.

**Final Output Masking (FOM).** A simpler alternative to EOM would be to mask the combined expert output for a random fraction of tokens, i.e., after the last stage in Figure 18b. We denote with $p_{\text{fom}}$ the fraction of tokens masked with this regularization method. Note that this type of masking is more generic as it can be applied to dense models as well — in testing it here, we validate the advantages of using an MoE-specific masking i.e. MoE Expert Output Masking.

**Conditional MoE Routing (CMR).** Instead of randomly dropping a proportion of activations or masking expert outputs, in this section, we consider the option of letting the model decide, and learn, which tokens need the extra capacity or specialization of MoE layers, and which tokens are better routed to a limited-capacity shared layer.

Inspired by Zhang et al. (2021)'s CLSR-Gate, we design Conditional MoE Routing layers (CMR for short). As depicted in Figure 19, we augment MoE layers with a binary gate that decides the weights associated with two branches of the computational graph: **(1)** a shared dense FFN sublayer ($\text{FFN}_{\text{shared}}$) and **(2)** an MoE layer with its own $E$ expert FFN sublayers. For an input token $x_t$, the output of CMR is evaluated as follows:

$$g(x_t) = \text{sigmoid}(W_{\text{CMR}} \cdot x_t), \tag{10}$$

$$\text{CMR}(x_t) = (1 - g(x_t)) \cdot \text{FFN}_{\text{shared}}(x_t) + g(x_t) \cdot \text{MoE}(x_t), \tag{11}$$

where $W_{\text{CMR}}$ are the weights of the CMR's binary gate. Unlike Zhang et al. (2021)'s CLSR-Gate, our CMR branches are FFN sublayers (dense or sparsely gated MoE) and not linear projections. Furthermore, our CMR does not have language-specfic parameters, but learned routing to experts using an MoE layer.

The CMR gate weights are learned by optimizing translation accuracy under a budget constraint $p$. For a mini-batch with $T$ tokens, this amounts to adding the following auxiliary loss term ($L_{\text{CMR}}$) to the loss function:

$$L_{\text{CMR}} = \frac{1}{T} \cdot \sum_{t=1}^{T} |g(x_t) - p| \,. \tag{12}$$

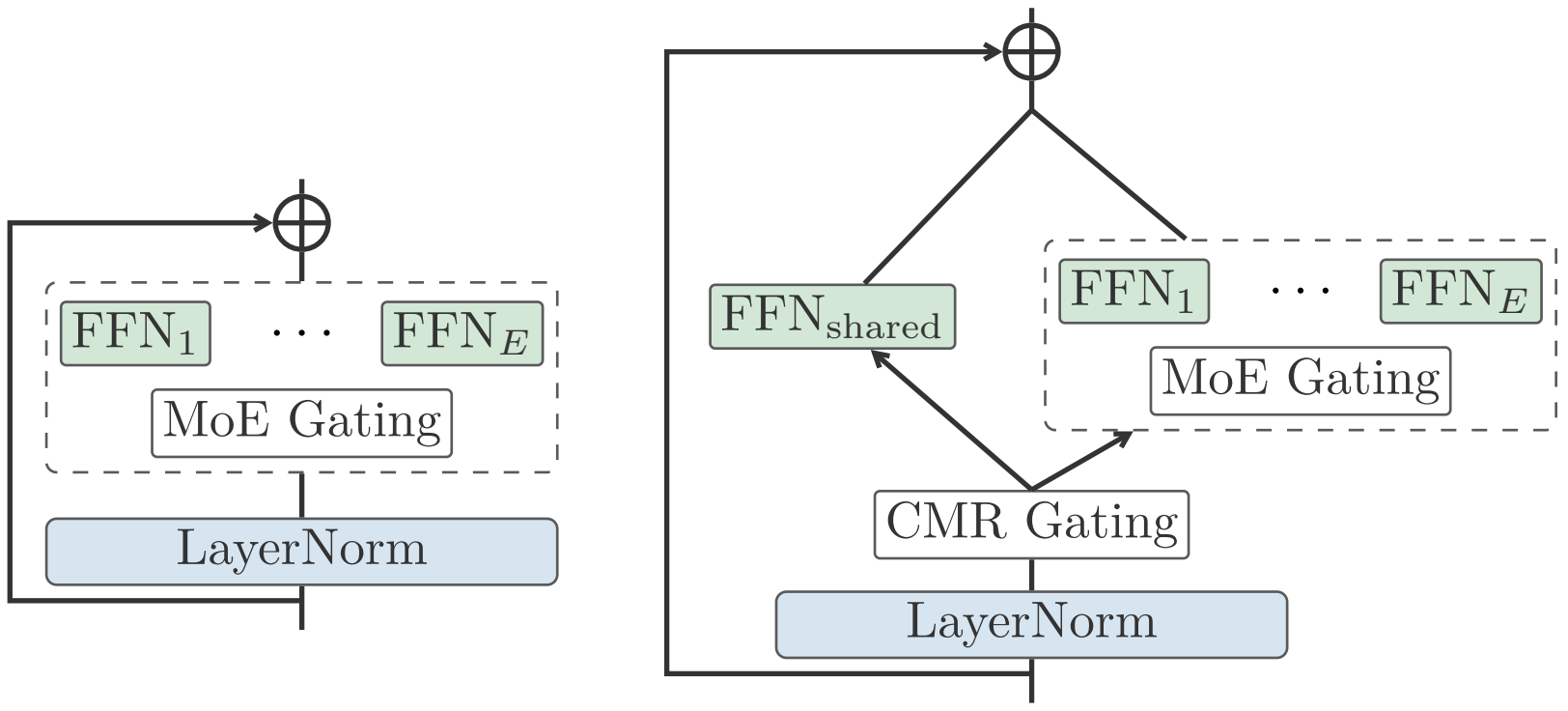

Figure 19: **Conditional MoE Routing (CMR)** - Residual block in a Transformer layer with regular MoE (left) *vs.* Conditional MoE Routing (right).

In Zhang et al. (2021), the budget $p$ controls language-specific capacity — here we use this budget constraint to limit the capacity induced by MoE layers; At $p$=0, the model is dense, practically pushing all tokens through $\text{FFN}_{\text{shared}}$, and at $p$=1, the model is free to always route tokens through the high-capacity MoE layer. This provides a regularizing effect that might help low-resource language pairs that are more likely to overfit with the large capacity of MoE layers. Initial experiments on other benchmarks show that the value of $p = 0.8$ achieves the desirable trade-off of improving the performance on low-resource pairs without hindering the performance on high-resource ones. An important ingredient in empirically helping CMR layers is zeroing out a fraction of the gates $g$ (Equation (11)) in the mini-batch; we denote this fraction with $p_{\text{cmr}}$. This means that we force $p_{\text{cmr}}$% tokens in the mini-batch to only take the route of $\text{FFN}_{\text{shared}}$. We note that CMR is also related to Rajbhandari et al. (2022), since both approaches combine the outputs of dense and MoE layers. The main difference is the absence of the auxiliary budget constraint and zeroing out a fraction of the gates $g$. From our initial experiments, we find that the addition of the auxiliary budget constraint $L_{\text{CMR}}$ and $p_{\text{cmr}}$ are important for improving accuracy.

**Experimental Setup.** Our baseline is a Sparsely Gated MoE model with model dimension 1024, FFN dimension 8192, 16 attention heads and 24 layers. We have 64 experts per MoE layer and we place an MoE layer in every alternate Transformer layer of the model ($f_{\text{MoE}}$=2). We compare this unregularized baseline to a variant with an overall dropout rate of 0.3 ($p_{\text{drop}}$=0.3), best performing after a sweep of $p_{\text{drop}} \in \{0.1, 0.2, 0.3\}$.

- For EOM, we sweep over the values of $(p_{\text{drop}}, p_{\text{eom}}) \in \{0.1, 0.2, 0.3\}^2$ and choose the best out of 9 variants based on the average chrF++ score on the validation set.
- For FOM, we set $p_{\text{drop}}$=0.3 and sweep over $p_{\text{fom}} \in \{0.1, 0.2, 0.3\}$ and choose the best out of the variants based on the average chrF++ score on the validation set.
- For CMR, we train the baseline augmented with a shared FFN sublayer with the same dimensions as the rest of the model. We sweep over the values of $(p_{\text{drop}}, p_{\text{cmr}}) \in \{0.1, 0.2, 0.3\}^2$ and choose the best out of 9 variants based on the average chrF++ score on the validation set.

| | eng_Latn-xx | | | | xx-eng_Latn | | | | xx-yy |
|---|---|---|---|---|---|---|---|---|---|
| | all | high | low | v.low | all | high | low | v.low | all |
| MoE-64 $p_{\text{drop}}$=0 | 43.3 | 55.9 | 38.2 | 29.6 | 52.9 | 63.9 | 48.4 | 43.7 | 39.3 |
| MoE-64 $p_{\text{drop}}$=0.3 † | 44.3 | 56.0 | 39.5 | 32.5 | 54.4 | 63.9 | 50.6 | 47.7 | 41.9 |
| MoE-64 FOM ($p_{\text{drop}}$=0.3, $p_{\text{fom}}$=0.1) † | 43.8 | 55.6 | 38.9 | 32.5 | 54.8 | 64.3 | 50.9 | **48.5** | 42.0 |
| MoE-64 EOM ($p_{\text{drop}}$=0.3, $p_{\text{eom}}$=0.1) † | **44.7** | **55.9** | **40.1** | **33.4** | 54.8 | 64.3 | 51.0 | 48.3 | **42.5** |
| MoE-64 CMR ($p_{\text{drop}}$=0.2, $p_{\text{cmr}}$=0.2) † | **44.8** | **56.1** | **40.2** | **33.4** | **55.2** | **64.6** | **51.4** | **48.5** | **42.6** |
| Gating Dropout($p_{\text{drop}}$=0.3, $p_{gd}$=0.2) † | 44.4 | 55.7 | 39.8 | 33.0 | 54.8 | 64.1 | 51.0 | **48.5** | 42.3 |

Table 16: **Comparison of Various Regularization Strategies**. We find that EOM and CMR strategies perform similarly, and both outperform the baseline MoE with overall dropout. †: best of sweep.

- For Gating Dropout (Liu et al., 2022), we sweep over the values of $(p_{\text{drop}}, p_{gd}) \in \{0.1, 0.2, 0.3\}^2$ and choose the best out of 9 variants based on the average chrF++ score on the validation set.

For each model, we report chrF++ averages on the validation set in 3 groups of directions: `eng_Latn-xx`, `xx-eng_Latn` and `xx-yy`, broken down w.r.t. to resource levels: high, low and very low (v.low) for `eng_Latn-xx` and `xx-eng_Latn`.

**Results.** In terms of alleviating the overfitting issue, the last column of Figure 17 shows that EOM leads to better regularization and less overfitting on low-resource tasks compared to overall dropout. In terms of translation quality, and as shown in Table 16, we observe gains of +0.4 chrF++ across all pairs into English and +0.6 chrF++ across non-English pairs for MoE EOM compared to vanilla MoE with overall dropout. Gains are larger on low and very low-resource languages — for out of English, there are improvements of +0.6 and 0.9 chrF++ with EOM.

The comparison between EOM and FOM proves that masking before combining the expert outputs is more beneficial than simply masking tokens in the final output. Our hypothesis is that this gain in performance stems from EOM strengthening the residual connection surrounding the MoE layer and reducing co-adaptation between selected top-2 experts, as well as co-adaptation between experts and the subsequent layers of the model.

We find that Gating Dropout performs better than vanilla MoE with overall dropout, but is outperformed by both EOM and CMR. For CMR, we see +0.8 chrF++ across all pairs into English and +0.7 chrF++ across non-English pairs. Similarly, the improvements are larger for low and very low-resource languages, with +0.7 and +1.0 chrF++ respectively.

These results demonstrate that both EOM and CMR strategies help improve on top of vanilla MoE with overall dropout. CMR is computationally more expensive by 18-25% at training time because of the additional shared FFN layer at the level of each MoE layer in the model. Given the additional computational overhead, we use the simpler MoE EOM strategy.

### 6.2.3 Curriculum Learning

To reduce overfitting on low-resource language pairs further, we next explore alternative means of adding additional regularization. We try a straightforward curriculum of introducing these low-resource language pairs in phases during model training. The language pairs that empirically overfit within K updates are introduced K updates before the end of training. This allows language pairs that tend to overfit with too many training updates to avoid overfitting, while allowing language pairs that benefit from additional training updates to continue training. Prior work explored, for instance, curriculum design for multilingual models based on data types (Kuwanto et al., 2021).

**Experimental Setup.** We train MoE-64 models with $f_{\text{MoE}}$=2 for a total of $T$ updates (see Section 6.2.1 for details of the base architecture). To derive the phases of the curriculum, we first train a regular model, i.e., without curriculum, then we partition language pairs into $n$ buckets $\{b_0, b_1, \ldots, b_{n-1}\}$ based on when they start to overfit. In the phased curriculum training, we introduce each bucket $b_i$ after exactly $T - k_i$ updates, where $k_i$ is the median number of updates after which all directions in bucket $b_i$ start to overfit. Based on observed overfitting patterns, we introduce pairs during training in $n = 3$ phases - we set $T = 100k$, $k_0 = 100k$, $k_1 = 40k$, $k_2 = 20k$, so $b_0$ is introduced first, $b_1$ is introduced at step $T - 40k$, and $b_2$ is introduced at step $T - 20k$. We compare to the baseline of an MoE model with overall dropout 0.3 without curriculum learning, i.e. introducing all pairs at the start of training and training for 100k updates.

**Results.** As shown in Table 17, for vanilla MoE, when translating out of English (`eng_Latn-xx`), there is an average improvement of +0.6 chrF++ on low-resource directions (low) and a +0.8 chrF++ improvement on very low-resource (v.low) directions. There is, however, no significant improvement on high-resource directions (high) or translation into English (`xx-eng_Latn`), most likely because there is no overfitting on these directions in the baseline. For MoE EOM, training with a curriculum actually hurts performance across `eng_Latn-xx`, `xx-eng_Latn` and `xx-yy`. Our analysis indicates that overfitting on the ablation dataset is already reduced by EOM, which has a higher fraction of language pairs with their best checkpoint by validation perplexity at $\geq$ 70000 steps. We hypothesize that adding a curriculum on top of EOM is not needed for the ablation dataset. However, results in Table 29 show that with the full dataset and larger model, combining curriculum learning and EOM improves performance, especially on low and very low-resource language pairs.

### 6.2.4 Analysis of Multilingual Sparsely Gated MoE Models.

MoE theoretically enables models to specialize expert capacity for different tasks, but what do these models actually learn? We now take a closer look at the routing of tokens to experts in MoE layers at different points of the encoder-decoder architecture. We take an MoE model ($E$=64, $p_{\text{drop}}$=0.3, $p_{\text{eom}}$=0.2) trained on our ablation dataset and do a forward pass on Flores-200 `dev` set data in teacher-forcing mode, i.e., we feed the true target prefix to predict the next target token. For each task (language pair), we log the routing decisions prior to Top-k-Gating, and depending on whether it is an encoder layer or a decoder layer, we average the routing vectors across multiple language pairs to estimate language-level routing vectors.

| | eng_Latn-xx | | | | xx-eng_Latn | | | | xx-yy |
|---|---|---|---|---|---|---|---|---|---|
| | all | high | low | v.low | all | high | low | v.low | all |
| MoE-64 ($p_{\text{drop}}$=0.3) | 44.3 | 56.0 | 39.5 | 32.5 | 54.4 | 63.9 | 50.6 | 47.7 | 41.9 |
| + Curriculum Learning (3 Phases) | 44.7 | 56.0 | 40.1 | 33.3 | 54.6 | 64.2 | 50.8 | 47.9 | 42.2 |
| $\Delta$ | +0.4 | +0 | +0.6 | +0.8 | +0.2 | +0.3 | +0.2 | +0.2 | +0.3 |
| MoE-64 EOM ($p_{\text{drop}}$=0.3, $p_{\text{eom}}$=0.1) | 44.7 | 55.9 | 40.1 | 33.4 | 54.8 | 64.3 | 51.0 | 48.3 | 42.5 |
| +Curriculum Learning (3 Phases) | 44.26 | 55.68 | 39.62 | 33.07 | 54.67 | 63.87 | 50.93 | 48.44 | 42.15 |
| $\Delta$ | -0.44 | -0.22 | -0.48 | -0.33 | -0.13 | -0.43 | -0.07 | +0.14 | -0.35 |

Table 17: **Curriculum Learning Results** demonstrate that for vanilla MoE, training on a curriculum reduces overfitting, particularly for `eng_Latn-xx` low and very low resource pairs. For MoE EOM, a curriculum does not help.

$$G_{<\text{lang}>} = \frac{1}{|\mathcal{T}_{<\text{lang}>}|} \sum_{x_t \in \mathcal{T}_{<\text{lang}>}} G_t, \tag{13}$$

where $\mathcal{T}_{<\text{lang}>}$ is the set of all tokens in <lang>, source-side for encoder layers and target-side for decoder layers. We plot in Figure 20 the cosine similarity scores between all 53 languages of the ablation dataset, in the first and last encoder MoE layer, and the first and last decoder MoE layer.

The similarity heatmaps demonstrate that in late decoder layers (see Figure 20d), the languages are being separated, i.e., dispatched to different set of experts. Languages within the same family are highly similar in their choice of experts, i.e., the late decoder MoE layers are language-specific. This is particularly the case for languages in the Atlantic-Congo family (the rows/columns from `cjk` to `yor`) and some pairs like {`snd_Arab`, `urd_Arab`} in the Indo-European family or {`yue_Hant`, `zho_Hans`} in the Sino-Tibetan family. To a lesser extent, the early encoder MoE layers (see Figure 20a), also show some language-expert specialization. The late encoder MoE layers and the early decoder MoE layers (see Figure 20b and Figure 20c) seem to be language-agnostic.

In Figure 21, we visualize the vectors of expert-distribution per language $G_{<\text{lang}>}$ (Equation (13)) using UMAP (McInnes et al., 2018). The first row displays languages by language family and the second row displays languages by script. Separation of languages is more discernible in the decoder's last layer (last column of Figure 21) particularly along language family lines, e.g., Atlantic-Congo in green and Dravidian in pink.

### 6.3 Self-Supervision Strategies on Large-scale Monolingual Corpora

For low-resource languages, there is generally limited or no bitext data available. In cases where bitext data is publicly available, the domain of the available data could be narrow (e.g. religious texts) or the bitext data could be noisy. In comparison, there is relatively more abundant monolingual data available in low-resource languages. In addition to bitext mining detailed in Section 5.3, another way of leveraging this monolingual data is via incorporating a self-supervised task into the process of training a multilingual machine translation model (Bapna et al., 2022; Chi et al., 2021; Ma et al., 2021). Self-supervised

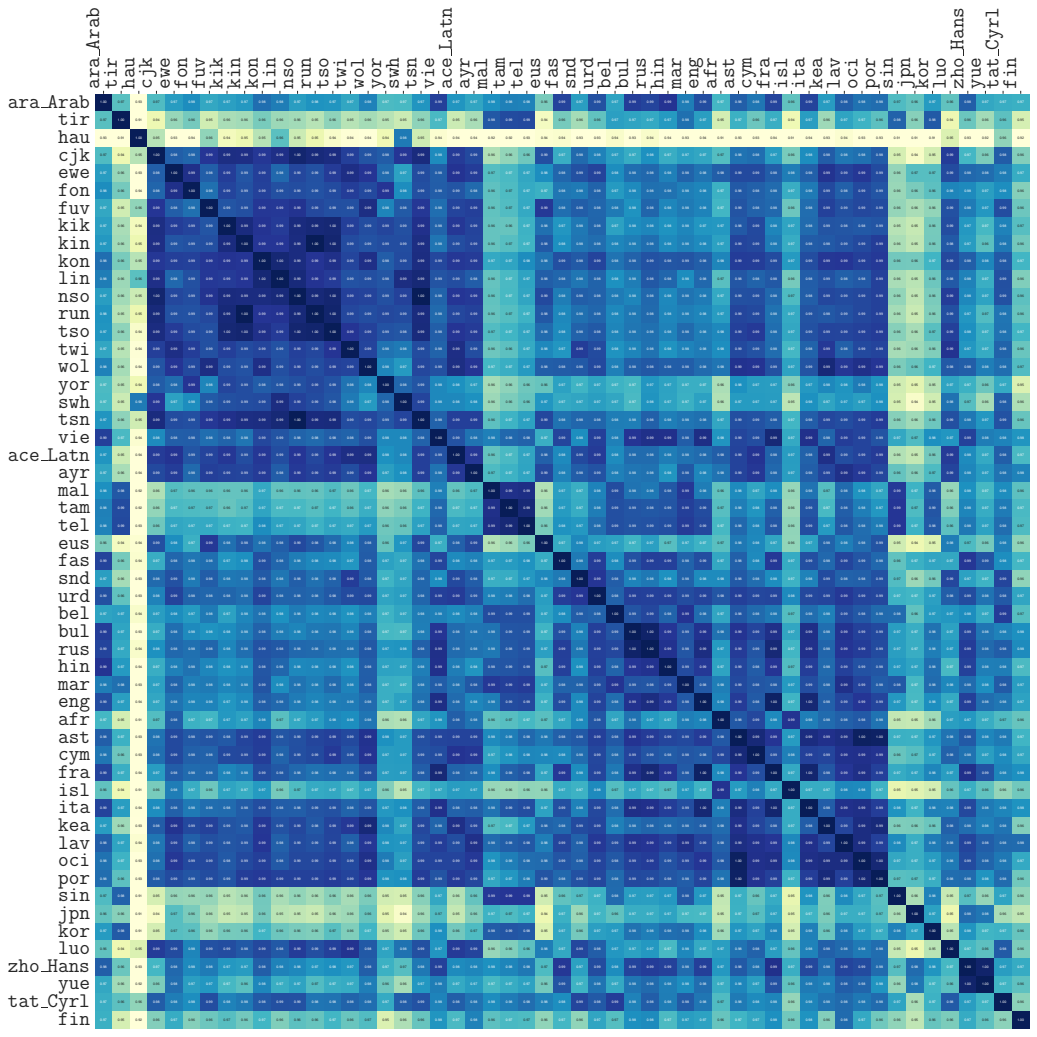

(a) First encoder MoE layer

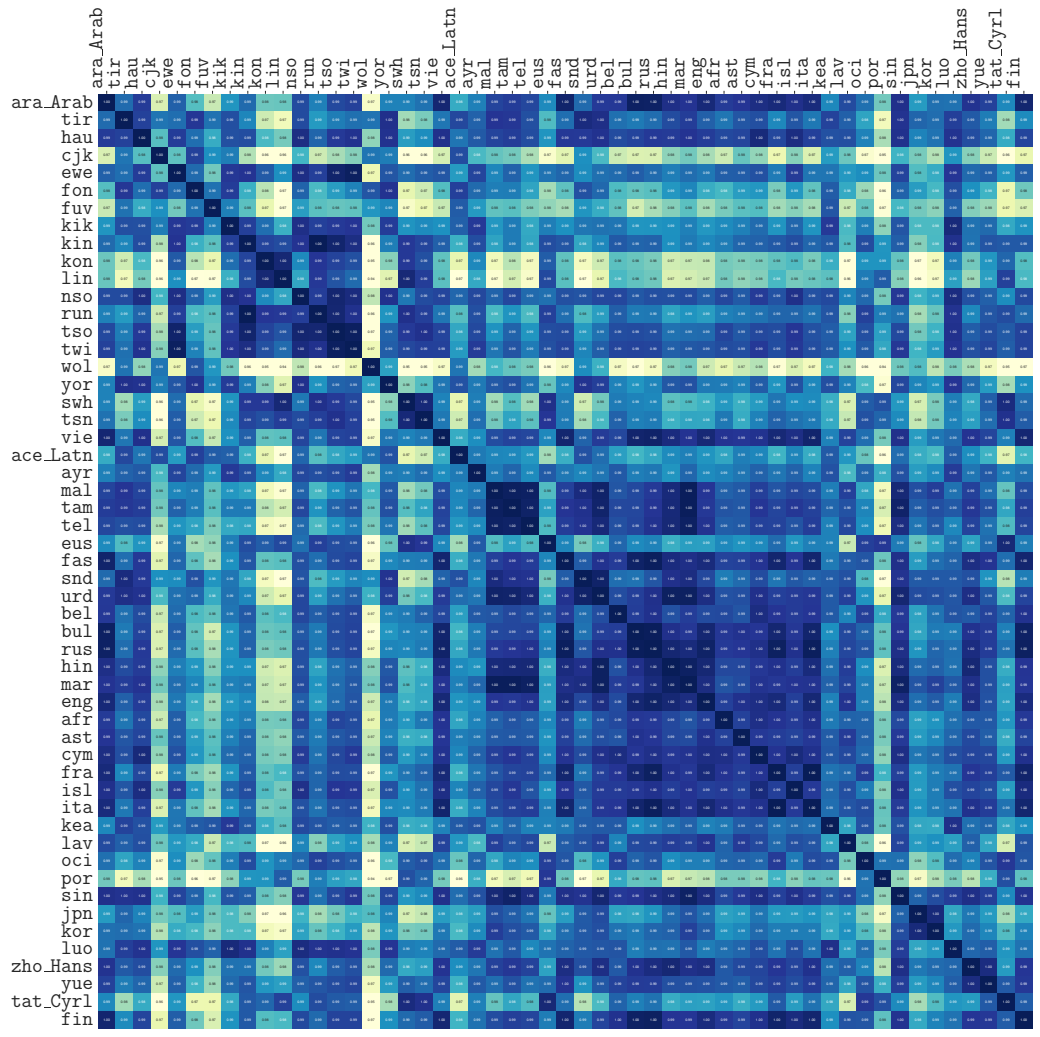

(b) Last encoder MoE layer

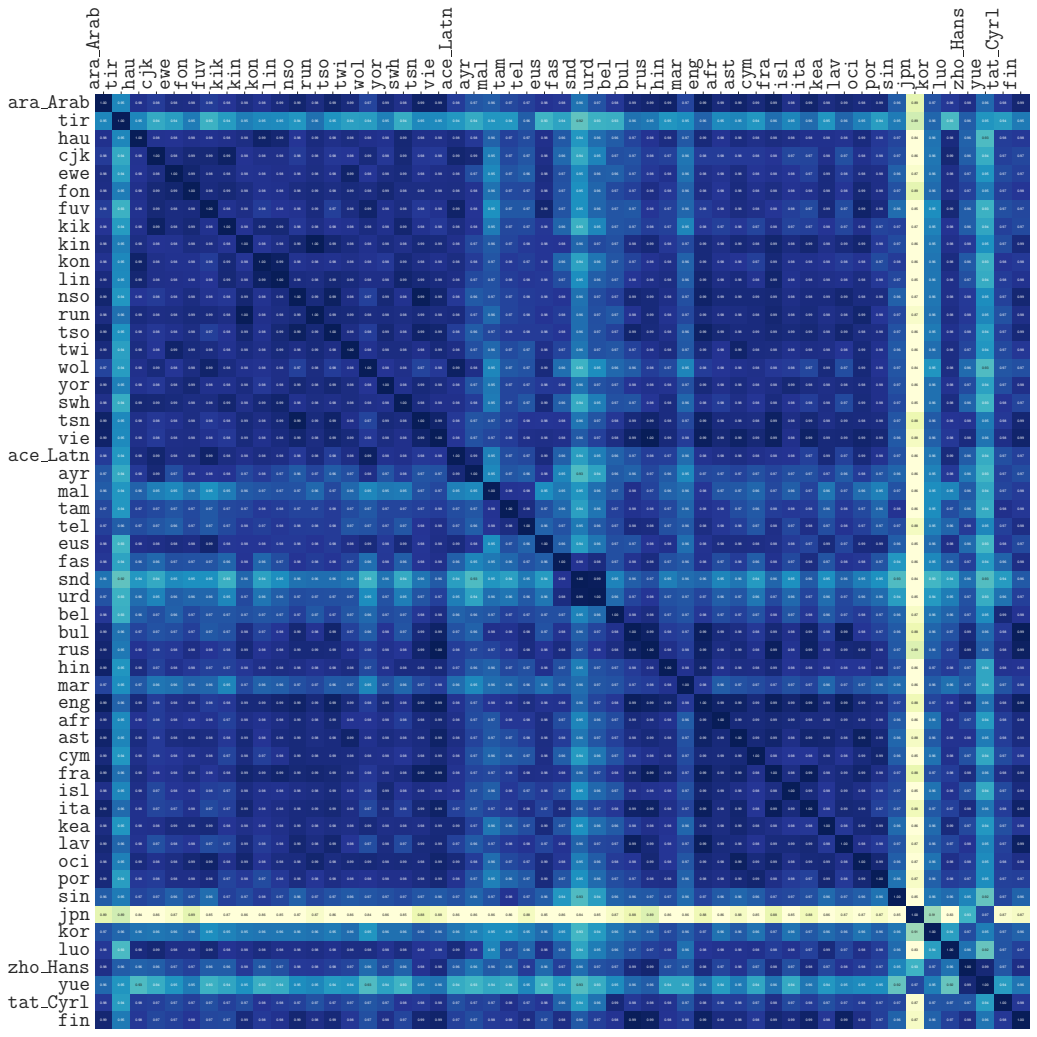

(c) First decoder MoE layer

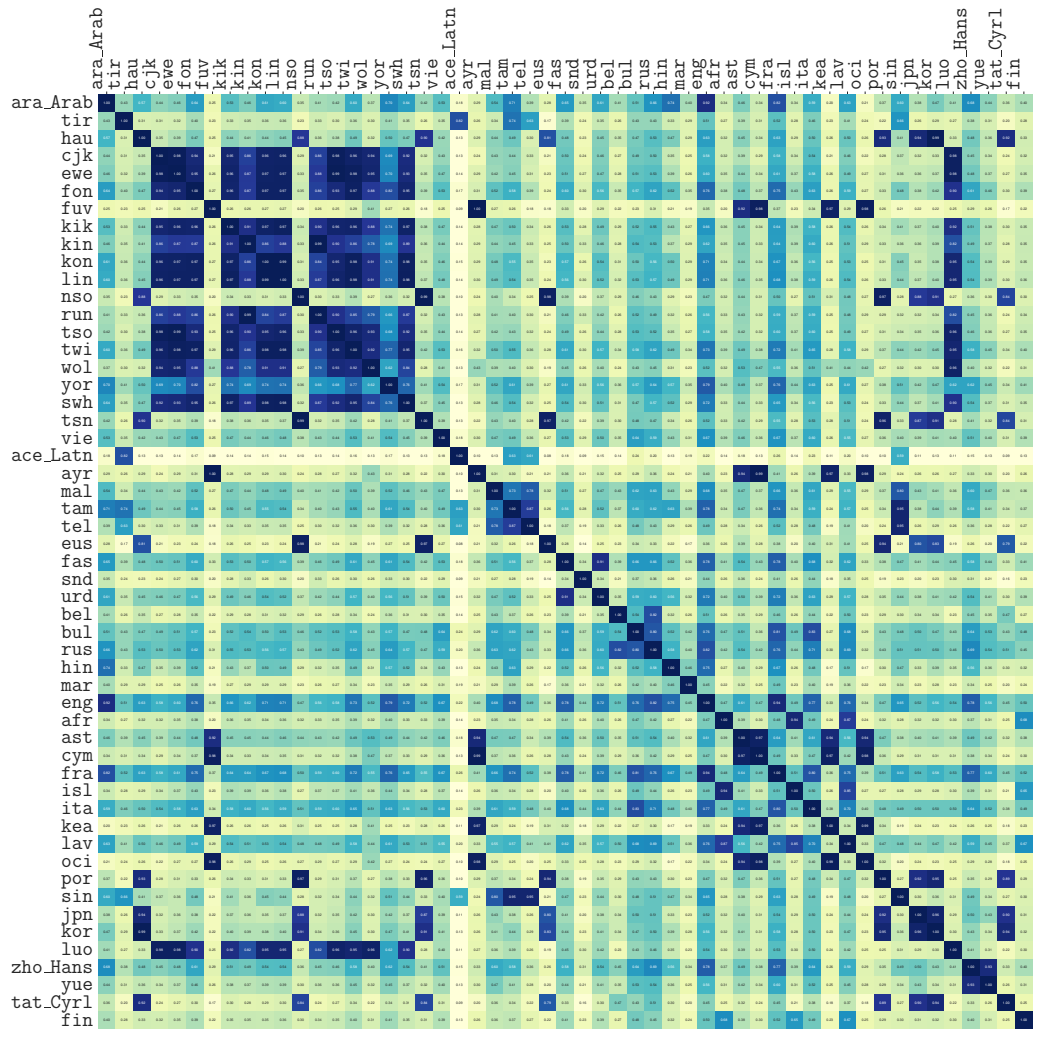

(d) Last decoder MoE layer

Figure 20: **Cosine Similarity Scores** between languages of the ablation dataset at different layers of the encoder-decoder architecture. The similarity is measured w.r.t. the gating decisions (expert choice) per language (source-side in the encoder and target-side in the decoder)

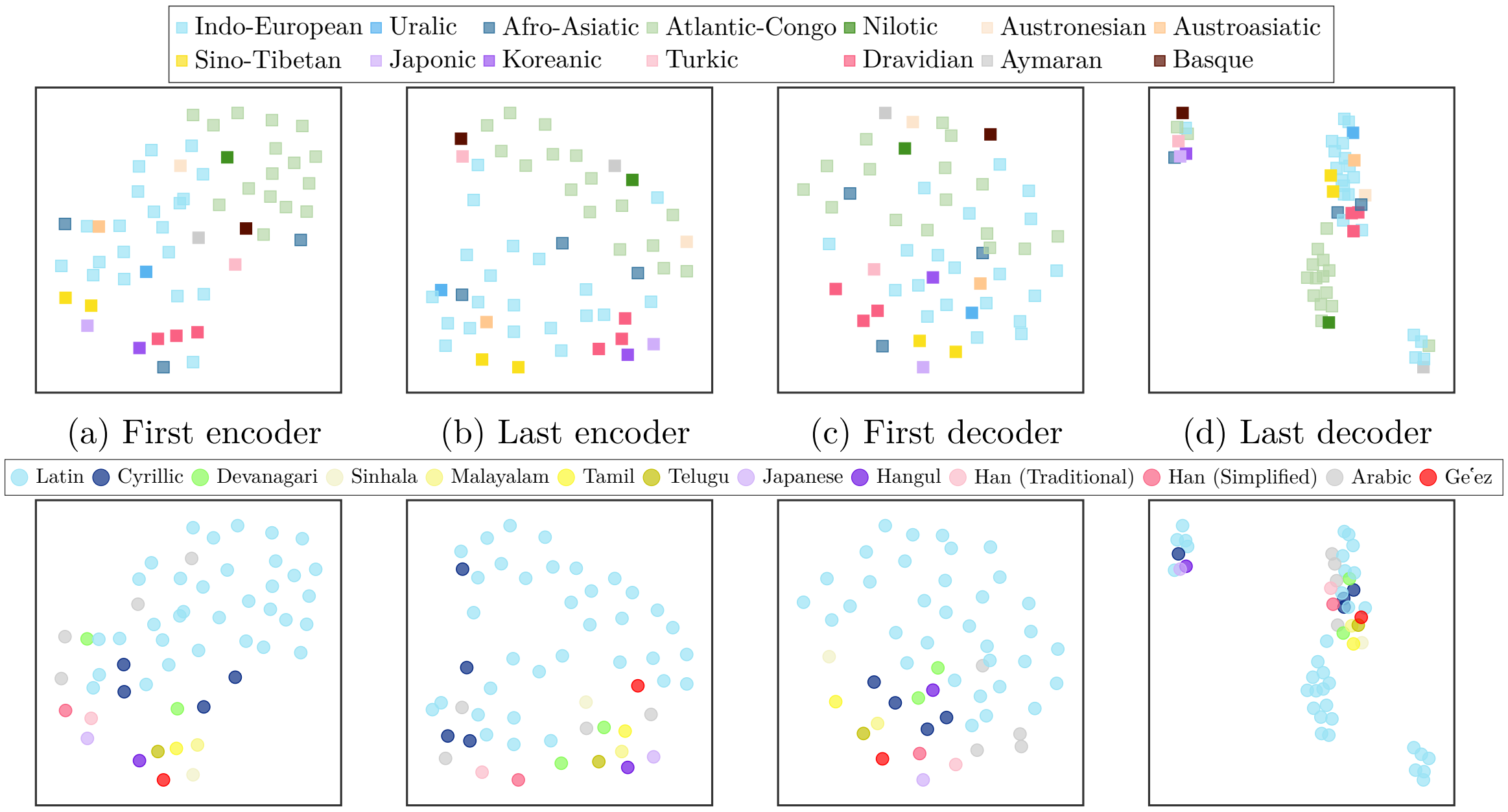

(a) First encoder (b) Last encoder (c) First decoder (d) Last decoder

(e) First encoder (f) Last encoder (g) First decoder (h) Last decoder

Figure 21: **UMAP Embeddings** of the languages in the ablation dataset. We color in the first row according to language family and in the second row according to script.

learning can learn patterns and constructs of a language from monolingual text. Hence, learning with self-supervised objectives on this additional data could help improve the performance of multilingual machine translation models.

We first study the effect of the choice of the self-supervised task: **(1)** language modeling objective, **(2)** denoising autoencoder objective (Liu et al., 2020), or **(3)** the combination of both. We follow that up with studying the optimal curriculum when combining the task of self-supervised learning (SSL) on monolingual data along with the task of training on multilingual machine translation (MMT) on bitext data. We use the recommendations from this study to decide on our self-supervision strategy for our final model on 200 languages.

#### 6.3.1 Incorporating Self-Supervised Objectives with Multilingual Machine Translation

There are different ways of incorporating self-supervision on monolingual data in the training procedure of multilingual machine translation models. Traditionally, self-supervised learning in NLP takes the form of pretraining with a self-supervised objective on monolingual data, followed by finetuning on the task-specific supervised data. Another strategy is to consider SSL on monolingual data and MMT on bitext data as separate tasks in a multi-task learning setup, where examples from both tasks are present in every batch during training. Our third strategy is a combination of the first two strategies. We can first pretrain with the multi-task setting of self-supervision on monolingual data and multilingual machine translation on bitext data, followed by finetuning on multilingual machine translation alone.

### 6.3.2 Self-Supervised Learning Objectives

There are different self-supervised objectives we can use on monolingual data.

**Denoising Autoencoder (DAE).** We follow Liu et al. (2020) and use a Transformer encoder-decoder architecture which is the same architecture used for multilingual machine translation. We set the target to be a sentence from the monolingual corpus. We set the source to be a noised version of the target monolingual sentence. We mask random spans of text from the target sentence. Rather than always replacing the masked token with the mask token (`<mask>`), with a specified probability, we replace the masked token with a random token from the vocabulary. The SSL objective here is to maximize the likelihood of predicting the target given the source which is a noised version of the target.

**Causal Language Modeling (LM).** Past work has shown some success with initializing components of a machine translation model with a pretrained decoder Transformer language model. So, we set up the language modeling task in a `seq2seq` setting where the source is empty and the target is a sentence from a monolingual corpus. The SSL objective here is to maximize the likelihood of predicting the target i.e. the causal language modeling objective.

### 6.3.3 Effect of Curriculum of Self-Supervision combined with Multilingual Machine Translation

In this section, we study the impact of our choice of curriculum when combining the Denoising Autoencoder(DAE) objective with the usual MMT task.

**Experimental Setup.** In our setup, we use the multilingual denoising autoencoder objective for self-supervision on the multilingual text corpus for languages in our ablation benchmark. Our baseline is a dense 1.3B model trained for 100K updates on the MMT task with a sampling temperature of 1.0. We train the SSL task variants with the same architecture and sampling temperature for 200K updates each to ensure that SSL task variants are exposed to as much multilingual parallel data as the baseline. We hope to find the best variant that improves the performance on low-resource pairs which do not have sufficient parallel bitext data but relatively abundant monolingual data. We consider three options:

- Pretraining on SSL, followed by finetuning on MMT (DAE⇒MMT). We train on SSL for the first 100K updates, followed by finetuning on MMT task for the next 100K updates. We use a sampling temperature of 1.25 for SSL pretraining.
- Multitask training on SSL and MMT (DAE+MMT). We combine the training examples for SSL and MMT and train the model for 200K updates on a combination of both tasks.
- Multitask training on SSL and MMT, followed by finetuning on MMT (DAE+MMT ⇒ MMT). We first perform multitask training on SSL and MMT for 150K updates, followed by finetuning on MMT only for 50K updates

**Results.** First, we see that pretraining on SSL, followed by finetuning on MMT (DAE ⇒ MMT) hurts performance (-0.7 chrF++ on `eng_Latn-xx`, -1.2 chrF++ on `xx-eng_Latn`) across all subsets of pairs compared to a baseline of training on MMT alone. This indicates that there is a fundamental mismatch in the two training tasks and it is hard for finetuning to recover from the initial state at the end of pretraining. Multi-task training on SSL and MMT (DAE+MMT) shows +0.4 chrF++ on low-resource eng_Latn-xx pairs and +1.1 chrF++ on very low-resource eng_Latn-xx pairs, thus confirming that the additional monolingual data in the low-resource target languages is useful when presented in a multitask framework. Similarly, we see +1.3 and +1.9 chrF++ improvements on low and very low-resource pairs in xx-eng_Latn directions as well as +1.1 chrF++ on xx-yy pairs. We hypothesize that we observe stronger performance translating into English as there is an abundance of very high-quality English monolingual data which is critical for SSL. Finally, we wish to confirm whether there are any benefits to pausing multitask training after the first K updates, and finetuning on MMT only (DAE+MMT⇒MMT). We see that this performs on par with multitask training on SSL and MMT (DAE+MMT) without finetuning on MMT only. This indicates that when SSL and MMT are jointly trained in a multi-task setup, they do not suffer from interference that could be countered by the final stage of finetuning purely on the final task (MMT).

| | eng_Latn-xx | | | | xx-eng_Latn | | | | xx-yy |
|---|---|---|---|---|---|---|---|---|---|
| | all | high | low | v.low | all | high | low | v.low | all |
| MMT | 43.3 | **55.4** | 38.4 | 31.6 | 53.5 | **63.6** | 49.4 | 46.5 | 41.3 |
| DAE⇒MMT | 42.6 | 55.0 | 37.6 | 30.8 | 52.3 | 62.2 | 48.3 | 45.4 | 40.4 |
| DAE+MMT | **43.5** | 55.2 | **38.8** | **32.7** | **54.4** | **63.6** | **50.7** | **48.4** | **42.4** |
| DAE+MMT⇒MMT | 43.4 | **55.4** | 38.5 | 32.2 | 54.3 | **63.6** | 50.5 | 48.0 | 42.2 |

Table 18: **Effect of SSL Curriculum.** We find DAE+MMT training jointly to be the most effective strategy.

#### 6.3.4 Effect of Self-Supervision Objectives

In this section, we compare the impact of choosing different self-supervised objectives.

**Experimental Setup.** We train all SSL task variants in this section in a multitask learning setup of training SSL and MMT tasks together, because that is the best performing curriculum as demonstrated in Section 6.3.3. Our baseline is a dense 1.3B model trained for 100K updates and SSL models trained for 200K updates. We train the SSL variants with the same architecture and sampling temperature for 200K updates each to ensure that SSL variants are exposed to as much supervised bitext data as the baseline. We hope to find the best self-supervised objective that improves the performance on low-resource pairs which do not have enough bitext data, and relatively abundant monolingual data. We consider three options. **(1)** Denoising Autoencoder (DAE), **(2)** Causal Language Modeling (LM), and **(3)** the combination of both (DAE+LM).

**Results.** First, we observe that the LM task as an SSL objective results in a decline in performance compared to the baseline of training MMT alone. One hypothesis is that self-supervision on the encoder-decoder attention plays an important role, and this is not present in the LM objective that focuses on training the decoder to generate the text in the monolingual corpus. Next, as also demonstrated in Section 6.3.3, we see significant improvements when using the DAE objective. Finally, we explore whether there are complementary gains of combining DAE along with the LM task. The results show a decline in performance compared to using only the DAE task. This suggests that there might be some interference between the different tasks, which reduces the overall performance when combining them.

| | eng_Latn-xx | | | | xx-eng_Latn | | | | xx-yy |
|---|---|---|---|---|---|---|---|---|---|
| | all | high | low | v.low | all | high | low | v.low | all |
| MMT | 43.3 | **55.4** | 38.4 | 31.6 | 53.5 | **63.6** | 49.4 | 46.5 | 41.3 |
| MMT+LM | 42.6 | 54.9 | 37.5 | 30.8 | 53.5 | **63.6** | 49.4 | 46.7 | 41.5 |
| MMT+DAE | **43.5** | 55.2 | **38.8** | **32.7** | **54.4** | **63.6** | **50.7** | **48.4** | **42.4** |
| MMT+DAE+LM | 42.6 | 55.0 | 37.6 | 31.4 | 53.4 | 62.7 | 49.6 | 47.0 | 40.8 |

Table 19: **Effect of Different SSL Objectives.** We find training MMT+DAE the most effective compared to adding the LM task.

#### 6.3.5 Discussion

Recent works (Bapna et al., 2022; Chi et al., 2021; Ma et al., 2021) have demonstrated that denoising and similar self-supervised objectives are very useful for improving model performance when trained along with machine translation task in a multitask setup. In our work, we try two SSL objectives, DAE and LM and experimented with different combinations of both along with the MMT task. We observed that only DAE performs best when trained with MMT. Benefits of the LM task in a multitask setup with MMT is still not evident and future work could reveal a deeper understanding regarding this finding. We also study different curriculum learning strategies with the SSL tasks and find that multitask learning of DAE and MMT is usually the best setup, similar to findings in (Chi et al., 2021; Ma et al., 2021). Section 8.1.2 further discusses how SSL+MMT multitask training improves model performance for generating higher quality *backtranslated* data. Self-supervised learning is a powerful technique for optimally utilizing monolingual data and there is scope to study and design better SSL objectives for this. In addition, designing curricula to combine these objectives in a multitask framework is also an interesting direction for future research.

### 6.4 Data Augmentation

Another way of leveraging monolingual data is through *backtranslation* (Edunov et al., 2018; Sennrich et al., 2016a), a technique which involves creating parallel corpora that are noisy on the source side via machine translation. However, when it comes to low-resource languages, the machine translation models that are used to generate backtranslation data are often not good enough, and hence the generated data is often noisy and degenerate. Hoang et al. (2018) proposed *iterative backtranslation* to offset this, as better models are used to generate backtranslations in every iteration. However, this is an expensive endeavor. Since the aim is to use the best possible models to generate backtranslated data, this means using massively multilingual models, which are computationally intensive to run both at training and inference time. In this section, we discuss how we can effectively generate high-quality backtranslation data for low-resource languages, using a single iteration to be computationally efficient.

#### 6.4.1 Different sources of data

Backtranslation is a source of synthetically augmented data for translation models. We contrast it with two other sources of data: **(1)** *primary* bitext data, i.e. high-quality parallel corpora that have been human-translated, and **(2)** mined bitext data, i.e. parallel corpora obtained via mining (see Section 5.3). Backtranslation relies on the availability of an initial translation system, a *teacher*, to produce a noisy parallel corpus from a monolingual corpus. A natural choice for a teacher system is to use a transformer-based model similar to our previous approaches. However, these neural systems are often thought to be data-inefficient when compared with traditional phrase or rule based statistical machine translation (Koehn and Knowles, 2017) models. In light of this, we also experiment with using a traditional phrase-based statistical translation model as a teacher for backtranslation (Schwenk, 2008). We will therefore distinguish four different sources of data in the following experiments, one primary and three augmented sources: human-translated data; mined data; backtranslated data via multilingual neural machine translation; and *backtranslated data via statistical machine translation.* For brevity and further use, we show the characteristics of these data sources in Table 20 and use the following abbreviated terms for the data sources:

- NLLB-Seed: our professionally-translated seed datasets as described in Section 4.2.
- PublicBitext: publicly available parallel corpora. These datasets may be human-translated but are often automatically aligned.
- Primary: the combination of the above two sources.
- Mined: mined data as described in Section 5.3.
- MmtBT: backtranslations obtained via a 1.3B-parameter dense multilingual neural model.
- SmtBT: backtranslations obtained via a series of bilingual MOSES models (Koehn et al., 2007) trained on Primary and Mined data. The optimal model hyperparameters were chosen via Flores-200 validation data.

**Experimental Setup.** First, we study the effect of using MmtBT in addition to Primary and Mined data. We use a dense 1.3B model and train on two data setups, namely Primary and Primary+Mined. We then use the model trained on the Primary+Mined dataset and generate backtranslation data for all the English-centric pairs in the dataset. This backtranslated data becomes our MmtBT. In the next experiment, we use the exact same model setup and train on a dataset comprising Primary+Mined+MmtBT. Our objective here is to observe the benefits of backtranslation over Primary+Mined. We train all the models for 200,000 updates and compare the best checkpoints on the Flores-200 development set using chrF++.

| Source | Human Aligned? | Noisy? | Limited Size? | Model-Dependent? | Models Used |
|---|---|---|---|---|---|
| NLLB-Seed | ✓ | ✗ | ✓ | ✗ | — |
| PublicBitext | ✗ | ✓ | ✓ | ✗ | — |
| Mined | ✗ | ✓ | ✗ | ✓ | Sentence Encoders |
| MmtBT | ✗ | ✓ | ✗ | ✓ | Multilingual |
| SmtBT | ✗ | ✓ | ✗ | ✓ | Bilingual MOSES |
| *Ideal Data* | ✓ | ✗ | ✗ | ✗ | — |

Table 20: **Dataset Characteristics** of the sources we compare in this section. Of these datasets, NLLB-Seed is by far the smallest. For low-resource languages, PublicBitext is often extremely limited. Mined, MmtBT, and SmtBT are limited only by the amount of available monolingual data and the quality of the models used to produce them.

| | eng_Latn-xx | | | | xx-eng_Latn | | | | xx-yy |
|---|---|---|---|---|---|---|---|---|---|
| | all | high | low | v.low | all | high | low | v.low | all |
| Primary | 41.0 | 52.8 | 36.3 | 28.1 | 47.4 | 60.5 | 42.1 | 36.7 | 39.2 |
| +Mined | 43.8 | 55.2 | 39.2 | **34.0** | 53.9 | 64.4 | 49.6 | 46.1 | 40.9 |
| +MmtBT | 44.0 | 55.1 | 39.5 | **34.0** | 55.7 | 64.8 | 52.0 | 50.8 | 40.6 |
| +SmtBT | **44.2** | **55.5** | **39.6** | **34.0** | **55.9** | **64.9** | **52.2** | **50.9** | **41.1** |

Table 21: **Comparison of aggregate Model Performance trained on Different Data Combinations**, evaluated on Flores-200 dev for ablation dataset directions. We observe that adding SmtBT data improves over the +Mined+MmtBT and overall gives the best performance across all language directions and resource level types.

In the second set of experiments, we try to understand the benefits of adding an additional source of backtranslated data, SmtBT. For this, we train bilingual statistical machine translation (SMT) models on Primary+Mined bitexts. We compare the performance of these models against the multilingual machine translation (MMT) model trained on the same data, and pick the directions where the SMT models are either better or comparable to the MMT models. For the directions we pick, we generate backtranslation data using the SMT *teacher* models. This gives us the SmtBT dataset. For comparing the complementary benefits of SmtBT, we combine all sources of data Primary+Mined+MmtBT+SmtBT, train a similar 1.3B dense model and compare its performance.

**Results.** In Table 21, we report the performance of the baseline model using only Primary data, and then the other three models trained by incrementally augmenting the training data with the Mined, MmtBT, and SmtBT datasets. We observe that the highest performance is achieved when using all sources of data. Despite recent advances which cast into doubt the supposed data inefficiency of neural machine translation (Sennrich and Zhang, 2019), we see that using SMT as a source of backtranslation still leads to improvements for very low-resource directions.

We can further probe into backtranslation quality by looking at the performance of our two teacher models, the MMT model and the set of bilingual SMT models, which are

| Method | avg chrF++ | #best |
|---|---|---|
| SMT | 23.1 | 8 |
| MMT | **43.7** | **30** |

(a) `eng_Latn-xx`

| Method | avg chrF++ | #best |
|---|---|---|
| SMT | 26.2 | 0 |
| MMT | **54.8** | **38** |

(b) `xx-eng_Latn`

Table 22: **Average Performance of Teacher Backtranslation Models**, evaluated on Flores-200 dev for the subset of backtranslated directions where both methods were used. We also report the number of directions at which each method does best.

trained on the same Primary+Mined data. In Table 22 the MMT teacher is, as expected, outperforming traditional SMT on all but a few directions. Although the average SMT performance might be low, we hypothesize that the combination of different, complementary sources of noise is the reason why its addition is still beneficial to the overall performance of the model. The differences between generations can be visualized by plotting the total token frequencies generated by the MMT and SMT teachers when translating the same corpus. We refer readers to Figure 42 in Appendix D for two such histograms.

#### 6.4.2 Data Tagging

Tagged backtranslation (Caswell et al., 2019) is a technique to help the model discern between the different sources of data it is being exposed to during training. This is achieved by pre-pending special tokens to backtranslated training examples, and has been shown to boost performance by helping the model distinguish noisy data and avoid overfitting on it (Marie et al., 2020). In the experiments of the previous section we used an extended tagging scheme, using special tokens for each of the three data sources: <MINED_DATA> for Mined, <MMT_BT_DATA> for MmtBT and <SMT_BT_DATA> for SmtBT. We study the effects of ablating away this tagging scheme, experimenting with using just a single tag to mark all secondary data (<SECONDARY_DATA>) as well as using no tags at all.

**Experimental Setup.** We train MMT models on the full dataset made up of Primary, Mined, MmtBT and SmtBT, but ablate different tagging schemes. The *no tags* setting does not use any tags at all, the *single tag* setting uses the same tag for Mined, MmtBT and SmtBT. Finally the *finegrained tags* setting uses separate tags for Mined, MmtBT and SmtBT.

**Results.** The results in Table 23 demonstrate the benefits of using finegrained tags. This provides further evidence to support the hypothesis of Caswell et al. (2019) that tagging is useful to help the model distinguish between synthetic and natural data. It also suggests that signaling the specific nature of synthetic data can further boost performance.

### 6.5 Bootstrapping models with NLLB-Seed

For a considerable number of the low-resource languages examined in this work, the parallel corpora which are publicly available for research often have only a few thousand sentences. They frequently come from sources with a highly specific domain such as scripture, and the level of quality assurance is often unclear. While translating millions of sentences with professional translators is unfeasible, translating a few thousand is possible. It is important to understand whether there is value in such small but high quality human-annotated *seed* datasets for low-resource languages. Is such a small dataset sufficient to bootstrap a machine translation system for a new low-resource language or finetune an existing machine translation system on a new domain? In this section, we investigate these questions and we aim to measure the effect of training translation models on such a small human annotated *seed* dataset like NLLB-Seed (see Section 4.2). We are interested in quantifying the importance of a dataset which has been professionally translated, covers a wider domain, and which can be confidently attributed to the specified language.

| | eng_Latn-xx | | | | xx-eng_Latn | | | | xx-yy |
|---|---|---|---|---|---|---|---|---|---|
| | all | high | low | v.low | all | high | low | v.low | all |
| No Tags | 42.8 | 54.5 | 38.0 | 31.9 | 54.8 | 64.2 | 50.9 | 48.4 | 40.8 |
| Single Tag | 44.0 | 55.2 | 39.4 | **34.2** | 55.5 | 64.6 | 51.8 | 50.5 | 40.7 |
| Finegrained Tags | **44.2** | **55.5** | **39.6** | 34.0 | **55.9** | **64.9** | **52.2** | **50.9** | **41.1** |

Table 23: **Comparing Different Tagging Schemes** on Flores-200 ablation dev set. We compare models trained on the ablation dataset using *no tags*, a *single tag*, and *finegrained tags*. We report chrF++ scores aggregated by language direction and resource level type. We observe that finegrained tagging gives the best performance.

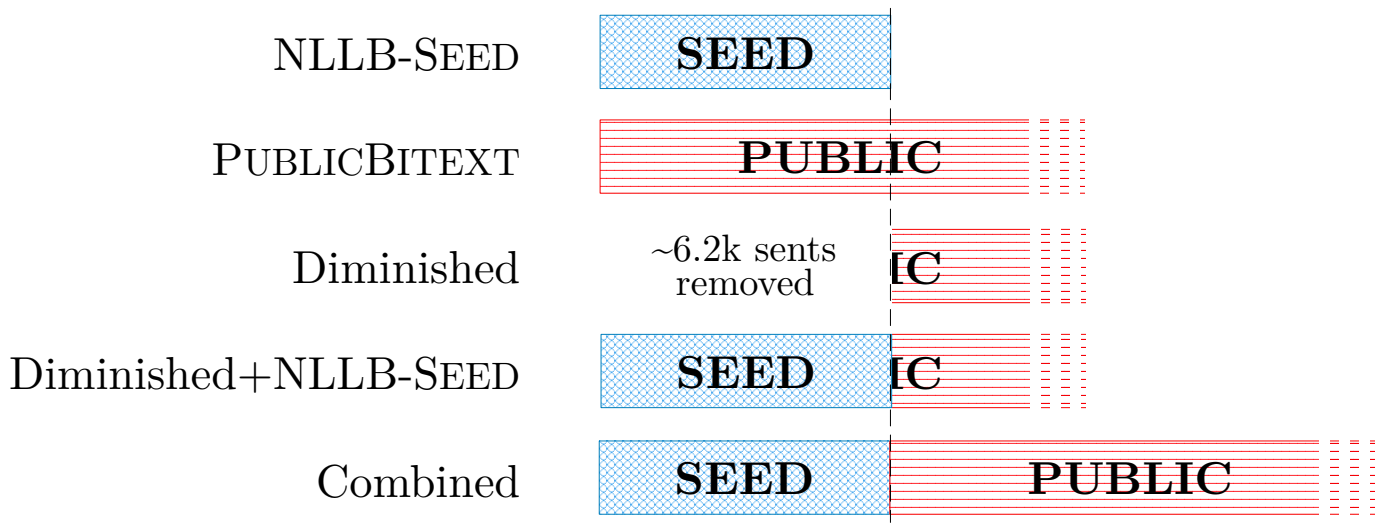

Figure 22: **Dataset Configurations** for the ablation experiments of Section 6.5. We quantify the impact of using NLLB-Seed in various experimental settings.

### 6.5.1 Usefulness of NLLB-Seed

First, we want to measure the performance of bilingual models trained on NLLB-Seed against those trained on publicly available data (PublicBitext) as well as on the combination of both (NLLB-Seed+PublicBitext). Secondly, given a certain amount of publicly available parallel data, we study the incremental effect of adding NLLB-Seed, compared to simply adding more data from the original domain of the public bitext. To answer this we create for each language a new dataset, which we call Diminished+NLLB-Seed. This is obtained by subtracting from PublicBitext a random sample of sentences of the same size as the seed dataset (∼6.2k), and swapping in NLLB-Seed in its place. The result is a dataset which is of the same size as PublicBitext, but which also contains NLLB-Seed. The dataset compositions that we study here are depicted in Figure 22.

| Direction | PublicBitext #train | PublicBitext chrF++ | NLLB-Seed #train | NLLB-Seed chrF++ | Combined chrF++ |
|---|---|---|---|---|---|
| *Average* | | 14.5 | | 17.4 | 20.5 |
| ban_Latn-eng_Latn | 10.2 k | 13.1 | 6.2 k | 20.8 | **22.2** |
| eng_Latn-ban_Latn | 10.2 k | 15.9 | 6.2 k | 20.6 | **21.9** |
| dik_Latn-eng_Latn | 16.9 k | 12.9 | 6.2 k | 16.1 | **17.0** |
| eng_Latn-dik_Latn | 16.9 k | 9.0 | 6.2 k | **13.7** | 13.1 |
| fuv_Latn-eng_Latn | 12.1 k | 15.6 | 6.2 k | 16.3 | **18.1** |
| eng_Latn-fuv_Latn | 12.1 k | 9.2 | 6.2 k | 9.8 | **13.5** |
| mri_Latn-eng_Latn | 31.3 k | 16.7 | 6.2 k | 17.4 | **26.8** |
| eng_Latn-mri_Latn | 31.3 k | 23.2 | 6.2 k | 24.3 | **31.5** |

Table 24: **Training Set Size and FLORES-200 Validation Performance of low-resource bilingual models.** For each direction we report figures for the model trained on PublicBitext, on NLLB-Seed, and on their combination.

**Experimental Setup.** We select eight low-resource directions covered by NLLB-Seed data. For fair comparison to existing datasets we only select directions for which we could also find a minimum of 10 k sentences of publicly available parallel text. These are `ban_Latn`, `dik_Latn`, `fuv_Latn` and `mri_Latn`, translated into and out of `eng_Latn`. We train bilingual models for each direction, using a transformer architecture with 6 encoder layers and 6 decoder layers trained with an inverse square-root learning rate schedule with warm-up. Each language pair uses a custom SentencePiece vocabulary of size 1k. To study the incremental effect of adding NLLB-Seed, we prepare Diminished+NLLB-Seed datasets for each of the above languages. Then we train bilingual models and compare against PublicBitext in a similar setup as described above.

**Results.** The results in Table 24 reveal that, despite the considerably larger size of publicly available training data, training on NLLB-Seed leads to markedly higher performance on average. Unsurprisingly, the best performance is obtained by combining all available data. This result is encouraging, especially in light of recent results showing that a larger MT model can be finetuned on a small but high quality dataset such as NLLB-Seed to adapt it to a new low-resource language easily (Adelani et al., 2022). Furthermore, from Table 25, we observe that the base performance of the Diminished+NLLB-Seed model is higher than that of the base PublicBitext model. This demonstrates that increasing the training data of a model with NLLB-Seed is beneficial compared to increasing it by the same amount of publicly available data. To control for noise, we repeat each experiment three times with different subsets of PublicBitext and report the averages and standard deviations for each. The small variance validates that the low scores on PublicBitext data are not due to any bias of sampling but the lower quality of the PublicBitext data compared to NLLB-Seed.

#### 6.5.2 Effect of NLLB-Seed on Backtranslation

We investigate whether using NLLB-Seed might additionally affect performance when backtranslation is used, for example, if using a small amount of human-translated data such as NLLB-Seed might increase backtranslation quality. For this, we start with base models trained on PublicBitext and on Diminished+NLLB-Seed data, use them to perform backtranslation, and then train a new set of models on this augmented data to quantify performance.

| Direction | PublicBitext chrF++ | PublicBitext BT chrF++ | Diminished+NLLB-Seed chrF++ | Diminished+NLLB-Seed BT chrF++ |
|---|---|---|---|---|
| *Average* | 17.8 | 19.4 | 26.6 | 30.1 |
| ban_Latn-eng_Latn | 12.6 (±0.6) | 14.0 (±0.3) | **22.5** (±0.2) | **24.0** (±0.1) |
| eng_Latn-ban_Latn | 16.8 (±1.0) | 18.7 (±0.3) | **22.2** (±0.2) | **25.2** (±0.3) |
| dik_Latn-eng_Latn | 12.1 (±0.7) | 13.6 (±0.2) | **16.6** (±0.2) | **17.8** (±1.0) |
| eng_Latn-dik_Latn | 8.0 (±1.0) | 10.3 (±0.2) | **13.1** (±0.4) | **14.9** (±0.1) |
| fuv_Latn-eng_Latn | 15.6 (±0.3) | 14.6 (±1.2) | **17.3** (±0.6) | **18.2** (±0.4) |
| eng_Latn-fuv_Latn | 10.0 (±0.9) | 13.6 (±0.3) | **11.8** (±0.4) | **13.1** (±0.2) |
| mri_Latn-eng_Latn | 16.7 (±0.5) | 20.6 (±0.2) | **25.9** (±0.1) | **31.8** (±0.4) |
| eng_Latn-mri_Latn | 23.0 (±0.2) | 24.8 (±0.2) | **30.6** (±0.4) | **36.2** (±0.5) |

Table 25: **Training Set Size and FLORES-200 Validation Performance of base low-resource bilingual models and their backtranslation-augmented versions.** We report figures for the model trained on publicly available parallel data and on the diminished public data concatenated with NLLB-Seed. Each experiment is repeated three times and we report averages, with standard deviation in brackets.

**Results.** From Table 25, we observe that the gap in performance between the models trained with and without NLLB-Seed is further increased when backtranslation is applied. Starting with PublicBitext data only, adding BT brings +1.6 chrF++ improvement, while starting with Diminished+NLLB-Seed data leads to +3.5 chrF++ gain (despite the total size of both datasets being the same). This indicates that a small amount of high-quality bitext significantly improves the effectiveness of model-based data augmentation such as backtranslation.

### 6.6 Human Evaluation

For rapid experimental iteration, the vast majority of modeling ablation decisions are assessed using automatic metrics such as BLEU or chrF++. However, performance improvements in automatic metrics may not translate to human-perceived quality, especially for minor improvements in scores such as BLEU. In this section, we conduct a human evaluation (see Section 7 for a description of our protocol) to understand if our described modeling improvements correlate with quality improvements detectable through human evaluation.

**Experimental Setting.** We evaluate four models: a dense model baseline, our best performing MoE variant, our best performing SSL variant, and our best performing BT variant. For simplicity, we evaluate the same 24 translation directions for all models with 506 source sentences translated per model. 9 directions are translation into English, 11 out of English, and 4 non-English directions for representativeness.

**Results.** We investigate the relationship between chrF++ score and human evaluation score to understand what quantity of automatic metric improvement in a model ablation would be detectable by human translators. Overall, we find that chrF++ improvements of +0.5 chrF++ usually correlate to statistically significant human evaluation improvements, with a score of +1 chrF++ almost always being detectable by human evaluators.

For our best MoE variant, there were 14 directions with chrF++ improvements more than +0.5 over the baseline and 10 of these were statistically significant improvements in human evaluation. For these 10 directions, the human evaluation score more than 0.2 (on a 5-point scale) over the baseline with a corresponding chrF++ improvement of +2.5

For our best SSL variant, there were 9 directions with chrF++ improvement more than +0.5 over the baseline and 4 directions were statistically significantly better in human evaluation. For these 4 directions, the human evaluation score improved +0.2 over the baseline with a corresponding chrF++ improvement of +2.7.

Finally, for our best BT variant, there were 14 directions with chrF++ improvement more than +0.5 over the baseline and 10 directions with statistically significant human evaluation improvements. For these 10 directions, the human evaluation score improved +0.18 over the baseline and chrF++ improved on average by +3.

In conclusion, we believe that based on our human evaluation studies of model ablations, that an improvement of +0.5 chrF++ is often detectable by human evaluators.

### 6.7 Conclusion

Improving the performance of low-resource translation in massively multilingual settings faces several challenges. Directly increasing model size is largely ineffective as low-resource pairs start to overfit. In this section, we studied how to most effectively increase capacity through Mixtures-of-Experts and presented multiple novel regularization strategies. These methods reduce the interference between unrelated language directions. Paired with a training curriculum that introduces higher-resourced pairs earlier in training, we achieved strong gains on low-resource directions while maintaining high-resource performance.

Beyond architectural challenges, low-resource performance is difficult to improve due to data scarcity. We demonstrated how to effectively utilize monolingual data through both self-supervised training and more effective data augmentation. Using multiple different sources of backtranslated data from MMT and SMT models, in combination with mined data, produces significant performance gains.

## 7. Evaluation

The ability to quantify performance is critical to the development of machine learning systems, because improving quality of such systems is impossible without a reliable way to track progress. Machine translation is commonly evaluated using automatic metrics as well as human evaluation, e.g., in the WMT evaluation campaign (Akhbardeh et al., 2021), in the AmericasNLP Shared Task (Mager et al., 2021). In this section, we describe the automatic metrics that we used and the methodology we followed to perform human evaluation. We present the results of various studies conducted on multilingual translation models and analyze the reliability of automatic metrics on such a varied and low-resource set of languages.

We further focus on analyzing quality along other axes. Metrics such as BLEU and human evaluation often focus on an axis of translation quality heavily grounded in accuracy and fluency. Beyond these, we work to quantify our translations from a user safety perspective, choosing to focus additionally on quantifying *toxicity* in the generated translations. As added toxic content is generally undesirable for the user, we focus first on detecting the appearance of such toxicity in all 200 languages, followed by an analysis of mitigation techniques and prevalence on Flores-200. We open source these novel toxicity lists for all 200 languages.

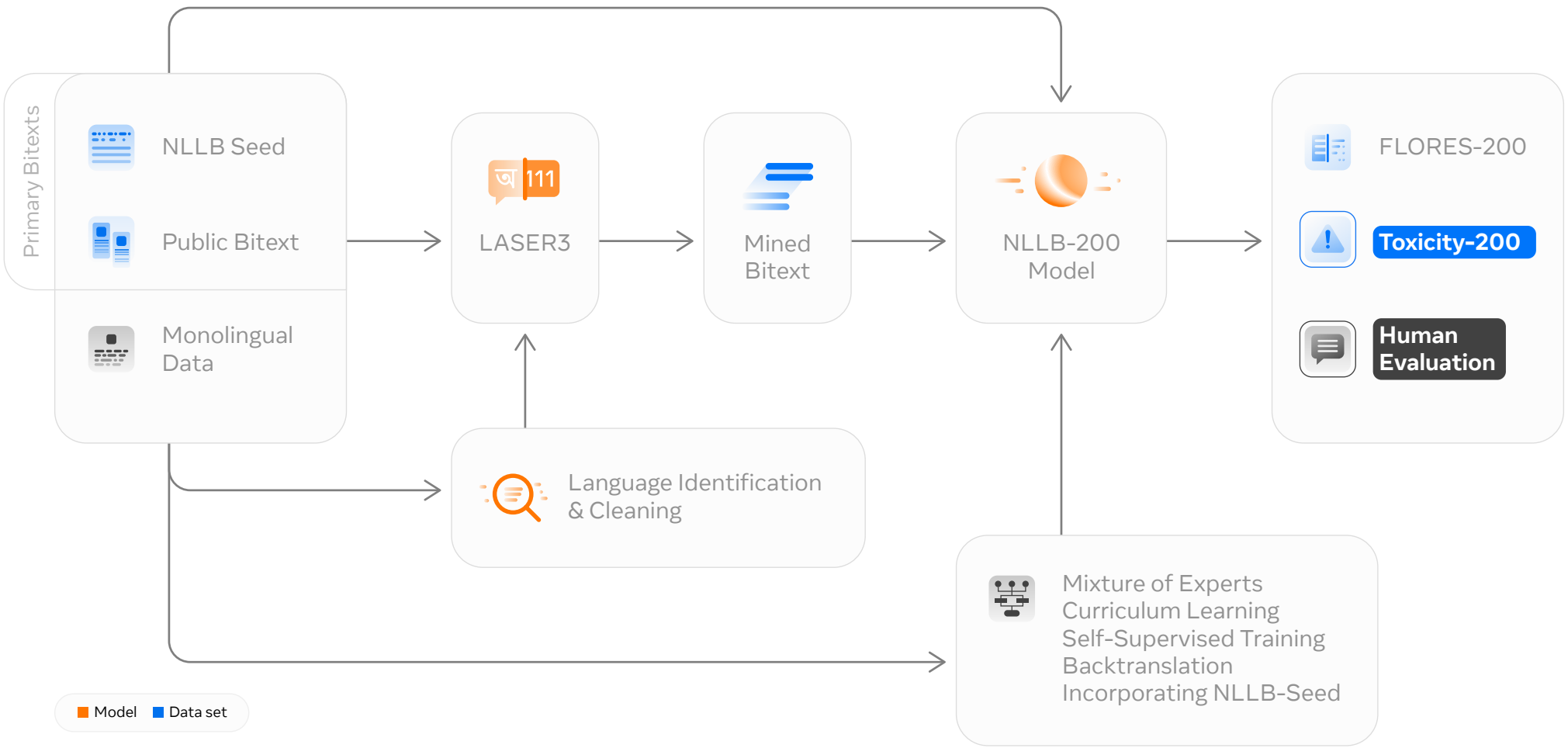

Figure 23: **Evaluation Contributions of No Language Left Behind**: As highlighted, we describe our procedure for Human Evaluation and the creation of Toxicity lists for 200+ languages.

### 7.1 Automatic Evaluation

Various metrics for automatic translation quality assessment exist, including *model-based* metrics such as COMET (Rei et al., 2020) and BLEURT (Sellam et al., 2020). While model-based metrics have shown better correlation with human judgment in recent metrics shared tasks (Freitag et al., 2021), they require training and are not easily extended to a large set of low-resource languages. Another approach is to use highly approximate metrics based on roundtrip translations such as RttLangIDChrF (Bapna et al., 2022), although roundtrip translation may not correlate well with translation quality (Koehn, 2005). While such methods are more easily scaled to new languages, they are highly dependent on factors which makes results difficult to replicate.[27] In this work, we therefore choose not make use of model-based and roundtrip-based metrics, and rely instead on BLEU and chrF++. Both measures rely on the core concept that translation quality can be quantified based on how

27. More specifically, for the case of RttLangIDChrF, neither the full language identification model nor the corpus of sentences used to compute the metric are made public. Furthermore, the main version of the metric used throughout Bapna et al.'s paper does not penalize a translation model's failure to produce text in the correct language, effectively cherry-picking which sentences to evaluate on. The stricter version of the metric, which includes a penalty, is shown to correlate poorly with human judgments.

similar a machine translation is compared to that produced by a human translator. We briefly describe both metrics and a variant of BLEU below.

**BLEU.** The BLEU score (Papineni et al., 2002) has been the standard metric for machine translation evaluation since its proposal two decades ago. It measures overlap between machine translation and a human reference translation by combining precision of 1-grams to 4-grams with a brevity penalty.

**spBLEU.** A major downside of BLEU is that it is *tokenization-dependent*. Efforts such as `sacrebleu` (Post, 2018) have taken strides towards standardization, supporting utilizing community-standard tokenizers under the hood. However, these tokenizers do not extend to many languages. Goyal et al. (2022) propose spBLEU, a BLEU metric based on a standardized SentencePiece model (SPM) covering 101 languages, released with Flores-101. In this work, we provide SPM-200 along with Flores-200 to enable measurement of spBLEU.[28] We describe this in greater detail in Section 8.

**chrF++.** The chrF++ score (Popović, 2017) overcomes the limitation of the BLEU score which requires that a sentence can be broken up into word tokens. However, some languages, such as Chinese or Thai, do not use spaces to separate words and word segmentation tools may not be readily available or even exist. There is also a concern about highly agglutinative languages where BLEU fails to assign any credit to morphological variants. chrF++ overcomes this weakness by basing the overlap calculation on character-level n-grams F-score ($n$ ranging from 1 to 6) and complementing with word unigrams and bi-grams. In this work, we primarily evaluate using chrF++ using the settings from `sacrebleu`. However, when comparing to other published works, we utilize BLEU and spBLEU where appropriate.

### 7.2 Human Evaluation

While automatic scores are a great tool to drive research, human evaluation is essential to ensure meaningful assessments of translation quality (Kocmi et al., 2021). We use two advances — the *XSTS evaluation protocol* and the use of *calibration sets* — to enable meaningful human evaluation scores that are comparable across language pairs.

#### 7.2.1 Methodology

When building machine translation systems for many different language pairs, a core question is which language pairs reach certain levels of quality. Hence, we need meaningful scores that are comparable across language pairs.

**XSTS Evaluation Protocol.** We adapt the recently proposed *crosslingual Semantic Text Similarity* (XSTS) methodology from Agirre et al. (2012). In short, XSTS is a human evaluation protocol that focuses on *meaning preservation* far more than fluency. For low-resource languages, translations are usually of weaker quality, and so we focus far more on usable (meaning-preserving) translations, even if they are not fully fluent. Compared to Direct Assessment (Graham et al., 2013) with a 5-point scale (the original direct assessment uses a

28. Our analyses demonstrate that there are minor differences between the SPM-200 from Flores-200 and SPM-100 model from Flores-101 when measuring on the Flores-101 languages. The major advantage of SPM-200 is that it covers 200+ languages.

100 point scale), work has found that XSTS yields higher interannotator agreement (Licht et al., 2022).

XSTS rates each source sentence and its machine translation on a five-point scale, where 1 is the lowest score and 5 is the highest score. Each point on the scale is as follows:

1. The two sentences are not equivalent, share very little details, and may be about different topics. If the two sentences are about similar topics, but less than half of the core concepts mentioned are the same, then 1 is still the appropriate score.
2. The two sentences share some details, but are not equivalent. Some important information related to the primary subject/verb/object differs or is missing, which alters the intent or meaning of the sentence.
3. The two sentences are mostly equivalent, but some unimportant details can differ. There cannot be any significant conflicts in intent or meaning between the sentences, no matter how long the sentences are.
4. The two sentences are paraphrases of each other. Their meanings are near-equivalent, with no major differences or missing information. There can only be minor differences in meaning due to differences in expression (e.g., formality level, style, emphasis, potential implication, idioms, common metaphors).
5. The two sentences are exactly and completely equivalent in meaning and usage expression (e.g., formality level, style, emphasis, potential implication, idioms, common metaphors).

**Calibration Set.** To enable meaningful scores that are comparable across language pairs, we ask each evaluator to provide assessments using the XSTS scale on exactly the same set of sentence pairs. The purpose of this is to identify which sets of annotators have a systemic tendency to be more harsh or generous in their scoring, and correct for this effect. While evaluators assess different languages, the calibration set consists of machine translation output into English paired with an English reference translation. Based on how evaluators use the XSTS scale on this calibration set, we adjust their raw scores on the actual evaluation task to ensure consistency across evaluators. While the monolingual task does not precisely mimic the bilingual XSTS task, it is a reasonable first approximation and has been shown to increase the correlation between human and automatic metrics, primarily by reducing one source of 'noise' in the human evaluations; the lack of score calibration between annotators.

**Obtaining Aggregate Human Quality Metrics from Multiple Studies.** To obtain an aggregate human quality metric for each language direction in an evaluation study, we take the majority XSTS score for each sentence and average these majority scores over all evaluated sentences. In a given study, the aggregate human evaluation score for a source, target language pair $l_s \to l_t$ is

$$H_{l_s \to l_t} = \frac{1}{|\mathcal{T}_{l_s \to l_t}|} \sum_{(S,T) \in \mathcal{T}_{l_s \to l_t}} \text{median}\{X_{l_s \to l_t, i}(S,T) \mid 1 \le i \le M_{l_s \to l_t}\}, \tag{14}$$

where $l_s \to l_t$ denotes a source language, target language pair, $X_{l_s \to l_t, i}(S,T)$ denotes the XSTS score of the $i$-th evaluator evaluating sentences in a given translation direction $l_s \to l_t$ for a source sentence $S$ and target sentence $T$. $M_{l_s \to l_t}$ denotes the total amount of evaluators evaluating the (source, translation) sentence pair $(S, T)$ for translation direction $l_s \to l_t$ and $\mathcal{T}_{l_s \to l_t} = \{(S_{l_s \to l_t, k}, T_{l_s \to l_t, k}) \mid 1 \leq k \leq N_{l_s \to l_t}\}$ is the set of $N_{l_s \to l_t}$ (source, translation) sentence pairs being evaluated for translation direction $l_s \to l_t$.

Every evaluator in a given study $s$ is also asked to provide ratings for all or part of a *calibration set*, $\mathcal{C}_s = \{(S_{s,k}, T_{s,k}) \mid 1 \leq k \leq K_s\}$, where $S_{s,k}$ denotes the $k$-th source sentence in the calibration set for evaluation study $s$, $T_{s,k}$ denotes the translated sentence corresponding to $S_{s,k}$, and $K_s = |\mathcal{C}_s|$ is the number of sentence pairs in the calibration set for evaluation study. The calibration sets for all evaluation studies are drawn from a set of source, target sentence pairs consisting of $K = 1000$ backtranslated Flores-200 sentences of varying quality.

For each language direction evaluated in a study, we obtain the mean median XSTS score ("majority score") on the calibration set:

$$C^{(s)}_{l_s \to l_t} = \frac{1}{|\mathcal{C}_s|} \sum_{(S,T) \in \mathcal{C}_s} \text{median}\{X^{(s)}_{l,i}(S, T) \mid 1 \leq i \leq M^{(s)}_{l_s \to l_t}\}, \tag{15}$$

where $X^{(s)}_{l,i}(S, T)$ denotes the XSTS score provided by the $i$-th evaluator for the language direction $l_s \to l_t$ in study $s$ that evaluated a given source sentence $S$ and a translated sentence $T$ in the study's calibration set $\mathcal{C}_s$.

To obtain aggregate calibrated XSTS scores on the language direction level, we explored several different calibration methodologies of the form

$$\widetilde{H}^{(s)}_{l_s \to l_t} = f(H^{(s)}_{l_s \to l_t}, C^{(s)}_{l_s \to l_t}) \tag{16}$$

Including a linear shift

$$\widetilde{H}^{(s)}_{l_s \to l_t}[\text{lin}] = H^{(s)}_{l_s \to l_t} - (C^{(s)}_{l_s \to l_t} - \bar{C}), \tag{17}$$

where

$$\bar{C} = \frac{\sum_s \sum_{l_s \to l_t} C^{(s)}_{l_s \to l_t}}{\sum_s \sum_{l_s \to l_t} 1} \tag{18}$$

is the mean majority XSTS score on the calibration set across all evaluated language directions across all studies, which in practice is close to 3 (3.01) and therefore for analysis of individual studies we often replace $\bar{C}$ with 3 to obviate the need for interacting with all evaluation data across all studies.

We also explored other calibration strategies, including clipping the strength of the calibration, adding a multiplicative factor $H^{(s)}_{l_s \to l_t} - \alpha(C^{(s)}_{l_s \to l_t} - \bar{C})$, as well as a more sophisticated heuristic calibration adjustment we name "moderated calibration" designed to keep the calibrated scores within the same $[1, 5]$ domain as the initial majority XSTS scores, to attenuate extreme calibration shifts, and to attentuate calibration shifts when the XSTS score is close to extreme values:

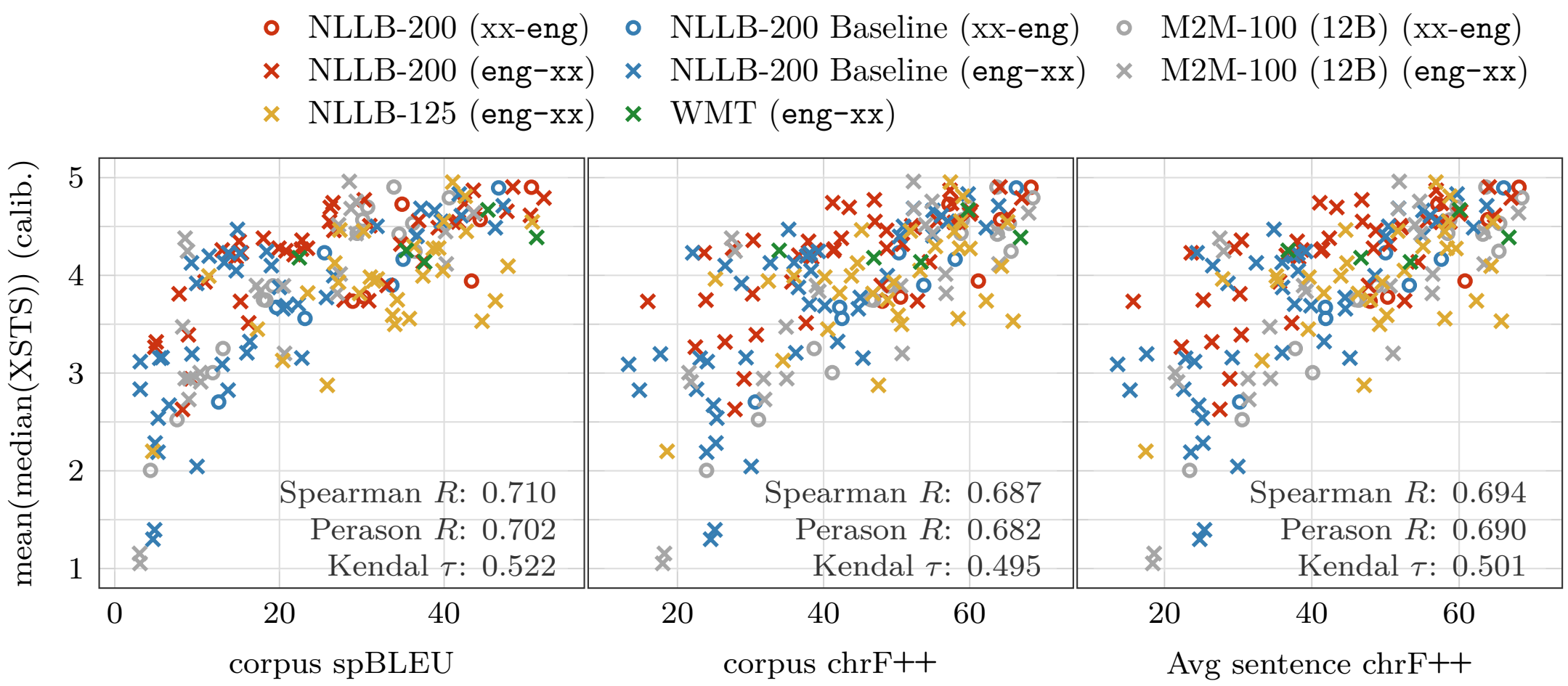

Figure 24: **Correlations between aggregated Human Quality Scores and Automated Metrics**. Left figure shows relationship between spBLEU and XSTS, middle figure shows relationship between chrF++ and XSTS and the right figure shows relationship between *average sentence-level* chrF++ and XSTS. All automated scores were computed only on the sentences evaluated for a given model and translation direction (either the full Flores-200 dataset or a subset). Note that NLLB-200 refers to a 55B parameter MoE model, and NLLB-200 Baseline refers to a dense 3.3B parameter model.

$$C = (C^{(s)}_{l_s \to l_t} - \bar{C}) \tag{19}$$

$$S = \tanh(-C) \tag{20}$$

$$E = \begin{Bmatrix} -\tanh\left(H^{(s)}_{l_s \to l_t} - 5\right), & \text{if } C \leq 0 \\ \tanh\left(H^{(s)}_{l_s \to l_t} - 1\right), & \text{if } C > 0 \end{Bmatrix} \tag{21}$$

$$\widetilde{H}[\text{mod}]^{(s)}_{l_s \to l_t} = H^{(s)}_{l_s \to l_t} + S \times E \tag{22}$$

None of the calibration methods we investigated showed a dramatic difference in correlation with automated scores, and all calibration methodologies we explored provided superior correlation compared with uncalibrated XSTS scores. For this paper, any references to calibrated scores refer to $\widetilde{H}[\text{mod}]^{(s)}_{l_s \to l_t}$.

### 7.2.2 Results

The performance of machine translation models according to human evaluators has been extensively analyzed for bilingual models and specific domains. For example, yearly evaluations at the Workshop for Machine Translation (Akhbardeh et al., 2021) examine a handful of translation directions in the news domain. Another prominent evaluation campaign (IWSLT) puts a focus on speech translation (Anastasopoulos et al., 2021). In contrast, we focus on multilingual translation. In this section, we analyze the correlation between human evaluation scores and automatic metrics such as chrF++, examine the difficulty of Flores-200 as judged by human evaluators in preliminary studies, and discuss variation in human evaluation scores across languages.

**Human Evaluation Studies of Translation Quality.** While human evaluation is the gold standard for understanding true translation quality, automatic evaluation is critical for model design. Comparing the performance of 10 models in a parameter sweep, for example, will rely on automatic metrics. We use aggregated results from three large-scale multilingual human evaluation studies (Study A, Study B, and Study C) to examine relationships between human measures of quality and automated scores like spBLEU and chrF++. These evaluation studies contain evaluations of translations from five distinct translation models (NLLB-200 (MoE 55B), **M2M-100** 12B (Fan et al., 2020), **NLLB-125** — a MoE model covering 125 languages — and an English-Centric multilingual **WMT2021** Submission covering 7 languages (Tran et al., 2021), and a dense 3.3B NLLB-200 model used as a baseline for NLLB-200 (MoE 55B)) and 86 distinct translation directions evaluated by up to 292 distinct human evaluators.

For each large-scale evaluation study, each combination of translation model and translation direction was assigned a group of evaluators to evaluate a set of source sentence and translation sentence pairs. Each (source, translation) pair was scored by 3 evaluators, though the evaluators may (rarely) change between different pairs of (source, translation) sentences. Study B was an exception: the study was conducted in two parts, with one set of evaluators evaluating the first half of the evaluations and another evaluating the second (though evaluator overlap was allowed). The evaluated sentence pairs, calibration sentence pairs, and evaluators differed in each part.

The source, translation pairs come from the Flores-200 dataset (1,000 sentences), however some language directions in some studies were evaluated on a randomly chosen subset of Flores-200 containing only 500 sentences.

Studies A and B shared the same calibration set of 1,000 items, and Study C contained a randomly chosen subset of 500 calibration sentences drawn from the original calibration set.

**How does Human Evaluation Correlate with chrF++ and spBLEU?** We find that automated metrics like spBLEU and chrF++ correlate reasonably well with calibrated human evaluations of translation quality, as seen in Figure 24. In particular, we find that the Spearman R correlation coefficients between aggregated XSTS and spBLEU, chrF++ (corpus) and chrF++ (average sentence-level) are 0.710, 0.687, and 0.694 respectively. Other correlation coefficients (Kendall's $\tau$ and Pearson's R) have the same ordering. Corpus spBLEU provides the best nominal correlation, followed by average sentence-level chrF++ with corpus chrF++ being the least well correlated out of the three.

We also find that calibrated human evaluation scores correlate more strongly with automated scores than uncalibrated human evaluation scores across all automated metrics and choices of correlation coefficient. In particular, uncalibrated human evaluation scores have a Spearman R correlation coefficient of 0.625, 0.607, and 0.611 for spBLEU, chrF++ (corpus) and chrF++ (average sentence-level), respectively.

**How do Human Evaluation scores differ across Languages?** We also inspect the individual score distributions for the NLLB-125 model. We observe three rough categories of XSTS score distribution. The first is *high performance across the board*, meaning that all sentences for that translation direction are rated highly. This is displayed in Figure 25's first column. Such a distribution indicates the translation for almost all sentences evaluated has strong performance. The second pattern is shown in Figure 25's second column, displaying *varied quality*. For these languages, while the average score can be high, there are many sentences that are rated poorly. Finally, the third pattern we observe is large numbers of poor-quality translations (XSTS scores of 1) along with high rates of *incoherent sentences*, meaning the evaluator specifically marked the translation as incoherent. These are shown in Figure 25's third column and often represents text that is mostly incomprehensible or has completely distorted wording.

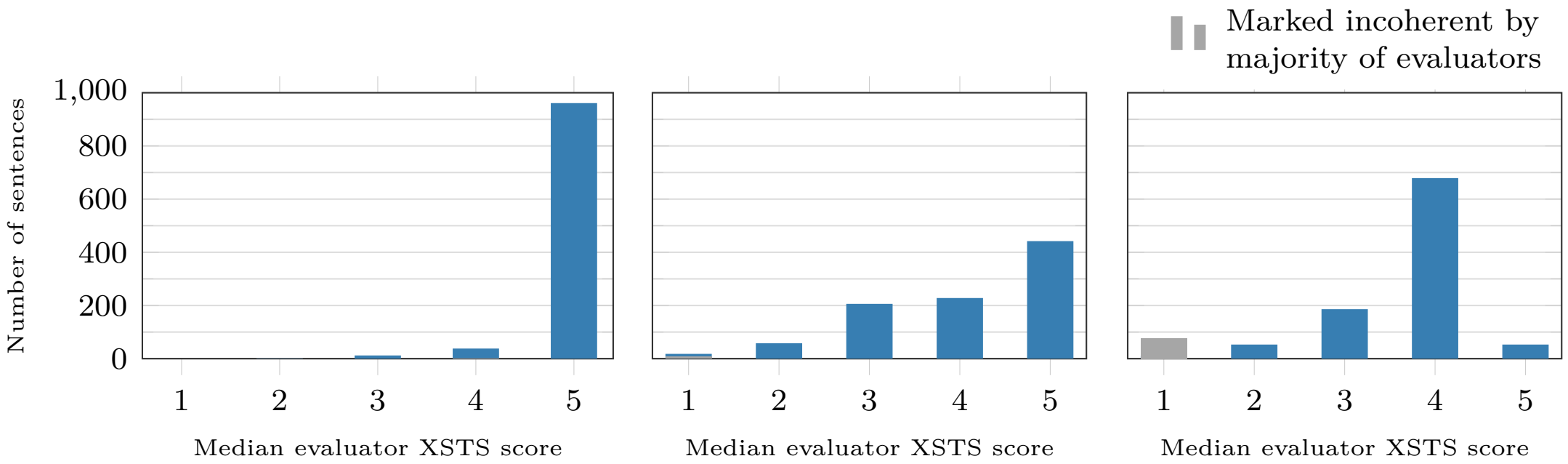

(a) eng_Latn-bos_Latn (b) eng_Latn-isl_Latn (c) eng_Latn-ssw_Latn

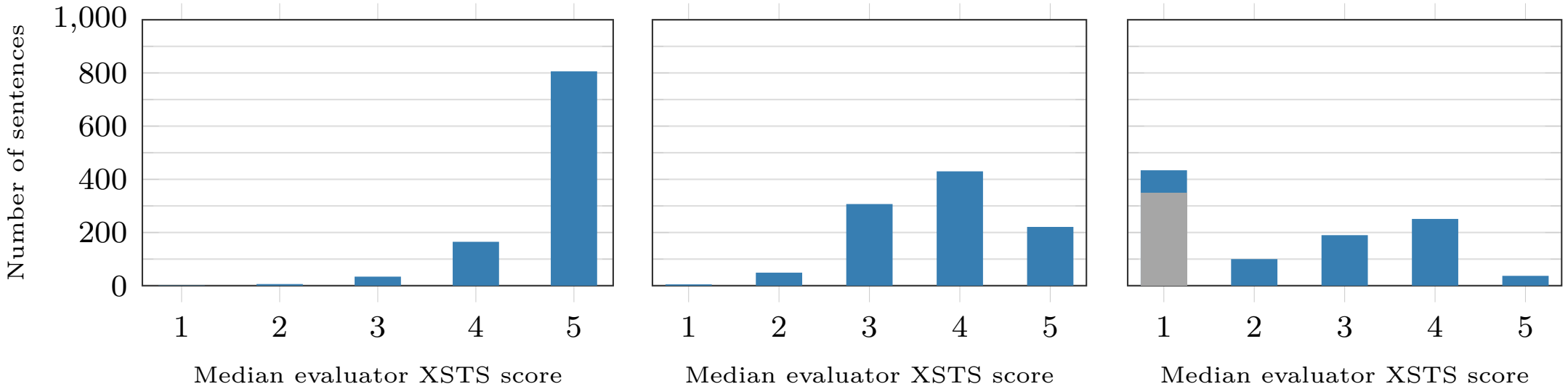

(d) eng_Latn-zul_Latn (e) eng_Latn-jpn_Jpan (f) eng_Latn-wol_Latn

Figure 25: **Selected distributions of median XSTS scores for different Translation Directions**. All translations were generated with the NLLB-125 model, and all scores are from a single evaluation study. We show six distributions that illustrate the three rough categories of score distributions seen in our dataset.

**Human Evaluation for Into English v. Out of English.** Several previous works (Arivazhagan et al., 2019) and our findings in Section 6 indicate that translation from various languages *into* English yields higher BLEU scores than translation *out of* English. We compare human evaluation differences in into English and out of English performance in Figure 26. Generally we find that, as suggested by automated scores like chrF++ and spBLEU, human evaluation scores seem to also reflect that into English translation quality is typically better than out of English translation quality, with some exceptions such as `snd` and `azj` where into English performance is notably worse on both automated metrics and human evaluation metrics.

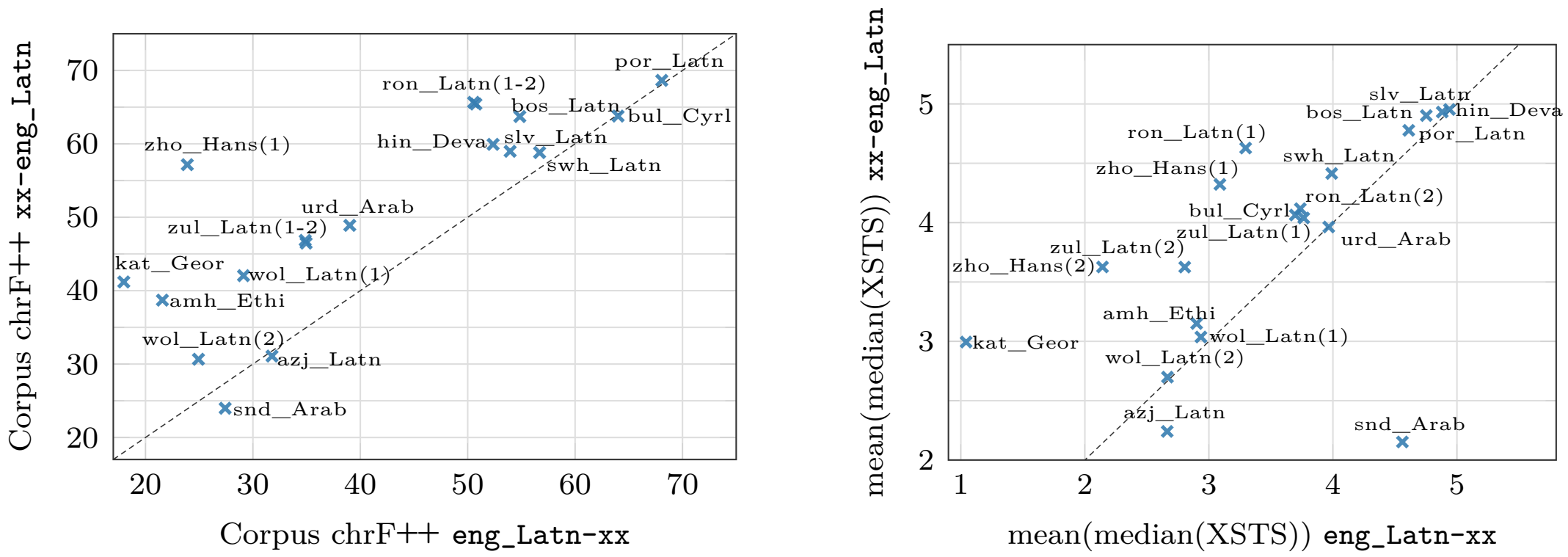

Figure 26: **Comparison of into English vs. out of English Translation Quality**. The left plot compares chrF++ scores and the right plot compares mean median XSTS scores between into English and out of English translation directions. All translations were generated using the M2M-100 (12B) translation model.

## 7.3 Toxicity

Toxicity detection in digital content has received significant attention in recent years, both for user-generated language (Kiritchenko et al., 2021; Mishra et al., 2019; Vidgen et al., 2019; Zampieri et al., 2019) and machine-generated text (Bender et al., 2021; Xu et al., 2020). The generation of toxic content has been explored for various sentence classification and dialogue tasks, but not extensively in translation. However, added toxicity terms, i.e., translated content containing toxic words that were not present in the original text, may have a significant negative impact beyond the overall translation quality. It could, for example, lead to decay of user trust. Our goal in this section is to provide an analysis of toxicity in multilingual MT models. We provide the first baseline to evaluate toxicity in a massive number of languages by collecting and releasing toxicity wordlists in 200 languages (Section 7.3.1). Subsequently, in Section 7.3.3 we propose and evaluate simple yet scalable toxicity detectors that can be optimized in precision or recall depending on the particular application (i.e., filtering or detection, respectively). Then, we propose a filtering strategy to mitigate toxicity imbalance in training data and visualize the source contributions of several examples with added toxicity in Section 7.3.3. Finally, we discuss open challenges and ethical considerations in Section 7.3.4. Note that in this section we will be giving translation examples that contain toxic language.

### 7.3.1 Preliminaries

**What is Toxicity?** Toxicity in natural language processing can be defined as the use of words or phrase structures that induce offensive utterances and bad sentiments (Google Jigsaw, 2017). In the context of translation, toxicity may originally be present in the source text or it can be generated de-novo in the target text (*added toxicity*). This added toxicity can come from a mistranslation (e.g., wrong lexical choice) or as a hallucination of a new target word from zero — both produce inaccurate translations.

**Goal of Toxicity Detection and Mitigation.** Reliable general-purpose MT systems should be able to translate any source content adequately regardless of the domain or register, which includes translating language that may be regarded as toxic. However, they should remain faithful to the source content, and should not add through the translation process any elements of toxicity that are not found in the source. Our main purpose is to improve translation safety through minimizing the probability of catastrophic mistranslations. Note that added toxicity represents one of several types of catastrophic mistranslations (Specia et al., 2021), along with mistranslations of named entities, genders (Levy et al., 2021), numbers and units, and reversal of semantic polarity.

### 7.3.2 Toxicity Lists for 200 Languages

To enable toxicity detection at scale, we focus on a wordlist-based approach. In this section, we describe what we consider a toxic item to be included in a list and how we scale the creation of these lists to hundreds of languages.

**What Content is Toxic?** Due to the subjective nature of toxicity, definitions of toxic language will vary. We include items that are commonly referred to as vulgar or profane language.[29]

In addition to these, we include items more specifically associated with depictions of pornographic content or sexual acts, some frequently used hate speech expressions, and some bullying expressions (including language that can cause trauma or be used with the purpose of silencing someone). We also include items, vulgar or not, referring to body parts that are commonly associated with sexual practices.

**Collection Methodology.** Toxicity is *culturally sensitive*, which constitutes a challenge when starting from one source language (in this case, English) and attempting to find equivalents in such a largely multilingual setting. Hate speech terms such as racial or ethnic slurs, for example, may well be the most challenging of all. We begin based on the professional translation of an initial list assembled in English, and allow additions to adapt to cultural specificities.

We iteratively designed a template for toxicity translation that provides information for disambiguation and contextualization purposes. In the latest iteration of the template, the additional information includes a breakdown into domains (e.g., slurs, sex-related terms, abbreviations), part-of-speech information, pointers to the dictionary definitions of the words in their toxic sense, indications as to the language register(s) (slang, vulgar, formal, etc.). In addition, the template provides clearly identified areas for morphological variants to be added (if the target language is morphologically rich). For polysemous terms, which may or may not be toxic depending on context, the template offers additional room and guidance as to best disambiguation practices through suggesting much less ambiguous, short n-grams (typically, $0 < n < 4$). For the purpose of reducing cultural blind spots, another section of the template gives translators the possibility to insert common toxic language for which it may be difficult to find direct English equivalence. The translators are asked to provide explanations or verbatim descriptions. Suggestions were limited to around forty

29. Note that vulgar or profane language is not always necessarily toxic (some common slang, for instance, may be considered vulgar but it is not necessarily toxic).

items without specific restrictions as to the number of derived word forms per item apart from the general guidance of keeping within the boundaries of frequently used word forms (i.e., steering clear of infrequent and archaic word forms). Translators were not asked to produce leetspeak or nonstandard spelling variants, yet not discouraged from including them where they saw fit. Finally, the template allows for a second translator to act as a peer-reviewer, and insert comments and additional suggestions.

The end product is a list of n-grams that mainly represent: common profanities, frequent insults, pornographic terms, frequent hate speech terms, some terms used for the purpose of bullying, and body parts generally associated with sexual activity.

**Toxicity Lists at a Glance.** To summarize our toxicity detection lists across all 200 languages, the average list length is 271 entries and the median number of entries is 143. The latter may be a better measure of central tendency than the mean average, given that languages with a rich inflectional morphology constitute extreme outliers (e.g., the Czech list has 2,534 entries, the Polish list 2,004). The shortest list has 36 entries and the longest 6,078.

**Related Work.** To detect toxicity in natural language, different approaches can be based on wordlists[30] or on machine-learning techniques.[31] Much recent toxicity analysis in NLP is based on the training of supervised classifiers at the sentence level mostly for English, and extended up to a few other languages (Lees et al., 2021) in multilingual classification. However, we are not aware of a machine translation analysis that evaluates added toxicity, or toxicity imbalances in parallel training datasets; and scales these analyses to a large number of languages. In general, this training of supervised classifiers requires large amounts of labeled data, which means they are often limited to high-resource languages. But, more importantly, recent studies have claimed the strong biases that these classifiers have in wrongly associating the language of underrepresented social groups with toxicity (Xu et al., 2021a). While we can not catch non-lexical toxicity nor toxicity terms that are not included in our wordlists, our approach is more scalable and thus capable of extending to hundreds of languages. Furthermore, using a wordlist-based approach provides higher degrees of transparency and explainability.[32] Finally, our wordlists can potentially be used for other applications in NLP. We discuss possible limitations of wordlists in Section 7.3.4.

### 7.3.3 Toxicity Detection

**Methodology.** Our toxicity detectors identify toxicity using the following two criteria: *number of toxic items* and *matched toxicity*. This is illustrated in Figure 27. We define toxic items as short n-grams (with $0 < n < 4$, most commonly) present in our lists described in Section 7.3.2. Among the number of toxic items we explore two cases: 1 or more toxic item and 2 or more toxic items. The toxicity can either be *matched* or *non-matched*, where non-matched indicates that there are toxic items either in the source or in the target part of the bitext. In contrast, *matched* toxicity indicates that toxic items are present on both sides of the bitext, but it does not necessarily mean that toxic words are correctly translated.

30. `https://cleanspeak.com/`
31. `https://www.perspectiveapi.com/`
32. `https://cyber.harvard.edu/publication/2020/principled-ai`

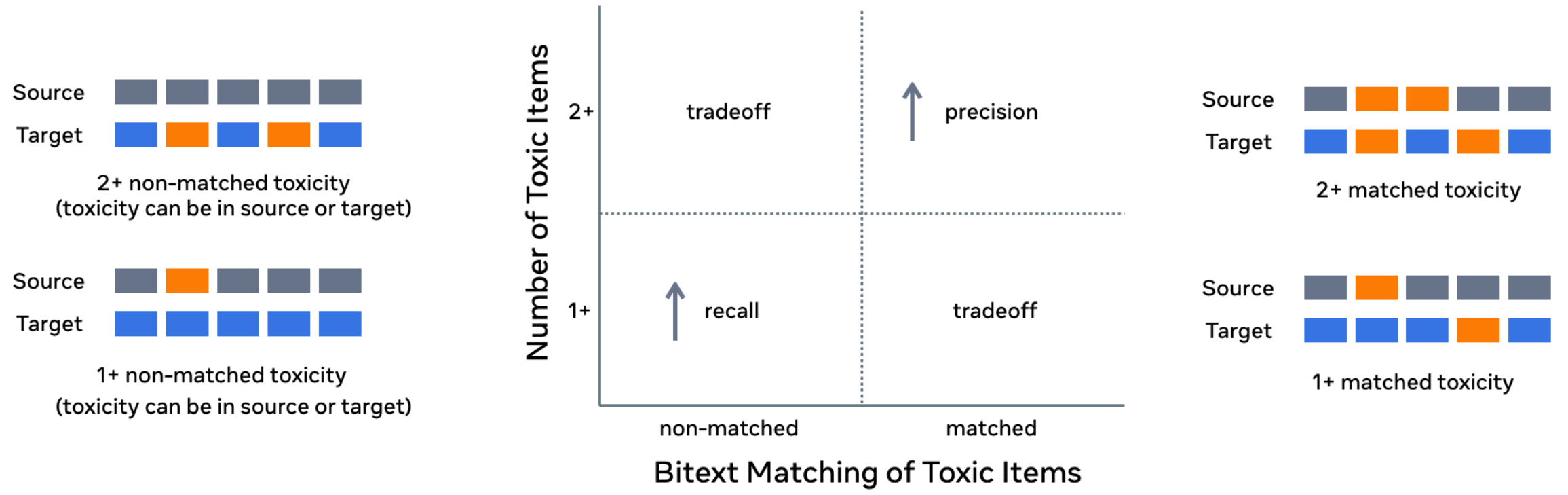

Figure 27: **Overview of Toxicity Detectors** which vary with number of toxic items and bitext matching of toxic items.

Therefore, we have the following toxicity detectors' options: (1) 1+ toxic item non-matched, (2) 2+ toxic items non-matched; (3) 1+ toxic items matched; (4) 2+ toxic items matched.

A toxic item (or phrase) is considered detected if it is present in a line to be examined, and it is surrounded by spaces or the start/end of a line (e.g., we do not detect *bass* if looking for *ass*). We track the number of unique toxic items found in a line, but do not count a phrase again if it has multiple occurrences. For some languages (i.e., Assamese, Burmese, Oriya, Korean, Chinese) space tokenization is not sufficient to distinguish words from one another. In those cases we utilize SentencePiece tokenization of both the sentences and the toxic wordlist.

**Evaluation.** To quantify the quality of our toxicity lists and detectors, we aim to utilize a general-purpose toxicity benchmark first before application to machine translation. We used the Jigsaw dataset,[33] which aims to take an input sentence and detect if this sentence is toxic or not. This dataset classifies toxicity among the following sub-classes: *toxic, any toxic, obscene, threat, insult, severe toxic, identity hate.*

We used the test set partition available from the first challenge and filtered comments by applying these criteria: lines with less than 600 characters, less than 100 words, and with at least 1 word token. Then, we filtered down to only those rows which actually contained at least 1 English letter to filter out only numbers, emoji, etc. Our final set is 86.2% benign and the remaining 13.8% sentences have 1 or more toxic labels (with non-exclusive classes). As no comparable baseline exists over so many languages, we provide the performance of better than chance 'random baseline model' which we compute as randomly generating the toxic/non-toxic labels (both at random 50/50 toxic/non-toxic rates, and at 13.8%/86.2% toxic/non-toxic rates). Then, we use NLLB-200 translation model to translate this English set into all other available languages in the model. Note that using a translation model to generate non-English toxicity references will affect the analysis of quality of our detectors as translation can modify the toxicity level in the references. However, we did not have the alternative of evaluating on toxic labelled data for our 200 languages.

Table 26 reports the results of our detectors in English, French (as an example of high-resource language), and an average over all languages with chrF++ $> 45$. The best

33. `https://www.kaggle.com/c/Jigsaw-toxic-comment-classification-challenge`

| | | F1 | Recall | Precision | FPR | FNR |
|---|---|---|---|---|---|---|
| Baseline | 50% | 22.2 | 51.4 | 14.1 | 49.8 | 48.7 |
| | 13.75% | 13.7 | 13.7 | 13.7 | 13.7 | 86.3 |
| `eng_Latn` (src) | 1+ toxic item | **66.8** | **71.4** | 62.8 | 6.6 | **28.5** |
| | 2+ toxic items | 32.3 | 19.7 | 88.1 | 0.4 | 80.2 |
| `fra_Latn` (tgt) | 1 toxic item | 61.6 | 63.6 | 59.6 | 6.7 | 36.3 |
| | 2+ toxic items | 24.1 | 14.2 | 79.2 | 0.6 | 85.8 |
| | 1+ toxic item matched | 61.2 | 55.9 | 67.6 | 4.1 | 44.0 |
| | 2+ toxic items matched | 19.0 | 10.6 | **88.4** | 0.2 | 89.3 |
| Mean (tgt) | 1+ toxic item | 29.0 | 22.8 | 56.0 | 2.6 | 77.1 |
| | 2+ toxic items | 5.8 | 3.2 | 76.0 | **0.0** | 96.7 |
| | 1+ toxic item matched | 27.8 | 20.0 | 69.5 | 1.2 | 79.9 |
| | 2+ toxic items matched | 4.4 | 2.4 | 86.2 | **0.0** | 97.5 |

Table 26: **Detecting Toxicity with Various Methods**. We display performance for random choice + majority class baseline, English as source, French as translation from English, and the mean score over all translation languages with a chrF++ score greater than 45. In bold best results.

performing toxicity detector has 66.8% F1 score in English and 61.6% of F1 score in the high-resource language French. These are far better than our baselines of better than chance. Figure 28 shows decreasing F1 scores for 1+ toxic item vs 2+ toxic items detectors. Dotted lines in the figure are random chance performance. The language set is a uniform sample of several high and low-resource languages. In order to minimize the confounding factor of quality of translation and detectors, we limited the sample to those with chrF++above 45 as model quality. The mean is computed both on languages with chrF++ above 45 and on all languages, without taking into account the chrF++ threshold. On average, our detectors are better than chance, even for low-resource languages, which demonstrates the usefulness of our detectors. However, there is a significant difference between high and low-resource languages. The results for the languages with the lowest performance in Figure 28, i.e., Hindi (`hin_Deva`), Kannada (`kan_Knda`), Maithili (`mai_Deva`), Telugu (`tel_Telu`), and Magahi (`mag_Deva`), may be partially explained by the fact that the scripts in which these languages are written are not always adequately tokenized by our detectors.

.

**Estimated Toxicity in the Training Data.** We used our detectors (1+ toxic item non-matched) to count potentially toxic items in the bitext training data. Figure 29 shows percentage of toxic items per corpora in the English-side of our bitext pairs only. Figure 30 shows the percentage of toxic items per corpora in the non-English-side. Since we are comparing across languages, we use the Christian Bible corpus to calibrate the baseline level. We use this corpus since it is highly multilingual and it should be the most consistent in content, quality and level of toxicity across languages, if the Christian Bible corpus was not available, results are left uncalibrated).

We see that the amount of potential toxicity varies among languages and corpora. Overall, OpenSubtitles and qed have a larger amount of potential toxicity in several languages and

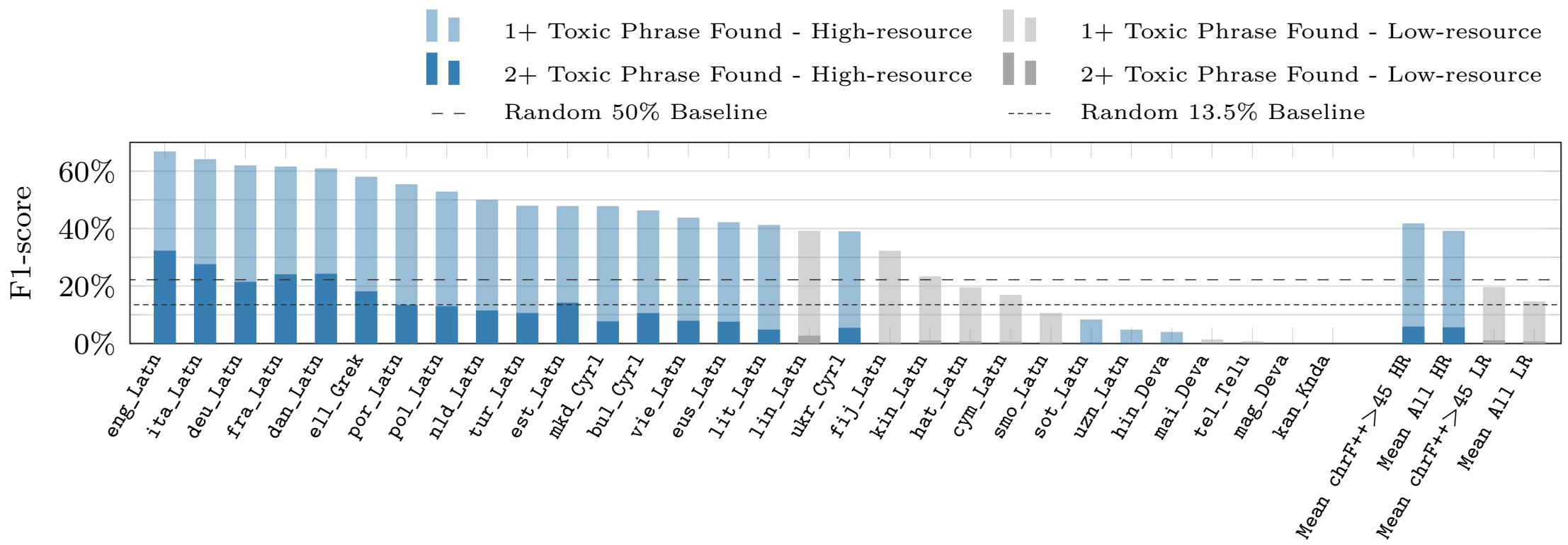

Figure 28: **Ordered F1 scores for list-based Toxicity Detectors.** 1+ toxic item vs 2+ toxic item. Dotted lines are random chance baselines performances. The language set is a uniform sample of several high and low-resource languages, limited to those with chrF++ > 45. Mean scores presented for both all languages without and with the chrF++ threshold for inclusion.

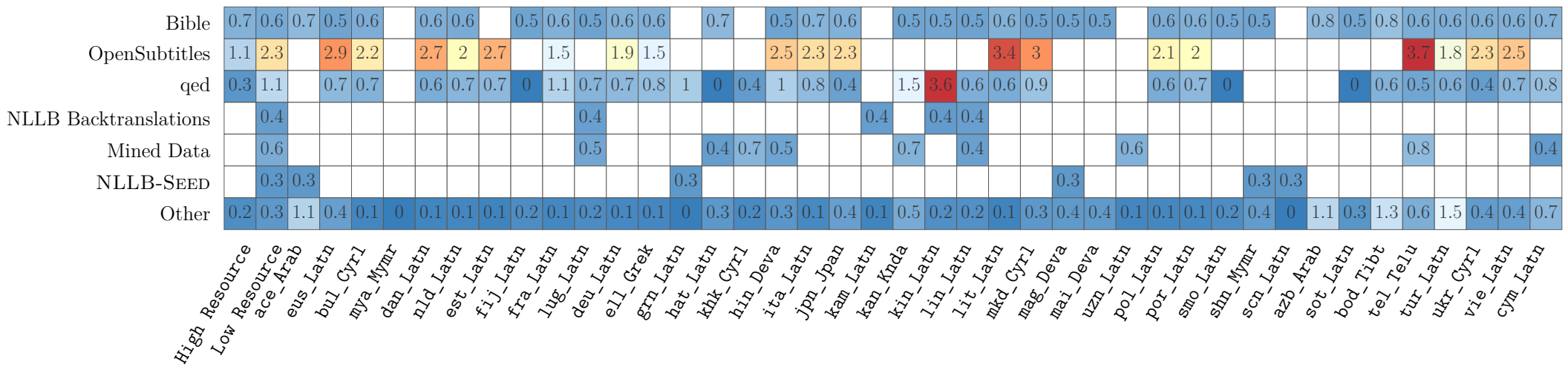

Figure 29: **Percentage of Toxic Items per found per corpora in the English-side.** The same list was used across all corpora language pairs.

Mined Data has a lower amount. We examined a sample of the potential toxicity of these corpora and found that a great proportion were misaligned bitext. Toxicity was present only in one side of these bitext, either through complete omission (by far the most common) or significant detoxification. Training in this kind of misaligned bitext can encourage mistranslations with added toxicity.

**Mitigating Toxicity by Filtering the Training Data.** Added toxicity can arise from toxic items present in the training data. The objective of data filtering is to detect imbalances in toxicity, which could eventually generate toxic mistranslations. Note that we do not want to filter out all toxicity from training corpora because this would introduce bias and prevent the model from correctly translating potentially toxic items, even under benign usage. Thus, we are interested in discarding sentences that are really toxic on one side and not on the other, optimizing for precision. The analysis on detector quality reveals that when our detectors detect multiple instances of toxicity (2+ toxic items) in a sentence, we have a higher precision in toxicity detection. Based on this, we propose to filter training sentences that contain a difference of multiple toxic items when comparing the source and target sides.

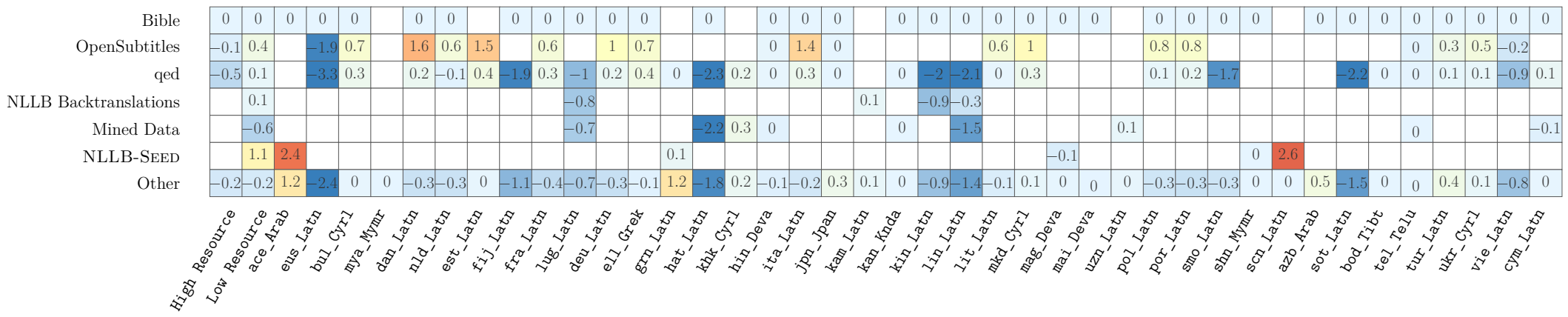

Figure 30: **Percentage of Toxic Items per found per corpora in the non-English-side**, calibrated by subtracting the value of the Christian Bible corpus, for consistency across languages

| | Before Filtering | | | After Filtering | | |
|---|---|---|---|---|---|---|
| Direction | # Train | chrF++ | Toxicity | # Train | chrF++ | Toxicity |
| `eng_Latn-smo_Latn` | 7.3 M | 48.7 | 33 | 3.8 M | **49.4** | **22** |
| `eng_Latn-sot_Latn` | 10.2 M | 44.4 | 9 | 6.2 M | **44.4** | **6** |
| `eng_Latn-twi_Latn` | 7.1 M | 35.7 | 138 | 2.9 M | **36.7** | **136** |
| `eng_Latn-umb_Latn` | 1.2 M | 13.1 | 1 | 0.8 M | **23.3** | **0** |
| `eng_Latn-vec_Latn` | 8.1 M | 37.7 | **14** | 3.8 M | **40.6** | 22 |
| `smo_Latn-eng_Latn` | 43.0 M | 49.0 | 4 | 32.2 M | **49.7** | **4** |
| `sot_Latn-eng_Latn` | 42.4 M | 51.8 | 2 | 33.4 M | **52.3** | **2** |
| `twi_Latn-eng_Latn` | 41.5 M | 38.4 | 1 | 30.1 M | **39.8** | **1** |
| `vec_Latn-eng_Latn` | 41.4 M | 53.8 | 2 | 30.1 M | **56.3** | **2** |
| `umb_Latn-eng_Latn` | 39.3 M | 27.1 | 4 | 29.9 M | **27.2** | **2** |

Table 27: **Experiments with Filtering** for bilingual translation models. We report for each direction the number of training sentences, chrF++ and added toxic items.

To test this methodology, we included toxicity filtering to the filtering pipeline which will be described in more detail in Section 8.1.4. We then trained bilingual models with and without this filtering using the architecture described in Section 6.5. Table 27 shows results on the Flores-200 `dev`. This table includes the number of training sentences, the translation quality (chrF++) and the number of toxic items in the translation output. All figures are reported before and after filtering. We observe that a filtering pipeline which includes toxicity filtering not only improves translation performance but also tends to reduce the number of toxic items in the translation.

**Interpretability of Added Toxicity.** Added toxicity in translation output may come from hallucination or mistranslation. Note that *hallucinated toxicity* means that there is a toxic item in the target output that appears without having any direct source correspondence. In contrast, *mistranslated toxicity* means that we are translating a non-toxic source word into a toxic target word. To differentiate between these two cases cases, we use ALTI+ (Ferrando et al., 2022), which enables examination of source and target prefix contributions to model predictions. Figure 31 reports the input attributions of a particular word. Each example contains the source sentence, the target output sentence and the translation English reference sentence.

- Sentence 1, which is a translation from Wolof to English, shows the input attributions of the word *idiots*. We observe that the input contributions of *idiots* is concentrated on the target prefix word of *self-absorbed*. The total source contribution for the predicted word *idiots* is 0.35, which is significantly lower than the total target prefix contribution of 0.65. Again, this is coherent with the fact that this word is **hallucinated**, since it does not have a direct correspondence to a source token.
- Sentence 2, which is a translation from Quechua to English, shows the input attributions of the word *ass*. We observe that the input contributions of *ass* is concentrated on the target prefix words of *pain, in, the*. The total source contribution for the predicted word *ass* is 0.20, which is significantly lower than the total target prefix contribution of 0.80. This is coherent with the fact that this word is **hallucinated**, since it does not have a direct correspondence to a source token.
- Sentence 3 shows the input contributions of the word *penis*, when translating from Northern Kurdish to English. We observe that these contributions clearly include the source word *Penceya*. The total source contribution for the predicted word *penis* is 0.68, which is significantly higher than the total target prefix contribution of 0.32. We conclude that this word is **mistranslated**, since it does have a direct correspondence to a source token *Penceya*, which should be translated as *claw*.
- Finally, Sentence 4 shows the input contributions of the word *Nazis*, when translating from German to English. The total source contribution for the predicted word *Nazis* is 0.77, which is significantly higher than the total target prefix contribution of 0.23. This example is added to compare the behaviour of ALTI+ in an **accurate** toxic translation.

### 7.3.4 Open Challenges in Toxicity for Machine Translation

There exist many open research questions for the challenge of detecting toxicity at scale for hundreds of languages. Since we are evaluating our approach in a translated dataset, the quality of translation may be a confounding factor that will be worth exploring. For example, the quality of the toxicity detection can be affected by the amount of resources available per language. Alternatively, the quality and efficiency of our detectors which detect or filter toxicity may vary depending on the wordlist length, the segmentation accuracy, the variation of toxicity class, and the amount of non-lexicalised toxicity. Challenges in the wordlists include morphological aspects such as case sensitivity. Differences in casing are often related to word order; i.e., the initial characters of words in sentence-initial position are often capitalized. Our detectors lowercase all items prior to detection, which makes it impossible to differentiate between certain homographs that differ only in the casing of their initial characters. The expansion and disambiguation of small toxicity lists are critical areas for future work, which likely require close collaboration with a larger number of native speakers. A first step towards disambiguation can be to contextualize polysemous words by replacing single tokens with n-grams that have a much higher probability of representing true toxic content. Finally, we know that added toxicity can be caused by the phenomena of hallucination or by an error in the translation. Our visualization examples with ALTI+ show that a low amount of source contribution in the toxic item, computed with this method, is a strong indicator for hallucination. We want to use this information to further quantify and mitigate added toxicity.

| **Hallucinated Toxicity** | **Mistranslated Toxicity** |
|---|---|
| Sentence 1 | Sentence 3 |
| Montaru biologie ay osiyatër yu yor seen bopp lañu yuy wey ci ab diir bu moom boppam donte amul ay juñju biti. | Penceya wî ya didoyê mezintir bû, û bû sebebê navê Hesperonychus-ê |
| It's about a bunch of self-absorbed **idiots** who live in their own little world, and they don't have time for anything else. | His dodo's **penis** was larger, and was the reason for the name Hesperonychus |
| Reference 1 | Reference 3 |
| Biological clocks are self sustaining oscillators which will continue a period of free-running cycling even in the absence of external cues. | Its second claw was larger, giving rise to the name Hesperonychus which means "western claw." |
| | **Correctly Translated Toxicity** |
| Sentence 2 | Sentence 4 |
| Kayqa nanaykuna tañichiqina imaymanamkunapi. | Machen Sie keine Witze über den Holocaust oder die Nazis. |
| It's a pain in the **ass.** | do not make jokes about the Holocaust or the **Nazis.** |
| Reference 2 | Reference 4 |
| This is just like symptomatic treatment in many cases. | Do not make jokes about the Holocaust or **Nazis**. |

Figure 31: **ALTI+ Visualizations of Source and Target Contributions**. Hallucinated toxicity (sentence 1) shows input contributions for the toxic item *idiots* in Wolof-to-English; (sentence 2) shows input contributions for the toxic item *ass* in Quechua-to-English; Mistranslated toxicity (sentence 3) shows input contributions for the toxic item *penis* in Kurdish-to-English. English reference is shown in each example; Correctly translated toxicity (sentence 4) shows a perfect translation for comparison with previous examples.

#### 7.3.5 Ethical Considerations for Toxicity Research

The evaluation of machine translation has been deeply studied in terms of quality assessment both using automatic and human annotation approaches. Recently, responsible MT evaluation is emerging, motivated by the more general responsible artificial intelligence area. This evaluation aims at measuring fairness, ethical, and social aspects of our technologies. For example, it allows us to quantify the toxicity or biases that our model keeps, generates, or potentially amplifies; e.g., Blodgett et al. (2020); Costa-jussà (2019); Renduchintala and Williams (2021). Among the different alternatives in responsible evaluation, we have started with the toxicity challenge because it contributes highly to harmful internet content (see Section 2). In this subsection, we discuss ethical issues related to toxicity detection in translation.

**Unintended use of Toxicity Lists.** It should not be assumed that our lists could be used to moderate content or suppress machine-generated toxic language. As is the case with human translators, we believe that machine-translation systems should remain faithful to the input text, which is why we focus solely on detecting added toxicity.[34] Even informative and

34. A single exception was made in the case of backtranslated data, due to the nature of web crawled data which contains a high proportion of pornographic and toxic content.

educational sources that contain either some degree of toxicity or other tokens that may not be toxic in all contexts but will be matched by our lists; for example, some Wikipedia pages contain descriptions of sexual acts, and others contain vocabulary describing intimate body parts for the purpose of providing information about human biology and health. However, we acknowledge that our toxicity lists could be used for other purposes that are different from, or opposed to, ours. They could be used to inform adversarial attacks against toxicity classifiers (Kurita et al., 2019), or for the enforcement of policies that aim to surveil or suppress toxic language. While acknowledging potential unintended uses of our lists, we remain mindful of the likely possibility that similar lists may have already been separately collected for the aforementioned purposes by entities who have not publicly disclosed them.

**Biases in List Building.** It is likely that toxicity lists themselves include biases due to errors, omissions, or insufficiently diverse cultural backgrounds (Davidson et al., 2019; Gehman et al., 2020; Ross et al., 2017; Sap et al., 2019). We hypothesize two major causes for these biases: (1) ambiguities that are inherent to toxic language itself and (2) cultural biases that can be introduced at any step of the list-building process, since our lists were for the most part translated from English.

First, we discuss ambiguities in toxic language itself. Due to the discomfort or potential harm toxic language can cause, there have always been attempts to curb its use in social interactions. This, in turn, has led social actors to design new means of expression to circumvent censorship or avoid opprobrium; metaphors and innuendos would be good representatives of such means of expression. For example, common linguistic camouflage schemes include referring to animal names for insults or references to body parts. For these reasons many words and phrases that make part of toxic language are also common words that can be used innocuously. Conversely, what may now be considered very specific technical terms (but were more widely used at different times in our history) have been known to sometimes take on a toxic meaning (e.g., the English noun *slag*). These linguistic phenomena cause ambiguities, which in turn lead to mistranslations. To prevent misunderstandings and resolve ambiguities, the latest mitigation steps included the creation of a one-hour translator training session and the search for more accurate n-grams as replacements for ambiguous single tokens. Periodic iterative testing on a variety of input texts followed by both quantitative and qualitative analyses of the results allow for more accurate n-gram substitutions, although it remains clear that such continuous improvement method eventually reaches a point of diminishing returns.

Apart from known ambiguities that are inherent to toxic language, cultural biases can also infiltrate toxicity lists either via the building of the reference list or via the translation process. Cultural biases can be introduced by the fact that the first drafts of almost all lists were translations from English (mostly American English). Some English entries have no clear direct equivalents in other languages. In this case, errors and biases can be introduced when expectations are not clearly stated and translators attempt to provide translations for all entries at all costs. In addition to clarifying expectations, this problem can be mitigated by asking the translators, as an intermediate step in the translation task, to indicate whether they think that a direct equivalent can be found in their native language, which helps them remember that they should refrain from providing translations if such is not the case. Conversely, some toxic items in languages other than English have no clear direct equivalents in English, which further increases the coverage discrepancy in between lists. Such discrepancy has proven more difficult to mitigate. Unrestricted additions generate noise, while restricted additions generate friction. This puts into perspective the efficiency of the alternative to list building through translation of a reference list, which would be building lists solely by collecting free-form suggestions.

Finally, there are also potential biases in the translation process. Biases can be introduced through unequal availability of translators and consultants across languages. For high-resource languages, numerous translators and consultants are available and compete on the language service market. This is unfortunately not the case for low-resource languages, where access to fewer professionals with diverse backgrounds and specializations increases the risk of biases in favor of a single worldview.

**Translator and Consultant Safety.** By definition, the toxicity area of work is one where translators and consultants may feel too uncomfortable to work (Welbl et al., 2021). Additionally, regardless of one's willingness to work on potentially toxic language, we are mindful of the fact that the production of toxic language, be it through translation, may not be perceived equally in all parts of the world. In parts of the world where free speech is generally protected by law, toxic language may be seen as offensive, triggering, or harmful, and its unsolicited use can be subject to consequences in certain venues (such as in the workplace), but anyone is theoretically free to voluntarily work on the collection, curation, translation, or analysis of toxic language. We cannot safely assume that such is the case in all other parts of the world. Bearing this in mind, translators and consultants are made aware that the work contains toxic language, and can reject it.

## 7.4 Conclusion

Evaluation plays a vital role in the development of machine translation systems, and also allows potential users to assess if translation quality is good enough for their purposes. For system development, we rely on automatic metrics. Due to the diverse set of languages we address, with some lacking visible indicators of word boundaries, we have to move beyond the common practice of using BLEU scores. We use two metrics: spBLEU, a modification to BLEU that uses a sentencepiece model to ensure tokenization of any text in any language, and chrF++ that operates primarily on character-level.

While automatic scores are an essential tool, they fail in one important aspect: providing meaningful scores that enable deployment decisions. Hence, we add human evaluation as a final system evaluation step. However, due to variance between human evaluators, this does not guarantee meaningful scores either. We achieve the goal of meaningful scores by a new and more clearly defined scoring metric (XSTS) and the use of a calibration set that not only allows us to adjust raw scores from evaluators working on the same language pair but also to obtain consistent scores across language pairs.

Moreover, we are concerned about harmful content generated from our translation models. We developed novel methods to detect toxicity, i.e., translated content containing toxic words that were not present in the original text. We were able to extend this work to all 200 languages of the NLLB effort due to a language resource that we created for all these languages: lists of toxic terms. Armed with this resource, we built classifiers to detect and mitigate toxicity in our translations.

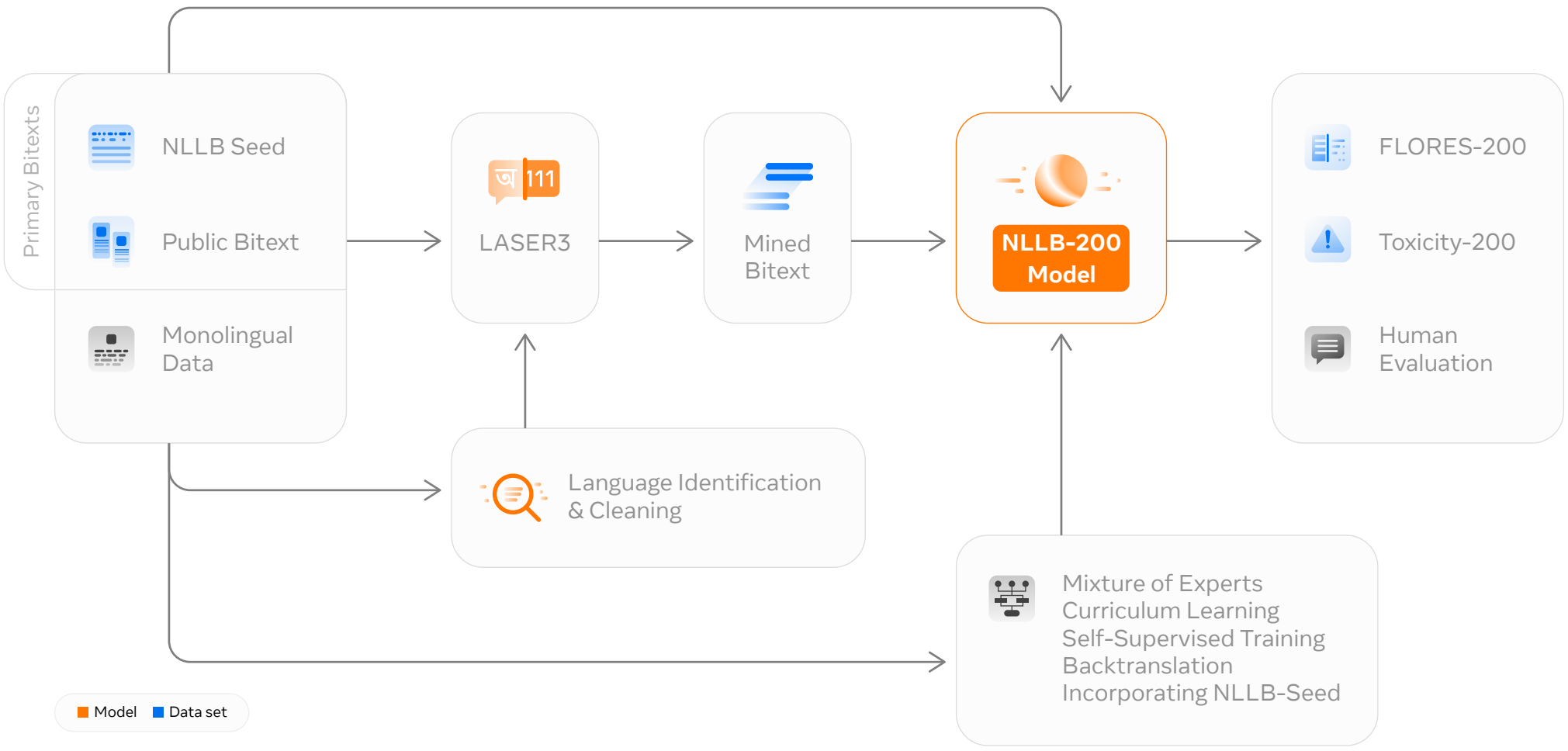

Figure 32: **Bringing it All Together**: We describe how we create our final NLLB-200 using all of the datasets and modeling techniques we created.

## 8. Bringing it All Together

We have seen in previous sections how multiple areas of interdisciplinary research have addressed different parts of the problem of scaling human-centered translation for low-resource languages. Buttressed by our guiding principles described in Section 2.2, we bring all the *interdisciplinary* work from the previous sections and combine them in a manner that improves the performance of multilingual translation systems for *low-resource languages from underserved communities*. We *share* everything transparently, our roadblocks and successes. To enable the community to leverage and build on top of NLLB, we open source all our evaluation benchmarks, models, datasets, training and inference code as well as other modeling and data artifacts. Driven by our guiding principle of *being reflexive*, we hope sharing our work and tools will help the community to examine the current practices and improve where we fail, in a mission towards the north star goal of *no language left behind*.

In the following sections, we describe how we combine our different datasets and different improvements from previous sections to build one massive multilingual machine translation model, NLLB-200 covering 202 languages. We compare the performance of NLLB-200 on Flores-200, with both automated metrics and human evaluation results and demonstrate we get state-of-the-art performance. We analyze toxicity in NLLB-200 generations and study the extent to which data and toxicity filtering methods can help reduce it. We also compare the performance of NLLB-200 on Flores-101 and several common MT benchmarks and show NLLB-200 generalizes well and achieves competitive performance. Next, we discuss model distillation and how some of our distilled models are providing translations for Wikipedia. Following that, we discuss transliteration and dialectal translations, and where multilingual systems have room to improve these. Finally, we discuss the environmental impact of NLLB-200 and the implications of compute-efficient decisions in scaling.

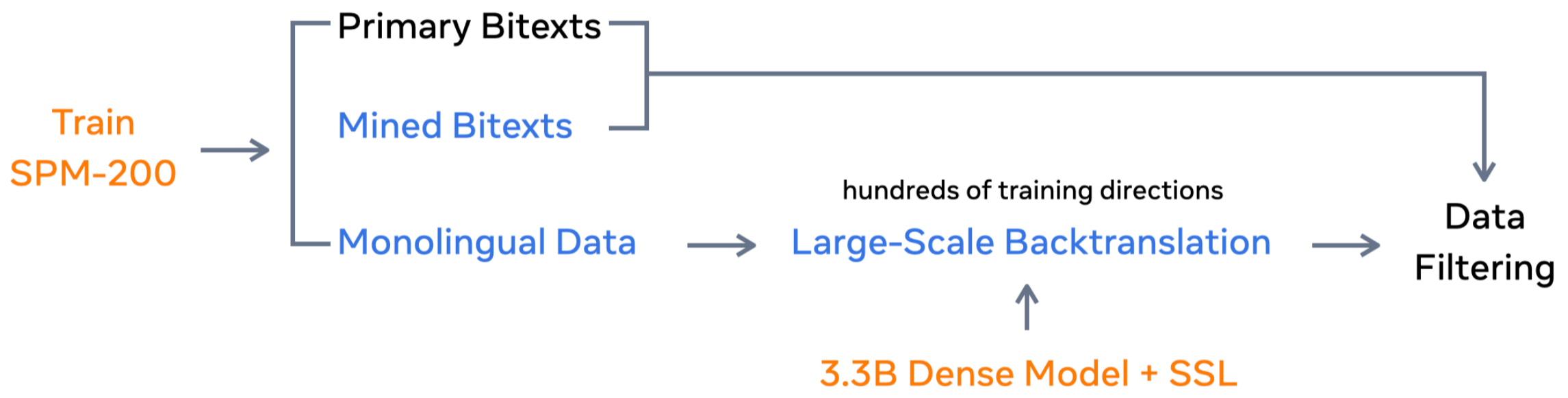

Figure 33: **Final Dataset Construction.** We depict how we combine together various data sources, data augmentation and data filtering techniques. The datasets created in our effort are shown in blue, and models in orange.

## 8.1 Preparing the Data

We describe how we aggregate various data sources discussed in the previous sections and augment with additional data through backtranslation. We experimentally demonstrate the advantages of leveraging large-scale mined and backtranslated data to significantly improve the performance of low-resource translation. Our overall process is summarized in Figure 33.

### 8.1.1 Training a Tokenizer for 200+ languages

To represent the 200+ languages of No Language Left Behind, we trained a new SentencePiece (SPM; Kudo and Richardson, 2018) tokenizer. To train this SentencePiece model, we sample a total of 100M sentences from primary bitext corpora. Given that most of the languages in NLLB are low-resource languages (150), uniform sampling would over-represent high-resource languages and under-represent low-resource languages, leading to too much fragmentation of low-resource language text. To mitigate this, we apply temperature sampling (with temperature $T = 5$), which effectively downsamples high-resource languages and upsamples low-resource languages. This results in a more balanced distribution of samples over all languages.

To validate the quality of the SPM, we first examine the rate of unknown tokens (`<unk>`) for each language. We observe that even after using a high temperature for sampling, certain languages such as `zho_Hans`, `zho_Hant` and `yue_Hant` had higher `<unk>` error rates, due to the very large character set of their scripts. To compensate, we further upsample those specific languages by a factor of 5 during training. With these modifications, the `<unk>` error rate for all languages is below 1%. Another important factor for quality is the *tokenization rate*, or the average number of tokens per sentence for each language (Mielke et al., 2021). Since SentencePiece identifies subword units based on language perplexity (roughly, frequency), underrepresented languages tend to have a higher tokenization rate than high-resource ones, leading to a near character-based model for those languages. This makes modeling more challenging, especially for long range dependencies and for synthesizing words from near character-level tokens. Based on the above two factors, we choose a vocabulary of size 256,000 for our SentencePiece model to allow for enough capacity to represent the wide spectrum and large number of languages we cover. As we achieve reasonable tokenization quality with a vocabulary size of 256k, we do not train SentencePiece models with even larger vocabulary sizes (e.g. 384k or more), as a larger vocabulary size would significantly increase the number of model parameters.

To evaluate with spBLEU, we use this SPM-200 as the tokenizer to better support the languages of Flores-200. We open source this SentencePiece model along with the Flores-200 dataset.

#### 8.1.2 Datasets

To train our systems, we leverage three different types of data, as listed below:

**Primary Bitext.** We use a set of publicly available parallel corpora from a variety of sources, including NLLB-Seed (Section 4.2). We add a total of 661 sets of primary bitext data. We choose all English-centric sets when available and also add non- English-centric pairs if they have a low resource language either as source, target, or both. Table 52 in Appendix E.1.1 provides further information on the list of public bitext corpora we used for training. We use the term Primary to refer to this type of data.

**Mined Bitext.** We use bitext corpora retrieved by large-scale *bitext mining*, as detailed in Section 5.3. We add mined data for a total of 784 directions. These include all the English-centric directions and a subset of non-English-centric directions. Non-English-centric mined data is effective in improving performance of multilingual translation systems (Fan et al., 2020). However, 200+ languages implies more than 40,000 non-English-centric pairs. Adding all the pairs could be detrimental, as some pairs will not have high quality mined bitexts. To subselect based on projected quality, we first pick the directions which have `xsim` error rate under 5. As a further restriction, we add mining data primarily for pairs containing low-resource languages within a given language family or a geographical region. This is an imperfect approximation to ensure improved transfer learning between similar languages. In Appendix E.1.3, we share the full list of bitext mined pairs used for training. We use Mined to refer to this type of data.

**Monolingual Text.** Details about our monolingual data can be found in Section 5.2. We use monolingual data for a total of 192 languages. This data is used for self-supervised learning and for generating backtranslated data (described in the next section). We use Monolingual to refer to this type of data.

#### 8.1.3 Large Scale Backtranslation

Backtranslated data provides a form of weak supervision which is crucial for improving translation performance of low-resource languages. As we observed in Section 6.4.1, combining backtranslation data generated from multiple sources improves performance of a translation model due to increased backtranslation diversity. Following this result, we generate backtranslated data from two models: **(1)** a multilingual neural machine translation model (MmtBT) and **(2)** a set of bilingual statistical machine translation models (SmtBT). We next describe how we chose the language pairs for backtranslation.

**Identifying Backtranslation Directions.** While effective, backtranslation is computationally expensive as it requires training massively multilingual as well as bilingual models and generating translations for up to tens of millions of sentences per direction. We describe how we identify which translation directions would benefit the most from augmented backtranslation.

We first train a baseline multilingual neural machine translation model (3.3B parameters, dense) on a dataset composed of all the primary bitext pairs and English-centric mined pairs. We train this model in a multitask learning setup, comprised of both MMT and SSL tasks, to produce the strongest possible model. We find that it is extremely important to backtranslate with the best possible model, as the quality of generated backtranslations is highly correlated to the performance of the model used to generate BT data (see Figure 43 for details). In Section 6.3, we observed that training on self-supervised objectives in addition to the MMT task in a multitask setup improves performance particularly when trained on Primary+Mined data (see Section 8.2.1). Improvements on `xx-eng_Latn` directions are more significant compared to `eng_Latn-xx` directions for low-resource languages. This is an added advantage for a model to generate backtranslations for `eng_Latn-xx`.

After training the above baseline model, we select the subset of languages for which we generate SmtBT data. We select every language `xx` for which the baseline model achieves spBLEU< 10 on `eng_Latn-xx` directions or spBLEU< 15 on `xx-eng_Latn` directions on Flores-200 `dev` set. These thresholds are chosen to keep the number of backtranslated directions manageable given computational constraints, and are also informed by previous preliminary experiments which showed that gains from using SmtBT were concentrated in directions on which the baseline model gave poor performance. This is usually true for very low and underserved languages. For each of the selected languages we then produce SmtBT data both in and out of English.

Finally, we identify the directions for which to obtain MmtBT data. As neural models are particularly effective for high-resource scenarios, we increase the threshold to capture a wider range of languages. We select every direction into and out of English where the Flores-200 `dev` set performance of the baseline model is below 30 spBLEU. Overall, using these criteria, we selected **76** English-centric directions for backtranslation through the SmtBT pipeline and **261** directions through MmtBT.

**Scaling Backtranslation Generation.** To perform MmtBT on a total of 261 directions with a 3.3B-parameter dense neural model, we leverage the model inference/generation framework in `fairseq` (Ott et al., 2019). For SmtBT, we backtranslate 76 directions using the same MOSES setup described in Section 6.4.1. This setup consists of individual CPU-bound bilingual models, and the cost of scaling is linear in the number of directions. By optimizing our pipelines to improve GPU/CPU utilization, we improve efficiency of this expensive process to some extent.

### 8.1.4 Filtering Strategy

The addition of data from backtranslation and mining can yield considerable gains in model performance. However, these processes typically yield sentence pairs that are much noisier than human-translated data. To benefit from such data while at the same time limiting the negative effects of noise, it is necessary to perform a series of *bitext filtering* steps. Corpus filtering has been used for many years in machine translation (Koehn et al., 2018, 2019, 2020). Our filtering pipeline performs several types of checks meant to determine whether a given data point is unlikely to be a real translation pair. We can divide these into several families.

**LASER Filtering.** The LASER filter operates on mined data and can remove bitext pairs whose LASER score falls below a given threshold (see Section 5.3). For all our mined bitext filtering we apply a threshold of 1.06 (Schwenk et al., 2021b).

**Length Filtering.** The *length-based* filter identifies sentences that are not within provided maximum and minimum length ratios. It also filters out sentence pairs with highly skewed length ratios, as such pairs are typically unlikely to represent real translations.

The concept of *length* is not straightforward when working with a massively multilingual model due to the inherent differences in sentence length distributions across different languages. In order to account for this phenomenon, we measure sentence lengths based on Unicode code points, and apply a language-specific correction factor. For a given language $\ell$, the factor is computed as $\alpha_\ell = \frac{N_{\texttt{eng_Latn}}}{N_\ell}$, where $N_\ell$ is the total length of the Flores-200 validation set for language $\ell$. Concretely, multiplying all lengths by this factor allows us to express a single length threshold in terms of typical English-language character lengths, and have a length discount or penalty applied for languages with generally shorter or longer sentences respectively. For mined and backtranslated data we filter out sentences with length ratios above 9.0. For backtranslations we additionally filter out sentences of length below 15 (corresponding to about three words in English) as we empirically found such short sentences to be a source of noise.

**LID Filtering.** Another important step is to discard pairs whose sentences do not appear to be written in the expected languages. This can be performed automatically using language identification (Section 5.1), with thresholds chosen appropriately based on the reliability of LID scores for each given language.

**Toxicity Filter.** Based on the techniques proposed in Section 7.3, we implement a toxicity filter. This removes sentence pairs which have *toxicity imbalance*, i.e., when the difference in number of toxic items detected in source and target is above a certain threshold. An alternative mode of operation supported by the filter is to remove all pairs in which one or both sentences contain toxic elements above a given threshold, regardless of their relative difference. Further details are available in Section 7.3.3.

**Deduplication.** Lee et al. (2021) demonstrates that training data deduplication is critical for large language model training. Deduplication is important for machine translation as well, so in the final step of the filtering pipeline we remove duplicates. In order to determine if two texts are duplicates, we apply a normalization process which removes punctuation and non-printing characters, then replaces all digits. The filtering can remove *pair duplicates*, defined as cases where two data points have identical source and target; or *source/target duplicates*, i.e. data points that have the same source side but might have different targets and vice versa. *Source deduplication* is useful for backtranslated data to catch cases where the backtranslation model decodes to the same sequence regardless of its input. *Target deduplication* is useful for Mined, where same source sentences are aligned with multiple target sentences.

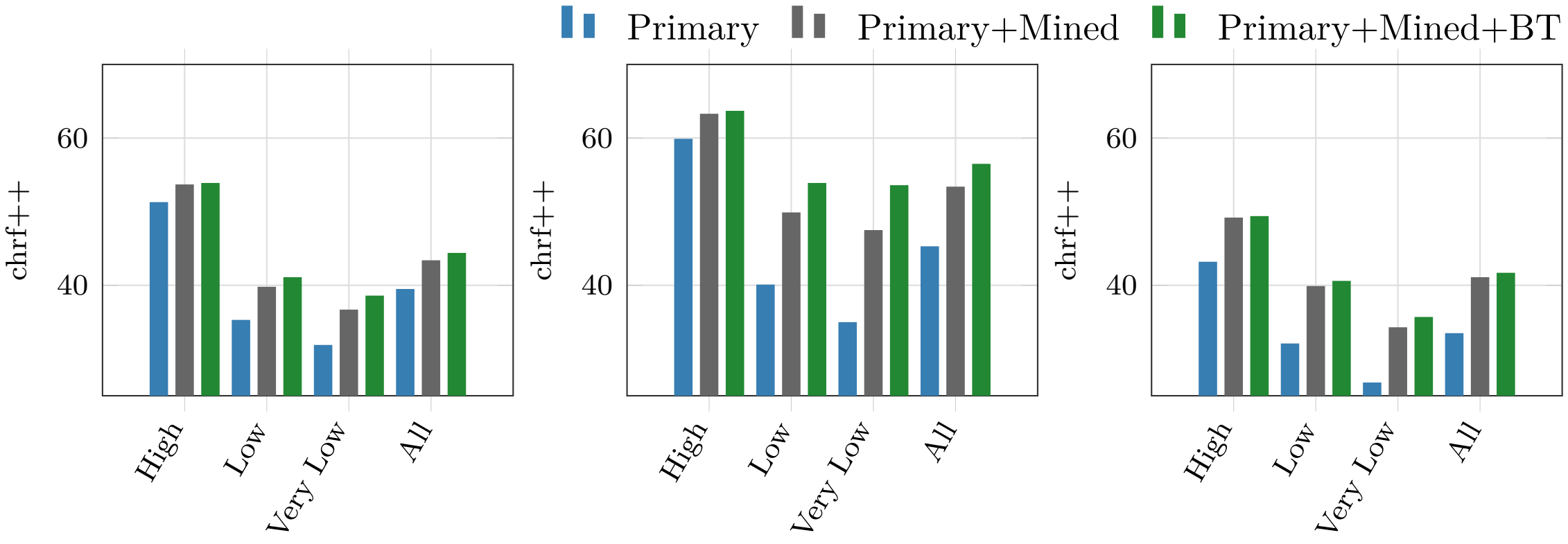

(a) eng_Latn-xx pairs (b) xx-eng_Latn pairs (c) xx-yy pairs

Figure 34: **Comparing Model Performance when trained on data from various sources.** We observe significant improvements on adding Mined and MmtBT+SmtBT backtranslated data for all type of language pairs and resource levels.

The effect of filtering is discussed in detail in Section 7.3.3, where we experiment with training bilingual models for six directions on filtered and unfiltered data, comparing their performance as well as the amount of hallucinated toxicity produced. The results in Table 27 confirm that filtering leads to an improvement in performance as well as a reduction in hallucinated toxicity. Our filtering configuration for various data sources is made available along with our other training scripts.[35]

#### 8.1.5 Effect of using Different Data Sources on Performance

We expect to have cumulative benefits by combining the different sources of data. We empirically investigate this hypothesis in this section.

**Experimental Setup.** We train dense 3.3B Transformer encoder-decoder models with model dimension 2048, FFN dimension 8192, 16 attention heads and 48 layers (24 encoder, 24 decoder) for these data ablation experiments. We train these models on three sets of data: **(1)** Primary, **(2)** Primary+Mined, and **(3)** Primary+Mined+MmtBT+SmtBT to compare the cumulative improvements coming from adding each source of data. All models are trained for a total of 300k iterations and we report the results with best chrF++ score checkpoints.

**Results.** In Figure 34, we demonstrate the impact of adding different data sources over Primary data. We aggregate results over language pair type and resource level. We observe that across all language pairs, performance improves significantly by adding Mined data and further by adding MmtBT+SmtBT backtranslated data. Focusing our observation on resource levels, we observe that low-resource languages improve more compared to high-resource languages. This is not surprising, as high-resource languages already have significant amounts of Primary bitext data publicly available.

35. https://github.com/facebookresearch/stopes

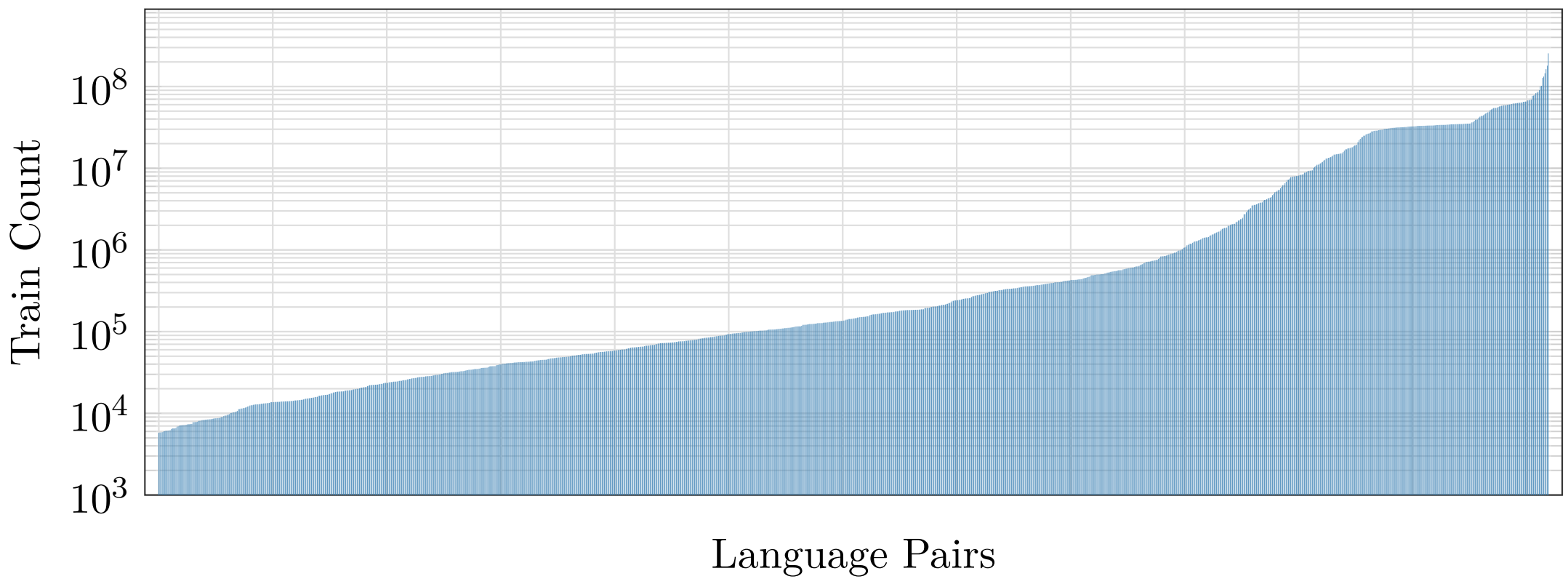

Figure 35: **Distribution of Amount of Training Sentence Pairs across 1220 language pairs in our dataset.** We observe that the majority of pairs have fewer than 1M sentences and are low-resource.

**Impact of Mining and Backtranslation on Very Low-Resource Languages.** Looking deeper at the results, we investigate how mined and backtranslated data sources impact very low-resource languages. We define *very low-resource* as languages with fewer than 100K unique sentence pairs across all language pairings available in public bitext corpora, with 84 total. On aggregate, our proposed techniques of mining and backtranslation improve low-resource and very low-resource language directions significantly (see Figure 34). Most prominently, very low-resource into English directions improve by +12.5 chrF++ with mined data and +6.1 chrF++ with additional BT data, with an overall improvement of +18.6 chrF++.

Similarly, we observe that out of English directions improve by +4.7 chrF++ when adding mined data and +1.9 chrF++ when adding backtranslated data, with an overall improvement of +6.6 chrF++. For non-English-centric pairs, we see an improvement of +7.5 chrF++ when adding mined data and +1.4 chrF++ when adding backtranslated data, with an overall improvement of +8.9 chrF++. These results show that our improvements in bitext mining and backtranslation increase the data quantity as well as quality for low-resource languages that are often underserved or excluded by existing translation systems.

#### 8.1.6 The 200 Language Dataset

We observed the benefits of both mining and backtranslation on low-resource languages. Combining all the sources of data, we prepare our final dataset, covering 202 languages.[36] The dataset comprises primary bitext for 661 language pairs, mined bitext for 784 language pairs, and 261 directions of backtranslated bitext. In total, there are **1220** language pairs or 2440 directions (`xx-yy` and `yy-xx`) for training. These 2440 directions sum to over **18B** total sentence pairs. Figure 35 displays the distribution of samples across the 1220 language pairs — the majority of the pairs have fewer than 1M sentences and are low-resource directions.

36. Two languages among the 204 in Flores-200, `arb_Latn` and `min_Arab`, have no available training data and hence we did not include them in the model training dataset.

## 8.2 Preparing the Model

In the previous section, we discuss how we improve data quantity and quality through mining, backtranslation, and filtering, leading to significant gains in model performance on low-resource languages. In this section, we discuss how we scale and adapt our model architecture and training procedure to build multilingual machine translation models for more than 200 languages and thousands of language directions. Training large models in a massively multilingual setting is a challenging problem due to the extreme data imbalance between language pairs as shown in Figure 35 and varying levels of translation difficulty. Learning objectives in the multilingual setting have complex and unknown dynamics and often compete with each other due to gradient interference (Wang et al., 2020c). Low-resource language pairs quickly overfit while high-resource language pairs usually benefit from longer training. Overall, these conflicting training dynamics make it a difficult optimization problem.

We addressed some of these challenges in Section 6.2 by showing how Sparsely Gated Mixture of Expert models with different regularization strategies and curriculum learning help improve the performance of massively multilingual machine translation models, especially for low-resource languages. In Section 6.3, we demonstrated how monolingual data can be leveraged to improve multilingual machine translation via self-supervision in the form of an additional denoising autoencoder task during training. In Section 6.4, we saw another way of leveraging monolingual data through large-scale backtranslation.

We now apply these strategies on the full training dataset as described in the previous section. First, we analyze the benefits of self-supervised learning (SSL) with the denoising autoencoder (DAE) task when training with and without backtranslated data. This helps us understand whether SSL helps further on top of mining and backtranslation, since all the approaches leverage the exact same monolingual data. Next, we apply the most promising regularization and curriculum learning strategies from Section 6.2 and train Sparsely Gated Mixture of Expert (MoE) models on the full dataset. We analyze the impact of MoE layer frequency in the model. With the best MoE layer frequency and regularization strategy, we then analyze the impact of introducing language pairs with a curriculum, based on the overfitting properties of each language pair. Based on these experiments, we propose the model architecture and training recipe to build our final model, NLLB-200, a massively multilingual machine translation model covering 202 languages and capable of translating 40k+ language directions.

### 8.2.1 Does Self-supervised Learning help on top of Mining and Backtranslation?

Using large scale *bitext mining* as discussed in Section 5.3, we already leverage a subset of the monolingual sentences in low-resource languages that can be successfully mapped to sentences in other languages. As detailed in Section 6.3, self-supervised learning objectives can help further leverage monolingual data to improve performance on low-resource languages (Bapna et al., 2022; Liu et al., 2021a; Ma et al., 2021). We saw in Section 8.1.3 that self-supervised learning brings strong improvements to the multilingual neural machine translation model trained on primary bitext and mined data, and that model is then used to generate *backtranslation* data. It is important to note that the same monolingual data is again used for backtranslation generation. Low-resource translations improve significantly as we add mined bitext and backtranslated bitext, as observed from the results in Section 8.1.5. Here we show some experiments and associated findings to understand how much more improvement self-supervised learning can bring when combined with training on mined as well as backtranslated data.

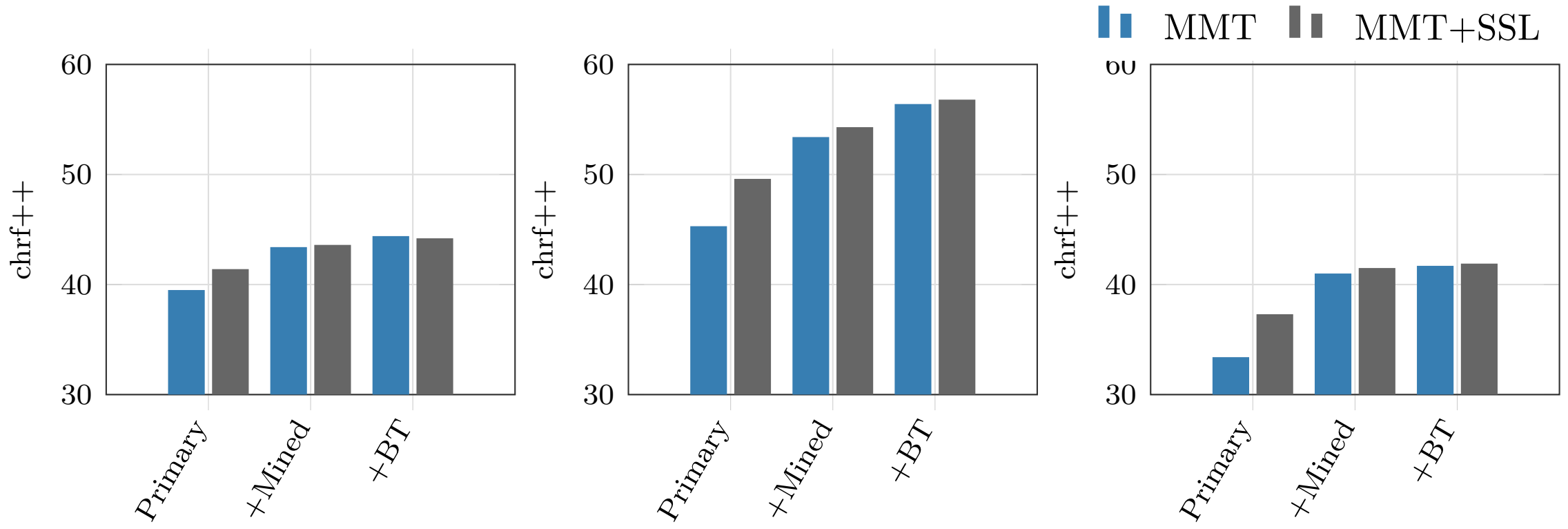

(a) `eng_Latn-xx` pairs (b) `xx-eng_Latn` pairs (c) `xx-yy` pairs

Figure 36: **Comparison of only MMT vs MMT+SSL Multitask Performance** on aggregate over Into English and Out of English directions, when trained on different sources of data. We compare on chrF++ scores. We observe that as we add more and more data from mining and/or backtranslation, we start to see diminishing improvements from the SSL task.

**Experimental Setup.** We use the setup similar to the experiments in Section 8.1.5 on different sets of data and train with and without self-supervised (SSL) multitasking. The datasets used are **(1)** Primary, **(2)** Primary + Mined, and **(3)** Primary + Mined + MmtBT + SmtBT. We follow a similar setup as in Section 6.3 for the multitask training and train 3.3B dense models for 300,000 updates. We compare the models on the best chrF++ checkpoints.

**Results.** In Figure 36, we observe that when using the additional self-supervised denoising autoencoder task (DAE) on the monolingual data in addition to the multilingual machine translation (MMT) task, the improvements decrease as we add Mined data, and become negligible once we add MmtBT+SmtBT backtranslated data. This demonstrates that mining and backtranslation provide good quality augmented bitexts from monolingual data, which improves low-resource translation. When combined with backtranslation, MMT+SSL multitask training begins to perform worse on `eng_Latn-xx` directions with no improvements on `xx-eng_Latn` and `xx-yy` directions. Based on these results, for our final model training, we do not use SSL and simply train on the MMT task. That said, there certainly could be situations when it is not feasible to get enough good quality mined or backtranslated data and some form of self-supervised learning on better quality and/or relatively larger monolingual corpora could be promising. The best strategy to utilize monolingual data is an active field of study and future work could reveal a deeper understanding of the relative benefits of mining, backtranslation, self-supervised training and the best ways to combine them.

### 8.2.2 Scaling Model Architecture

As we prepare to train on the final 202 language dataset comprising of over 18B sentence pairs and 2440 language directions, we need to scale up our models to ensure we have enough capacity to model all languages and language directions well. We increase our dense model size moderately to 3.3B parameters. Beyond that, we scale model size using the conditional compute strategy of Sparsely Gated Mixture of Expert (MoE) models for two reasons: **(1)** Adding expert capacity helps high-resource languages due to increased model capacity and also helps low-resource languages by minimizing interference with unrelated languages. **(2)** The computational overhead (FLOPs) of Sparsely Gated MoE models does not increase proportional to the number of parameters. However, large models, especially MoE, are prone to rapid overfitting on low-resource directions, so we apply the regularization and curriculum learning strategies discussed in Section 6.2 to optimize the complex training dynamics while training 2440 different language pairs with varying resource levels, varying difficulty and varying relatedness to other language pairs.

**Sparsely Gated MoE Layers.** We now describe how we incorporate Sparsely Gated MoE layers into our models. By default, existing literature (Artetxe et al., 2021; Fedus et al., 2022; Lepikhin et al., 2020) replaces the Feed Forward Network (FFN) layer at every alternate Transformer block with an MoE layer. The number of experts chosen are dependent on how much we want to increase MoE model capacity and the nature of the task being trained. Scaling the number of experts in MoE models comes with an increase in distributed communication cost during training, due to the expensive `All-to-All` communication primitive, which is performed twice (*dispatch* and *combine*) for every layer on each model update (Lepikhin et al., 2020). Another disadvantage is that the `All-to-All` communication primitive scales sub-linearly with the distributed world size (Lepikhin et al., 2020). To improve communication efficiency of our models during training, we use 128 experts. This decision is based on the fact that we observe that 32 experts perform as well as 64 experts on the 53 language benchmark when the overall dropout is tuned separately for both settings in Section 6.2. Another approach to increase training efficiency is to reduce the frequency of MoE layers by placing MoE layers at wider intervals in the encoder and decoder of the model. We experiment with placing MoE layers at every $2^{nd}$, $4^{th}$ and $6^{th}$ layer of a model with 24 encoder and 24 decoder layers. We then pick the variant with the best efficiency-accuracy trade off.

**Choosing the Optimal Regularization Strategy.** Dropout (Srivastava et al., 2014) is a strong regularization technique to reduce overfitting in deep neural networks. In Section 6.2.2, we expand and improve upon overall dropout and design specific regularization techniques for Mixture of Expert (MoE) layers. Based on the results and analyses of the variants we tried, we choose the regularization strategy of EOM, which provides the best trade-off between performance/accuracy and efficiency. EOM requires 23% fewer FLOPs compared to CMR. Although CMR is slightly better in terms of performance, we optimize for training efficiency as opposed to marginal performance gains.

**Experimental Setup.** We compare MoE-128 models in three setups varying the frequency of placement of MoE layers: $f_{\mathrm{MoE}} \in \{2, 4, 6\}$. We use the EOM dropout strategy for all these models and keep all other parameters same. All the models are trained for 200k updates and we pick the checkpoint with the best chrF++ score.

| | eng_Latn-xx | | | | xx-eng_Latn | | | | xx-yy | Avg. |
|---|---|---|---|---|---|---|---|---|---|---|
| | all | high | low | v.low | all | high | low | v.low | all | all |
| $f_{\text{MoE}}$=2 | 44.6 | **54.3** | 41.2 | 38.8 | 56.0 | **63.8** | 53.2 | 52.1 | 41.4 | 47.3 |
| $f_{\text{MoE}}$=4 | **44.8** | **54.2** | **41.4** | **39.1** | **56.3** | **63.8** | **53.5** | **52.8** | **42.0** | **47.7** |
| $f_{\text{MoE}}$=6 | 44.6 | 54.0 | 41.2 | 39.0 | 56.1 | 63.7 | 53.4 | 52.7 | 41.7 | 47.5 |

Table 28: **Effect of Frequency of Insertion of MoE Layers.** We report chrF++ scores on Flores-200 dev set on different types of language pairs. For eng_Latn-xx and xx-eng_Latn we include all 201 pairs each. For xx-yy we randomly choose 200 directions. We observe that placing MoE layers with a frequency 4 provides the best performance across all types of pairs and overall average.

**Results.** We see in Table 28 that inserting MoE layers at an interval of every 4 Transformer blocks exhibits the best performance, in particular improving performance in very-low resource settings. In terms of training efficiency, the MoE model with $f_{\text{MoE}}$=4 is 28% more efficient than $f_{\text{MoE}}$=2.

#### 8.2.3 Designing an Optimized Training Curriculum

Agnostic to the model architecture, an optimal curriculum in a massively multitask setting helps to find a better local minimum and acts as a strong regularizer (Bengio et al., 2009; Lu et al., 2020). There are several types of curriculum learning strategies based on how we order tasks — in terms of their difficulty level or their convergence patterns. For massively multilingual machine translation, high-resource pairs continue to converge when trained longer, whereas low-resource pairs overfit relatively quickly given smaller training datasets and higher model capacity. We propose a simple strategy where we introduce high-resource directions first, and introduce pairs that overfit later on in the training process. Section 6.2.3 indicates the benefit of this strategy on low-resource and very low-resource directions. Based on observing when different directions overfit at different points during training, we experiment with how to optimally bucket language pairs to introduce at different *phases* during training.

**Experimental Setup.** We compare three curriculum learning strategies here: **(1)** no curriculum, **(2)** curriculum with naive bucketing based on training example counts, and **(3)** curriculum with bucketing into multiple phases based on observed overfitting when training with no curriculum. We train 128-expert MoE models with $f_{\text{MoE}}$=4 with each strategy, and set $T = 300k$, where $T$ is the total number of updates for the entire training.

In the naive bucketing curriculum of variant **(2)**, we divide language pairs into **(2a)** those with more than 9M training sentence pairs, and **(2b)** those with fewer than 9M sentence pairs. We begin training with the language pairs with more than 9M sentences (261 language pairs, including 128 low-resource ones). At 200k updates, we then introduce language pairs with fewer than 9M sentences (2179 language pairs, including 2010 low-resource ones) and train all the pairs for a total of $T$ updates. This threshold was chosen empirically based on the observed relationship between each language pair's training example count and number of updates before overfitting.

| | eng_Latn-xx | | | | xx-eng_Latn | | | | xx-yy | Avg. |
|---|---|---|---|---|---|---|---|---|---|---|
| | all | high | low | v.low | all | high | low | v.low | all | all |
| **(1)** No CL | 44.8 | 54.3 | 41.4 | 39.0 | 56.2 | 64.0 | 53.4 | 52.5 | 41.9 | 47.6 |
| **(2)** Naive CL | **45.4** | **54.9** | **42.0** | **39.8** | 57.2 | **64.5** | 54.6 | 54.4 | **42.6** | 48.4 |
| **(3)** 4-Phase CL | 45.2 | 54.6 | 41.8 | 39.5 | **57.6** | 64.2 | **55.3** | **55.6** | **42.6** | **48.5** |

Table 29: **Improvements from different Curriculum Learning (CL) Strategies.** We report chrF++ scores on Flores-200 dev set on different types of language pairs. For eng_Latn-xx and xx-eng_Latn we include all 201 pairs each. For xx-yy we randomly choose 200 directions. We observe that on average, the variant **(3)** 4-phase CL, performs best.

In the phased curriculum variant **(3)**, we first train the model with variant **(1)** no curriculum, and then divide language pairs into $n$ different buckets $b_0, b_1, \ldots, b_{n-1}$ based on when they start to overfit. Then we restart training, introducing a particular bucket $b_i$ at $T - k_i$ updates, where $k_i$ is the median number of updates after which all directions in bucket $b_i$ start to overfit. In our particular instantiation, we set $k_0 = 300k$, $k_1 = 130k$, $k_2 = 70k$, $k_3 = 30k$. Pairs in $b_0$ are introduced first at the start of model training, pairs in $b_1$ are introduced at step $T - 130k$, and so on. The exact composition of each bucket in the phased curriculum is explained in Appendix E.1.3. Briefly, there are 822 pairs, with 673 low-resource pairs in $b_0$. There are 506 pairs, with 449 low-resource pairs in $b_1$. There are 455 pairs, with 444 low-resource pairs in $b_2$. Finally, there are 657 pairs, with 612 low-resource pairs in $b_3$.

**Results.** From Table 29, we observe that using a curriculum i.e. variants **(2)** or **(3)**, is better on this massively multilingual dataset compared to using no curriculum, variant **(1)**. This is in contrast with our observation from Section 6.2.3, where we saw no benefits of curriculum learning on top of MoE EOM. This difference is likely due to training on a dataset with 4x more languages and thousands of more directions, the majority of which are low to very low-resource and are prone to overfitting even after strong regularization techniques are applied. eng_Latn-xx and xx-yy directions see improvements of >+0.6 chrF++ with the majority of improvements coming from low and very low-resource directions. On xx-eng_Latn we see +1 chrF++ improvement and most significantly on very low-resource(+1.9 chrF++). Next, we observe that, on average, the 4-phase curriculum strategy performs best among the two variants of curriculum we test. Compared to the **(2)** naive curriculum, the **(3)** phased curriculum shows significant improvements on low (+0.7 chrF++) and very low-resource (+1.2 chrF++) pairs for xx-eng_Latn directions. Both variants show comparable performance on eng_Latn-xx and non-English-centric directions. We analyze the effect of phased curricula in more depth in Section 8.5.2.

#### 8.2.4 The 200 Language Model: NLLB-200

Based on the experiments and analyses detailed in this section, we now summarize the best recipe we found to build the NLLB-200 model. Our final model is a Transformer encoder-decoder model in which we replace the Feed Forward Network (FFN) layer in every $4^{th}$ Transformer block with a Sparsely Gated Mixture of Experts layer containing 128 experts. We use model dimension 2048, FFN dimension 8192, 16 attention heads, 24 encoder layers and 24 decoder layers. We use Pre-LayerNorm (Xiong et al., 2020) as described in Section 6.1.1. We share the embedding weights of the encoder input embedding, decoder input embedding and decoder output embedding layers. We use an overall dropout of 0.3, attention dropout 0.1 and EOM with $p_{\text{eom}}$=0.2. The model has a total of 54.5B parameters and FLOPs similar to that of a 3.3B dense model.

**Training Details.** We train the model for 300k steps using the 4 phase curriculum described in Section 8.2.3. We use an effective batch size of 1M tokens per update. The maximum sequence length during training is 512 for both the encoder and the decoder. We use the Adam optimizer (Kingma and Ba, 2015) with $\beta_1 = 0.9$ and $\beta_2 = 0.98$ and $\epsilon = 10^{-6}$. For training efficiency, we use memory efficient FP16 for training as implemented in the `fairseq` library and also maintain Adam optimizer states in FP16 (Dhariwal et al., 2020). We linearly increase the learning rate from $10^{-7}$ to 0.002 for 8000 warmup steps and then follow the inverse square root learning rate schedule. The loss is cross-entropy with label smoothing of 0.1 (Szegedy et al., 2015). For balancing expert utilization, we use the additional load balancing loss as described in Section 6.2 with a weight of 0.01. The capacity factor of each expert is set to 2, i.e., each expert can process up to $2 \times T/E$ tokens, where $T$ is the number of tokens in the mini-batch and $E$ is the number of experts. During generation, we set the capacity to be equal to the batch size so that all tokens can be routed to a single expert if needed. This would be useful for low-resource pairs that might prefer a small subset of experts. Finally, the model is trained to accept the source language token as prefix for the source sequence and the target language token as the first token input to the decoder as explained in Section 6.1.1.

### 8.3 Results on Flores-200

We present results comparing NLLB-200's performance against existing state-of-the-art models on Flores-101 languages. We dive into performance on Flores-200 languages, focusing on English-centric, non-English-centric and zero-shot performance. We primarily show results with the chrF++ metric, but also list spBLEU for convenience. We follow this with a human evaluation on NLLB-200's translations on Flores-200. Finally, we discuss the prevalence of hallucinated toxicity in NLLB-200's generations.

#### 8.3.1 Performance on Flores-101 and Comparison to State-of-the-Art

We evaluate NLLB-200 on Flores-101 to compare its performance with state-of-the-art models. We use the same evaluation method as used for Flores-101, with the 101 language SentencePiece (SPM) tokenizer officially provided. In Table 30, we observe that NLLB-200 outperforms the nearest state-of-the-art by almost +7.3 spBLEU on average — a 44% improvement. We then compare with a few other state-of-the-art models such as Deepnet (Wang et al., 2022) and M2M-100 (Fan et al., 2020), which report scores on a subset of 87 languages of Flores-101. On this smaller subset, NLLB-200 again outperforms by +6.9 spBLEU on average. Overall, the results show that NLLB-200 improves upon state-of-the-art systems by a significant margin despite covering 200+ languages — twice as many languages (or more than 30k additional directions) compared to any previous work.

| | eng_Latn-xx | xx-eng_Latn | xx-yy | Avg. |
|---|---|---|---|---|
| 87 languages | | | | |
| M2M-100 | -/- | -/- | -/- | 13.6/- |
| Deepnet | -/- | -/- | -/- | 18.6/- |
| NLLB-200 | **35.4**/52.1 | **42.4**/62.1 | **25.2**/43.2 | **25.5**/43.5 |
| 101 languages | | | | |
| DeltaLM | 26.6/- | 33.2/- | 16.4/- | 16.7/- |
| NLLB-200 | **34.0**/50.6 | **41.2**/60.9 | **23.7**/41.4 | **24.0**/41.7 |

Table 30: **Comparison on FLORES-101 devtest**. We evaluate over full Flores-101 10k directions. We report both spBLEU/chrF++ where available. All spBLEU numbers are computed with Flores-101 SPM tokenizer. Scores for DeltaLM are taken from Flores-101 leaderboard. M2M-100 and Deepnet average is only over 87 languages that overlap with Flores-101, we also show NLLB-200 performance on that subset of languages. NLLB-200 outperforms previous state of the art models by a significant margin, even after supporting twice as many languages.

| | eng_Latn-xx | | | xx-eng_Latn | | |
|---|---|---|---|---|---|---|
| | MMTAfrica | M2M-100* | NLLB-200 | MMTAfrica | M2M-100* | NLLB-200 |
| hau_Latn | -/- | 4.0/- | **33.6/53.5** | -/- | 16.3/- | **38.5/57.3** |
| ibo_Latn | 21.4/- | 19.9/- | **25.8/41.4** | 15.4/- | 12.0/- | **35.5/54.4** |
| lug_Latn | -/- | 7.6/- | **16.8/39.8** | -/- | 7.7/- | **27.4/46.7** |
| luo_Latn | -/- | 13.7/- | **18.0/38.5** | -/- | 11.8/- | **24.5/43.7** |
| swh_Latn | **40.1**/- | 27.1/- | 37.9/**58.6** | 28.4/- | 25.8/- | **48.1/66.1** |
| wol_Latn | -/- | 8.2/- | **11.5/29.7** | -/- | 7.5/- | **22.4/41.2** |
| xho_Latn | 27.1/- | -/- | **29.5/48.6** | 21.7/- | -/- | **41.9/59.9** |
| yor_Latn | 12.0/- | 13.4/- | **13.8/25.5** | 9.0/- | 9.3/- | **26.6/46.3** |
| zul_Latn | -/- | 19.2/- | **36.3/53.3** | -/- | 19.2/- | **43.4/61.5** |

Table 31: **Comparison on FLORES-101 devtest on African Languages.** We compare to two recent works MMTAfrica and M2M-100* finetuned on MAFAND-MT dataset. We report spBLEU/chrF++ and bold the best score. NLLB-200 outperforms previous state-of-the-art by significant margins on most translation directions.

**Performance on African and Indian Languages** Next, we compare NLLB-200's performance for two specific language groups — African and Indian languages. Several recent works in NLP have focused on African languages (Abbott and Martinus, 2019; Adelani et al., 2022; Azunre et al., 2021c; Dabre and Sukhoo, 2022; Emezue and Dossou, 2021, 2020; Hacheme, 2021; Nekoto et al., 2020; Siminyu et al., 2021). Here we compare against two recent works: MMTAfrica (Emezue and Dossou, 2021) and Mafand-MT (Adelani et al., 2022). Mafand-MT uses an M2M-100 model finetuned on the MAFAND dataset. For MMTAfrica, we take the max score of their `BT` and `BT&REC` methods. In Table 31, we observe that NLLB-200 outperforms both models significantly on most of the `eng_Latn-xx` and all of xx-eng_Latn directions. On eng_Latn-yor_Latn and eng_Latn-swh_Latn, MMTAfrica is slightly better than NLLB-200. NLLB-200 improves significantly on all other directions as it benefits from multilingual transfer by handling 55 African languages, in addition to data and modeling improvements. In comparison, the other works train on only 6 to 10 African languages. In Table 53, we further compare NLLB-200's performance on non-English-centric African language directions.

| | eng_Latn-xx | | | | | xx-eng_Latn | | | | |
|---|---|---|---|---|---|---|---|---|---|---|
| | **(a)** | **(b)** | **(c)** | **(d)** | NLLB-200 | **(a)** | **(b)** | **(c)** | **(d)** | NLLB-200 |
| asm | -/ 6.9/- | -/-/- | -/-/- | -/**13.6**/- | 7.9/11.7/35.9 | 23.3/-/- | -/-/- | -/-/- | 24.9/-/- | **33.9**/-/57.8 |
| ben | -/20.3/- | 17.3/-/- | -/**23.7**/- | -/22.9/- | 19.4/22.1/50.0 | 32.2/-/- | 30.7/-/- | 33.6/-/- | 31.2/-/- | **38.7**/-/62.2 |
| guj | -/22.6/- | 22.6/-/- | -/26.6/- | -/**27.7**/- | 25.0/25.2/53.3 | 34.3/-/- | 33.6/-/- | 39.5/-/- | 35.4/-/- | **44.6**/-/66.6 |
| hin | -/34.5/- | 31.3/-/- | -/**38.8**/- | -/31.8/- | 34.6/36.7/57.3 | 37.9/-/- | 36.0/-/- | 42.7/-/- | 36.9/-/- | **44.4**/-/66.5 |
| kan | -/18.9/- | 16.7/-/- | -/**23.6**/- | -/22.0/- | 21.3/22.1/53.4 | 28.8/-/- | 27.4/-/- | 31.7/-/- | 30.5/-/- | **36.9**/-/61.0 |
| mal | -/16.3/- | 14.2/-/- | -/**21.6**/- | -/21.1/- | 17.1/18.3/51.6 | 31.7/-/- | 30.4/-/- | 33.4/-/- | 34.1/-/- | **39.1**/-/62.9 |
| mar | -/16.1/- | 14.7/-/- | -/**20.1**/- | -/18.3/- | 17.6/17.9/48.0 | 30.8/-/- | 30.0/-/- | 35.5/-/- | 32.7/-/- | **40.3**/-/63.8 |
| ory | -/13.9/- | 10.1/-/- | -/**22.7**/- | -/20.9/- | 15.1/16.9/45.7 | 30.1/-/- | 28.6/-/- | 30.3/-/- | 31.0/-/- | **41.6**/-/64.4 |
| pan | -/26.9/- | 21.9/-/- | -/**29.2**/- | -/28.5/- | 24.5/27.7/49.0 | 35.8/-/- | 34.2/-/- | 37.8/-/- | 35.1/-/- | **44.8**/-/66.3 |
| tam | -/16.3/- | 14.9/-/- | -/**20.6**/- | -/20.0/- | 19.8/19.8/53.7 | 28.6/-/- | 27.7/-/- | 31.2/-/- | 29.8/-/- | **36.8**/-/60.8 |
| tel | -/22.0/- | 20.4/-/- | -/**26.3**/- | -/30.5/- | 24.8/25.3/55.9 | 33.5/-/- | 32.7/-/- | 38.3/-/- | 37.3/-/- | **43.6**/-/65.5 |

Table 32: **Comparison on FLORES-101 devtest on Indian Languages.** We report BLEU (with default 13a Moses tokenizer)/BLEU (with IndicNLP tokenizer)/chrF++ where available, and bold the best score. **(a)** IndicTrans (Ramesh et al., 2022), **(b)** IndicBART (Dabre et al., 2021), **(c)** Google Translate, **(d)** Microsoft Translate. Numbers for **(d)** are taken from (Ramesh et al., 2022). NLLB-200 outperforms other translation systems on all the xx-eng_Latn directions. On eng_Latn-xx, NLLB-200 outperforms **(a)** and **(b)**, but performs worse compared to **(c)** and **(d)**.

In recent years, Indian languages have seen a lot of progress in low-resource multilingual NLP (Bhattacharjee et al., 2022; Dabre et al., 2021; Kumar et al., 2022; Ramesh et al., 2022). We compare NLLB-200's translation performance with **(a)** IndicTrans (Ramesh et al., 2022), **(b)** IndicBART (Dabre et al., 2021), and commercial translation systems such as **(c)** Google Translate, **(d)** Microsoft Translate. IndicTrans reports results using the default 13a Moses tokenizer from SacreBLEU (Post, 2018) for xx-eng_Latn and IndicNLP tokenizer[37] for eng_Latn-xx directions. IndicBART reports scores with the default 13a Moses tokenizer from SacreBLEU for all directions. We report the scores with both these variants as well as chrF++. From Table 32 we observe that NLLB-200 outperforms all the models significantly on the xx-eng_Latn directions. On eng_Latn-xx directions, NLLB-200 performs better than **(a)** and **(b)** but worse than commercial systems **(c)** and **(d)**. Overall, on an average over all directions, NLLB-200 outperforms all the above systems. NLLB-200's training dataset includes 25 Indian languages[38], which is almost twice the languages covered by **(a)** and **(b)**. The performance improvements can be attributed to more multilingual transfer, along with improved mined and backtranslated data quality for the Indian language family.

### 8.3.2 Performance on Flores-200

We present the performance of NLLB-200 on the full Flores-200 devtest set. For the 202 languages in NLLB-200, there are a possible **40,602** translation directions in Flores-

37. Available in https://github.com/anoopkunchukuttan/indic_nlp_library.
38. Languages recognized by the Indian Constitution and a few other recognized minority languages of India.

| | eng_Latn-xx | | | | xx-eng_Latn | | | | xx-yy | Average |
|---|---|---|---|---|---|---|---|---|---|---|
| | all | high | low | v.low | all | high | low | v.low | all | all |
| chrF++ | 45.3 | 54.9 | 41.9 | 39.5 | 56.8 | 63.5 | 54.4 | 54.4 | 35.6 | 35.7 |
| spBLEU | 27.1 | 38.3 | 23.1 | 21.3 | 38.0 | 44.7 | 35.5 | 35.6 | 17.3 | 17.5 |

| | xx-yy (supervised) | | | | xx-yy (zero-shot) | | | |
|---|---|---|---|---|---|---|---|---|
| | all | high | low | v.low | all | high | low | v.low |
| chrF++ | 39.7 | 43.9 | 39.3 | 38.6 | 35.4 | 46.3 | 34.6 | 33.3 |
| spBLEU | 20.3 | 24.3 | 19.9 | 20.0 | 17.2 | 28.3 | 16.4 | 15.3 |

Table 33: **Performance of NLLB-200 on FLORES-200 devtest set.** We report both chrF++ and spBLEU scores. All spBLEU scores are computed with our newly trained SPM tokenizer in Section 8.1.1.

200. We categorize our results first into high-resource, low-resource, and very low-resource directions, and then we further subdivide each of these into *out of English* (eng_Latn-xx), *into English* (xx-eng_Latn), and *non-English-centric* directions (xx-yy).

**Evaluation Setup.** We evaluate NLLB-200 on 201 eng_Latn-xx directions, 201 xx-eng_Latn directions, and 40,200 xx-yy directions — for a total of **40602** directions. There are 2,862 high-resource pairs and 37,740 low-resource[39] pairs, out of which 26,796 are also very low-resource.[40] Out of these, our training set contains parallel data for 2,440 directions, meaning 38,162 directions were never seen by the model during training (*zero-shot*). We present the results in Table 33 with both chrF++ and spBLEU metrics.

**English-Centric Performance.** We first discuss performance on English-centric directions. We observe that, on average, xx-eng_Latn directions perform much better than eng_Latn-xx directions — a general trend we have previously observed in Section 6. We hypothesize this is due to several reasons: **(1)** xx-eng_Latn directions all require the decoder to decode into the same language, and English is the majority language in the training dataset; **(2)** Due to abundant and high-quality monolingual data in English, bitext mining and backtranslation produce higher quality data on the English side, and; **(3)** The non-English side of Flores-200 are usually human translated, and might therefore not have as good fluency as naturally occurring text due to the effect of *translationese* (Zhang and Toral, 2019). From Table 33, we observe that high-resource eng_Latn-xx directions on average are +13.0 chrF++ better than low-resource ones. For xx-eng_Latn, the gap is smaller, with high-resource only +9.0 chrF++ better on average. This also demonstrates that into English directions perform better for low resource languages. One benefit of a model having better into English performance is that it can be used to improve out of English performance using backtranslation (Section 8.1.3).

We further break down the English-centric performance into Flores-101 languages and new languages added to Flores-200. On Flores-101 directions, NLLB-200 achieves 50.7

39. A direction is defined as low-resource if any one language in the pair is low-resource, otherwise it is considered high-resource.
40. A language is defined as very low-resource if it has fewer than 100k samples across all pairings with any other language in our dataset.

| | eng_Latn-xx | | xx-eng_Latn | | Average | |
|---|---|---|---|---|---|---|
| | low | v.low | low | v.low | low | v.low |
| Google Translate | **32.3/50.3** | **27.0/46.5** | 35.9/57.1 | 35.8/57.0 | 34.1/53.7 | 31.3/51.7 |
| NLLB-200 | 30.3/48.2 | 25.7/45.0 | **41.3/60.4** | **41.1/60.3** | **35.8/54.3** | **33.4/52.6** |

Table 34: **Comparison on 102 Low-Resource Directions on FLORES-200 devtest against commercial translation systems.** We evaluate on all English-centric low-resource directions that overlap between Flores-200 and Google's Translation API as of this writing. We report both spBLEU/chrF++ and bold the best score. We observe that NLLB-200 outperforms significantly on xx-eng_Latn and overall average.

chrF++ on eng_Latn-xx and 60.9 chrF++ on xx-eng_Latn directions. On the subset of the new languages (Flores-200 - Flores-101), NLLB-200 achieves 39.9 chrF++ on eng_Latn-xx and 52.6 chrF++ on xx-eng_Latn directions. The new languages in Flores-200 are all low-resource, and on average have significantly worse performance than the languages in Flores-101 — overall, translation systems still have lot of room for improvement.

**Non-English-Centric and Zero-Shot Performance.** We next focus on the non-English-centric performance of NLLB-200. The Flores-200 evaluation dataset has complete many-to-many support, meaning the majority of directions it covers are non-English-centric. Compared to models evaluated on Flores-101, Flores-200 supports evaluation on ∼30k more non-English-centric directions. We evaluate on 40,200 xx-yy directions, where NLLB-200 achieves on average 35.6 chrF++. This is significantly lower than performance compared to xx-eng_Latn(-21.2 chrF++) or eng_Latn-xx(-9.7 chrF++) because most directions are unsupervised and the majority of the pairs have languages which are not lexically or syntactically similar, making translation more challenging.

As previously mentioned, 38,162 directions in total in Flores-200 are never seen by NLLB-200. On these *zero-shot* pairs, the model achieves 35.4 chrF++. In contrast with the ∼2k *supervised* pairs, where the model achieves 39.5 chrF++, we see only a -4.1 chrF++ drop on average over the ∼38k zero-shot pairs. Zero-shot low and very low-resource directions are on average only around -5.0 chrF++ worse compared to supervised pairs. The model has reasonable zero-shot translation performance likely due to its massively multilingual nature. Following Fan et al. (2020), we added non-English-centric mined bitext data to our training set, which also contributes to improving zero-shot. Further, we carefully chose the mined bitext directions which have better aligned bitexts and are from languages which are similar to each other.

**Comparison against Google Translate.** We compare NLLB-200's performance against commercial translation systems like Google Translate. Since these systems are proprietary, fair comparisons are challenging as there exists little information about model architecture, training settings, or number of models. Thus, we provide these results only for approximate comparison. We focus our observation on low and very low-resource language directions that overlap between Flores-200 and the Google Translate API[41] — note that NLLB-200 covers far more languages. There are 102 directions which are low-resource and among those 30 are

41. https://cloud.google.com/translate/docs/languages

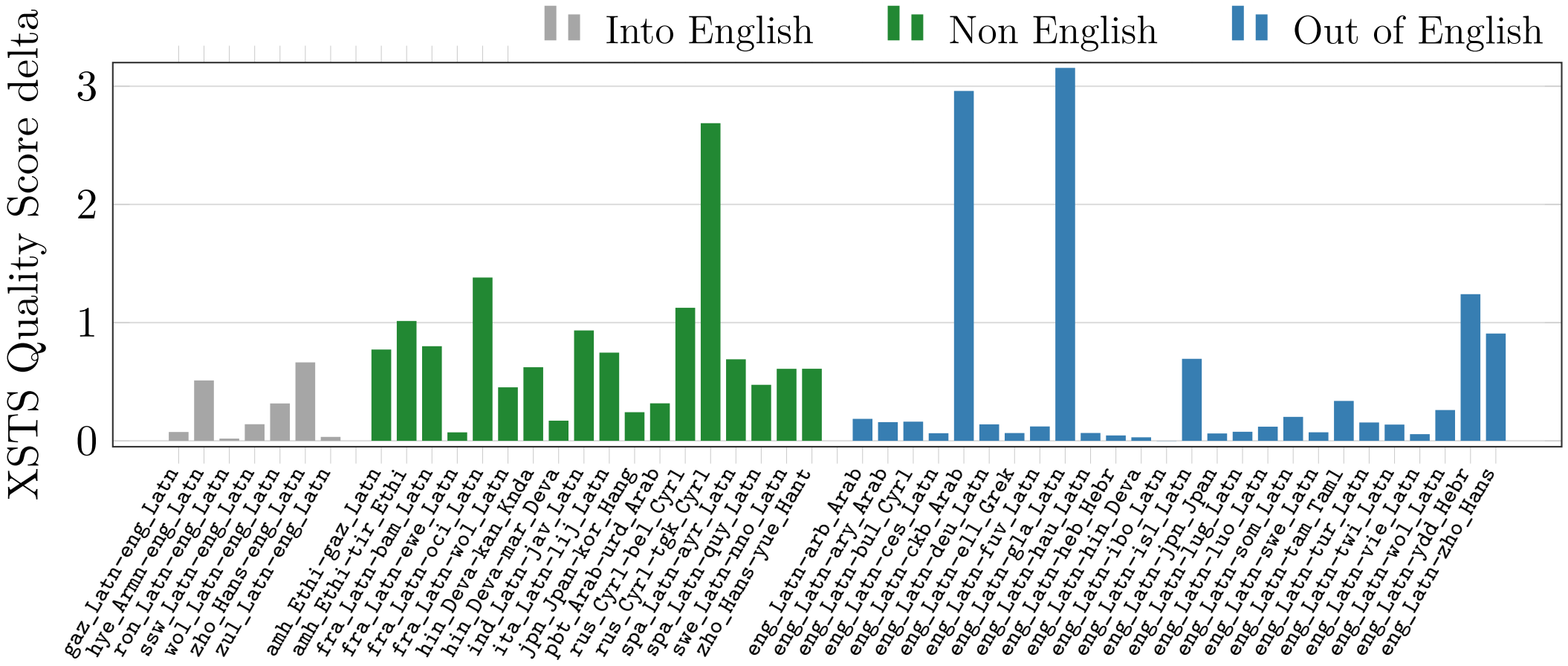

Figure 37: **XSTS Human Evaluation Quality Delta.** Delta between NLLB-200 model and dense baseline — both trained on our NLLB dataset. On average, NLLB-200 outperforms the dense baseline by more than 0.5 XSTS score (on a 5 point scale). Average performance improvement into English is 0.25, average performance improvement translation out of English is 0.44, and average performance improvement of non-English directions is 0.76.

very low resource. Table 34 demonstrates that NLLB-200 significantly outperforms on `xx-eng_Latn` directions with +3.3 chrF++ for both low-resource and very low-resource directions. On `eng_Latn-xx` directions, NLLB-200 lags behind by -2.1 chrF++ on low-resource and -1.5 chrF++ on very low-resource directions. Overall, NLLB-200 outperforms by +0.6 chrF++ on low and +0.9 chrF++ on very low-resource pairs despite covering 202 languages. We provide performance comparison on high-resource pairs and additional analysis in Appendix E.2.2.

#### 8.3.3 Human Evaluation

We next evaluate the quality of translations with professional human translators following the XSTS evaluation protocol and calibration methodology described in Section 7.

**Experimental Setting.** We compare two models: NLLB-200 and a baseline 3.3B parameter dense model. Both are trained on the same data and cover 202 languages. To understand model performance across a diverse set of language pairs, we evaluate **51** different translation directions. 26 out of English, 18 non-English, and 7 into English directions that represent many everyday translation needs that were described in our survey studies with low-resource language speakers (see Section 2). Each evaluation uses the 1,012 sentences of Flores-200 `devtest`.

**Results.** Results are shown in Figure 37. Overall, NLLB-200 achieves an average XSTS score of 4.22 averaged across all directions, and the dense model achieves an average XSTS score of 3.66. Compared to the baseline dense model, the performance of NLLB-200 is statistically significantly stronger. NLLB-200 achieves an average XSTS score of 4.09 for translation into English, 4.33 for translation out of English, and 4.12 for translation for non-English directions. Certain directions have a significant difference, such as `rus_Cyrl-tgk_Cyrl` and `eng_Latn-gla_Latn`.

### 8.3.4 Prevalence of Toxicity

Our goal is to produce high-quality safe translations for each of our 200 languages. Deviating in meaning from the source sentence is not desired behavior, but adding toxicity is worse. We use the toxicity detectors proposed in Section 7.3 (in particular, the 1+ toxic item detector) to evaluate the added toxicity in the NLLB-200 translations. We conduct our evaluation on the Flores-200 `devtest` set. Note that this data has an extremely low prevalence of true toxicity (only 3 toxic items in the `devtest` set) due to its Wikimedia domain.

Using NLLB-200, we evaluated translation outputs into and out of English. Potential added toxicity with the 1+ toxic item detector was detected in 130 out of 201 `eng_Latn-xx` directions (1,636 sentences) and 200 out of 201 `xx-eng_Latn` directions (555 sentences). Figure 44 in Appendix E.3 contains a more detailed breakdown. Overall, our analysis indicates a low prevalence of potential added toxicity in the translation of Flores-200. However, it does not mean that our models generate low toxicity in general. They could generate higher levels of toxicity in other domains (i.e., different topics, registers, or discourse types) than those found in Flores-200. These other domains may contain a higher prevalence of potential triggers, such as out of vocabulary tokens, infrequent words or word forms used with an unusual part-of-speech value (e.g., *a doff grandpa*) or in unusual constructions (e.g., *potbellied veterans*). In addition, we observe that there is a big difference in the potential added toxicity when translating out of English compared to translating into English. In out of English translations, we have more than 3 times as many potential added toxicity examples (0.80% on average) than into English (0.27% on average). Whether these differences are due to an over-detection by our toxicity lists or due to actual added toxicity by our models when translating particular translation directions is a direction for future work.

## 8.4 Out-of-domain Generalization: Performance on non-Flores-200 Domains

We next evaluate our model's generalization capability on some non-FLORES MT benchmarks and compare to existing works. We further analyze how NLLB-200 can be adapted to specialize in various domains and discuss the significant performance improvement that comes from in-domain finetuning.

Our goal in this section is to examine if we are developing a robust general-purpose MT system capable of translating in various domains. We first evaluate the capability of NLLB-200 to generalize on a wide selection of non-FLORES MT benchmarks from different domains (news, health, governmental, *etc.*). Then, we leverage our own NLLB-MD dataset (see Section 4.3) to validate the transferability of NLLB-200 to other domains by finetuning on small quantities of high-quality bitexts.

### 8.4.1 Public Benchmarks

We use publicly available and widely used MT evaluation benchmarks to assess the performance of NLLB-200 on domains other than the Wikimedia text of Flores-200, and to compare the performance of NLLB-200 to previous state-of-the-art models.[42] We select a total of 238 directions from 8 benchmarks. We describe each of these benchmarks and the set of selected directions below.

42. Note that training setups vary, and thus models are not directly comparable

**Flores(v1)**: with a total of 8 directions, the original Flores dataset (Guzmán et al., 2019) pairs four low-resource languages (khm_Khmr, npi_Deva, pbt_Arab, sin_Sinh) with eng_Latn in the Wikimedia domain.

**WAT**: we select 3 languages (hin_Deva, khm_Khmr and mya_Mymr ) paired with English (6 directions) from the WAT competition.

**WMT**: we evaluate on the 15 WMT languages selected in Siddhant et al. (2020). This set overlaps with the 10 languages selected in Wang et al. (2020a) and both are frequently used for benchmarking MMT models (Kim et al., 2021; Kudugunta et al., 2021). The 15 languages paired with English in this set are: (ces_Latn, deu_Latn, est_Latn, fin_Latn, fra_Latn, guj_Gujr, hin_Deva, kaz_Cyrl, lvs_Latn, lit_Latn, ron_Latn, rus_Cyrl, spa_Latn, tur_Latn and zho_Hans).

**IWSLT**: we select 24 directions from the IWSLT translation competition. With bitexts based on aligned TED talks, the selected directions come from different campaigns (see Table 55 in the appendix for more details on each direction).

**TICO**: sampled from a variety of public sources containing COVID-19 related content (Anastasopoulos et al., 2020), this dataset comes from different domains (medical, news, conversational, *etc.*) and covers 36 languages. We pair 28 languages with English for a total of 56 directions.

**Mafand-MT**: an African news corpus that covers 16 languages (Adelani et al., 2022). 8 languages are paired with English (hau_Latn, ibo_Latn, lug_Latn, luo_Latn, swh_Latn, tsn_Latn, yor_Latn, zul_Latn ) and 5 other languages are paired with French (bam_Latn, ewe_Latn, fon_Latn, mos_Latn and wol_Latn ) for a total of 26 directions.

**Autshumato**: an evaluation set for machine translation of South African languages (McKellar, 2017), it consists of 500 sentences from South African governmental data, translated separately by four different professional human translators for each of the 11 official South African languages. 9 of these languages are covered by NLLB-200: afr_Latn, eng_Latn, nso_Latn, sot_Latn, ssw_Latn, tsn_Latn, tso_Latn, xho_Latn and zul_Latn. There is no standard valid/test split, so we use the first half (250 sentences yielding 1000 pairs) for validation and the second half for testing following Fan et al. (2020).

**MADAR**: created by translating select sentences from the Basic Traveling Expression Corpus (BTEC) (Bouamor et al., 2018). This corpus covers dialects from 25 Arabic-speaking cities, in addition to English, French and Modern Standard Arabic (MDA). We map 16 out of these dialects to the 8 Arabic dialects in NLLB-200 (aeb_Arab, acm_Arab, acq_Arab, ajp_Arab, apc_Arab, ars_Arab, ary_Arab and arz_Arab ) and pair each with Modern Standard Arabic (arb_Arab). We use the MADAR shared task test split (corpus_6_test_corpus_26_test) for evaluation (Bouamor et al., 2019).

| | eng-xx | | xx-eng | |
|---|---|---|---|---|
| | Published | NLLB-200 | Published | NLLB-200 |
| khm | [b]**5.9**/- | 0.4/27.4 | [b]10.7/- | **16.8**/36.5 |
| npi | [c]7.4/- | **10.4**/39.0 | [c]14.5/- | **29.3**/54.8 |
| pbt | [b]9.3/- | **10.5**/34.3 | [b]15.7/- | **22.0**/46.8 |
| sin | [c]3.3/- | **11.6**/40.9 | [c]13.7/- | **23.7**/49.8 |

(a) Flores(v1)

| | eng-xx | | xx-eng | |
|---|---|---|---|---|
| | Published | NLLB-200 | Published | NLLB-200 |
| hin | [l]22.1/- | **27.2**/51.5 | [l]32.9/- | **37.4**/61.9 |
| khm | [l]43.9/- | **45.8**/42.3 | [l]27.5/- | **39.1**/61.1 |
| mya | [c]**39.2**/- | 23.5/31.5 | [c]**34.9**/- | 32.7/57.9 |

(b) WAT

Table 35: **Comparison to State-of-the-Art on FLORES (v1) (devtest) and WAT's Test Sets**. We report BLEU/chrF++ scores where available and bold the best score. Low-resource languages are underlined. In each direction, we display the best performing model from published work: [b] Tang et al. (2020), [c] Liu et al. (2020), and [l] Nakazawa et al. (2019)

| | eng-xx | | xx-eng | |
|---|---|---|---|---|
| | Published | NLLB-200 | Published | NLLB-200 |
| ces | [b]**26.5**/- | 25.2/50.6 | [d]**35.3**/- | 33.6/56.8 |
| deu | [a]**44.9**/- | 33.0/59.2 | [a]**42.6**/- | 37.7/60.5 |
| est | [a]26.5/- | **27.0**/55.7 | [a]**38.6**/- | 34.7/59.1 |
| fin | [a]**32.1**/- | 27.7/57.7 | [a]**40.5**/- | 28.8/53.7 |
| fra | [a]**46.7**/- | 44.2/65.7 | [a]**43.9**/- | 41.9/63.9 |
| guj | [d]**17.8**/- | 17.6/46.6 | [f]25.1/- | **31.2**/56.5 |
| hin | [f]25.5/- | **26.0**/51.5 | [f]29.7/- | **37.4**/61.9 |
| kaz | [i]15.5/- | **34.8**/61.5 | [i]**30.5**/- | 30.2/56.0 |
| lit | [a]17.0/- | **37.0**/63.9 | [a]**36.8**/- | 29.7/56.4 |
| lvs | [a]**25.0**/- | 21.3/50.8 | [a]**28.6**/- | 24.8/50.8 |
| ron | [a]41.2/- | **41.5**/58.0 | [h]**43.8**/- | 43.4/64.7 |
| rus | [a]31.7/- | **44.8**/65.1 | [a]39.8/- | **39.9**/61.9 |
| spa | [e]33.5/- | **37.2**/59.3 | [e]34.5/- | **37.6**/59.9 |
| tur | [a]**32.7**/- | 23.3/54.2 | [a]**35.0**/- | 34.3/58.3 |
| zho | [b]**35.1**/- | 33.9/22.7 | [a]**28.9**/- | 28.5/53.9 |

(a) WMT

| | eng-xx | | xx-eng | |
|---|---|---|---|---|
| | Published | NLLB-200 | Published | NLLB-200 |
| arb | [b]22.0/- | **25**/47.2 | [b]44.5/- | **44.7**/63.7 |
| deu | [k]25.5/- | **31.6**/57.8 | [k]28.0/- | **36.5**/57.5 |
| fra | [g]40.0/- | **43.0**/65.6 | [g]39.4/- | **45.8**/64.8 |
| ita | [b]38.1/- | **42.5**/64.4 | [b]43.3/- | **48.2**/66.5 |
| jpn | [c]19.4/- | **19.5**/21.5 | [c]19.1/- | **22.6**/46.1 |
| kor | [c]**22.6**/- | 22.5/27.9 | [c]24.6/- | **25.4**/48.0 |
| nld | [c]34.8/- | **34.9**/60.2 | [c]**43.3**/- | 41.0/60.9 |
| pes | [j]06.5/- | **15.5**/39.2 | [j]18.4/- | **42.3**/61.3 |
| pol | [j]16.1/- | **21.1**/48.3 | [j]18.3/- | **27.1**/48.2 |
| ron | [k]25.2/- | **29.4**/55.5 | [k]31.8/- | **42.0**/62.0 |
| rus | [j]11.2/- | **24.0**/47.0 | [j]19.3/- | **30.1**/51.3 |
| vie | [c]**35.4**/- | 34.8/53.7 | [c]36.1/- | **36.6**/57.1 |

(b) IWSLT

Table 36: **Comparison to State-of-the-Art on WMT & IWSLT Test Sets.** We report BLEU/chrF++ scores where available and bold the best score. Low-resource languages are underlined. In each direction, we display the best performing model from published work: [a] Kim et al. (2021), [b] Tang et al. (2020), [c] Liu et al. (2020), [e] Kudugunta et al. (2021), [f] Ramesh et al. (2022), [g] Provilkov et al. (2020), [i] Bojar et al. (2016). [j] Cettolo et al. (2014) and [k] Cettolo et al. (2017)

**Evaluation on Other Benchmarks.** We evaluate the translations' accuracy with BLEU, spBLEU, and chrF++ (choosing to match the evaluation methodology of the individual benchmarks). To measure BLEU, we first detokenize the hypotheses. Then, for each evaluation corpus and for each language, we conform to the tokenization or normalization used in the current state-of-the-art. Once the hypotheses and references are tokenized, we compute BLEU. In most languages, we pass the detokenized output to SacreBLEU (Post, 2018) and use the default `13a` Moses tokenizer. See Table 56 in the appendix for a breakdown of the pre-processing steps for each irregular evaluation direction.

### 8.4.2 Results

**FloresV1.** On the in-domain test sets of Flores(v1) in Table 35a, we outperform the state of the art (Liu et al., 2020; Tang et al., 2020) on all 8 directions but 1 (`khm_Khmr-eng_Latn`)

**WAT.** Next, in Table 35b, we evaluate NLLB-200 on 6 directions from WAT. The `mya_Mymr` and `khm_Khmr` test sets are part of the ALT corpus (news articles) and the `hin_Deva` test set comes from the news domain as well (IITB: Newswire). NLLB-200 outperforms the state of the art when translating into English with an average of 3.5 BLEU points. Translating from English is better by 3.5 BLEU points on average, excluding the drop on `eng_Latn-mya_Latn` which puts the average at -3 BLEU.

**WMT.** Similar to WAT in domain, WMT's test sets cover news articles. On this particular benchmark (see Table 36a), Kim et al. (2021) achieved accuracies that outperform the previous state of the art by a large margin. Compared to Kim et al. (2021), NLLB-200 improves on `eng_Latn-xx` directions by approximately 1.9 BLEU point. However, when translating into English, NLLB-200 scores worse on average by 0.6 BLEU points.

**IWSLT.** On the 12 high-resource languages paired with English in IWSLT (Table 36b), NLLB-200 outperforms state of the art on 21 out of 24 directions. Translating from English improves on average by 3.5 BLEU points and translating into English by 6.3 BLEU points.

**TICO.** Table 37 shows the scores of NLLB-200 and those of the best baseline trained in the original TICO paper (Anastasopoulos et al., 2020). We see a considerable gain in accuracy on low and high-resource languages alike. Additional results on TICO can be found in Table 58.

**MAFAND.** We compare in Table 38 the performance of NLLB-200 on the MAFAND-MT test set to that of the best M2M-100 model finetuned on the news domain with MAFAND data, as reported in Adelani et al. (2022). NLLB-200 outperforms the previous state-of-the-art on 14 out of the 26 tested directions, proving that the model can generalize well to other domains. Translating into English sees an average improvement of +4.0 BLEU points. However, translating into French (high-resource) is worse on average by -0.9 BLEU, with a significant drop in the performance of the `bam_Latn-fra_Latn` direction. Translating from English is worse on average by -0.6 BLEU and translating from French is worse on average by -4.4 BLEU.

**Non-English-Centric Evaluation on Autshumato and MADAR.** Additionally, we evaluate NLLB-200 on datasets with non-English-centric pairs; MADAR for Arabic dialects and Autshumato for African languages, in Table 57 and Table 58 of the appendix.

### 8.4.3 Effective Domain Adaptation with NLLB-MD

In the following, we study if NLLB-200 can effectively transfer to other domains and if it lends itself to the common strategy of single-task finetuning with small quantities of in-domain high quality translations (Adelani et al., 2022; Lee et al., 2022; Liu et al., 2021b; Tang et al., 2020).

| | eng-xx | | xx-eng | |
|---|---|---|---|---|
| | Published | NLLB-200 | Published | NLLB-200 |
| arb | 15.2/- | **34.1**/59.4 | 28.6/- | **49.6**/70.3 |
| fra | 37.6/- | **44.9**/64.4 | 39.4/- | **47.3**/65.4 |
| gaz | 0.6/- | **10.7**/44.0 | 2.1/- | **35.9**/57.2 |
| hin | 6.4/- | **46.2**/65.8 | 18.9/- | **58.0**/76.2 |
| ind | 41.3/- | **55.1**/74.8 | 34.9/- | **54.3**/73.5 |
| lin | 7.8/- | **24.6**/51.5 | 6.7/- | **33.7**/54.1 |
| lug | 3.0/- | **22.1**/48.6 | 5.6/- | **39.0**/58.2 |
| mar | 0.2/- | **16.1**/46.3 | 1.2/- | **44.3**/66.9 |
| pes | 8.5/- | **30.0**/55.6 | 15.1/- | **45.5**/67.5 |
| por | 47.3/- | **52.9**/72.9 | 48.6/- | **58.7**/76.5 |
| rus | 28.9/- | **35.7**/59.1 | 28.5/- | **41.2**/65.1 |
| spa | 48.7/- | **57.2**/74.9 | 46.8/- | **57.5**/75.9 |
| swh | 22.6/- | **34.1**/59.1 | 0.0/- | **49.6**/68.1 |
| urd | 2.8/- | **27.4**/53.3 | 0.0/- | **44.7**/66.9 |
| zho | 33.7/- | **42.0**/33.3 | 28.9/- | **37.6**/61.9 |
| zsm | 6.3/- | **52.4**/73.4 | 0.0/- | **58.8**/76.1 |
| zul | 11.7/- | **22.4**/55.1 | 25.5/- | **50.6**/68.4 |

Table 37: **Comparison on TICO**. We report BLEU/chrF++ scores, where available, from Anastasopoulos et al. (2020) and bold the best score. Low-resource languages are underlined.

**Experimental Setup.** We experiment with the NLLB-MD dataset (see Section 4.3). It provides high-quality translations in four domains (news, scripted formal speech (scripted), unscripted informal speech (chat) and health), translating from English to 6 languages (5 of which are low-resource). We hold 500 sentences in each language for testing, finetune on 2000 sentences, and use the remainder for validation. In each translation direction (into and out of English), we finetune NLLB-200 on that single task for 50 updates (15-20 epochs) with a learning rate of 5e-5 following an inverse square-root schedule after warming up for 10 updates. We consider two options for finetuning NLLB-200 for the new task: **(1)** finetuning with the original training objective (label-smoothed cross-entropy with an additional load balancing regularization term) (see Section 6.2) and **(2)** finetuning without regularization, thus, leaving the MoE's load distribution unconstrained.

**Results.** Figure 38 shows validation chrF++ scores in the chat domain tasks of the pretrained NLLB-200, the similarly finetuned model with load balancing (NLLB-200+FN+LB), and the finetuned model without load balancing (NLLB-200+FN).

On average, finetuning (FN+LB) improves the accuracy by +6.1 chrF++ points. The performance gain is more considerable when translating into high-resource languages (eng and rus) with an average +8.9 chrF++ points and an average +2.0 points when translating into the 5 low-resource languages in NLLB-MD. When switching off the load balancing regularization, NLLB-200+FN improves by +7.2 chrF++ and it is particularly interesting when translating into low-resource languages with an increase of +3.7.

We next finetune with our best strategy (NLLB-200+FN) on the other 3 domains of NLLB-MD and report chrF++ scores on the test sets in Figure 39. On average, by finetuning

| | eng-xx | | xx-eng | |
|---|---|---|---|---|
| | Adelani et al. (2022) | NLLB-200 | Adelani et al. (2022) | NLLB-200 |
| hau_Latn | **15.9/42.1** | 8.2/34.8 | **18.2/40.2** | 13.5/37.9 |
| ibo_Latn | **26.0/51.3** | 23.9/50.4 | **21.9/48.0** | **21.9**/46.1 |
| lug_Latn | 15.7/46.9 | **25.8/55.2** | 22.4/48.5 | **30.9/54.4** |
| luo_Latn | 12.0/39.4 | **14.0/40.4** | 14.3/38.3 | **15.9/38.4** |
| swh_Latn | 27.7/**57.2** | **30.7**/56.0 | 30.6/55.8 | **39.3/60.8** |
| tsn_Latn | **31.9/59.5** | 28.5/55.6 | 27.8/54.0 | **37.3/60.2** |
| yor_Latn | 13.9/**37.4** | **14.4**/36.3 | 18.0/41.0 | **24.4/46.7** |
| zul_Latn | **22.9/56.3** | 16.1/47.3 | 38.1/57.7 | **40.3**/59.7 |
| | fra-xx | | xx-fra | |
| | Adelani et al. (2022) | NLLB-200 | Adelani et al. (2022) | NLLB-200 |
| bam_Latn | **24.7/49.9** | 7.7/29.9 | **25.8/49.0** | 14.6/37.5 |
| ewe_Latn | **8.9/37.5** | 8.3/36.4 | 11.6/37.2 | **19.4/42.6** |
| fon_Latn | **7.4/28.5** | 3.4/21.8 | **9.9/28.9** | 8.9/28.7 |
| mos_Latn | 2.2/16.8 | **5.4/27.6** | 4.1/18.8 | **6.1/23.5** |
| wol_Latn | **12.7/35.8** | 9.1/29.9 | **11.5/35.3** | 9.5/30.2 |

Table 38: **Comparison to State-of-the-Art on MAFAND-MT's Test Set.** We report BLEU/chrF++ for NLLB-200 and the best BLEU and chrF++ from Adelani et al. (2022). The best scores in each direction are bolded. Low-resource languages are underlined.

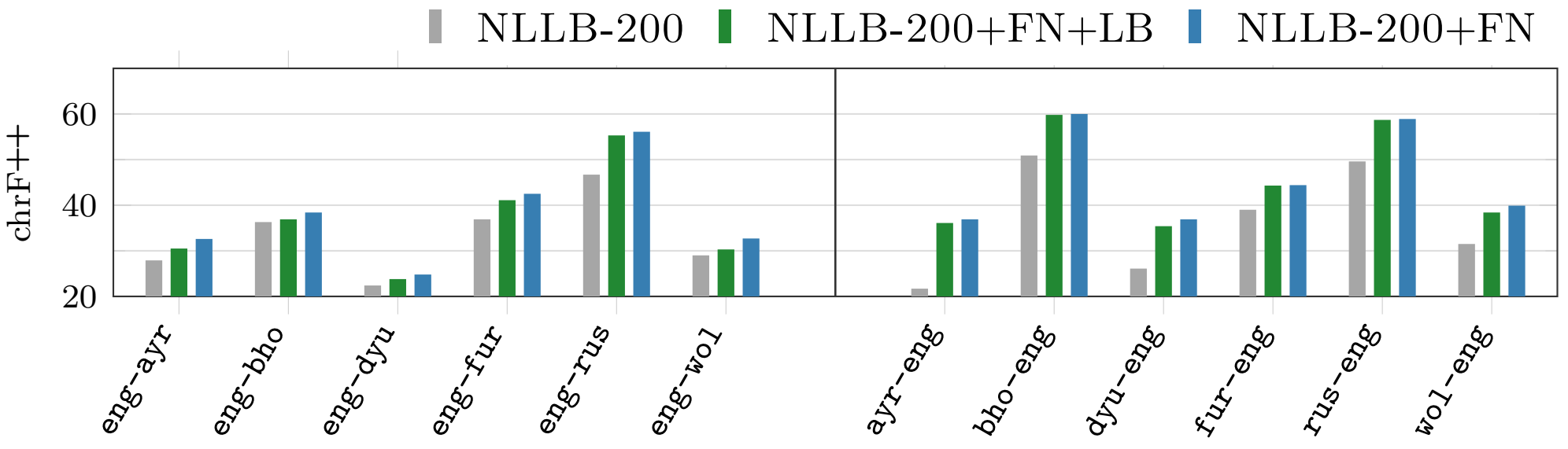

Figure 38: **Comparison of NLLB-200 with and without Finetuning on the 12 English-centric tasks of NLLB-MD.** NLLB-200+FN+LB and +FN refer to finetuning with and without load balancing (LB). We report accuracy in terms of chrF++ on the validation set.

NLLB-200, we can improve translation accuracy in new domains by +7.7 in chat, +3.1 in news, +4.1 in health and +5.8 in scripted (in terms of chrF++). These results are evidence of NLLB-200's transferability and adaptability to other domains.

The issue of finetuning sparsely activated large models has been raised in prior work (Artetxe et al., 2021; Fedus et al., 2022; Zoph et al., 2022). These large models are more prone to overfitting than their dense counterparts, and in some cases they performed poorly when finetuning (Artetxe et al., 2021; Fedus et al., 2022). Fedus et al. (2022) suggests increasing regularization with *expert dropout*, effectively applying stronger regularization to the expert parameters, while Zoph et al. (2022) combat overfitting by updating only a subset of model parameters. With MoE Expert Output Masking (EOM), NLLB-200 is heavily regularized and exhibits less overfitting on downstream tasks. We hypothesize that without load balancing, we are allowing the model to drop experts, practically activating a few that will be finetuned for the downstream task. This is particularly relevant when finetuning on a single-task for which NLLB-200 has learned to assign specific experts (see Section 8.5.1); adding load balancing loss when the mini-batches are not mixed will considerably shift this learned assignment. We leave the exploration of MoE finetuning strategies with added regularization, selective finetuning and relaxed optimization for future work.

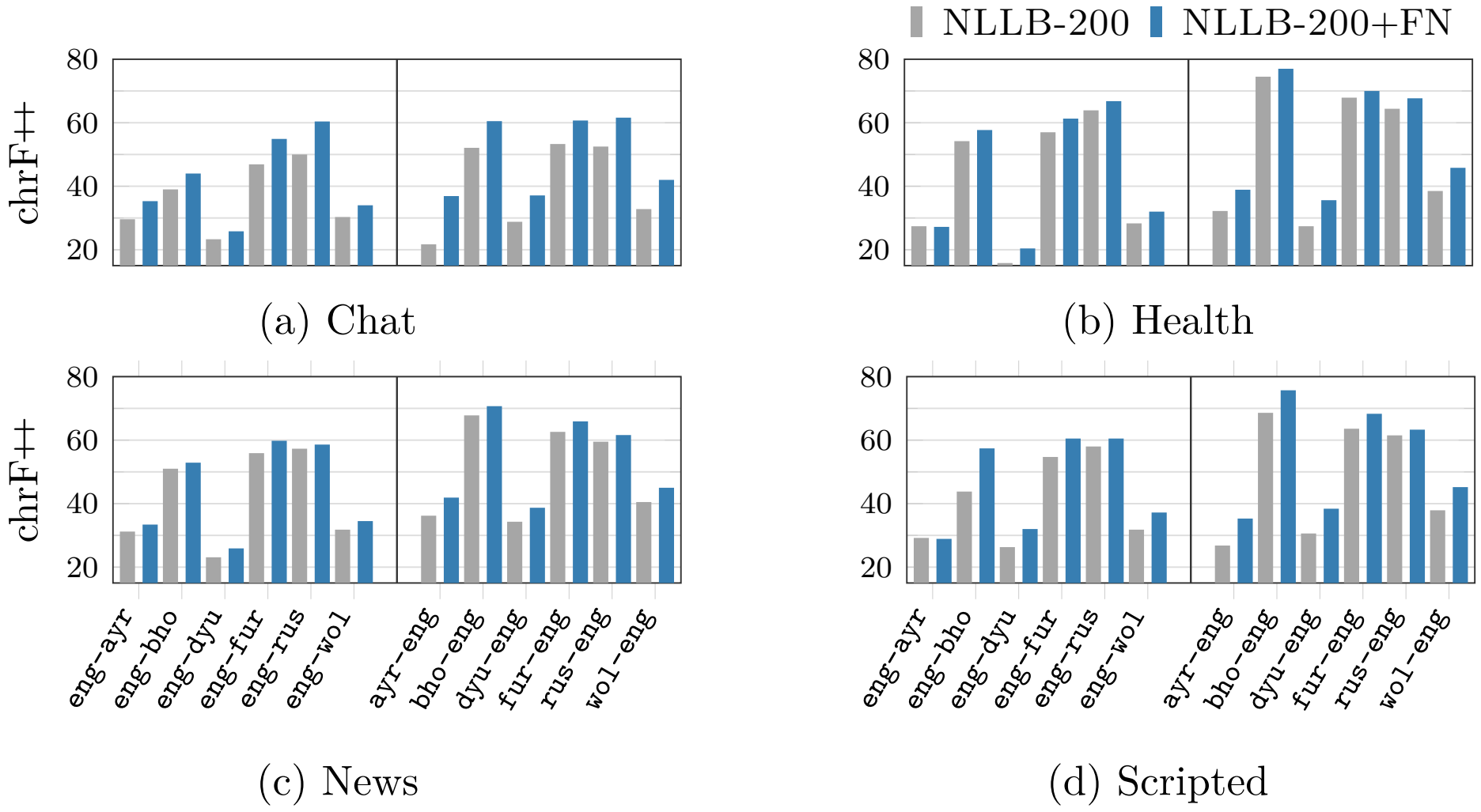

Figure 39: **Performance on NLLB-MD Test Sets** (12 tasks in 4 domains) of NLLB-200 and the single-task finetuned models NLLB-200+FN (without load balancing).

## 8.5 Analysis of NLLB-200

In this section, we analyze several properties of NLLB-200. We first discuss the language co-location of the massively multilingual NLLB-200's experts. Then, we examine how different curriculum learning strategies address the issue of overfitting on low-resource directions while also maintaining performance of high resource directions. Finally we dive into the impact of multilingual transfer on low-resource languages.

### 8.5.1 Language Co-location in NLLB-200 experts

Similar to our analysis of MoE models on the ablation dataset with its 53 languages (see Section 6.2.4), we aim to analyze language co-location in NLLB-200's experts. Following the same steps, we compute a per-language distribution across the 128 experts, evaluated on the `dev` set of Flores-200, then we embed all the trained languages in 2D with UMAP. Similar to our observations on the ablation dataset, we see in Figure 40 that late decoder layers and early encoder layers of NLLB-200 process tokens from separate languages by dispatching them to different sets of experts. Languages within the same family are assigned similar sets of experts, and families that are geographically proximate (e.g., Nilotic, Saharan and Atlantic-Congo) or are genealogically related (e.g., Arabic and Hebrew) also get assigned to similar MoE experts of NLLB-200. Detailed similarity scores between the 200 languages in each of the 4 depicted layers can be found in Figure 45.

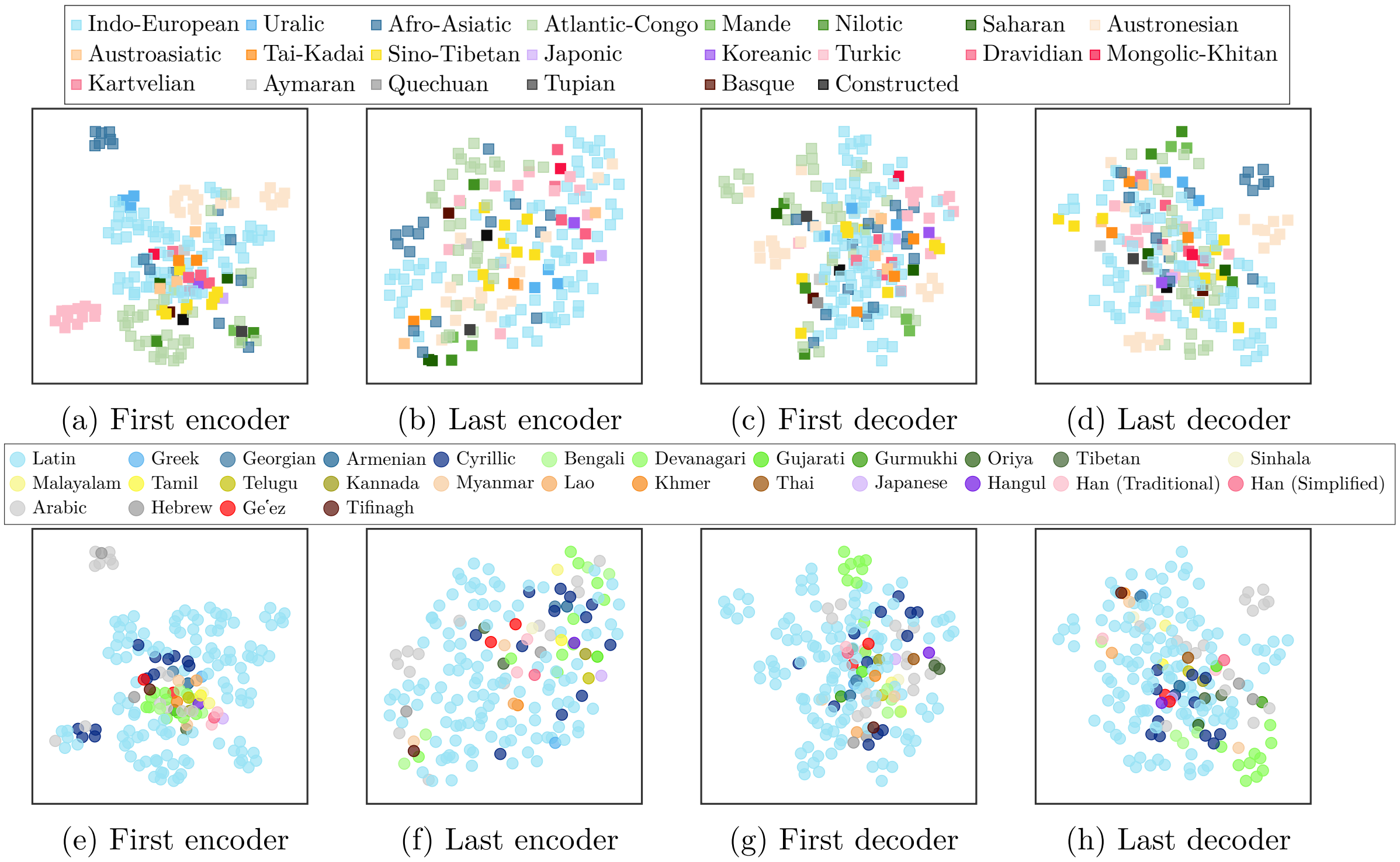

Figure 40: **UMAP embedding of the FLORES-200 languages.** We embed vectors of expert-language assignment in 2D with UMAP. Languages with similar expert choice (averaged on FLORES-200 `dev` set) are adjacent in the 2D-projected space. We color in the first row according to language family and in the second row according to script.

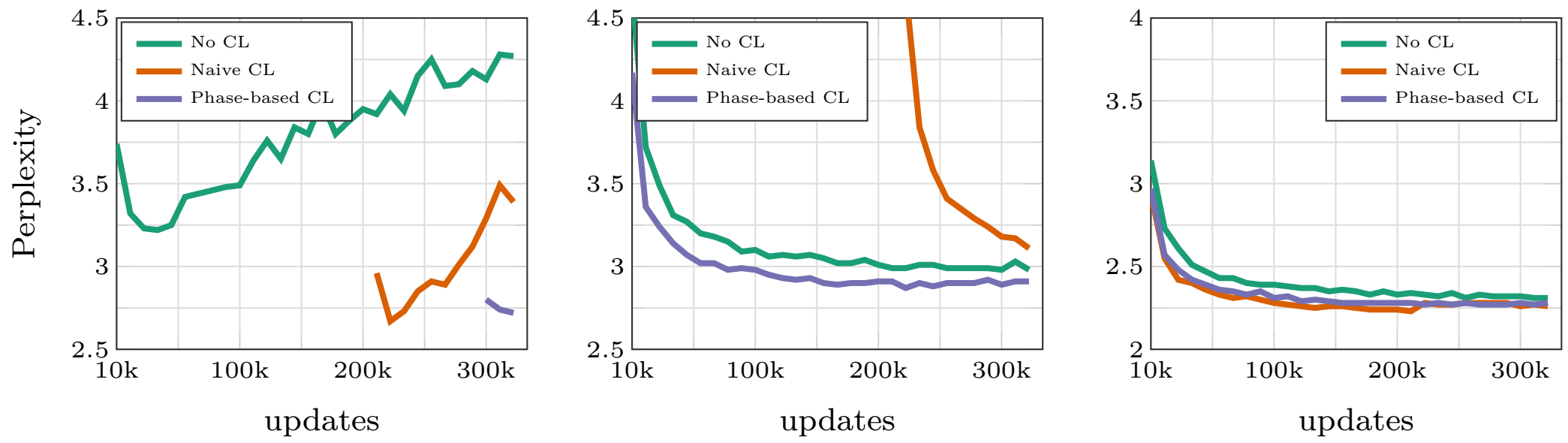

(a) `arz_Arab-eng_Latn` (very low) (b) `eng_Latn-hye_Armn` (low) (c) `eng_Latn-rus_Cyrl` (high)

Figure 41: **Validation Perplexity with Various Curriculum Strategies**. Our proposed phased curriculum is particularly beneficial for low-resource and very low-resource pairs, mitigating the overfitting observed in the baseline and naive curriculum variants. With the naive curriculum, introducing pairs too early (a) or too late (b) hurts performance. The curriculum does not affect high-resource pairs much (c).

#### 8.5.2 Effect of Phased Curriculum on Low-Resource Overfitting

In Figure 41, we compare the effect of no curriculum, naive curriculum, and the phased curriculum training of NLLB-200. We observe that while naive curriculum helps alleviate some low-resource overfitting, some low-resource language pairs are introduced too early, while others are introduced too late. These effects are mitigated with the phased curriculum. The effects are more prominent for very low-resource pairs, which start to overfit significantly if introduced early, as seen in Figure 41(a). Although phased curriculum strategy in general helps alleviate lot of overfitting issues we see with naive curriculum, it is still not optimal. Designing automated curricula that monitor training overfitting patterns can further improve performance, and we keep this direction of research for future work.

| | eng_Latn-xx | | | | | xx-eng_Latn | | | | |
|---|---|---|---|---|---|---|---|---|---|---|
| | Bilingual | Multilingual Dense | | Multilingual MoE | | Bilingual | Multilingual Dense | | Multilingual MoE | |
| | chrF++ | chrF++ | Δ | chrF++ | Δ | chrF++ | chrF++ | Δ | chrF++ | Δ |
| aka_latn | 16.8 | 35.6 | **18.8** | 36.2 | **19.4** | 36.6 | 45.6 | **9.0** | 46.7 | **10.1** |
| smo_Latn | 49.4 | 50.5 | **1.1** | 50.2 | **0.8** | 49.7 | 57.1 | **7.4** | 58.1 | **8.4** |
| sot_Latn | 44.4 | 46.4 | **2.0** | 46 | **1.6** | 52.3 | 61.9 | **9.6** | 62.5 | **10.2** |
| twi_Latn | 36.7 | 38.7 | **2.0** | 39.1 | **2.4** | 39.8 | 46.6 | **6.8** | 48 | **8.2** |
| umb_Latn | 23.3 | 23.8 | **0.5** | 24.1 | **0.8** | 27.2 | 31.6 | **4.4** | 32.7 | **5.5** |
| vec_Latn | 40.6 | 47.1 | **6.5** | 48.9 | **8.3** | 56.3 | 67.9 | **11.6** | 63.7 | **7.4** |
| guj_Gujr | 51.7 | 50.6 | -1.1 | 51.4 | -0.3 | 57.4 | 65.3 | **7.9** | 66.4 | **9.0** |
| mya_Mymr | 27.1 | 32.7 | **5.6** | 32.6 | **5.5** | 43.5 | 56 | **12.5** | 57.2 | **13.7** |
| npi_Deva | 32.4 | 47.9 | **15.5** | 48.6 | **16.2** | 56.2 | 67.3 | **11.1** | 67.6 | **11.4** |
| pbt_Arab | 37.1 | 35.3 | -1.8 | 36.5 | -0.6 | 47.2 | 56.6 | **9.4** | 56.9 | **9.7** |
| sin_Sinh | 41.8 | 43.1 | **1.3** | 44.1 | **2.3** | 51.0 | 61.7 | **10.7** | 63.0 | **12.0** |

Table 39: **Comparison of FLORES-200 devtest Performance between Bilingual and Multilingual dense and MoE models.** We see that most low-resource languages benefit from multilingual transfer, leading to significant chrF++ increase. Both dense and MoE models perform better and the gains are more consistent when translating into eng_Latn.

#### 8.5.3 Impact of Multilingual Transfer

One of the main benefits of multilingual models lies in the strong transfer between the languages. To quantify the gains obtained with the massively multilingual NLLB-200, we trained bilingual models on a few low resource directions with the same training data and compared their performance. The results are presented in Table 39. We observe impressive gains over almost all directions we compare on, which demonstrates the benefits of multilingual transfer. We also notice that the gains are more consistent on xx-eng_Latn directions. For eng_Latn-xx language pairs, improvements are less consistent and we can observe some directions where huge gains (+19.4 chrF++) are seen, but also other directions where slight decreases(-0.6 chrF++) are observed.

### 8.6 Making Large Models More Accessible through Distillation

While large models often have strong performance, their sheer size limits their utility and makes inference expensive. Adding additional languages, finetuning, or even generating translations can require significant GPU compute. In this section, we explore *distillation*, a common technique for training smaller *student* models from larger, better performing *teacher* models (Buciluă et al., 2006; Hinton et al., 2015; Tan et al., 2018). Such techniques yield stronger performance compared to training a small model from scratch (Gordon and Duh, 2020; Kim and Rush, 2016). Various approaches have been explored extensively to compress translation models (Freitag et al., 2017; Zhang et al., 2018; Zhou et al., 2020), including in multilingual (Sun et al., 2020) and low-resource (Ahia et al., 2021; Saleh et al., 2020) settings. We explore and discuss two types of distillation approaches, *online* and *offline* distillation. We detail how we apply these techniques to create specialized models for the Wikipedia Content Translation tool (Laxström et al., 2015). These models serve translations for 25 languages and 74 directions, translating on-demand with low latency. Finally, we apply a compute efficient approach for distilling our 54B parameter NLLB-200 across all 200 languages. For machine learning practitioners, we hope that making such models available will enable translations for far more languages in various applications.

#### 8.6.1 Knowledge Distillation

Broadly, knowledge distillation is the process of transferring knowledge from a larger *teacher* model to a smaller *student* model (Hinton et al., 2015). We investigate two forms of distillation: online and offline. We describe both these approaches briefly here.

**Online Distillation.** We explore *Word-Level* knowledge distillation (Hinton et al., 2015). In this setting, the student model is trained on the training data but with an additional objective: to minimize the cross-entropy with respect to the word-level distribution of the teacher model.

**Offline Distillation.** We explore *Sequence-Level* knowledge distillation (Kim and Rush, 2016), which can be thought of as a form of data augmentation (Gordon and Duh, 2019; Xu et al., 2021b) similar to backtranslation. The teacher model is used to generate translations, taking monolingual data as input. The student model is then trained on data generated by the teacher. This approach has the benefit that the student model can learn to mimic the teacher at the sentence level, but not local predictions over individual words.

#### 8.6.2 Creating Models Specialized for the Wikipedia Domain

In our interviews with native speakers of low-resource languages, a common theme was *access* to information online, such as educational content. While knowledge has been curated at a global scale on online platforms such as Wikipedia, these platforms remain accessible primarily by those who speak one of the handful of languages that dominate the web — such as English. Driven by our principle of *prioritizing underserved communities* and *sharing*, we next explore how our work could be put into practice. We describe how we build multilingual translation models for the Wikipedia Content Translation tool. This is an online interface that Wikipedia editors can use to translate an article that already exists in another language as a starting point for writing a new article. This aids the creation of new articles for underserved languages. We leverage distillation process to create models that are fast enough to meet the latency requirements of an on-demand service, while producing high-quality translations. Such work is broadly useful for machine learning practitioners who may be interested in adapting general-purpose systems to specific domains and deployed applications.

**Language Pairs.** The language pairs include `eng_Latn`, `fra_Latn`, `spa_Latn` as source languages, and 25 target languages where the languages are either **(1)** not yet supported in Wikipedia's Content Translation tool, **(2)** supported but produced low quality translations,

| Size | teacher 1.3B | teacher ft 1.3B | offline student 500M | online student 500M |
|---|---|---|---|---|
| eng_Latn-xx | 40.0 | 43.0 | **43.4** | 43.0 |
| fra_Latn-xx | 36.3 | 39.2 | **39.6** | 39.1 |
| spa_Latn-xx | 35.1 | 36.9 | **37.3** | 36.8 |

Table 40: **Performance of Offline and Online Distillation in the Wikipedia Setting.** We show the average Flores-200 devtest chrf++ performance for English, French, and Spanish source for teacher, fine-tuned teacher, and student models. A full list of results is available in Table 59

or **(3)** had high article deletion rates. Additionally, we include four other directions of interest to Wikipedia: por_Latn-oci_Latn, cat_Latn-oci_Latn, zho_Hans-yue_Hant, and rus_Cyrl-bak_Cyrl.

**Teacher Training and Fine-Tuning.** We train a 1.3B parameter dense *teacher* model on the ablation dataset (see Section 6.4), for all the language pairs listed above. The teacher model is smaller than our full 54B NLLB-200 model and is dense rather than sparse, reducing the time required to generate the student model training dataset for offline distillation. We further fine-tune this model on Wikipedia domain bitexts, which include Wikipedia translation content edits. We filter the fine-tuning data using LID and length-based filtering, similar to the filtering described in Section 8.1.4.

**Applying Offline Distillation.** Using offline sequence-level distillation (Kim and Rush, 2016), we distill the same fine-tuned teacher model into a smaller 500M parameter dense model. We prepare the distilled model's training dataset by generating translations for monolingual Wikipedia corpus data, using the fine-tuned teacher model. To generate this training dataset for the student model, we run beam search with a beam size of 4 with the teacher model. The source sentences come from monolingual Wikipedia data dumps[43] for eng_Latn, fra_Latn, spa_Latn, rus_Cyrl, por_Latn, cat_Latn. Finally, we train the student model on this distilled training dataset. Note that while the teacher model is trained on both directions, the distilled student model is only trained on the eng_Latn-xx, spa_Latn-xx, fra_Latn-xx, por_Latn-oci_Latn, cat_Latn-oci_Latn, zho_Hans-yue_Hant, and rus_Cyrl-bak_Cyrl directions.

We open-source an end-to-end distillation pipeline in the stopes[44] library which can be used to apply offline distillation given a trained teacher model.

**Applying Online Distillation.** We use online word-level distillation (Hinton et al., 2015) to distill the fine-tuned 1.3B parameter dense teacher model into a smaller 500M parameter dense model for inference efficiency. We use the same monolingual Wikipedia data dumps for source sentences, and we use only the soft cross-entropy with respect to the teacher distribution $L_{KD}$ because we use the monolingual Wikipedia data dumps as source sentences.

**Comparing Online and Offline Distillation.** We first examine the performance of online distillation. From Table 40, we see that the 515 million parameter student model

43. https://dumps.wikimedia.org/other/cirrussearch/
44. https://github.com/facebookresearch/stopes

performs on par with the fine-tuned teacher model, within 0.1 chrF++ for `eng_Latn-xx`, `fra_Latn-xx`, and `spa_Latn-xx` directions.

For offline distillation, Table 40 indicates that the 515 million parameter student model performs better than the fine-tuned 1.3 billion parameter teacher model on average across all pairs. Improvements are observed on `eng_Latn-xx` (+0.4 chrf++), `fra_Latn-xx` (+0.4 chrf++), and `spa_Latn-xx` (+0.3 chrf++) directions. We hypothesize that these improvements come in part from the in-domain Wikipedia monolingual source data used for the distillation training dataset. Further, distilled models often produce more translationese which can have effects on the reliability of reference-based metrics (Riley et al., 2020).

Based on these results, we conclude that offline sequence-level distillation performs marginally better than online word-level distillation when distilling our models for Wikipedia Content Translation. However, online distillation still performs well and is more compute efficient as a very large MoE model is not required to autoregressively generate millions of translations, which is important when distilling larger models to more languages. In the next sections, we explore compute efficient online distillation for very large scale multilingual models like NLLB-200.

**Comparison with Bapna et al. (2022)** While significant differences prevent fair comparison, we briefly compare our Wikipedia distilled model to Bapna et al. (2022)'s distilled models on Flores-101 `devtest`. Apart from the major difference of training data for both the teacher and student models, there are several other notable distinctions. Our model begins with a 1.3B parameter dense teacher rather than a 6B parameter model. We cover 25 target languages across `eng_Latn`, `fra_Latn`, `spa_Latn` source languages, and 79 directions in **one** student model rather than having two student models `eng_Latn-xx` and `xx-eng_Latn` each supporting 30 languages and directions. Our final student model is only 500M parameters, compared to the 850M parameter encoder model with a Transformer based encoder and LSTM decoder presented in (Bapna et al., 2022). In spite of these, when averaged over the performance of 6 overlapping directions on Flores-101 (`eng_Latn-xx` for `asm_Beng`, `ckb_Arab`, `lin_Latn`, `lug_Latn`, `nso_Latn`, and `gaz_Latn`), our 500M parameter distilled model has similar performance as the 850M parameter distilled model from (Bapna et al., 2022). Across the 6 overlapping directions, both models achieve 24.9 spBLEU on Flores-101 `devtest`.

#### 8.6.3 Distillation of NLLB-200, a 54B Parameter MoE Model

We finally explore distillation of NLLB-200, a 54B parameter Mixture-of-Experts model, with the help of online distillation. The final distilled model retains full translation support of all 202 languages. Since inference for the full NLLB-200 is slower than the 1.3B parameter dense model and we distill to more than 30 times the number of language pairs of the Wikipedia model, we choose to use online rather than offline distillation due to the time and compute required to generate sequence-level teacher outputs. We train baseline and distilled dense models, each with 1.3B and 615M parameters, all for 200,000 updates.

**Results.** Table 41 shows that on average, the distilled dense 1.3B student model performs better than the dense baseline 1.3B model by +0.5 chrF++, and the 615M student model performs better than the baseline 615M model by +0.3 chrF++. We observe that distillation provides larger improvements over the corresponding baseline dense models for low and very low-resource languages. For example, for the 1.3B model setting, the distilled model performs better than the dense baseline by +0.6 chrF++. However, on average, a gap of -1.4 chrF++ for 1.3B and -3.7 chrF++ for 615M remains between the student model and NLLB-200 performance.

| | size | eng_Latn-xx | | | | xx-eng_Latn | | | | xx-yy | Avg. |
|---|---|---|---|---|---|---|---|---|---|---|---|
| | | all | high | low | v.low | all | high | low | v.low | all | all |
| NLLB-200 | 54B | 45.3 | 54.9 | 41.9 | 39.5 | 56.8 | 63.5 | 54.4 | 54.4 | 42.7 | 48.3 |
| dense baseline | 1.3B | 43.5 | 52.8 | 40.1 | 37.6 | 54.7 | 61.8 | 52.2 | 51.9 | 41.0 | 46.4 |
| dense distilled | 1.3B | **44.0** | **53.2** | **40.8** | **38.4** | **55.1** | **61.9** | **52.6** | **52.5** | **41.5** | **46.9** |
| dense baseline | 615M | 41.4 | 50.7 | 38.1 | 35.1 | 52.2 | 59.7 | 49.6 | 49.1 | 39.3 | 44.3 |
| dense distilled | 615M | **41.8** | **50.9** | **38.5** | **35.8** | **52.3** | 59.7 | **49.7** | **49.3** | **39.5** | **44.6** |

Table 41: **Distillation of NLLB-200.** We report chrF++ scores on Flores-200 devtest set for the full NLLB-200, dense baselines, and dense distilled models. For eng_Latn-xx and xx-eng_Latn we include all 201 pairs each. For xx-yy we randomly choose 200 directions. We observe that distilled models perform better than dense baseline models trained from scratch without distillation.

#### 8.6.4 Conclusion.

Practically deploying machine learning models, particularly neural methods, is extremely difficult and an active area of research. Our investigation indicates that distillation is a very promising avenue for leveraging multilingual models and specializing them to *a subset of desired language directions* and *to the desired domain*. This enables the Wikipedia translation model we create to have strong performance (despite the teacher model being only 1.3B parameters). We hope that such work, and open-sourcing these distilled models, enables others to access translation for their own native languages.

### 8.7 Effectively Including Languages with Multiple Scripts and Related Languoids

Languages are not monolithic units, but fluid and full of variation. Some languages are written naturally in multiple scripts — Serbian is a well-known example that uses both Latin script and Cyrillic script. Other languages have a large amount of variation — we term these *languoids*. This variation can be fairly small differences in spelling (British English compared to American English, for instance) or render the languages mutually unintelligible. In Flores-200 (see Section 3), we include multiple scripts for languages and multiple Arabic languoids. In this section, we study how to best represent natural language variation and produce the most accurate, localized translations. We focus on two cases: transliteration between different scripts and translation of closely related Arabic languoids.

#### 8.7.1 Transliteration

The languages of the world use a wide variety of writing systems. Examples include *logosyllabaries* such as the Han script, which use ideographs to represent words or morphemes; *syllabaries* such as Katakana, which represent syllables; *abugidas* such as the Devanagari script, whose base units are consonant-vowel pairs; *abjads* like the Arabic script, which instead only require that consonants be written and may allow for vowels to be represented as diacritics; and *alphabets* like the Latin and Cyrillic writing systems, whose symbols denote both vowels and consonants; and many more. In many situations, a single language may be commonly communicated through different writing systems. Such cases arise due to historical, geopolitical, religious, or technological reasons. Language boundaries rarely overlap neatly with the borders of geopolitical entities such as nation states. The coexistence of multiple writing systems for a single language leads to two main challenges for creators of language technologies, especially when the language in question falls into the *low-resource* or *endangered* classifications: data challenges and ethical considerations.

**Data Challenges.** Obtaining transliteration models can be challenging. Whether the technique being tested is rule-based or model-based, the amount of readily available data may not be sufficient. To go from an abjad to an alphabet using a rule-based technique, simple rules which are deductively derived from character-to-character mapping tables, where the mapping of diacritics proves particularly challenging, do not produce satisfactory results. Similarly, model-based techniques require fairly large quantities of textual resources (Madhani et al., 2022), which are by definition unavailable for low-resource languages. Having linguistic data spread across different scripts can further reduce the amount of text which can be used for training, leading to poor performance.

**Ethical Considerations.** Developing technologies that favor one script might inadvertently further relegate minorities using a different writing system. In this section, we explore the question of whether a technological solution can be devised to alleviate these concerns. How effective are our translation models at generating as well as assimilating content in any script? Can they be used to perform transliteration, so as to bridge any gaps between writing systems, or are traditional transliteration methods more effective?

**Experimental Setting.** For evaluation, we choose the challenging task of transliterating from an abjad to an alphabet, which requires the recovery of vowels. We evaluate transliteration from the Arabic script to the Latin script for Acehnese and Banjar, as well as from the Tifinagh script to the Latin script for Tamasheq. The systems we compare are:

- The out-of-the box universal romanization tool `uroman` (Hermjakob et al., 2018)
- Two online transliteration tools: `Ejawi`[45] and an `ALA-LC` transliterator for the Tifinagh script[46]
- Our own NLLB-200 translation model, which is capable of translating between any pair of 200 languages
- A custom neural transliteration model. The model is trained on NLLB-Seed and is based on a standard transformer architecture, with 4-layer encoder and decoder, 4

45. `https://www.ejawi.net/` We are aware that Acehnese and Banjar use two different varieties of Arabic script (Jawi and Pegon, respectively), and that Acehnese is not part of the Malay group of languages, while eJawi is optimized for Malay languages written in Jawi. We use this approach to get a reasonably approximated transliteration.
46. `https://www.translitteration.com/transliteration/en/tamazight/ala-lc/`

| | ace_Arab-ace_Latn | bjn_Arab-bjn_Latn | taq_Tfng-taq_Latn |
|---|---|---|---|
| uroman | 0.47 | 0.40 | 0.23 |
| Ejawi/ALA-LC | 0.66 | 0.72 | 0.24 |
| Ejawi/ALA-LC + rules | 0.37 | 0.35 | **0.22** |
| Custom Model | 0.32 | 0.31 | 0.29 |
| Rules + Custom Model | **0.25** | **0.20** | **0.22** |
| NLLB-200 translation model | 0.54 | 0.41 | 0.66 |

Table 42: **Transliteration Performance of Various Approaches**, measured by Character Error Rate (CER, lower is better) on Flores-200 devtest set.

attention heads and 256-dimensional embeddings. The size of the SentencePiece vocabulary, which for each language is joint across source and target script, is determined as a function of the character set size. We experiment with vocabularies of size 1.5, 2.5 and 10 times the character set size, and determine the best choice for a given language through validation. Similar architectures have already been shown to be effective at character transduction and low-resource text normalization tasks (Lusito et al., 2022; Wu et al., 2021).

- A rule-based approach which naively replaces individual characters according to a transliteration table. We combine this approach with Ejawi as a postprocessing step, as we noticed a number of words still remained in the source script after using this online tool. We also attempt to use this rule-based approach as a preprocessing step for the neural transliterator, with the aim of bringing source and target embedding representations closer to each other.

**Evaluation.** We evaluate using CER (character error rate), which is computed as the ratio of edits to reference characters, and is common in tasks such as Optical Character Recognition. All systems are evaluated on Flores-200 devtest.

**Results.** Results are shown in Table 42. We observe that specialized transliteration models perform best at this task. Among these tools, we note that the Ejawi online transliteration service achieves a high error rate on its own, but performs much better when followed by a simple rule-based character replacement step.[47] While the neural transliteration systems we trained do not in general achieve the lowest error rates, their performance could be plausibly improved by collecting larger training datasets. Regardless of the approach, error rates remain relatively high. This highlights the importance of being mindful when designing linguistic technologies, so as to reduce as much as possible the differences which might otherwise be introduced for users of different scripts.

#### 8.7.2 Multidialectal Translation

We want to consider the possibility of translating from or into different Arabic languoids. To test the feasibility and efficacy of multilingual translation models for this task, we focus here on Arabic languoids, presented in Table 43. Arabic has the advantage of being rich in

47. Based on Andries (2004) for the Tifinagh script, and on the Arabic ICU transliteration rules for Jawi.

| Language | Variety | sDL | cDL | spBLEU | chrF++ |
|---|---|---|---|---|---|
| acm_Arab | Mesopotamian | 22.84% | 6.07% | 61.8 | 70.5 |
| acq_Arab | Ta‘izzi-Adeni | 15.96% | 3.88% | 73.2 | 79.0 |
| aeb_Arab | Tunisian | 32.34% | 8.50% | 50.5 | 59.6 |
| ars_Arab | Najdi | 3.01% | 2.40% | 94.8 | 96.5 |
| ajp_Arab | South Levantine | 42.13% | 17.43% | 32.6 | 47.3 |
| arz_Arab | Egyptian | 37.31% | 8.56% | 41.9 | 53.5 |

Table 43: **Quantification of Differences between six Arabic Languoids** in Flores-200 devtest, measured via sentence- and corpus-level DL, spBLEU and chrF++ of each languoid's reference against the MSA (arb_Arab) reference.

dialectal variation that has been well studied and documented (Alshargi et al., 2019; Habash, 2010; Habash et al., 2013; Salameh et al., 2018).

**Arabic Languoids.** While NLP systems have historically focused on Modern Standard Arabic (MSA), there has been a considerable amount of research on translating Arabic languoids in recent years (Baniata et al., 2021; Kumar et al., 2021; Sajjad et al., 2020). NLLB-200 includes several other varieties of Arabic beyond MSA: Egyptian, Moroccan, Najdi, North Levantine, South Levantine, Ta‘izzi-Adeni, Mesopotamian and Tunisian. This expansionary effort needs a clear way of quantifying how different the latter languoids are from MSA. We should note that a commonly stated characteristic of Arabic is the sociolinguistic phenomenon termed *diglossia*. In diglossic situations, users code-switch between a more vernacular dialect used for informal and casual tasks (e.g., day-to-day conversations, social media chats or posts) and another dialect used for tasks that are typically performed at a higher register of language (e.g., technical or scientific discourse, educational or informative content; see Ferguson, 1959). Such is the case for Arabic languoids, where MSA serves for higher-register use. The direct implication of this phenomenon is that sentences being produced in any Arabic languoid at a higher register will likely have more in common with MSA than sentences produced at a lower register (e.g., informal social media chats or posts).

**Dialectness Level Metric and Calculation on Flores-200.** Dialectness Level (DL; see Sajjad et al., 2020) serves as a way of measuring the degree of lexical variation present in these languoids when compared to MSA. Specifically, the DL is a representation of the amount of tokens that are present in the languoid which are not present in MSA, including instances of semantic differences for a token. This means that in instances where the meaning of a given token in MSA is different from that of the same token in another languoid, it will not be taken into account as part of the DL score. The DL is represented as a range from 0% to 100% wherein the higher the range, the higher the level of dialectness in that specific dialect.

The DL is calculated in two ways for our purposes: at the corpus and the sentence levels. The first metric, *corpus-level DL* (henceforth *cDL*), is computed for a given non-MSA languoid as the fraction of all tokens in its Flores-200 devtest dataset which are not present in MSA. The second DL measurement, *sentence-level DL* (henceforth *sDL*) is computed analogously but at a higher level of granularity. We work out, for each line in a languoid's Flores-200 devtest dataset, the fraction of its tokens which are not present in the corresponding MSA sentence. This number is then averaged across all lines in the dataset. As in Sajjad et al. (2020), to compute these metrics we normalize our Arabic text by replacing different forms of Alif and Hamzah as well as Indo-Arabic numerals. Additionally, we also use the same metrics used throughout this paper for evaluating machine translation, spBLEU and chrF++, as described in Section 7.

| | Transl. perf. | | against generated MSA | | | |
|---|---|---|---|---|---|---|
| Direction | spBLEU | chrF++ | sDL | cDL | spBLEU | chrF++ |
| eng_Latn-acm_Arab | 11.8 | 31.9 | 43.54% | 7.89% | 21.7 | 46.2 |
| eng_Latn-acq_Arab | 26.9 | 42.2 | 27.97% | 13.16% | 47.3 | 58.6 |
| eng_Latn-aeb_Arab | 19.9 | 38.2 | 38.90% | 10.23% | 31.6 | 50.5 |
| eng_Latn-ars_Arab | 36.7 | 50.5 | 20.38% | 11.46% | 62.8 | 70.8 |
| eng_Latn-ajp_Arab | 36.3 | 51.3 | 24.50% | 9.07% | 58.5 | 67.1 |
| eng_Latn-arz_Arab | 32.1 | 46.8 | 28.33% | 7.79% | 54.5 | 64.9 |

Table 44: **Comparison of Translations Generated by NLLB-200** from English FLORES-200 devtest data. We report the translation performance for each direction. We further compare against the MSA reference as well as the translation of English into MSA.

| | Transl. perf. | | against generated MSA | | | |
|---|---|---|---|---|---|---|
| Direction | spBLEU | chrF++ | sDL | cDL | spBLEU | chrF++ |
| fra_Latn-acm_Arab | 3.7 | 18.4 | 65.80% | 8.47% | 7.3 | 25.4 |
| fra_Latn-acq_Arab | 19.1 | 34.3 | 33.06% | 17.54% | 37.1 | 49.4 |
| fra_Latn-aeb_Arab | 11.5 | 30.2 | 49.89% | 10.49% | 18.2 | 37.9 |
| fra_Latn-ars_Arab | 25.8 | 40.5 | 25.71% | 11.16% | 50.3 | 60.6 |
| fra_Latn-ajp_Arab | 27.0 | 42.8 | 29.41% | 10.41% | 49.0 | 59.2 |
| fra_Latn-arz_Arab | 25.3 | 41.2 | 29.86% | 9.26% | 49.8 | 62.5 |

Table 45: **Comparison of Translations Generated by NLLB-200 from French.** FLORES-200 devtest data, analogously to Table 44.

**Results.** Given that FLORES-200 is aligned, we start by measuring the differences between the FLORES-200 reference data for six Arabic languoids: Mesopotamian, Ta'izzi-Adeni, Tunisian, South Levantine, Najdi and Egyptian. It should be noted that data for South Levantine (ajp_Arab) and Egyptian (arz_Arab) was obtained by translating English, whereas the datasets of the other four languoids were adapted from MSA directly. While we decide to include these two languoids in our analyses to provide additional context, one should bear in mind that any results involving them might be skewed due to the slightly different data collection process. For this reason, we keep them separate in results tables. The results in Table 43 show that, with the exception of the two languoids whose data was obtained via translation, all other languoids are close to MSA, with spBLEU scores exceeding 50 points. Of the adapted datasets, Tunisian (aeb_Arab) is the one diverging the most from MSA, and Najdi (ars_Arab) matches MSA almost exactly.

| | into eng_Latn | | into fra_Latn | |
|---|---|---|---|---|
| Language | spBLEU | chrF++ | spBLEU | chrF++ |
| acm_Arab | 43.3 | 63.1 | 39.2 | 57.6 |
| acq_Arab | 45.1 | 64.9 | 40.2 | 58.5 |
| aeb_Arab | 38.6 | 59.5 | 36.0 | 55.1 |
| ars_Arab | 46.8 | 66.0 | 41.6 | 59.5 |
| ajp_Arab | 48.2 | 67.4 | 41.8 | 59.9 |
| arz_Arab | 40.7 | 60.8 | 37.8 | 56.1 |
| arb_Arab | 48.3 | 66.9 | 42.3 | 59.8 |

Table 46: **Languoid Performance** on FLORES-200 devtest. We evaluate translation into English and French for seven Arabic languoids, including MSA (bottom).

**NLLB-200 Performance.** We next shift our focus to how NLLB-200 generates Arabic languoids. Tables 44 and 45 look at the model's translation from English and French respectively. The first two columns measure the model's translation performance against each languoid's FLORES-200 reference. We see that, among the languoids whose evaluation data was adapted from the MSA dataset (first four rows), the trends observed in Table 43 are still visible. Performance is low for Tunisian (aeb_Arab) and Mesopotamian (acm_Arab), and best for Najdi (ars_Arab), which can likely be explained by how much each model benefits from transfer learning via the much higher resourced MSA. Conversely, the two directions whose evaluation data was directly translated from English (bottom two rows) are achieving relatively high translation scores, despite their references showing the highest differences from MSA. We can formulate two hypotheses as to why. First, regardless of the dialectal variation shown for these languages in Table 43, they might in fact be relatively close to MSA, and their perceived dialectal differences might largely be explained away by the slightly different procedure which was used to collect their evaluation data. Second, since their evaluation data was translated directly from English (as opposed to being adapted from MSA which was itself translated from English), it may actually be closer semantically and syntactically to the English source, making the task of translating it easier.

We then analyze how distinct the dialects generated by our translation model are. In the rightmost part of Tables 44 and 45, we measure the differences between the generated non-MSA Arabic text and the generated MSA. For the first four languoids in the group, the major trends observed in Table 43 still hold, with Tunisian and Mesopotamian being the farthest away from MSA, and Najdi being the closest. More generally however, we observe a slight flattening of the differences in dialectness levels of the text translated from English compared to the human-annotated evaluation data. Indeed, the sentence-level DL scores have a standard deviation of 14.6 in the evaluation data, which reduces to 8.8 for the data generated by the model.

Additionally, we examine the translation model's ability to process various Arabic languoids, by looking at its performance when translating them into English and French. The results are reported in Table 46. As expected, for both target languages the high-resource MSA (arb_Arab) performs best, and is followed by the low-resource South Levantine (ajp_Arab) and Najdi (ars_Arab) languoids which benefit from transfer learning. Of the four languoids whose evaluation data was collected by adaptation, we see that the relative ranking of translation performance figures matches the order of the dialectness level rankings in Table 43: dialects closest to MSA are the ones most effectively understood by the model.

| Directions | Custom | | | NLLB-200 | | |
|---|---|---|---|---|---|---|
| | spBLEU | chrF++ | #best | spBLEU | chrF++ | #best |
| {Arabic dialects}-eng_Latn | 34.1 (4.7) | 56.7 (3.7) | 0 | **44.4** (3.8) | **64.1** (3.1) | **7** |
| {Arabic dialects}-fra_Latn | 19.9 (1.4) | 41.0 (1.3) | 0 | **39.8** (2.3) | **58.1** (1.9) | **7** |
| eng_Latn-{Arabic dialects} | **32.0** (6.3) | **47.5** (5.7) | **4** | 29.5 (10.8) | 45.4 (8.6) | 3 |
| fra_Latn-{Arabic dialects} | **31.8** (5.7) | **47.1** (5.5) | **7** | 21.1 (10.5) | 36.7 (10.2) | 0 |
| *Overall* | 29.4 (7.3) | 48.1 (7.1) | 11 | **33.7** (11.9) | **51.1** (12.7) | **17** |

Table 47: **Comparison of a Custom Arabic model and NLLB-200** on FLORES-200 devtest. We report average performance and standard deviation for the translation of Arabic languoids into and out of eng_Latn and fra_Latn. We also report the number of directions in each set for which a given model achieves the top performance.

Finally, in Table 47 we compare the performance of the large NLLB-200 model against that of a more targeted, smaller scale model that only focuses on translating between Arabic languoids, eng_Latn and fra_Latn. This custom model uses a smaller dense transformer architecture with 12 encoder layers and 12 decoder layers, FFN dimension 4096, 16 attention heads, and which otherwise follows the setup of the baseline model used in Section 6.2.1. We see that on average NLLB-200 outperforms the custom model, achieving the top performance for 17/28 directions as well as a higher average score. Performance for the large model is especially much higher when translating into French. This can be explained by difference in number of fra_Latn primary training sentences that each model is exposed to. The NLLB-200 model, having access to a much wider number of directions, can count on over 53M unique primary French sentences. Limiting training corpora to English-centric and Arabic-centric directions instead reduces this number to under 38M for the custom model. On the other hand, the smaller model outperforms NLLB-200 when translating into Arabic languoids, showing that a large multilingual model is not uniformly better for this set of languages.

**Conclusion.** In conclusion, these results mirror other similar efforts, such as the AraBench benchmark (Sajjad et al., 2020), highlighting the benefits of translations for Arabic languoids made using an MSA system. Their conclusions, correlating Dialectness Level and translation quality of generations, can also be seen in our results. As stated previously, of the Arabic languoids that we focused on, Najdi show this best as it has consistently low DL, and subsequently, it also represents the high end of translation quality. Conversely Mesopotamian and Tunisian, showing high DL in Table 43, achieve consistently low translation performance. These results, as well as the AraBench benchmark findings, wherein they conclude that for translations made using an MSA system, a languoid with a lower DL will generate a higher quality translation, provide a way of overcoming the resource gap that might exist for these languoids. Their short linguistic distance to MSA allows them to benefit largely through transfer learning serving as a means of improving low-resource MT and multidialectal translation. Finally, our comparison of NLLB-200 and a more targeted model shows that while a massively multilingual model achieves the best average score, the smaller model can still outperform it on specific directions, highlighting the importance of more focused research on closely related languages.

### 8.8 Environmental Impact of NLLB

Carbon emission estimates are not precise as the community lacks tools to accurately measure the factors that contribute to the emissions. Previous works have reported estimates and recommendations in Bender et al. (2021); Dodge et al. (2022); Patterson et al. (2021); Wu et al. (2022). In this work, we rely on the best available power consumption estimates of GPU devices and carbon efficiency. Note that estimates of cloud providers are still inexact. There are several factors that affect the accuracy of these measurements: the real GPU power usage depends on GPU utilization and is likely different from Thermal Design Parameter(TDP) that we use as GPU power. Additionally, we did not include additional power costs, such as InfiniBand (IB) power consumption or non-GPU power consumption of the servers or datacenter cooling. Furthermore, manufacturing carbon cost for AI systems, such as GPUs, can introduce additional carbon footprint (Gupta et al., 2022a,b). We hope the carbon footprint analysis for NLLB helps provide transparency to understand the environmental implications of AI technologies.

**Carbon Emissions for Training NLLB-200.** The training of NLLB-200 was performed on NVIDIA A100 GPUs. Using the NVIDIA A100 system specifications (Choquette et al., 2021), we use TDP 400W as the power per processor. To train NLLB-200, a cumulative of 51968 GPU hours of computation was performed on hardware of type A100-SXM-80GB (TDP of 400W). We estimate the total emissions for training NLLB-200 to be **8.39 $tCO_2eq$** of which 100% were directly offset by the provider's sustainability program.[48]

**Total Carbon Footprint of the entire No Language Left Behind Effort.** The above only captures the carbon footprint of our final model. However, there are several steps in the research process before training a final model and steps afterwards (such as producing translations human evaluation steps) which we must also consider (Wu et al., 2022). Most previous works simply report the carbon footprint of training their largest models and multiply it with a factor (usually 2x) to report the total emissions. Instead, we try to report the carbon footprint for all the steps that have GPU utilization. This also provides useful insights to the community about the compute requirements and efficiency of each stage. Our detailed report includes steps for data preparation comprising large scale bitext mining (Section 5.3) and backtranslation (Section 8.1.3), all modeling experiments to design our architecture and training methods (Section 6), final model ablations for all 200 languages (Section 8.2), and model evaluations (Section 8.3). In Table 48, we report each calculation in detail and observe that the experimentation phase of our research is the most compute expensive of all. Total emissions for the NLLB project as a whole is estimated to be **104.31 $tCO_2eq$** of which **100% were directly offset by the provider**.

Sparse Mixture-of-Expert models may have a huge number of total parameters, but they are only sparsely activated when processing tokens during training. Hence, they can have greater compute efficiency compared to their dense counterparts and scaling

48. `https://sustainability.fb.com/2021-sustainability-report/`

| | Time (h) | Power Consumption (W) | Carbon Emitted ($tCO_2eq$) |
|---|---|---|---|
| Data Mining | 108,366 | 400 | 17.55 |
| Backtranslation | 18,000 | 300 | 2.17 |
| Modeling | 196,608 | 400 | 31.74 |
| Final Ablations | 224,000 | 400 | 36.17 |
| Evaluations | 51,200 | 400 | 8.26 |
| NLLB-200 | 51,968 | 400 | 8.39 |
| Total | | | 104.31 |

Table 48: **Total Carbon Footprint for No Language Left Behind.** We provide detailed estimates for all the steps that use GPUs for computation. Here we list factors that went into computation: Time in hours - total GPU time required for the step. Power Consumption - power consumption per GPU device for the GPUs used adjusted for power usage efficiency. 100% of the emissions are directly offset by the providers.

such models with careful considerations can help keep power consumption and carbon emissions lower for large workloads. It is a paramount responsibility for machine learning researchers to measure and report carbon emission impact of their work thoroughly, optimize their workloads towards energy efficient architectures and training paradigms, and keep in mind the negative environmental impact of inefficient workloads. Large scale machine learning workloads often come at a cost to the environment and if such works are not open sourced, similar efforts will have to be duplicated across multiple research groups which further results in more carbon emissions. To help reduce the need for any duplication of similar workloads, we open source the NLLB-200 and other smaller dense models, distilled models, our optimized training/inference code, evaluation results and benchmarks, and metadata for mined bitext data. All materials and code are accessible at `https://github.com/facebookresearch/fairseq/tree/nllb`.

## 9. No Language Left Behind: Social Impact & Concluding Thoughts

In this effort, we took on the challenge of creating high-quality machine translation systems for 200+ languages. Faced with major obstacles such as the lack of reliable evaluation and training data, progress in low-resource translation has been slow compared to its high-resource counterpart. In NLLB, we use novel approaches to make several major contributions aimed at bridging these gaps: **(1)** Flores-200, a high-quality human-translated evaluation dataset, and NLLB-Seed, a dataset comprising of human-translated bitext for 43 languages, **(2)** a novel bitext mining method that creates hundreds of millions of aligned training sentences for low-resource languages using our open-source mining library `stopes` and language identification model, and **(3)** various modeling techniques specifically devised to dramatically improve low-resource multilingual translation by reducing over-fitting. Beyond these, we also created smaller, distilled models so that the research community and various machine learning practitioners can more easily deploy this work.

To conclude, we discuss the potential social impact of our work. As is the case with most AI advancements, measuring NLLB's social impact requires a systematic evaluation framework and a longitudinal outlook. While its delivery could bring benefits to several stakeholders, including low-resource language groups and the scientific community at large, we also recognize that such an intervention has its potential downsides. As such, we reflect on the possibilities and limitations of NLLB, and ways to maximize its benefits while minimizing harm.

### 9.1 Expanding Information Access

In the summer of 2016, the United Nations declared internet access as a basic human right (Howell and West, 2016). While the intent of this declaration was to compel countries to limit censorship and allow for information and ideas to flow without interference, much of the internet remains inaccessible to many due to language barriers. NLLB has the potential to alter the status quo by making the internet more accessible for many.

For many low-resource language communities, NLLB's offering would be the first model designed to support translation of their languages. Adopters of NLLB's tooling might be able to access content previously unavailable to them, allowing bolstered exposure to information and media. While its impact could cut across many domains of everyday lives, its impact on education, which other machine translation studies have also examined (Lee, 2020), could be significant. In formal educational settings, for instance, students and educators belonging to low-resource language groups would be able to tap into more books, research articles, and archives than before. Within the realms of informal learning, low-resource language speakers could experience greater access to information from global news outlets and social media platforms, as well as online encyclopedias such as Wikipedia. In these latter spaces, where the production of knowledge and content moves at a breakneck speed, the value of translation cannot be downplayed (Bywood et al., 2017; Singh et al., 2012).

The benefits of better quality translation are not exclusive for underserved communities. For communities currently being served by other translation services, the improvement in translation quality would boost their overall accessibility and utilization of the web's offerings. Such quality improvements could also lead to more streamlined knowledge acquisition and communicative processes. The cognitive energy one saves from deciphering poorly translated content could then be channeled to performing other more important tasks.

Because language is intrinsically tied to culture, for many low-resource languages facing endangerment, the threat of losing one's language could also mean the erosion of one's heritage (Sallabank, 2013). NLLB could motivate more low-resource language writers or content creators to share localized knowledge or various aspects of their culture with both cultural insiders and outsiders through social media platforms or websites like Wikipedia. Giving individuals access to new translation tools could thus open up a valuable avenue for bidirectional learning. In the long run, such generative processes could create dents on the global knowledge system, challenge Western-centric modes of knowledge production and dissemination, and aid in the revitalization of certain minority cultures and languages (Bird, 2019; Bird and Chiang, 2012).

### 9.2 The Janus-faced Nature of Digital Participation

The benefits of a technological intervention like NLLB needs to be carefully weighed against the costs and risks it might incur on low-resource language groups and other stakeholders.

An increase in digital participation and linguistic representation, for example, heightens the visibility of a group (Bucher, 2012). Such visibility may amplify the odds of groups becoming targeted for surveillance and censorship (Treré, 2016; Zuboff, 2019). Relatedly, affected communities may also become more susceptible to misinformation, online scams, or hate speech (Gereme et al., 2021; Hossain et al., 2020). In other words, the expansion of language and information access renders certain groups more vulnerable to longstanding issues plaguing digital communities at large. While no simple solution exists for these complex issues, we hope that NLLB could be leveraged for its cross-lingual potential (Conneau et al., 2018) to strengthen existing (and typically monolingual) tooling designed to detect and classify hate speech, phishing, and other socially harmful online texts in low-resource languages (Khonji et al., 2013; MacAvaney et al., 2019). Recognizing that such tools act as a first defense, we believe that long term, structural investments aimed at curbing nefarious digital activities and improving online literacy need to go hand in hand with the introduction of new tools such as NLLB.

While access to translation could boost overall digital participation, it could also exacerbate existing digital inequities at a local or community-level. For one, those with technological know-how will benefit from NLLB more than those without. Demographically, these patterns are reflected through differences in factors such as age, education, economic standing, and rurality (Elena-Bucea et al., 2021; Hindman, 2000). Moreover, because technological infrastructure is unevenly distributed in many parts of the world, communities that are already lagging behind when it comes to internet access may experience aggregated information gaps compared to their better-served counterparts. In other words, the disparities in knowledge access, social connectivity, and economic mobility could deepen if the structural measures needed to rectify existing challenges that affect low-resource language communities are not in place.

Given that the primary goal of NLLB is to reduce language inequities in a global context, more and more low-resource languages will be incorporated into the project (or others alike) in the long run. Along this trajectory, those within this research space will inevitably encounter an increasing number of vulnerable communities that may resist the idea of letting technological entities they have little ties to capitalize on their languages (Coffey, 2021). To this end, being reflexive in our approach and prioritizing relationships with local institutions and community members to better understand their needs and concerns is of utmost importance in any expansion efforts. This motivation further explains why we have developed long-term in-depth interview and fieldwork studies with speakers of low-resource languages to understand how our intervention might impact their day-to-day lives. Moreover, collaborations with research groups that already possess vested interest in the topic at hand are imperative to any future success. We hope that spotlighting mutual interests and shared moral visions would facilitate resource and knowledge pooling, paving the way for long-term cooperation amongst various stakeholders.

### 9.3 The Future of NLLB: A Collective Responsibility

Recognizing that solving language disparities through machine translation is a mammoth task, NLLB's decision to make datasets and models publicly available encourages innovation through community production and collaboration (Weber, 2004). Open-sourcing our datasets and models not only advocates for transparency in the development of AI technologies, it further prevents the duplication of effort and allows machine translation practitioners to devote their energy at identifying gaps and building on the work we have done (Kogut and Metiu, 2001). Furthermore, we are actively developing mechanisms that would allow us to support data scientists and researchers who wish to use and adapt our models to meet their own needs, and provide the necessary assistance when needed. We are cognizant that open-sourcing does not translate into equitable access; the technological and infrastructural barriers to deploying our models and datasets remain high for researchers in many parts of the world. As such, to alleviate issues around the uneven distribution of computing power, we plan to develop toolkits and issue grants to under-resourced labs to assist them in their research endeavors. We believe that a collaborative mindset, alongside systematic and long-term documentation, will allow us to better assess the impact we have on the various communities implicated in our project.

Moreover, sharing NLLB with the larger scientific and research community will allow those with diverse expertise to contribute to the advancement of the project. In many ways, the composition of the NLLB effort speaks to the centrality of interdisciplinarity in shaping our vision. Machine translation lies at the intersection of technological, cultural, and societal development, and thus requires scholars with disparate training and standpoints to fully comprehend every angle (Kusters et al., 2020). It is our hope in future iterations, NLLB continues to expand to include of scholars from fields underrepresented in the world of machine translation and AI, particularly those from humanities and social sciences background. More importantly, we hope that teams developing such initiatives would come from a wide range of race, gender, and cultural identities, much like the communities whose lives we seek to improve.

Finally, we want to stress that overcoming the challenges that prevent the web from being truly accessible to speakers of all languages requires a multifaceted approach. As a single technological intervention, NLLB is all but one piece in a massive puzzle. Policy interventions aimed at more fundamental issues surrounding education, internet access, and digital literacy are imperative to eradicating the structural problem of language disparities. We are committed to working together with various stakeholders as we continue our path to materialize translation technologies that make the web a more accessible place, regardless of the language one speaks.

## 10. Contributions

We outline the contributions of each member of No Language Left Behind, grouped by section and sorted alphabetically by last name. Each person is only mentioned once even though many contributed to several areas. No amount of space could fully describe the passion and contributions of every single person involved in bringing this effort to life.

### Data

**Bapi Akula** - monolingual data pipeline to go from CommonCrawl to deduplicated, filtered sentences

**Pierre Andrews** - engineering lead for data, led the development of `stopes`, mining and monolingual cleaning pipelines

**Onur Çelebi** - LID for 200+ languages, open source of mined data, implementation and discussions to develop LASER3 and monolingual pipeline
**Kenneth Heafield** - technical feedback on monolingual data quality, open source of mined data
**Kevin Heffernan** - implementation and experimentation to create LASER3, produced mined bitext
**Semarley Jarrett** - worked on data partnerships
**Holger Schwenk** - research lead for data, also led the development of LASER3 and bitext mining
**Guillaume Wenzek** - technical feedback on data open-source and monolingual data, implemented infrastructure for Wikipedia model deployment

**Modeling**

**Loic Barrault** - visualization of model scores
**Shruti Bhosale** - research lead for modeling, led experimentation and research direction on MoE
**James Cross** - experimentation for incorporating SSL
**Maha Elbayad** - implementation and experimentation with various MoE architectures, analysis of model quality and properties of MoE models
**Vedanuj Goswami** - research lead for modeling, led experimentation and research direction on BT, SSL, led execution on the 200 languages goal
**Jean Maillard** - creation of bitext filtering pipeline, experimentation on backtranslation, effect of NLLB-Seed, and transliteration
**Kaushik Ram Sadagopan** - experimentation on data quality on model training, effect of NLLB-Seed
**Anna Sun** - experimentation with effective MoE regularization, creation of Wikipedia Translation models, model distillation, experimentation curriculum learning
**Chau Tran** - experimentation with self-supervised learning, model distillation

**Evaluation**

**Marta R. Costa-jussà** - analysis, mitigation and interpretability of toxicity, created data&model sheets, worked on ethics research
**Cynthia Gao** - led and worked on all human data collection and annotations (FLORES 200, NLLB-Seed, Human Evaluations)
**John Hoffman** - analysis of translation quality and human evaluation experiments
**Elahe Kalbassi** - worked on all human data collection and annotations (FLORES 200, NLLB-Seed, Human Evaluations)
**Philipp Koehn** - technical feedback on mining and monolingual data quality, development of XSTS and human evaluation study analysis
**Daniel Licht** - analysis of toxicity in translation and human evaluation experiments, development of XSTS
**Dirk Rowe** - designed figures and UI for human studies
**Shannon Spruit** - advised ethics research on the creation of machine translation models
**Skyler Wang** - helped design and conducted interview studies to understand the impact of translation, advised research on ethics of translation and development of models
**Al Youngblood** - designed and conducted interview studies to understand the impact of translation on people, worked on ethics research

**Linguistics**

**Gabriel Mejia Gonzalez** - Arabic dialectal variation, transliteration
**Prangthip Hansanti** - transliteration, model output quality, toxicity detection
**Janice Lam** - LID improvement, FLORES 200 language information and codes, FLORES data quality, toxicity detection
**Christophe Ropers** - linguist lead, FLORES 200 language information and codes, model output quality, toxicity detection, ethics

**Organization**

**Necip Fazil Ayan** - research director, helped with the overall direction and strategy
**Sergey Edunov** - manager lead, provided engineering and open-source direction
**Angela Fan** - research and project lead, provided research direction for the entire project
**Francisco Guzmán** - research and engineering manager, provided direction for evaluation research
**Alexandre Mourachko** - engineering manager, provided direction for data research
**Safiyyah Saleem** - technical program manager
**Jeff Wang** - product manager, led Wikimedia Foundation collaboration

## 11. Acknowledgements

We thank our interns for the energy and discussions they brought in: Christos Baziotis, Dheeru Dua, Alex Guo, Oana Ignat, Ammar Kamran, Tasnim Mohiuddin, Andre Niyongabo Rubungo, Simeng Sun, Steven Tan, Haoran Xu, Shijie Wu, Yuwei Zhang. We thank the Wikimedia Foundation staff and Wikimedia volunteers who worked with us and provided feedback to our model. We thank Vishrav Chaudhary for help with the data pipeline. We thank Edouard Grave for his help in scaling fasttext to all Flores-200 languages. We thank Mona Diab for XSTS work and Lucia Specia for discussions on Toxicity and XSTS. We thank Javier Ferrando and Carlos Escolano for their invaluable help in using the ALTI+ method. We thank Brian O'Horo and Justine Kao for their insights and guidance. We thank Gloria Chang, Carole-Jean Wu and Ramya Raghavendra for helping us compute the $CO_2$ cost of our models. We thank Anjali Sridhar for help with FSDP. We thank Scott Jeschonek, Giri Anantharaman, Diego Sarina, Joaquin Colombo, Sanjana Krishnan, Dinesh Kannappan, Kalyan Saladi, Vivek Pai, Amit Yajurvedi, and Shubho Sengupta for their help with training infrastructure. We thank Kyle Johnson for his help with UXR studies and model evaluation. We thank Antoine Bordes, Marina Zannoli, and Chris Moghbel for supporting this project. We thank Pascale Fung for inspirational and generative discussions on the human-centered objectives of the project. We thank Nicolas Usunier, Sebastian Riedel, Shubho Sengupta, and Emily Dinan for helpful feedback on the paper. Finally, we thank all of the translators, reviewers, human evaluators, linguists, as well as the translation and quality assurance agencies we partnered with, for helping create Flores-200, NLLB-Seed, NLLB-MD, Toxicity-200, performing our human evaluations, and teaching us about their native languages.

## References

Julien Abadji, Pedro Javier Ortiz Suárez, Laurent Romary, and Benoît Sagot. Towards a cleaner document-oriented multilingual crawled corpus. *CoRR*, abs/2201.06642, 2022. URL https://arxiv.org/abs/2201.06642.

Solomon Teferra Abate, Michael Melese, Martha Yifiru Tachbelie, Million Meshesha, Solomon Atinafu, Wondwossen Mulugeta, Yaregal Assabie, Hafte Abera, Binyam Ephrem Seyoum, Tewodros Abebe, et al. Parallel corpora for bi-directional statistical machine translation for seven ethiopian language pairs. In *Proceedings of the First Workshop on Linguistic Resources for Natural Language Processing*, pages 83–90, 2018.

Jade Abbott and Laura Martinus. Benchmarking neural machine translation for Southern African languages. In *Proceedings of the 2019 Workshop on Widening NLP*, pages 98–101, Florence, Italy, August 2019. Association for Computational Linguistics. URL https://aclanthology.org/W19-3632.

Ahmed Abdelali, Francisco Guzman, Hassan Sajjad, and Stephan Vogel. The AMARA corpus: Building parallel language resources for the educational domain. In *Proceedings of the Ninth International Conference on Language Resources and Evaluation (LREC'14)*, pages 1856–1862, Reykjavik, Iceland, May 2014. European Language Resources Association (ELRA). URL http://www.lrec-conf.org/proceedings/lrec2014/pdf/877_Paper.pdf.

Sadaf Abdul-Rauf and Holger Schwenk. On the Use of Comparable Corpora to Improve SMT performance. In *EACL*, pages 16–23, 2009. URL http://www.aclweb.org/anthology/E09-1003.

David Adelani, Dana Ruiter, Jesujoba Alabi, Damilola Adebonojo, Adesina Ayeni, Mofe Adeyemi, Ayodele Esther Awokoya, and Cristina España-Bonet. The effect of domain and diacritics in Yoruba–English neural machine translation. In *Proceedings of the 18th Biennial Machine Translation Summit (Volume 1: Research Track)*, pages 61–75, Virtual, August 2021. Association for Machine Translation in the Americas. URL https://aclanthology.org/2021.mtsummit-research.6.

David Ifeoluwa Adelani, Jesujoba Oluwadara Alabi, Angela Fan, Julia Kreutzer, Xiaoyu Shen, Machel Reid, Dana Ruiter, Dietrich Klakow, Peter Nabende, Ernie Chang, Tajuddeen Gwadabe, Freshia Sackey, Bonaventure F. P. Dossou, Chris Chinenye Emezue, Colin Leong, Michael Beukman, Shamsuddeen Hassan Muhammad, Guyo Dub Jarso, Oreen Yousuf, Andre Niyongabo Rubungo, Gilles Hacheme, Eric Peter Wairagala, Muhammad Umair Nasir, Benjamin Ayoade Ajibade, Tunde Oluwaseyi Ajayi, Yvonne Wambui Gitau, Jade Abbott, Mohamed Ahmed, Millicent Ochieng, Anuoluwapo Aremu, Perez Ogayo, Jonathan Mukiibi, Fatoumata Ouoba Kabore, Godson Koffi Kalipe, Derguene Mbaye, Allahsera Auguste Tapo, Victoire Memdjokam Koagne, Edwin Munkoh-Buabeng, Valencia Wagner, Idris Abdulmumin, Ayodele Awokoya, Happy Buzaaba, Blessing Sibanda, Andiswa Bukula, and Sam Manthalu. A few thousand translations go a long way! leveraging pre-trained models for african news translation. *CoRR*, abs/2205.02022, 2022. doi: 10.48550/ARXIV.2205.02022. URL https://arxiv.org/abs/2205.02022.

Eneko Agirre, Daniel Cer, Mona Diab, and Aitor Gonzalez-Agirre. SemEval-2012 task 6: A pilot on semantic textual similarity. In **SEM 2012: The First Joint Conference on Lexical and Computational Semantics – Volume 1: Proceedings of the main conference and the shared task, and Volume 2: Proceedings of the Sixth International Workshop on Semantic Evaluation (SemEval 2012)*, pages 385–393, Montréal, Canada, June 2012. Association for Computational Linguistics. URL https://aclanthology.org/S12-1051.

Orevaoghene Ahia, Julia Kreutzer, and Sara Hooker. The low-resource double bind: An empirical study of pruning for low-resource machine translation. *CoRR*, abs/2110.03036, 2021. URL https://arxiv.org/abs/2110.03036.

Benjamin Akera, Jonathan Mukiibi, Lydia Sanyu Naggayi, Claire Babirye, Isaac Owomugisha, Solomon Nsumba, Joyce Nakatumba-Nabende, Engineer Bainomugisha, Ernest Mwebaze, and John Quinn. Machine translation for african languages: Community creation of datasets and models in uganda. In *3rd Workshop on African Natural Language Processing*, 2022. URL https://openreview.net/forum?id=BK-z5qzEU-9.

Farhad Akhbardeh, Arkady Arkhangorodsky, Magdalena Biesialska, Ondřej Bojar, Rajen Chatterjee, Vishrav Chaudhary, Marta R. Costa-jussa, Cristina España-Bonet, Angela Fan, Christian Federmann, Markus Freitag, Yvette Graham, Roman Grundkiewicz, Barry Haddow, Leonie Harter, Kenneth Heafield, Christopher Homan, Matthias Huck, Kwabena Amponsah-Kaakyire, Jungo Kasai, Daniel Khashabi, Kevin Knight, Tom Kocmi, Philipp Koehn, Nicholas Lourie, Christof Monz, Makoto Morishita, Masaaki Nagata, Ajay Nagesh, Toshiaki Nakazawa, Matteo Negri, Santanu Pal, Allahsera Auguste Tapo, Marco Turchi, Valentin Vydrin, and Marcos Zampieri. Findings of the 2021 conference on machine translation (WMT21). In *Proceedings of the Sixth Conference on Machine Translation*, pages 1–88, Online, November 2021. Association for Computational Linguistics. URL https://aclanthology.org/2021.wmt-1.1.

Amjad Almahairi, Nicolas Ballas, Tim Cooijmans, Yin Zheng, Hugo Larochelle, and Aaron Courville. Dynamic capacity networks. In *Proceedings of the 33rd International Conference on International Conference on Machine Learning - Volume 48*, ICML'16, page 2091–2100. JMLR.org, 2016.

Faisal Alshargi, Shahd Dibas, Sakhar Alkhereyf, Reem Faraj, Basmah Abdulkareem, Sane Yagi, Ouafaa Kacha, Nizar Habash, and Owen Rambow. Morphologically annotated corpora for seven Arabic dialects: Taizi, sanaani, najdi, jordanian, syrian, iraqi and Moroccan. In *Proceedings of the Fourth Arabic Natural Language Processing Workshop*, 2019.

Charity Delmus Alupo, Daniel Omeiza, and David Vernon. Realizing the potential of ai in africa. *Towards Trustworthy Artificial Intelligence Systems*, 2021.

Antonios Anastasopoulos, Alessandro Cattelan, Zi-Yi Dou, Marcello Federico, Christian Federmann, Dmitriy Genzel, Franscisco Guzmán, Junjie Hu, Macduff Hughes, Philipp Koehn, Rosie Lazar, Will Lewis, Graham Neubig, Mengmeng Niu, Alp Öktem, Eric Paquin, Grace Tang, and Sylwia Tur. TICO-19: the translation initiative for COvid-19. In *Proceedings of the 1st Workshop on NLP for COVID-19 (Part 2) at EMNLP 2020*,

Online, December 2020. Association for Computational Linguistics. doi: 10.18653/v1/2020.nlpcovid19-2.5. URL https://aclanthology.org/2020.nlpcovid19-2.5.

Antonios Anastasopoulos, Ondřej Bojar, Jacob Bremerman, Roldano Cattoni, Maha Elbayad, Marcello Federico, Xutai Ma, Satoshi Nakamura, Matteo Negri, Jan Niehues, Juan Pino, Elizabeth Salesky, Sebastian Stüker, Katsuhito Sudoh, Marco Turchi, Alexander Waibel, Changhan Wang, and Matthew Wiesner. Findings of the IWSLT 2021 evaluation campaign. In *Proceedings of the 18th International Conference on Spoken Language Translation (IWSLT 2021)*, pages 1–29, Bangkok, Thailand (online), August 2021. Association for Computational Linguistics. doi: 10.18653/v1/2021.iwslt-1.1. URL https://aclanthology.org/2021.iwslt-1.1.

Patrick Andries. *Proposition d'ajout de l'écriture tifinaghe. Organisation internationale de normalisation.* Jeu universel des caractères codés sur octets (JUC). ORGANISATION INTERNATIONALE DE NORMALISATION, 2004.

Mohd Zeeshan Ansari, M. M. Sufyan Beg, Tanvir Ahmad, Mohd Jazib Khan, and Ghazali Wasim. Language identification of hindi-english tweets using code-mixed BERT. *CoRR*, abs/2107.01202, 2021. URL https://arxiv.org/abs/2107.01202.

Naveen Arivazhagan, Ankur Bapna, Orhan Firat, Dmitry Lepikhin, Melvin Johnson, Maxim Krikun, Mia Xu Chen, Yuan Cao, George F. Foster, Colin Cherry, Wolfgang Macherey, Zhifeng Chen, and Yonghui Wu. Massively multilingual neural machine translation in the wild: Findings and challenges. *CoRR*, abs/1907.05019, 2019. URL http://arxiv.org/abs/1907.05019.

Nigel Armstrong and Ian E. Mackenzie. *Social levelling, or anti-standardization*, pages 161–207. Palgrave Macmillan UK, London, 2013. ISBN 978-1-137-28439-6. doi: 10.1057/9781137284396_6. URL https://doi.org/10.1057/9781137284396_6.

Alejandro Barredo Arrieta, Natalia Díaz-Rodríguez, Javier Del Ser, Adrien Bennetot, Siham Tabik, Alberto Barbado, Salvador García, Sergio Gil-López, Daniel Molina, Richard Benjamins, et al. Explainable artificial intelligence (xai): Concepts, taxonomies, opportunities and challenges toward responsible ai. *Information fusion*, 58:82–115, 2020.

Mikel Artetxe and Holger Schwenk. Margin-based parallel corpus mining with multilingual sentence embeddings. In *Proceedings of the 57th Annual Meeting of the Association for Computational Linguistics*, pages 3197–3203, 2019a.

Mikel Artetxe and Holger Schwenk. Massively multilingual sentence embeddings for zero-shot cross-lingual transfer and beyond. *TACL*, pages 597–610, 2019b.

Mikel Artetxe, Shruti Bhosale, Naman Goyal, Todor Mihaylov, Myle Ott, Sam Shleifer, Xi Victoria Lin, Jingfei Du, Srinivasan Iyer, Ramakanth Pasunuru, Giri Anantharaman, Xian Li, Shuohui Chen, Halil Akin, Mandeep Baines, Louis Martin, Xing Zhou, Punit Singh Koura, Brian O'Horo, Jeff Wang, Luke Zettlemoyer, Mona T. Diab, Zornitsa Kozareva, and Ves Stoyanov. Efficient large scale language modeling with mixtures of experts. *CoRR*, abs/2112.10684, 2021. URL https://arxiv.org/abs/2112.10684.

Andoni Azpeitia, Thierry Etchegoyhen, and Eva Martínez Garcia. Weighted Set-Theoretic Alignment of Comparable Sentences. In *BUCC*, pages 41–45, 2017. URL http://aclweb.org/anthology/W17-2508.

Andoni Azpeitia, Thierry Etchegoyhen, and Eva Martínez Garcia. Extracting Parallel Sentences from Comparable Corpora with STACC Variants. In *BUCC*, May 2018.

Paul Azunre, Lawrence Adu-Gyamfi, Esther Appiah, Felix Akwerh, Salomey Osei, Cynthia Amoaba, Salomey Afua Addo, Edwin Buabeng-Munkoh, Nana Boateng, Franklin Adjei, and Bernard Adabankah. English-akuapem twi parallel corpus, January 2021a. URL https://doi.org/10.5281/zenodo.4432117.

Paul Azunre, Salomey Osei, Salomey Addo, Lawrence Asamoah Adu-Gyamfi, Stephen Moore, Bernard Adabankah, Bernard Opoku, Clara Asare-Nyarko, Samuel Nyarko, Cynthia Amoaba, et al. English-twi parallel corpus for machine translation. *arXiv preprint arXiv:2103.15625*, 2021b.

Paul Azunre, Salomey Osei, Salomey Addo, Lawrence Asamoah Adu-Gyamfi, Stephen Moore, Bernard Adabankah, Bernard Opoku, Clara Asare-Nyarko, Samuel Nyarko, Cynthia Amoaba, et al. NLP for ghanaian languages. *arXiv preprint arXiv:2103.15475*, 2021c.

Lei Jimmy Ba, Jamie Ryan Kiros, and Geoffrey E. Hinton. Layer normalization. *CoRR*, abs/1607.06450, 2016. URL http://arxiv.org/abs/1607.06450.

Claire Babirye, Joyce Nakatumba-Nabende, Andrew Katumba, Ronald Ogwang, Jeremy Tusubira Francis, Jonathan Mukiibi, Medadi Ssentanda, Lilian D Wanzare, and Davis David. Building text and speech datasets for low resourced languages: A case of languages in east africa. In *3rd Workshop on African Natural Language Processing*, 2022. URL https://openreview.net/forum?id=SO-U99z4U-q.

Dzmitry Bahdanau, Kyung Hyun Cho, and Yoshua Bengio. Neural machine translation by jointly learning to align and translate. In *3rd International Conference on Learning Representations, ICLR 2015*, 2015.

Loretta Baldassar, Mihaela Nedelcu, Laura Merla, and Raelene Wilding. Ict-based co-presence in transnational families and communities: Challenging the premise of face-to-face proximity in sustaining relationships. *Global Networks*, 16(2):133–144, 2016.

Laith H. Baniata, Isaac. K. E. Ampomah, and Seyoung Park. A transformer-based neural machine translation model for arabic dialects that utilizes subword units. *Sensors*, 21 (19), 2021. ISSN 1424-8220. doi: 10.3390/s21196509. URL https://www.mdpi.com/1424-8220/21/19/6509.

Marta Bañón, Pinzhen Chen, Barry Haddow, Kenneth Heafield, Hieu Hoang, Miquel Esplà-Gomis, Mikel L. Forcada, Amir Kamran, Faheem Kirefu, Philipp Koehn, Sergio Ortiz Rojas, Leopoldo Pla Sempere, Gema Ramírez-Sánchez, Elsa Sarrías, Marek Strelec, Brian Thompson, William Waites, Dion Wiggins, and Jaume Zaragoza. ParaCrawl: Web-scale acquisition of parallel corpora. In *Proceedings of the 58th Annual Meeting of the Association for Computational Linguistics*, pages 4555–4567, Online, July 2020.

Association for Computational Linguistics. doi: 10.18653/v1/2020.acl-main.417. URL https://aclanthology.org/2020.acl-main.417.

Ankur Bapna, Isaac Caswell, Julia Kreutzer, Orhan Firat, Daan van Esch, Aditya Siddhant, Mengmeng Niu, Pallavi Baljekar, Xavier Garcia, Wolfgang Macherey, Theresa Breiner, Vera Axelrod, Jason Riesa, Yuan Cao, Mia Xu Chen, Klaus Macherey, Maxim Krikun, Pidong Wang, Alexander Gutkin, Apurva Shah, Yanping Huang, Zhifeng Chen, Yonghui Wu, and Macduff Hughes. Building machine translation systems for the next thousand languages, 2022. URL https://arxiv.org/abs/2205.03983.

Emily Bender. The #benderrule: On naming the languages we study and why it matters. *The Gradient*, 2019.

Emily M Bender, Timnit Gebru, Angelina McMillan-Major, and Margaret Mitchell. On the dangers of stochastic parrots: Can language models be too big? In *Proceedings of the 2021 ACM Conference on Fairness, Accountability, and Transparency*, pages 610–623, 2021.

Yoshua Bengio, Jérôme Louradour, Ronan Collobert, and Jason Weston. Curriculum learning. In *Proceedings of the 26th annual international conference on machine learning*, pages 41–48, 2009.

Yoshua Bengio, Nicholas Léonard, and Aaron C. Courville. Estimating or propagating gradients through stochastic neurons for conditional computation. *CoRR*, abs/1308.3432, 2013. URL http://arxiv.org/abs/1308.3432.

Abhik Bhattacharjee, Tahmid Hasan, Wasi Uddin Ahmad, and Rifat Shahriyar. Banglanlg: Benchmarks and resources for evaluating low-resource natural language generation in bangla. *arXiv preprint arXiv:2205.11081*, 2022.

Steven Bird. Designing for language revitalisation. In Gilles Adda, Khalid Choukri, Irmgarda Kasinskaite-Buddeberg, Joseph Mariani, Hélène Mazo, and Sakriani Sakti, editors, *Language Technologies for All (LT4All)*, pages 296–299. European Language Resources Association (ELRA), 2019. URL https://en.unesco.org/LT4All. International Conference Language Technologies for All, LT4All ; Conference date: 04-12-2019 Through 06-12-2019.

Steven Bird and David Chiang. Machine translation for language preservation. In *Proceedings of COLING 2012: Posters*, pages 125–134, 2012.

Su Lin Blodgett, Solon Barocas, Hal Daumé III, and Hanna Wallach. Language (technology) is power: A critical survey of "bias" in NLP. In *Proceedings of the 58th Annual Meeting of the Association for Computational Linguistics*, pages 5454–5476, Online, July 2020. Association for Computational Linguistics. doi: 10.18653/v1/2020.acl-main.485. URL https://aclanthology.org/2020.acl-main.485.

Su Lin Blodgett, Q Vera Liao, Alexandra Olteanu, Rada Mihalcea, Michael Muller, Morgan Klaus Scheuerman, Chenhao Tan, and Qian Yang. Responsible language technologies:

Foreseeing and mitigating harms. In *CHI Conference on Human Factors in Computing Systems Extended Abstracts*, pages 1–3, 2022.

Piotr Bojanowski, Edouard Grave, Armand Joulin, and Tomas Mikolov. Enriching word vectors with subword information. *Transactions of the Association for Computational Linguistics*, 5:135–146, 2017. ISSN 2307-387X.

Ondřej Bojar, Rajen Chatterjee, Christian Federmann, Yvette Graham, Barry Haddow, Matthias Huck, Antonio Jimeno Yepes, Philipp Koehn, Varvara Logacheva, Christof Monz, Matteo Negri, Aurélie Névéol, Mariana Neves, Martin Popel, Matt Post, Raphael Rubino, Carolina Scarton, Lucia Specia, Marco Turchi, Karin Verspoor, and Marcos Zampieri. Findings of the 2016 conference on machine translation. In *Proceedings of the First Conference on Machine Translation: Volume 2, Shared Task Papers*, pages 131–198, Berlin, Germany, August 2016. Association for Computational Linguistics. doi: 10.18653/v1/W16-2301. URL `https://aclanthology.org/W16-2301`.

Marcel Bollmann. A large-scale comparison of historical text normalization systems. In *Proceedings of the 2019 Conference of the North American Chapter of the Association for Computational Linguistics: Human Language Technologies, Volume 1 (Long and Short Papers)*, pages 3885–3898, Minneapolis, Minnesota, June 2019. Association for Computational Linguistics. doi: 10.18653/v1/N19-1389. URL `https://aclanthology.org/N19-1389`.

Houda Bouamor and Hassan Sajjad. H2@BUCC18: Parallel Sentence Extraction from Comparable Corpora Using Multilingual Sentence Embeddings. In *BUCC*, May 2018.

Houda Bouamor, Nizar Habash, Mohammad Salameh, Wajdi Zaghouani, Owen Rambow, Dana Abdulrahim, Ossama Obeid, Salam Khalifa, Fadhl Eryani, Alexander Erdmann, and Kemal Oflazer. The MADAR Arabic dialect corpus and lexicon. In *Proceedings of the Eleventh International Conference on Language Resources and Evaluation (LREC 2018)*, Miyazaki, Japan, May 2018. European Language Resources Association (ELRA). URL `https://aclanthology.org/L18-1535`.

Houda Bouamor, Sabit Hassan, and Nizar Habash. The MADAR shared task on Arabic fine-grained dialect identification. In *Proceedings of the Fourth Arabic Natural Language Processing Workshop*, pages 199–207, Florence, Italy, August 2019. Association for Computational Linguistics. doi: 10.18653/v1/W19-4622. URL `https://aclanthology.org/W19-4622`.

Pierre Bourdieu. *Distinction: A social critique of the judgement of taste*. Harvard University Press, 1987.

Samuel R. Bowman, Gabor Angeli, Christopher Potts, and Christopher D. Manning. A large annotated corpus for learning natural language inference. In *EMNLP*, pages 632–642, Lisbon, Portugal, September 2015. Association for Computational Linguistics. doi: 10.18653/v1/D15-1075. URL `https://aclanthology.org/D15-1075`.

Peter F Brown, Vincent J Della Pietra, Stephen A Della Pietra, and Robert L Mercer. The mathematics of statistical machine translation: Parameter estimation. *Computational linguistics*, 1993.

Ralf D Brown. Non-linear mapping for improved identification of 1300+ languages. In *Proceedings of the 2014 Conference on Empirical Methods in Natural Language Processing (EMNLP)*, pages 627–632, 2014.

Taina Bucher. Want to be on the top? algorithmic power and the threat of invisibility on facebook. *New media & society*, 14(7):1164–1180, 2012.

Cristian Buciluǎ, Rich Caruana, and Alexandru Niculescu-Mizil. Model compression. In *Proceedings of the 12th ACM SIGKDD international conference on Knowledge discovery and data mining*, pages 535–541, 2006.

Christian Buck and Philipp Koehn. Findings of the wmt 2016 bilingual document alignment shared task. In *Proceedings of the First Conference on Machine Translation*, pages 554–563, Berlin, Germany, August 2016. Association for Computational Linguistics. URL `http://www.aclweb.org/anthology/W/W16/W16-2347`.

Lindsay Bywood, Panayota Georgakopoulou, and Thierry Etchegoyhen. Embracing the threat: machine translation as a solution for subtitling. *Perspectives*, 25(3):492–508, 2017.

Isaac Caswell, Ciprian Chelba, and David Grangier. Tagged back-translation. In *Proceedings of the Fourth Conference on Machine Translation (Volume 1: Research Papers)*, pages 53–63, 2019.

Isaac Caswell, Theresa Breiner, Daan van Esch, and Ankur Bapna. Language ID in the wild: Unexpected challenges on the path to a thousand-language web text corpus. In *Proceedings of the 28th International Conference on Computational Linguistics*, pages 6588–6608, Barcelona, Spain (Online), December 2020. International Committee on Computational Linguistics. doi: 10.18653/v1/2020.coling-main.579. URL `https://aclanthology.org/2020.coling-main.579`.

Mauro Cettolo, Jan Niehues, Sebastian Stüker, Luisa Bentivogli, and Marcello Federico. Report on the 11th IWSLT evaluation campaign. In *Proceedings of the 11th International Workshop on Spoken Language Translation: Evaluation Campaign*, pages 2–17, Lake Tahoe, California, December 2014. URL `https://aclanthology.org/2014.iwslt-evaluation.1`.

Mauro Cettolo, Marcello Federico, Luisa Bentivogli, Jan Niehues, Sebastian Stüker, Katsuhito Sudoh, Koichiro Yoshino, and Christian Federmann. Overview of the IWSLT 2017 evaluation campaign. In *Proceedings of the 14th International Conference on Spoken Language Translation*, pages 2–14, Tokyo, Japan, December 2017. International Workshop on Spoken Language Translation. URL `https://aclanthology.org/2017.iwslt-1.1`.

Guanhua Chen, Shuming Ma, Yun Chen, Dongdong Zhang, Jia Pan, Wenping Wang, and Furu Wei. Towards making the most of multilingual pretraining for zero-shot neural machine translation. *CoRR*, abs/2110.08547, 2021. URL `https://arxiv.org/abs/2110.08547`.

Zewen Chi, Li Dong, Shuming Ma, Shaohan Huang Xian-Ling Mao, Heyan Huang, and Furu Wei. mt6: Multilingual pretrained text-to-text transformer with translation pairs. *arXiv preprint arXiv:2104.08692*, 2021.

Kyunghyun Cho, Bart van Merriënboer, Dzmitry Bahdanau, and Yoshua Bengio. On the properties of neural machine translation: Encoder–decoder approaches. In *Proceedings of SSST-8, Eighth Workshop on Syntax, Semantics and Structure in Statistical Translation*, pages 103–111, 2014.

Jack Choquette, Wishwesh Gandhi, Olivier Giroux, Nick Stam, and Ronny Krashinsky. Nvidia a100 tensor core gpu: Performance and innovation. *IEEE Micro*, 41(2):29–35, 2021.

Brian Christian. *The alignment problem: Machine learning and human values.* WW Norton & Company, 2020.

Christopher Cieri, Mike Maxwell, Stephanie Strassel, and Jennifer Tracey. Selection criteria for low resource language programs. In *Proceedings of the Tenth International Conference on Language Resources and Evaluation (LREC'16)*, pages 4543–4549, 2016.

Donavyn Coffey. Māori are trying to save their language from big tech, April 2021. URL `https://www.wired.co.uk/article/maori-language-tech`.

Alexis Conneau and Guillaume Lample. Cross-lingual language model pretraining. In H. Wallach, H. Larochelle, A. Beygelzimer, F. d'Alché-Buc, E. Fox, and R. Garnett, editors, *Advances in Neural Information Processing Systems*, volume 32. Curran Associates, Inc., 2019. URL `https://proceedings.neurips.cc/paper/2019/file/c04c19c2c2474dbf5f7ac4372c5b9af1-Paper.pdf`.

Alexis Conneau, Guillaume Lample, Ruty Rinott, Adina Williams, Samuel R Bowman, Holger Schwenk, and Veselin Stoyanov. Xnli: Evaluating cross-lingual sentence representations. *arXiv preprint arXiv:1809.05053*, 2018.

Alexis Conneau, Kartikay Khandelwal, Naman Goyal, Vishrav Chaudhary, Guillaume Wenzek, Francisco Guzmán, Edouard Grave, Myle Ott, Luke Zettlemoyer, and Veselin Stoyanov. Unsupervised cross-lingual representation learning at scale. In Dan Jurafsky, Joyce Chai, Natalie Schluter, and Joel R. Tetreault, editors, *Proceedings of the 58th Annual Meeting of the Association for Computational Linguistics, ACL 2020, Online, July 5-10, 2020*, pages 8440–8451. Association for Computational Linguistics, 2020. doi: 10.18653/v1/2020.acl-main.747. URL `https://doi.org/10.18653/v1/2020.acl-main.747`.

Marta R. Costa-jussà. An analysis of gender bias studies in natural language processing. *Nature Machine Intelligence*, 1(11):495–496, 2019.

Raj Dabre and Aneerav Sukhoo. Morisienmt: A dataset for mauritian creole machine translation. *arXiv preprint arXiv:2206.02421*, 2022.

Raj Dabre, Himani Shrotriya, Anoop Kunchukuttan, Ratish Puduppully, Mitesh M. Khapra, and Pratyush Kumar. Indicbart: A pre-trained model for natural language generation of indic languages. *CoRR*, abs/2109.02903, 2021. URL `https://arxiv.org/abs/2109.02903`.

Thomas Davidson, Debasmita Bhattacharya, and Ingmar Weber. Racial bias in hate speech and abusive language detection datasets. *CoRR*, abs/1905.12516, 2019. URL `http://arxiv.org/abs/1905.12516`.

Tullio De Mauro. *Storia linguistica dell'Italia repubblicana: dal 1946 ai nostri giorni*. Laterza, 2014. ISBN 9788858113622.

Kevin Degila, Godson Kalipe, Jamiil Touré Ali, and Momboladji Balogoun. Parallel text dataset for Neural Machine Translation (French -> Fongbe, French -> Ewe), November 2020. URL `https://doi.org/10.5281/zenodo.4266935`.

Stefano Demichelis and Jorgen W Weibull. Language, meaning, and games: A model of communication, coordination, and evolution. *American Economic Review*, 98(4):1292–1311, 2008.

Jacob Devlin, Ming-Wei Chang, Kenton Lee, and Kristina Toutanova. BERT: Pre-training of deep bidirectional transformers for language understanding. In *NAACL*, pages 4171–4186, 2019. URL `https://aclanthology.org/N19-1423`.

Prafulla Dhariwal, Heewoo Jun, Christine Payne, Jong Wook Kim, Alec Radford, and Ilya Sutskever. Jukebox: A generative model for music. *arXiv preprint arXiv:2005.00341*, 2020.

Jesse Dodge, Taylor Prewitt, Remi Tachet Des Combes, Erika Odmark, Roy Schwartz, Emma Strubell, Alexandra Sasha Luccioni, Noah A Smith, Nicole DeCario, and Will Buchanan. Measuring the carbon intensity of ai in cloud instances. *arXiv preprint arXiv:2206.05229*, 2022.

Liam Donaldson and Paul Rutter. Healthier, fairer, safe: the global health journey 2007 – 2017. Technical report, World Health Organization, May 2017.

Nan Du, Yanping Huang, Andrew M. Dai, Simon Tong, Dmitry Lepikhin, Yuanzhong Xu, Maxim Krikun, Yanqi Zhou, Adams Wei Yu, Orhan Firat, Barret Zoph, Liam Fedus, Maarten Bosma, Zongwei Zhou, Tao Wang, Yu Emma Wang, Kellie Webster, Marie Pellat, Kevin Robinson, Kathy Meier-Hellstern, Toju Duke, Lucas Dixon, Kun Zhang, Quoc V. Le, Yonghui Wu, Zhifeng Chen, and Claire Cui. Glam: Efficient scaling of language models with mixture-of-experts. *CoRR*, abs/2112.06905, 2021. URL `https://arxiv.org/abs/2112.06905`.

Jonathan Dunn. Mapping languages: The corpus of global language use. *Language Resources and Evaluation*, 54(4):999–1018, 2020.

Bernardt Duvenhage. Short text language identification for under resourced languages. *CoRR*, abs/1911.07555, 2019. URL `http://arxiv.org/abs/1911.07555`.

Abteen Ebrahimi, Manuel Mager, Arturo Oncevay, Vishrav Chaudhary, Luis Chiruzzo, Angela Fan, John Ortega, Ricardo Ramos, Annette Rios, Ivan Vladimir Meza Ruiz, Gustavo Giménez-Lugo, Elisabeth Mager, Graham Neubig, Alexis Palmer, Rolando Coto-Solano, Thang Vu, and Katharina Kann. AmericasNLI: Evaluating zero-shot natural language understanding of pretrained multilingual models in truly low-resource languages. In *Proceedings of the 60th Annual Meeting of the Association for Computational Linguistics (Volume 1: Long Papers)*, pages 6279–6299, Dublin, Ireland, May 2022. Association for Computational Linguistics. doi: 10.18653/v1/2022.acl-long.435. URL `https://aclanthology.org/2022.acl-long.435`.

Sergey Edunov, Myle Ott, Michael Auli, and David Grangier. Understanding back-translation at scale. In *Proc. of EMNLP*, 2018.

Ahmed El-Kishky, Vishrav Chaudhary, Francisco Guzman, and Philipp Koehn. A massive collection of cross-lingual web-document pairs. In *EMNLP*, pages 5960–5969, 2020.

Anca Elena-Bucea, Frederico Cruz-Jesus, Tiago Oliveira, and Pedro Simões Coelho. Assessing the role of age, education, gender and income on the digital divide: evidence for the european union. *Information Systems Frontiers*, 23(4):1007–1021, 2021.

Chris Chinenye Emezue and Bonaventure F. P. Dossou. MMTAfrica: Multilingual machine translation for African languages. In *Proceedings of the Sixth Conference on Machine Translation*, pages 398–411, Online, November 2021. Association for Computational Linguistics. URL `https://aclanthology.org/2021.wmt-1.48`.

Chris Chinenye Emezue and Femi Pancrace Bonaventure Dossou. Ffr v1. 1: Fon-french neural machine translation. In *Proceedings of the The Fourth Widening Natural Language Processing Workshop*, pages 83–87, 2020.

Cristina España-Bonet, Ádám Csaba Varga, Alberto Barrón-Cedeño, and Josef van Genabith. An Empirical Analysis of NMT-Derived Interlingual Embeddings and their Use in Parallel Sentence Identification. *IEEE Journal of Selected Topics in Signal Processing*, pages 1340–1348, 2017.

Thierry Etchegoyhen and Andoni Azpeitia. Set-Theoretic Alignment for Comparable Corpora. In *ACL*, pages 2009–2018, 2016. doi: 10.18653/v1/P16-1189. URL `http://www.aclweb.org/anthology/P16-1189`.

Angela Fan, Shruti Bhosale, Holger Schwenk, Zhiyi Ma, Ahmed El-Kishky, Siddharth Goyal, Mandeep Baines, Onur Celebi, Guillaume Wenzek, Vishrav Chaudhary, Naman Goyal, Tom Birch, Vitaliy Liptchinsky, Sergey Edunov, Edouard Grave, Michael Auli, and Armand Joulin. Beyond english-centric multilingual machine translation. *The Journal of Machine Learning Research*, 2020.

William Fedus, Barret Zoph, and Noam Shazeer. Switch transformers: Scaling to trillion parameter models with simple and efficient sparsity. *Journal of Machine Learning Research*, 23(120):1–39, 2022. URL `http://jmlr.org/papers/v23/21-0998.html`.

Fangxiaoyu Feng, Yinfei Yang, Daniel Cer, Naveen Arivazhagan, and Wei Wang. Language-agnostic bert sentence embedding, 2020. URL `https://arxiv.org/abs/2007.01852`.

Charles A Ferguson. Diglossia. *word*, 15(2):325–340, 1959.

Javier Ferrando, Gerard I. Gállego, Belen Alastruey, Carlos Escolano, and Marta R. Costa-jussà. Towards opening the black box of neural machine translation: Source and target interpretations of the transformer, 2022. URL https://arxiv.org/abs/2205.11631.

Markus Freitag, Yaser Al-Onaizan, and Baskaran Sankaran. Ensemble distillation for neural machine translation. *CoRR*, abs/1702.01802, 2017. URL http://arxiv.org/abs/1702.01802.

Markus Freitag, Ricardo Rei, Nitika Mathur, Chi-kiu Lo, Craig Stewart, George Foster, Alon Lavie, and Ondřej Bojar. Results of the WMT21 metrics shared task: Evaluating metrics with expert-based human evaluations on TED and news domain. In *Proceedings of the Sixth Conference on Machine Translation*, pages 733–774, Online, November 2021. Association for Computational Linguistics. URL https://aclanthology.org/2021.wmt-1.73.

Batya Friedman and David G Hendry. *Value sensitive design: Shaping technology with moral imagination*. MIT Press, 2019.

Pascale Fung and Percy Cheung. Multi-level bootstrapping for extracting parallel sentences from a quasi-comparable corpus. In *COLING 2004, 20th International Conference on Computational Linguistics, Proceedings of the Conference, 23-27 August 2004, Geneva, Switzerland*, 2004. URL https://aclanthology.org/C04-1151/.

Samuel Gehman, Suchin Gururangan, Maarten Sap, Yejin Choi, and Noah A. Smith. Realtoxicityprompts: Evaluating neural toxic degeneration in language models. *CoRR*, abs/2009.11462, 2020. URL https://arxiv.org/abs/2009.11462.

Fantahun Gereme, William Zhu, Tewodros Ayall, and Dagmawi Alemu. Combating fake news in "low-resource" languages: Amharic fake news detection accompanied by resource crafting. *Information*, 12(1):20, 2021.

Jeff Good and Calvin Hendryx-Parker. Modeling contested categorization in linguistic databases. In *Proceedings of the EMELD 2006 Workshop on Digital Language Documentation: Tools and standards: The state of the art*, pages 20–22, 2006.

Google Jigsaw. Perpective api. https://www.perspectiveapi.com/, 2017. Accessed: 2022-05-03.

Mitchell A Gordon and Kevin Duh. Explaining sequence-level knowledge distillation as data-augmentation for neural machine translation. *arXiv preprint arXiv:1912.03334*, 2019.

Mitchell A. Gordon and Kevin Duh. Distill, adapt, distill: Training small, in-domain models for neural machine translation. *CoRR*, abs/2003.02877, 2020. URL https://arxiv.org/abs/2003.02877.

Cyril Goutte, Serge Léger, and Marine Carpuat. The NRC system for discriminating similar languages. In *Proceedings of the first workshop on applying NLP tools to similar languages, varieties and dialects*, pages 139–145, 2014.

Cyril Goutte, Serge Léger, Shervin Malmasi, and Marcos Zampieri. Discriminating similar languages: Evaluations and explorations. *CoRR*, abs/1610.00031, 2016. URL `http://arxiv.org/abs/1610.00031`.

Thamme Gowda, Zhao Zhang, Chris Mattmann, and Jonathan May. Many-to-English machine translation tools, data, and pretrained models. In *Proceedings of the 59th Annual Meeting of the Association for Computational Linguistics and the 11th International Joint Conference on Natural Language Processing: System Demonstrations*, pages 306–316, Online, August 2021. Association for Computational Linguistics. doi: 10.18653/v1/2021.acl-demo.37. URL `https://aclanthology.org/2021.acl-demo.37`.

Naman Goyal, Cynthia Gao, Vishrav Chaudhary, Peng-Jen Chen, Guillaume Wenzek, Da Ju, Sanjana Krishnan, Marc'Aurelio Ranzato, Francisco Guzmán, and Angela Fan. The Flores-101 evaluation benchmark for low-resource and multilingual machine translation. *Transactions of the Association for Computational Linguistics*, 10:522–538, 2022. doi: 10.1162/tacl_a_00474. URL `https://aclanthology.org/2022.tacl-1.30`.

Yvette Graham, Timothy Baldwin, Alistair Moffat, and Justin Zobel. Continuous measurement scales in human evaluation of machine translation. In *Proceedings of the 7th Linguistic Annotation Workshop and Interoperability with Discourse*, pages 33–41, Sofia, Bulgaria, August 2013. Association for Computational Linguistics. URL `https://aclanthology.org/W13-2305`.

Édouard Grave, Piotr Bojanowski, Prakhar Gupta, Armand Joulin, and Tomáš Mikolov. Learning word vectors for 157 languages. In *Proceedings of the Eleventh International Conference on Language Resources and Evaluation (LREC 2018)*, 2018.

Jiatao Gu, Hany Hassan, Jacob Devlin, and Victor O.K. Li. Universal neural machine translation for extremely low resource languages. In *Proceedings of the 2018 Conference of the North American Chapter of the Association for Computational Linguistics: Human Language Technologies, Volume 1 (Long Papers)*, pages 344–354, New Orleans, Louisiana, June 2018. Association for Computational Linguistics. doi: 10.18653/v1/N18-1032. URL `https://aclanthology.org/N18-1032`.

Jiatao Gu, Yong Wang, Kyunghyun Cho, and Victor OK Li. Improved zero-shot neural machine translation via ignoring spurious correlations. In *Proceedings of the 57th Annual Meeting of the Association for Computational Linguistics*, pages 1258–1268, 2019.

Mandy Guo, Qinlan Shen, Yinfei Yang, Heming Ge, Daniel Cer, Gustavo Hernandez Abrego, Keith Stevens, Noah Constant, Yun-Hsuan Sung, Brian Strope, and Ray Kurzweil. Effective parallel corpus mining using bilingual sentence embeddings, 2018. URL `https://arxiv.org/abs/1807.11906`.

Udit Gupta, Mariam Elgamal, Gage Hills, Gu-Yeon Wei, Hsien-Hsin S Lee, David Brooks, and Carole-Jean Wu. Act: designing sustainable computer systems with an architectural carbon modeling tool. In *Proceedings of the 49th Annual International Symposium on Computer Architecture*, pages 784–799, 2022a.

Udit Gupta, Young Guen Kim, Sylvia Lee, Jordan Tse, Hsien-Hsin Sean Lee, Gu-Yeon Wei, David Brooks, and Carole-Jean Wu. Chasing carbon: The elusive environmental footprint of computing. *IEEE Micro*, 2022b.

Francisco Guzmán, Peng-Jen Chen, Myle Ott, Juan Pino, Guillaume Lample, Philipp Koehn, Vishrav Chaudhary, and Marc'Aurelio Ranzato. The FLORES evaluation datasets for low-resource machine translation: Nepali–English and Sinhala–English. In *Proceedings of the 2019 Conference on Empirical Methods in Natural Language Processing and the 9th International Joint Conference on Natural Language Processing (EMNLP-IJCNLP)*, pages 6098–6111, Hong Kong, China, November 2019. Association for Computational Linguistics. doi: 10.18653/v1/D19-1632. URL https://aclanthology.org/D19-1632.

René Haas and Leon Derczynski. Discriminating between similar nordic languages. In *Proceedings of the Eighth Workshop on NLP for Similar Languages, Varieties and Dialects*, pages 67–75, Kiyv, Ukraine, April 2021. Association for Computational Linguistics. URL https://aclanthology.org/2021.vardial-1.8.

Nizar Habash. Introduction to arabic natural language processing. In *Introduction to Arabic Natural Language Processing*, 2010.

Nizar Habash, Ryan Roth, Owen Rambow, Ramy Eskander, and Nadi Tomeh. Morphological analysis and disambiguation for dialectal Arabic. In *NAACL*, pages 426–432, Atlanta, Georgia, June 2013. Association for Computational Linguistics. URL https://aclanthology.org/N13-1044.

Gilles Hacheme. English2gbe: A multilingual machine translation model for {Fon/Ewe}gbe. *CoRR*, abs/2112.11482, 2021. URL https://arxiv.org/abs/2112.11482.

Barry Haddow, Rachel Bawden, Antonio Valerio Miceli Barone, Jindřich Helcl, and Alexandra Birch. Survey of Low-Resource Machine Translation. *Computational Linguistics*, pages 1–67, 06 2022. ISSN 0891-2017. doi: 10.1162/coli_a_00446. URL https://doi.org/10.1162/coli_a_00446.

Asmelash Teka Hadgu, Gebrekirstos G. Gebremeskel, and Abel Aregawi. HornMT: Machine translation benchmark dataset for languages in the horn of africa. https://github.com/asmelashteka/HornMT, 2021.

Joan Kelly Hall. *Teaching and researching: Language and culture*. Routledge, 2013.

Harald Hammarström, Robert Forkel, Martin Haspelmath, and Sebastian Bank. Glottolog database 4.6, 2022.

Hany Hassan, Anthony Aue, Chang Chen, Vishal Chowdhary, Jonathan Clark, Christian Federmann, Xuedong Huang, Marcin Junczys-Dowmunt, William Lewis, Mu Li, Shujie Liu, Tie-Yan Liu, Renqian Luo, Arul Menezes, Tao Qin, Frank Seide, Xu Tan, Fei Tian, Lijun Wu, Shuangzhi Wu, Yingce Xia, Dongdong Zhang, Zhirui Zhang, and Ming Zhou. Achieving human parity on automatic chinese to english news translation, 2018. URL https://arxiv.org/abs/1803.05567.

Einar Haugen. Planning for a standard language in modern norway. *Anthropological Linguistics*, 1(3):8–21, 1959. ISSN 00035483, 19446527. URL http://www.jstor.org/stable/30022188.

Kaiming He, Xiangyu Zhang, Shaoqing Ren, and Jian Sun. Deep Residual Learning for Image Recognition. In *Proc. of CVPR*, 2015.

Kenneth Heafield. KenLM: faster and smaller language model queries. In *Proceedings of the EMNLP 2011 Sixth Workshop on Statistical Machine Translation*, 2011.

Kevin Heffernan, Çelebi, and Holger Schwenk. Bitext mining using distilled sentence representations for low-resource languages, 2022. URL https://arxiv.org/abs/2205.12654.

Ulf Hermjakob, Jonathan May, and Kevin Knight. Out-of-the-box universal Romanization tool uroman. In *Proceedings of ACL 2018, System Demonstrations*, pages 13–18, Melbourne, Australia, July 2018. Association for Computational Linguistics. doi: 10.18653/v1/P18-4003. URL https://aclanthology.org/P18-4003.

Douglas Blanks Hindman. The rural-urban digital divide. *Journalism & Mass Communication Quarterly*, 77(3):549–560, 2000.

Geoffrey Hinton, Oriol Vinyals, Jeff Dean, et al. Distilling the knowledge in a neural network. *arXiv preprint arXiv:1503.02531*, 2015. URL https://arxiv.org/abs/1503.02531.

Phan Viet Hoang. Khmer natural language processing tookit. https://github.com/VietHoang1512/khmer-nltk, 2020.

Vu Cong Duy Hoang, Philipp Koehn, Gholamreza Haffari, and Trevor Cohn. Iterative back-translation for neural machine translation. In *Proceedings of the 2nd Workshop on Neural Machine Translation and Generation*, pages 18–24, 2018.

Md Zobaer Hossain, Md Ashraful Rahman, Md Saiful Islam, and Sudipta Kar. BanFakeNews: A dataset for detecting fake news in Bangla. In *Proceedings of the 12th Language Resources and Evaluation Conference*, pages 2862–2871, Marseille, France, 2020. European Language Resources Association.

Catherine Howell and Darrell M. West. The internet as a human right, November 2016. URL https://www.brookings.edu/blog/techtank/2016/11/07/the-internet-as-a-human-right/.

Junjie Hu, Sebastian Ruder, Aditya Siddhant, Graham Neubig, Orhan Firat, and Melvin Johnson. XTREME: A massively multilingual multi-task benchmark for evaluating cross-lingual generalization. In *ICML*, pages 4411–4421, 2020.

Changho Hwang, Wei Cui, Yifan Xiong, Ziyue Yang, Ze Liu, Han Hu, Zilong Wang, Rafael Salas, Jithin Jose, Prabhat Ram, et al. Tutel: Adaptive mixture-of-experts at scale. *arXiv preprint arXiv:2206.03382*, 2022.

Tommi Jauhiainen, Krister Lindén, and Heidi Jauhiainen. Evaluation of language identification methods using 285 languages. In *Proceedings of the 21st Nordic Conference on Computational Linguistics*, pages 183–191, 2017.

Tommi Jauhiainen, Marco Lui, Marcos Zampieri, Timothy Baldwin, and Krister Lindén. Automatic language identification in texts: A survey. *Journal of Artificial Intelligence Research*, 65:675–782, 2019.

Isaac Johnson and Emily Lescak. Considerations for multilingual wikipedia research. *arXiv preprint arXiv:2204.02483*, 04 2022.

Melvin Johnson, Mike Schuster, Quoc V Le, Maxim Krikun, Yonghui Wu, Zhifeng Chen, Nikhil Thorat, Fernanda Viégas, Martin Wattenberg, Greg Corrado, et al. Google' s multilingual neural machine translation system: Enabling zero-shot translation. *Transactions of the Association for Computational Linguistics*, 5:339–351, 2017.

Pratik Joshi, Christain Barnes, Sebastin Santy, Simran Khanuja, Sanket Shah, Anirudh Srinivasan, Satwik Bhattamishra, Sunayana Sitaram, Monojit Choudhury, and Kalika Bali. Unsung challenges of building and deploying language technologies for low resource language communities. *arXiv preprint arXiv:1912.03457*, 2019.

Pratik Joshi, Sebastin Santy, Amar Budhiraja, Kalika Bali, and Monojit Choudhury. The state and fate of linguistic diversity and inclusion in the nlp world. In *Proceedings of the 58th Annual Meeting of the Association for Computational Linguistics*, pages 6282–6293, 2020.

Armand Joulin, Edouard Grave, Piotr Bojanowski, and Tomas Mikolov. Bag of tricks for efficient text classification. In *Proceedings of the 15th Conference of the European Chapter of the Association for Computational Linguistics: Volume 2, Short Papers*, pages 427–431. Association for Computational Linguistics, April 2017.

Nal Kalchbrenner and Phil Blunsom. Recurrent continuous translation models. In *Proceedings of the 2013 Conference on Empirical Methods in Natural Language Processing*, pages 1700–1709, Seattle, Washington, USA, October 2013. Association for Computational Linguistics. URL `https://aclanthology.org/D13-1176`.

Shivani Kapania, Oliver Siy, Gabe Clapper, Azhagu Meena SP, and Nithya Sambasivan. " because ai is 100% right and safe" : User attitudes and sources of ai authority in india. In *CHI Conference on Human Factors in Computing Systems*, pages 1–18, 2022.

Alina Karakanta, Jon Dehdari, and Josef Genabith. Neural machine translation for low-resource languages without parallel corpora. *Machine Translation*, 32(1 – 2):167 – 189, jun 2018. ISSN 0922-6567. doi: 10.1007/s10590-017-9203-5. URL `https://doi.org/10.1007/s10590-017-9203-5`.

Jacob Devlin Ming-Wei Chang Kenton and Lee Kristina Toutanova. Bert: Pre-training of deep bidirectional transformers for language understanding. In *Proceedings of NAACL-HLT*, pages 4171–4186, 2019.

Mahmoud Khonji, Youssef Iraqi, and Andrew Jones. Phishing detection: a literature survey. *IEEE Communications Surveys & Tutorials*, 15(4):2091–2121, 2013.

Elaine C Khoong and Jorge A Rodriguez. A research agenda for using machine translation in clinical medicine. *Journal of General Internal Medicine*, 37(5):1275–1277, 2022.

Yoon Kim and Alexander M Rush. Sequence-level knowledge distillation. In *EMNLP*, 2016.

Young Jin Kim, Ammar Ahmad Awan, Alexandre Muzio, Andrés Felipe Cruz-Salinas, Liyang Lu, Amr Hendy, Samyam Rajbhandari, Yuxiong He, and Hany Hassan Awadalla. Scalable and efficient moe training for multitask multilingual models. *CoRR*, abs/2109.10465, 2021. URL `https://arxiv.org/abs/2109.10465`.

Diederik P. Kingma and Jimmy Ba. Adam: A method for stochastic optimization. In Yoshua Bengio and Yann LeCun, editors, *3rd International Conference on Learning Representations, ICLR 2015, San Diego, CA, USA, May 7-9, 2015, Conference Track Proceedings*, 2015. URL `http://arxiv.org/abs/1412.6980`.

Svetlana Kiritchenko, Isar Nejadgholi, and Kathleen C Fraser. Confronting abusive language online: A survey from the ethical and human rights perspective. *Journal of Artificial Intelligence Research*, 71:431–478, 2021.

Tom Kocmi, Dominik Macháček, and Ondřej Bojar. *The Reality of Multi-Lingual Machine Translation*. UFAL, Prague, Czechia, 2021.

Philipp Koehn. Europarl: A parallel corpus for statistical machine translation. In *MT Summit*, 2005.

Philipp Koehn. *Statistical machine translation*. Cambridge University Press, 2009.

Philipp Koehn and Ulrich Germann. The impact of machine translation quality on human post-editing. In *Proceedings of the EACL 2014 Workshop on Humans and Computer-assisted Translation*, pages 38–46, 2014.

Philipp Koehn and Rebecca Knowles. Six challenges for neural machine translation. In *Proceedings of the First Workshop on Neural Machine Translation*, pages 28–39, Vancouver, August 2017. Association for Computational Linguistics. URL `http://www.aclweb.org/anthology/W17-3204`.

Philipp Koehn, Hieu Hoang, Alexandra Birch, Chris Callison-Burch, Marcello Federico, Nicola Bertoldi, Brooke Cowan, Wade Shen, Christine Moran, Richard Zens, et al. Moses: Open source toolkit for statistical machine translation. In *Proceedings of the 45th annual meeting of the ACL on interactive poster and demonstration sessions*, pages 177–180. Association for Computational Linguistics, 2007.

Philipp Koehn, Huda Khayrallah, Kenneth Heafield, and Mikel L. Forcada. Findings of the wmt 2018 shared task on parallel corpus filtering. In *Proceedings of the Third Conference on Machine Translation: Shared Task Papers*, pages 726–739, Belgium, Brussels, October 2018. Association for Computational Linguistics. URL `https://www.aclweb.org/anthology/W18-6453`.

Philipp Koehn, Francisco Guzmán, Vishrav Chaudhary, and Juan Pino. Findings of the WMT 2019 shared task on parallel corpus filtering for low-resource conditions. In *Proceedings of the Fourth Conference on Machine Translation (Volume 3: Shared Task Papers, Day 2)*, 2019.

Philipp Koehn, Vishrav Chaudhary, Ahmed El-Kishky, Naman Goyal, Peng-Jen Chen, and Francisco Guzmán. Findings of the WMT 2020 shared task on parallel corpus filtering and alignment. In *Proceedings of the Fifth Conference on Machine Translation*, pages 726–742, Online, November 2020. Association for Computational Linguistics. URL `https://aclanthology.org/2020.wmt-1.78`.

Bruce Kogut and Anca Metiu. Open-source software development and distributed innovation. *Oxford review of economic policy*, 17(2):248–264, 2001.

Anastasia Kozyreva, Philipp Lorenz-Spreen, Ralph Hertwig, Stephan Lewandowsky, and Stefan M Herzog. Public attitudes towards algorithmic personalization and use of personal data online: Evidence from germany, great britain, and the united states. *Humanities and Social Sciences Communications*, 8(1):1–11, 2021.

Julia Kreutzer, Isaac Caswell, Lisa Wang, Ahsan Wahab, Daan van Esch, Nasanbayar Ulzii-Orshikh, Allahsera Tapo, Nishant Subramani, Artem Sokolov, Claytone Sikasote, Monang Setyawan, Supheakmungkol Sarin, Sokhar Samb, Benoît Sagot, Clara Rivera, Annette Rios, Isabel Papadimitriou, Salomey Osei, Pedro Ortiz Suarez, Iroro Orife, Kelechi Ogueji, Andre Niyongabo Rubungo, Toan Q. Nguyen, Mathias Müller, André Müller, Shamsuddeen Hassan Muhammad, Nanda Muhammad, Ayanda Mnyakeni, Jamshidbek Mirzakhalov, Tapiwanashe Matangira, Colin Leong, Nze Lawson, Sneha Kudugunta, Yacine Jernite, Mathias Jenny, Orhan Firat, Bonaventure F. P. Dossou, Sakhile Dlamini, Nisansa de Silva, Sakine Çabuk Ballı, Stella Biderman, Alessia Battisti, Ahmed Baruwa, Ankur Bapna, Pallavi Baljekar, Israel Abebe Azime, Ayodele Awokoya, Duygu Ataman, Orevaoghene Ahia, Oghenefego Ahia, Sweta Agrawal, and Mofetoluwa Adeyemi. Quality at a Glance: An Audit of Web-Crawled Multilingual Datasets. *Transactions of the Association for Computational Linguistics*, 10:50–72, January 2022. ISSN 2307-387X. doi: 10.1162/tacl_a_00447. URL `https://doi.org/10.1162/tacl_a_00447`.

Taku Kudo and John Richardson. Sentencepiece: A simple and language independent subword tokenizer and detokenizer for neural text processing. In Eduardo Blanco and Wei Lu, editors, *Proceedings of the 2018 Conference on Empirical Methods in Natural Language Processing, EMNLP 2018: System Demonstrations, Brussels, Belgium, October 31 - November 4, 2018*, pages 66–71. Association for Computational Linguistics, 2018. doi: 10.18653/v1/d18-2012. URL `https://doi.org/10.18653/v1/d18-2012`.

Sneha Kudugunta, Yanping Huang, Ankur Bapna, Maxim Krikun, Dmitry Lepikhin, Minh-Thang Luong, and Orhan Firat. Beyond distillation: Task-level mixture-of-experts for efficient inference. In *Findings of the Association for Computational Linguistics: EMNLP 2021*, pages 3577–3599, Punta Cana, Dominican Republic, November 2021. Association for Computational Linguistics. doi: 10.18653/v1/2021.findings-emnlp.304. URL `https://aclanthology.org/2021.findings-emnlp.304`.

Aman Kumar, Himani Shrotriya, Prachi Sahu, Raj Dabre, Ratish Puduppully, Anoop Kunchukuttan, Amogh Mishra, Mitesh M Khapra, and Pratyush Kumar. Indicnlg suite: Multilingual datasets for diverse nlg tasks in indic languages. *arXiv preprint arXiv:2203.05437*, 2022.

Sachin Kumar, Antonios Anastasopoulos, Shuly Wintner, and Yulia Tsvetkov. Machine translation into low-resource language varieties. In *Proceedings of the 59th Annual Meeting of the Association for Computational Linguistics and the 11th International Joint Conference on Natural Language Processing (Volume 2: Short Papers)*, pages 110–121, Online, August 2021. Association for Computational Linguistics. doi: 10.18653/v1/2021.acl-short.16. URL `https://aclanthology.org/2021.acl-short.16`.

Anoop Kunchukuttan. The IndicNLP Library. `https://github.com/anoopkunchukuttan/indic_nlp_library/blob/master/docs/indicnlp.pdf`, 2020.

Keita Kurita, Anna Belova, and Antonios Anastasopoulos. Towards robust toxic content classification. *CoRR*, abs/1912.06872, 2019. URL `http://arxiv.org/abs/1912.06872`.

Remy Kusters, Dusan Misevic, Hugues Berry, Antoine Cully, Yann Le Cunff, Loic Dandoy, Natalia Díaz-Rodríguez, Marion Ficher, Jonathan Grizou, Alice Othmani, et al. Interdisciplinary research in artificial intelligence: Challenges and opportunities. *Frontiers in Big Data*, page 45, 2020.

Garry Kuwanto, Afra Feyza Akyürek, Isidora Chara Tourni, Siyang Li, and Derry Wijaya. Low-resource machine translation for low-resource languages: Leveraging comparable data, code-switching and compute resources. *CoRR*, abs/2103.13272, 2021. URL `https://arxiv.org/abs/2103.13272`.

Ivana Kvapilíková, Mikel Artetxe, Gorka Labaka amd Eneko Agirre, and Ondřej Bojar. Unsupervised multilingual sentence embeddings for parallel corpus mining. In *ACL*, 2020.

Niklas Laxström, Pau Giner, and Santhosh Thottingal. Content translation: Computer-assisted translation tool for wikipedia articles. *CoRR*, abs/1506.01914, 2015. URL `http://arxiv.org/abs/1506.01914`.

En-Shiun Lee, Sarubi Thillainathan, Shravan Nayak, Surangika Ranathunga, David Adelani, Ruisi Su, and Arya McCarthy. Pre-trained multilingual sequence-to-sequence models: A hope for low-resource language translation? In *Findings of the Association for Computational Linguistics: ACL 2022*, pages 58–67, Dublin, Ireland, May 2022. Association for Computational Linguistics. doi: 10.18653/v1/2022.findings-acl.6. URL `https://aclanthology.org/2022.findings-acl.6`.

Katherine Lee, Daphne Ippolito, Andrew Nystrom, Chiyuan Zhang, Douglas Eck, Chris Callison-Burch, and Nicholas Carlini. Deduplicating training data makes language models better. *CoRR*, 2021. URL `https://arxiv.org/abs/2107.06499`.

Sangmin-Michelle Lee. The impact of using machine translation on efl students' writing. *Computer Assisted Language Learning*, 33(3):157–175, 2020.

Alyssa Lees, Daniel Borkan, Ian Kivlichan, Jorge Nario, and Tesh Goyal. Capturing covertly toxic speech via crowdsourcing. In *Proceedings of the First Workshop on Bridging Human–Computer Interaction and Natural Language Processing*, Online, April 2021. Association for Computational Linguistics. URL https://aclanthology.org/2021.hcinlp-1.3.

Heather Lent, Kelechi Ogueji, Miryam de Lhoneux, Orevaoghene Ahia, and Anders Søgaard. What a creole wants, what a creole needs. *arXiv preprint arXiv:2206.00437*, 2022.

Dmitry Lepikhin, HyoukJoong Lee, Yuanzhong Xu, Dehao Chen, Orhan Firat, Yanping Huang, Maxim Krikun, Noam Shazeer, and Zhifeng Chen. Gshard: Scaling giant models with conditional computation and automatic sharding. *CoRR*, abs/2006.16668, 2020. URL https://arxiv.org/abs/2006.16668.

Peggy Levitt and B Nadya Jaworsky. Transnational migration studies: Past developments and future trends. *Annual review of sociology*, 33:129, 2007.

Peggy Levitt and Deepak Lamba-Nieves. Social remittances revisited. *Journal of ethnic and migration studies*, 37(1):1–22, 2011.

Shahar Levy, Koren Lazar, and Gabriel Stanovsky. Collecting a large-scale gender bias dataset for coreference resolution and machine translation. *arXiv preprint arXiv:2109.03858*, 2021.

M. Paul Lewis, editor. *Ethnologue: Languages of the World*. SIL International, Dallas, TX, USA, sixteenth edition, 2009.

Mike Lewis, Shruti Bhosale, Tim Dettmers, Naman Goyal, and Luke Zettlemoyer. Base layers: Simplifying training of large, sparse models. In *International Conference on Machine Learning*, pages 6265–6274. PMLR, 2021.

Daniel Licht, Cynthia Gao, Janice Lam, Francisco Guzman, Mona Diab, and Philipp Koehn. Consistent human evaluation of machine translation across language pairs, 2022. URL https://arxiv.org/abs/2205.08533.

Pierre Lison and Jörg Tiedemann. Opensubtitles2016: Extracting large parallel corpora from movie and tv subtitles, 2016.

Rui Liu, Young Jin Kim, Alexandre Muzio, Barzan Mozafari, and Hany Hassan Awadalla. Gating dropout: Communication-efficient regularization for sparsely activated transformers. *arXiv preprint arXiv:2205.14336*, 2022.

Xuebo Liu, Longyue Wang, Derek F. Wong, Liang Ding, Lidia S. Chao, Shuming Shi, and Zhaopeng Tu. On the complementarity between pre-training and back-translation for neural machine translation. In *Findings of the Association for Computational Linguistics: EMNLP 2021*, pages 2900–2907, Punta Cana, Dominican Republic, November 2021a. Association for Computational Linguistics. doi: 10.18653/v1/2021.findings-emnlp.247. URL https://aclanthology.org/2021.findings-emnlp.247.

Yinhan Liu, Jiatao Gu, Naman Goyal, Xian Li, Sergey Edunov, Marjan Ghazvininejad, Mike Lewis, and Luke Zettlemoyer. Multilingual denoising pre-training for neural machine translation. *Transactions of the Association for Computational Linguistics*, 8:726–742, 2020.

Zihan Liu, Genta Indra Winata, and Pascale Fung. Continual mixed-language pre-training for extremely low-resource neural machine translation. In *Findings of the Association for Computational Linguistics: ACL-IJCNLP 2021*, pages 2706–2718, Online, August 2021b. Association for Computational Linguistics. doi: 10.18653/v1/2021.findings-acl.239. URL https://aclanthology.org/2021.findings-acl.239.

Adam Lopez. Statistical machine translation. *ACM Computing Surveys (CSUR)*, 40(3): 1–49, 2008.

Jiasen Lu, Vedanuj Goswami, Marcus Rohrbach, Devi Parikh, and Stefan Lee. 12-in-1: Multi-task vision and language representation learning. In *Proceedings of the IEEE/CVF Conference on Computer Vision and Pattern Recognition*, pages 10437–10446, 2020.

Stefano Lusito, Edoardo Ferrante, and Jean Maillard. Text normalization for endangered languages: the case of Ligurian. *arXiv preprint arXiv:2206.07861*, 2022.

Shuming Ma, Li Dong, Shaohan Huang, Dongdong Zhang, Alexandre Muzio, Saksham Singhal, Hany Hassan Awadalla, Xia Song, and Furu Wei. Deltalm: Encoder-decoder pre-training for language generation and translation by augmenting pretrained multilingual encoders. *CoRR*, abs/2106.13736, 2021. URL https://arxiv.org/abs/2106.13736.

Sean MacAvaney, Hao-Ren Yao, Eugene Yang, Katina Russell, Nazli Goharian, and Ophir Frieder. Hate speech detection: Challenges and solutions. *PloS one*, 14(8):e0221152, 2019.

Yash Madhani, Sushane Parthan, Priyanka Bedekar, Ruchi Khapra, Vivek Seshadri, Anoop Kunchukuttan, Pratyush Kumar, and Mitesh M Khapra. Aksharantar: Towards building open transliteration tools for the next billion users. *arXiv preprint arXiv:2205.03018*, 2022.

Manuel Mager, Arturo Oncevay, Abteen Ebrahimi, John Ortega, Annette Rios, Angela Fan, Ximena Gutierrez-Vasques, Luis Chiruzzo, Gustavo Giménez-Lugo, Ricardo Ramos, Ivan Vladimir Meza Ruiz, Rolando Coto-Solano, Alexis Palmer, Elisabeth Mager-Hois, Vishrav Chaudhary, Graham Neubig, Ngoc Thang Vu, and Katharina Kann. Findings of the AmericasNLP 2021 shared task on open machine translation for indigenous languages of the Americas. In *Proceedings of the First Workshop on Natural Language Processing for Indigenous Languages of the Americas*, pages 202–217, Online, June 2021. Association for Computational Linguistics. doi: 10.18653/v1/2021.americasnlp-1.23. URL https://aclanthology.org/2021.americasnlp-1.23.

Alexandre Magueresse, Vincent Carles, and Evan Heetderks. Low-resource languages: A review of past work and future challenges. *arXiv preprint arXiv:2006.07264*, 2020.

Laurette Marais, Ilana Wilken, Nina Van Niekerk, and Karen Calteaux. Mburisano covid-19 multilingual corpus. https://hdl.handle.net/20.500.12185/536, 2021.

Benjamin Marie, Raphael Rubino, and Atsushi Fujita. Tagged back-translation revisited: Why does it really work? In *Proceedings of the 58th Annual Meeting of the Association for Computational Linguistics*, pages 5990–5997, Online, July 2020. Association for Computational Linguistics. doi: 10.18653/v1/2020.acl-main.532. URL `https://aclanthology.org/2020.acl-main.532`.

Arya D. McCarthy, Rachel Wicks, Dylan Lewis, Aaron Mueller, Winston Wu, Oliver Adams, Garrett Nicolai, Matt Post, and David Yarowsky. The Johns Hopkins University Bible corpus: 1600+ tongues for typological exploration. In *Proceedings of the 12th Language Resources and Evaluation Conference*, pages 2884–2892, Marseille, France, May 2020. European Language Resources Association. ISBN 979-10-95546-34-4. URL `https://aclanthology.org/2020.lrec-1.352`.

Leland McInnes, John Healy, Nathaniel Saul, and Lukas Grossberger. Umap: Uniform manifold approximation and projection. *The Journal of Open Source Software*, 3(29):861, 2018.

Cindy A. McKellar. Autshumato machine translation evaluation set. In *Centre for Text Technology (CTexT)*, 2017.

Sharon J McLennan. Techno-optimism or information imperialism: Paradoxes in online networking, social media and development. *Information Technology for Development*, 22 (3):380–399, 2016.

Paul McNamee. Language identification: a solved problem suitable for undergraduate instruction. *Journal of computing sciences in colleges*, 20(3):94–101, 2005.

Sabrina J Mielke, Zaid Alyafeai, Elizabeth Salesky, Colin Raffel, Manan Dey, Matthias Gallé, Arun Raja, Chenglei Si, Wilson Y Lee, Benoît Sagot, et al. Between words and characters: A brief history of open-vocabulary modeling and tokenization in nlp. *arXiv preprint arXiv:2112.10508*, 2021.

Tomas Mikolov and Geoffrey Zweig. Context dependent recurrent neural network language model. In *2012 IEEE Spoken Language Technology Workshop (SLT)*, pages 234–239. IEEE, 2012.

Jamshidbek Mirzakhalov, Anoop Babu, Duygu Ataman, Sherzod Kariev, Francis Tyers, Otabek Abduraufov, Mammad Hajili, Sardana Ivanova, Abror Khaytbaev, Antonio Laverghetta Jr, et al. A large-scale study of machine translation in the turkic languages. *arXiv preprint arXiv:2109.04593*, 2021.

Pushkar Mishra, Helen Yannakoudakis, and Ekaterina Shutova. Tackling online abuse: A survey of automated abuse detection methods. *CoRR*, abs/1908.06024, 2019. URL `http://arxiv.org/abs/1908.06024`.

Margaret Mitchell, Simone Wu, Andrew Zaldivar, Parker Barnes, Lucy Vasserman, Ben Hutchinson, Elena Spitzer, Inioluwa Deborah Raji, and Timnit Gebru. Model cards for model reporting. In *Proceedings of the Conference on Fairness, Accountability, and Transparency*, FAT* '19, page 220 – 229, New York, NY, USA, 2019. Association for

Computing Machinery. ISBN 9781450361255. doi: 10.1145/3287560.3287596. URL https://doi.org/10.1145/3287560.3287596.

Aaron Mueller, Garrett Nicolai, Arya D McCarthy, Dylan Lewis, Winston Wu, and David Yarowsky. An analysis of massively multilingual neural machine translation for low-resource languages. In *Proceedings of The 12th language resources and evaluation conference*, pages 3710–3718, 2020.

Namrata Mukhija, Monojit Choudhury, and Kalika Bali. Designing language technologies for social good: The road not taken. *arXiv preprint arXiv:2110.07444*, 2021.

Dragos Stefan Munteanu and Daniel Marcu. Improving Machine Translation Performance by Exploiting Non-Parallel Corpora. *Computational Linguistics*, 31(4):477–504, 2005. URL http://www.aclweb.org/anthology/J05-4003.

Toshiaki Nakazawa, Chenchen Ding, Raj Dabre, Anoop Kunchukuttan, Nobushige Doi, Yusuke Oda, Ondřej Bojar, Shantipriya Parida, Isao Goto, and Hidaya Mino, editors. *Proceedings of the 6th Workshop on Asian Translation*, Hong Kong, China, November 2019. Association for Computational Linguistics. URL https://aclanthology.org/D19-5200.

Toshiaki Nakazawa, Hideki Nakayama, Chenchen Ding, Raj Dabre, Shohei Higashiyama, Hideya Mino, Isao Goto, Win Pa Pa, Anoop Kunchukuttan, Shantipriya Parida, Ondřej Bojar, Chenhui Chu, Akiko Eriguchi, Kaori Abe, Yusuke Oda, and Sadao Kurohashi. Overview of the 8th workshop on Asian translation. In *Proceedings of the 8th Workshop on Asian Translation (WAT2021)*, pages 1–45, Online, August 2021. Association for Computational Linguistics. doi: 10.18653/v1/2021.wat-1.1. URL https://aclanthology.org/2021.wat-1.1.

Wilhelmina Nekoto, Vukosi Marivate, Tshinondiwa Matsila, Timi Fasubaa, Taiwo Fagbohungbe, Solomon Oluwole Akinola, Shamsuddeen Muhammad, Salomon Kabongo Kabenamualu, Salomey Osei, Freshia Sackey, Rubungo Andre Niyongabo, Ricky Macharm, Perez Ogayo, Orevaoghene Ahia, Musie Meressa Berhe, Mofetoluwa Adeyemi, Masabata Mokgesi-Selinga, Lawrence Okegbemi, Laura Martinus, Kolawole Tajudeen, Kevin Degila, Kelechi Ogueji, Kathleen Siminyu, Julia Kreutzer, Jason Webster, Jamiil Toure Ali, Jade Abbott, Iroro Orife, Ignatius Ezeani, Idris Abdulkadir Dangana, Herman Kamper, Hady Elsahar, Goodness Duru, Ghollah Kioko, Murhabazi Espoir, Elan van Biljon, Daniel Whitenack, Christopher Onyefuluchi, Chris Chinenye Emezue, Bonaventure F. P. Dossou, Blessing Sibanda, Blessing Bassey, Ayodele Olabiyi, Arshath Ramkilowan, Alp Öktem, Adewale Akinfaderin, and Abdallah Bashir. Participatory research for low-resourced machine translation: A case study in African languages. In *Findings of the Association for Computational Linguistics: EMNLP 2020*, pages 2144–2160, Online, November 2020. Association for Computational Linguistics. doi: 10.18653/v1/2020.findings-emnlp.195. URL https://aclanthology.org/2020.findings-emnlp.195.

Toan Q. Nguyen and David Chiang. Transfer learning across low-resource, related languages for neural machine translation. In *Proceedings of the Eighth International Joint Conference on Natural Language Processing (Volume 2: Short Papers)*, pages 296–301, Taipei,

Taiwan, November 2017. Asian Federation of Natural Language Processing. URL https://aclanthology.org/I17-2050.

Abu Sadat Nurullah. Globalisation as a challenge to islamic cultural identity. *International Journal of Interdisciplinary Social Sciences*, 3(6):45–52, 2008.

Akintunde Oladipo, Odunayo Ogundepo, Kelechi Ogueji, and Jimmy Lin. An exploration of vocabulary size and transfer effects in multilingual language models for african languages. In *3rd Workshop on African Natural Language Processing*, 2022. URL https://openreview.net/forum?id=HOZmF9MV8Wc.

Iroro Orife, Julia Kreutzer, Blessing Sibanda, Daniel Whitenack, Kathleen Siminyu, Laura Martinus, Jamiil Toure Ali, Jade Z. Abbott, Vukosi Marivate, Salomon Kabongo, Musie Meressa, Espoir Murhabazi, Orevaoghene Ahia, Elan Van Biljon, Arshath Ramkilowan, Adewale Akinfaderin, Alp Öktem, Wole Akin, Ghollah Kioko, Kevin Degila, Herman Kamper, Bonaventure Dossou, Chris Emezue, Kelechi Ogueji, and Abdallah Bashir. Masakhane - machine translation for africa. *CoRR*, abs/2003.11529, 2020. URL https://arxiv.org/abs/2003.11529.

Pedro Javier Ortiz Suárez, Benoît Sagot, and Laurent Romary. Asynchronous Pipeline for Processing Huge Corpora on Medium to Low Resource Infrastructures. In Piotr Bański, Adrien Barbaresi, Hanno Biber, Evelyn Breiteneder, Simon Clematide, Marc Kupietz, Harald Lüngen, and Caroline Iliadi, editors, *7th Workshop on the Challenges in the Management of Large Corpora (CMLC-7)*, pages 9 – 16, Cardiff, United Kingdom, July 2019. Leibniz-Institut für Deutsche Sprache. doi: 10.14618/IDS-PUB-9021. URL https://hal.inria.fr/hal-02148693.

Myle Ott, Sergey Edunov, Alexei Baevski, Angela Fan, Sam Gross, Nathan Ng, David Grangier, and Michael Auli. fairseq: A fast, extensible toolkit for sequence modeling. In *Proceedings of the 2019 Conference of the North American Chapter of the Association for Computational Linguistics (Demonstrations)*, pages 48–53, Minneapolis, Minnesota, June 2019. Association for Computational Linguistics. doi: 10.18653/v1/N19-4009. URL https://aclanthology.org/N19-4009.

Kishore Papineni, Salim Roukos, Todd Ward, and Wei-Jing Zhu. Bleu: a method for automatic evaluation of machine translation. In *Proceedings of the 40th annual meeting of the Association for Computational Linguistics*, pages 311–318, 2002.

David Patterson, Joseph Gonzalez, Quoc Le, Chen Liang, Lluis-Miquel Munguia, Daniel Rothchild, David So, Maud Texier, and Jeff Dean. Carbon emissions and large neural network training. *arXiv preprint arXiv:2104.10350*, 2021.

Amandalynne Paullada, Inioluwa Deborah Raji, Emily M Bender, Emily Denton, and Alex Hanna. Data and its (dis) contents: A survey of dataset development and use in machine learning research. *Patterns*, 2(11):100336, 2021.

Barbara Plank. What to do about non-standard (or non-canonical) language in NLP. In *Proceedings of the 13th Conference on Natural Language Processing (KONVENS 2016)*,

pages 13–20, 2016. doi: 10.18653/v1/D16-1163. URL https://aclanthology.org/D16-1163.

Maja Popović. chrf++: words helping character n-grams. In *Proceedings of the Second Conference on Machine Translation, Volume 2: Shared Task Papers*, pages 612–618, Copenhagen, Denmark, September 2017. Association for Computational Linguistics. URL http://www.aclweb.org/anthology/W17-4770.

Matt Post. A call for clarity in reporting BLEU scores. In *Proceedings of the Third Conference on Machine Translation: Research Papers*, pages 186–191, Brussels, Belgium, October 2018. Association for Computational Linguistics. doi: 10.18653/v1/W18-6319. URL https://aclanthology.org/W18-6319.

Manasa Prasad, Theresa Breiner, and Daan van Esch. Mining training data for language modeling across the world's languages. In *SLTU*, pages 61–65, 2018.

Ivan Provilkov, Dmitrii Emelianenko, and Elena Voita. BPE-dropout: Simple and effective subword regularization. In *Proceedings of the 58th Annual Meeting of the Association for Computational Linguistics*, pages 1882–1892, Online, July 2020. Association for Computational Linguistics. doi: 10.18653/v1/2020.acl-main.170. URL https://www.aclweb.org/anthology/2020.acl-main.170.

Alec Radford, Jeffrey Wu, Rewon Child, David Luan, Dario Amodei, Ilya Sutskever, et al. Language models are unsupervised multitask learners. *OpenAI blog*, 1(8): 9, 2019. URL https://d4mucfpksywv.cloudfront.net/better-language-models/language-models.pdf.

Alexandre Rafalovitch and Robert Dale. United Nations general assembly resolutions: A six-language parallel corpus. In *Proceedings of the MT Summit XII*, pages 292–299, Ottawa, Canada, 2014.

Jenalea Rajab. Effect of tokenisation strategies for low-resourced southern african languages. In *3rd Workshop on African Natural Language Processing*, 2022. URL https://openreview.net/forum?id=SpMeq5M48W9.

Samyam Rajbhandari, Conglong Li, Zhewei Yao, Minjia Zhang, Reza Yazdani Aminabadi, Ammar Ahmad Awan, Jeff Rasley, and Yuxiong He. Deepspeed-moe: Advancing mixture-of-experts inference and training to power next-generation ai scale. *arXiv preprint arXiv:2201.05596*, 2022.

Pranav Rajpurkar, Jian Zhang, Konstantin Lopyrev, and Percy Liang. Squad: 100, 000+ questions for machine comprehension of text. In *EMNLP*, 2016.

Gowtham Ramesh, Sumanth Doddapaneni, Aravinth Bheemaraj, Mayank Jobanputra, Raghavan AK, Ajitesh Sharma, Sujit Sahoo, Harshita Diddee, Mahalakshmi J, Divyanshu Kakwani, Navneet Kumar, Aswin Pradeep, Srihari Nagaraj, Kumar Deepak, Vivek Raghavan, Anoop Kunchukuttan, Pratyush Kumar, and Mitesh Shantadevi Khapr. Samanantar: The largest publicly available parallel corpora collection for 11 indic languages.

*Transactions of the Association for Computational Linguistics*, 10:145–162, 2022. doi: 10.1162/tacl_a_00452. URL https://aclanthology.org/2022.tacl-1.9.

Ricardo Rei, Craig Stewart, Ana C Farinha, and Alon Lavie. Comet: A neural framework for mt evaluation. In *Proceedings of the 2020 Conference on Empirical Methods in Natural Language Processing (EMNLP)*, pages 2685–2702, 2020.

Nils Reimers and Iryna Gurevych. Sentence-BERT: Sentence embeddings using siamese bert-networks. In *EMNLP*, pages 3982–3992, 2019.

Nils Reimers and Iryna Gurevych. Making monolingual sentence embeddings multilingual using knowledge distillation. In *EMNLP*, pages 4512–4525, 2020.

Shuo Ren, Zhirui Zhang, Shujie Liu, Ming Zhou, and Shuai Ma. Unsupervised neural machine translation with smt as posterior regularization. In *Proceedings of the AAAI Conference on Artificial Intelligence*, pages 241–248, 2019.

Adithya Renduchintala and Adina Williams. Investigating failures of automatic translation in the case of unambiguous gender. *CoRR*, abs/2104.07838, 2021. URL https://arxiv.org/abs/2104.07838.

Philip Resnik. Mining the Web for Bilingual Text. In *ACL*, 1999. URL http://www.aclweb.org/anthology/P99-1068.

Philip Resnik and Noah A. Smith. The Web as a Parallel Corpus. *Computational Linguistics*, 29(3):349–380, 2003. URL http://www.aclweb.org/anthology/J03-3002.

Felix Richter. Infographic: English is the internet's universal language, February 2022. URL https://www.statista.com/chart/26884/languages-on-the-internet/.

John R. Rickford. Standard and non-standard language attitudes in a creole continuum. In Nessa Wolfson and Joan Manes, editors, *Language of Inequality*, pages 145–160. De Gruyter Mouton, 2012. doi: doi:10.1515/9783110857320.145. URL https://doi.org/10.1515/9783110857320.145.

Parker Riley, Isaac Caswell, Markus Freitag, and David Grangier. Translationese as a language in "multilingual" nmt. In *Proceedings of the 58th Annual Meeting of the Association for Computational Linguistics*, pages 7737–7746, 2020.

Samantha Robertson, Wesley Hanwen Deng, Timnit Gebru, Margaret Mitchell, Daniel J Liebling, Michal Lahav, Katherine Heller, Mark Díaz, Samy Bengio, and Niloufar Salehi. Three directions for the design of human-centered machine translation. *Google Research*, 2021.

Björn Ross, Michael Rist, Guillermo Carbonell, Benjamin Cabrera, Nils Kurowsky, and Michael Wojatzki. Measuring the reliability of hate speech annotations: The case of the european refugee crisis. *CoRR*, abs/1701.08118, 2017. URL http://arxiv.org/abs/1701.08118.

Hassan Sajjad, Ahmed Abdelali, Nadir Durrani, and Fahim Dalvi. AraBench: Benchmarking dialectal Arabic-English machine translation. In *Proceedings of the 28th International Conference on Computational Linguistics*, pages 5094–5107, Barcelona, Spain (Online), December 2020. International Committee on Computational Linguistics. doi: 10.18653/v1/2020.coling-main.447. URL `https://aclanthology.org/2020.coling-main.447`.

Mohammad Salameh, Houda Bouamor, and Nizar Habash. Fine-grained Arabic dialect identification. In *Coling*, pages 1332–1344, Santa Fe, New Mexico, USA, August 2018. Association for Computational Linguistics. URL `https://aclanthology.org/C18-1113`.

Fahimeh Saleh, Wray Buntine, and Gholamreza Haffari. Collective wisdom: Improving low-resource neural machine translation using adaptive knowledge distillation. In *Proceedings of the 28th International Conference on Computational Linguistics*, pages 3413–3421, 2020.

Julia Sallabank. *Attitudes to endangered languages: Identities and policies*. Cambridge University Press, 2013.

Nithya Sambasivan. Seeing like a dataset from the global south. *Interactions*, 28(4):76–78, 2021.

Nithya Sambasivan and Jess Holbrook. Toward responsible AI for the next billion users. *Interactions*, 26(1):68–71, 2018.

Nithya Sambasivan, Erin Arnesen, Ben Hutchinson, Tulsee Doshi, and Vinodkumar Prabhakaran. Re-imagining algorithmic fairness in india and beyond. In *Proceedings of the 2021 ACM conference on fairness, accountability, and transparency*, pages 315–328, 2021.

Maarten Sap, Dallas Card, Saadia Gabriel, Yejin Choi, and A Noah Smith. The risk of racial bias in hate speech detection. In *ACL*, 2019.

Kevin P Scannell. The Crúbadán Project: Corpus building for under-resourced languages. *Cahiers du Cental*, 5:1, 2007.

Holger Schwenk. Investigations on large-scale lightly-supervised training for statistical machine translation. In *Proceedings of the 5th International Workshop on Spoken Language Translation: Papers*, 2008.

Holger Schwenk. Filtering and mining parallel data in a joint multilingual space. In *ACL*, pages 228–234, 2018.

Holger Schwenk, Vishrav Chaudhary, Shuo Sun, Hongyu Gong, and Francisco Guzmán. WikiMatrix: Mining 135m parallel sentences in 1620 language pairs from wikipedia. In *ACL*, pages 1351–1361, 2021a.

Holger Schwenk, Guillaume Wenzek, Sergey Edunov, Édouard Grave, Armand Joulin, and Angela Fan. CCMatrix: Mining billions of high-quality parallel sentences on the web. In *Proceedings of the 59th Annual Meeting of the Association for Computational Linguistics and the 11th International Joint Conference on Natural Language Processing (Volume 1: Long Papers)*, pages 6490–6500, 2021b.

Thibault Sellam, Dipanjan Das, and Ankur Parikh. Bleurt: Learning robust metrics for text generation. In *Proceedings of the 58th Annual Meeting of the Association for Computational Linguistics*, pages 7881–7892, 2020.

Rico Sennrich and Biao Zhang. Revisiting low-resource neural machine translation: A case study. In *Proceedings of the 57th Annual Meeting of the Association for Computational Linguistics*, pages 211–221, Florence, Italy, July 2019. Association for Computational Linguistics. doi: 10.18653/v1/P19-1021. URL https://www.aclweb.org/anthology/P19-1021.

Rico Sennrich, Barry Haddow, and Alexandra Birch. Improving neural machine translation models with monolingual data. *Conference of the Association for Computational Linguistics (ACL)*, 2016a.

Rico Sennrich, Barry Haddow, and Alexandra Birch. Edinburgh neural machine translation systems for wmt 16. In *Proceedings of the First Conference on Machine Translation: Volume 2, Shared Task Papers*, pages 371–376, 2016b.

Noam Shazeer, Azalia Mirhoseini, Krzysztof Maziarz, Andy Davis, Quoc Le, Geoffrey Hinton, and Jeff Dean. Outrageously large neural networks: The sparsely-gated mixture-of-experts layer. In *Proceedings of International Conference on Learning Representations (ICLR)*, 2017. URL https://openreview.net/pdf?id=B1ckMDqlg.

Shashi Shekhar, Dilip Kumar Sharma, and MM Sufyan Beg. Language identification framework in code-mixed social media text based on quantum lstm—the word belongs to which language? *Modern Physics Letters B*, 34(06):2050086, 2020.

Aditya Siddhant, Ankur Bapna, Yuan Cao, Orhan Firat, Mia Xu Chen, Sneha Kudugunta, Naveen Arivazhagan, and Yonghui Wu. Leveraging monolingual data with self-supervision for multilingual neural machine translation. In *Proceedings of the 58th Annual Meeting of the Association for Computational Linguistics*, pages 2827–2835, 2020.

Aditya Siddhant, Ankur Bapna, Orhan Firat, Yuan Cao, Mia Xu Chen, Isaac Caswell, and Xavier Garcia. Towards the next 1000 languages in multilingual machine translation: Exploring the synergy between supervised and self-supervised learning. *CoRR*, abs/2201.03110, 2022. URL https://arxiv.org/abs/2201.03110.

Kathleen Siminyu, Godson Kalipe, Davor Orlic, Jade Abbott, Vukosi Marivate, Sackey Freshia, Prateek Sibal, Bhanu Neupane, David I. Adelani, Amelia Taylor, Jamiil Toure ALI, Kevin Degila, Momboladji Balogoun, Thierno Ibrahima DIOP, Davis David, Chayma Fourati, Hatem Haddad, and Malek Naski. AI4D – african language program. *arXiv preprint arXiv:2104.02516*, 2021.

Nitish Singh, Kevin Lehnert, and Kathleen Bostick. Global social media usage: Insights into reaching consumers worldwide. *Thunderbird International Business Review*, 54(5): 683–700, 2012.

Raivis Skadiņš, Mārcis Pinnis, Andrejs Vasiļjevs, Inguna Skadiņa, and Tomas Hudik. Application of machine translation in localization into low-resourced languages. In *Proceedings of the 17th Annual conference of the European Association for Machine Translation*, pages 209–216, Dubrovnik, Croatia, June 2014a. European Association for Machine Translation. URL https://aclanthology.org/2014.eamt-1.43.

Raivis Skadiņš, Jörg Tiedemann, Roberts Rozis, and Daiga Deksne. Billions of parallel words for free: Building and using the EU bookshop corpus. In *Proceedings of the Ninth International Conference on Language Resources and Evaluation (LREC'14)*, pages 1850–1855, Reykjavik, Iceland, May 2014b. European Language Resources Association (ELRA). URL http://www.lrec-conf.org/proceedings/lrec2014/pdf/846_Paper.pdf.

Lucia Specia, Frédéric Blain, Marina Fomicheva, Chrysoula Zerva, Zhenhao Li, Vishrav Chaudhary, and André F. T. Martins. Findings of the WMT 2021 shared task on quality estimation. In *Proceedings of the Sixth Conference on Machine Translation*, pages 684–725, Online, November 2021. Association for Computational Linguistics. URL https://aclanthology.org/2021.wmt-1.71.

Nitish Srivastava, Geoffrey Hinton, Alex Krizhevsky, Ilya Sutskever, and Ruslan Salakhutdinov. Dropout: a simple way to prevent neural networks from overfitting. *The journal of machine learning research*, 15(1):1929–1958, 2014.

Haipeng Sun, Rui Wang, Kehai Chen, Masao Utiyama, Eiichiro Sumita, and Tiejun Zhao. Knowledge distillation for multilingual unsupervised neural machine translation. In *Proceedings of the 58th Annual Meeting of the Association for Computational Linguistics*, pages 3525–3535, 2020.

Cass Robert Sunstein and Richard Thaler. Libertarian paternalism. *American Economic Review*, 2003.

Christian Szegedy, Vincent Vanhoucke, Sergey Ioffe, Jonathon Shlens, and Zbigniew Wojna. Rethinking the inception architecture for computer vision. *CoRR*, abs/1512.00567, 2015. URL http://arxiv.org/abs/1512.00567.

Xu Tan, Yi Ren, Di He, Tao Qin, Zhou Zhao, and Tie-Yan Liu. Multilingual neural machine translation with knowledge distillation. In *International Conference on Learning Representations*, 2018.

Yuqing Tang, Chau Tran, Xian Li, Peng-Jen Chen, Naman Goyal, Vishrav Chaudhary, Jiatao Gu, and Angela Fan. Multilingual translation with extensible multilingual pretraining and finetuning. *CoRR*, abs/2008.00401, 2020. URL https://arxiv.org/abs/2008.00401.

Martin Thoma. The wili benchmark dataset for written language identification. *CoRR*, abs/1801.07779, 2018. URL http://arxiv.org/abs/1801.07779.

J. Tiedemann. Parallel data, tools and interfaces in OPUS. In *LREC*, 2012.

Chau Tran, Shruti Bhosale, James Cross, Philipp Koehn, Sergey Edunov, and Angela Fan. Facebook AI's WMT21 news translation task submission. In *Proceedings of the Sixth Conference on Machine Translation*, pages 205–215, Online, November 2021. Association for Computational Linguistics. URL https://aclanthology.org/2021.wmt-1.19.

Emiliano Treré. The dark side of digital politics: Understanding the algorithmic manufacturing of consent and the hindering of online dissidence. *Opening Governance*, 2016.

Masao Utiyama and Hitoshi Isahara. Reliable Measures for Aligning Japanese-English News Articles and Sentences. In *ACL*, 2003. URL `http://www.aclweb.org/anthology/P03-1010`.

Jeroen Van Der Hoven and Noemi Manders-Huits. Value-sensitive design. In *The Ethics of Information Technologies*, pages 329–332. Routledge, 2020.

Jack Vance. *The Eyes of the overworld*, volume 3. Macmillan Reference USA, 1977.

Ashish Vaswani, Noam Shazeer, Niki Parmar, Jakob Uszkoreit, Llion Jones, Aidan N Gomez, Łukasz Kaiser, and Illia Polosukhin. Attention is all you need. In *Advances in neural information processing systems*, pages 5998–6008, 2017.

Bertie Vidgen, Alex Harris, Dong Nguyen, Rebekah Tromble, Scott Hale, and Helen Margetts. Challenges and frontiers in abusive content detection. In *Proceedings of the third workshop on abusive language online*, pages 80–93. Association for Computational Linguistics, 2019.

Vered Volansky, Noam Ordan, and Shuly Wintner. On the features of translationese. *Digital Scholarship in the Humanities*, 30(1):98–118, 2015.

Alex Wang, Amanpreet Singh, Julian Michael, Felix Hill, Omer Levy, and Samuel R Bowman. Glue: A multi-task benchmark and analysis platform for natural language understanding. In *International Conference on Learning Representations*, 2018.

Hongyu Wang, Shuming Ma, Li Dong, Shaohan Huang, Dongdong Zhang, and Furu Wei. Deepnet: Scaling transformers to 1,000 layers, 2022. URL `https://arxiv.org/abs/2203.00555`.

Yiren Wang, ChengXiang Zhai, and Hany Hassan. Multi-task learning for multilingual neural machine translation. In *Proceedings of the 2020 Conference on Empirical Methods in Natural Language Processing (EMNLP)*, pages 1022–1034, Online, November 2020a. Association for Computational Linguistics. doi: 10.18653/v1/2020.emnlp-main.75. URL `https://www.aclweb.org/anthology/2020.emnlp-main.75`.

Zihan Wang, Karthikeyan K, Stephen Mayhew, and Dan Roth. Extending multilingual bert to low-resource languages, 2020b. URL `https://arxiv.org/abs/2004.13640`.

Zirui Wang, Yulia Tsvetkov, Orhan Firat, and Yuan Cao. Gradient vaccine: Investigating and improving multi-task optimization in massively multilingual models. In *International Conference on Learning Representations*, 2020c.

Steven Weber. *The success of open source*. Harvard University Press, 2004.

Johannes Welbl, Amelia Glaese, Jonathan Uesato, Sumanth Dathathri, John Mellor, Lisa Anne Hendricks, Kirsty Anderson, Pushmeet Kohli, Ben Coppin, and Po-Sen Huang. Challenges in detoxifying language models. *CoRR*, abs/2109.07445, 2021. URL `https://arxiv.org/abs/2109.07445`.

Guillaume Wenzek, Marie-Anne Lachaux, Alexis Conneau, Vishrav Chaudhary, Francisco Guzmán, Armand Joulin, and Édouard Grave. CCNet: Extracting high quality monolingual datasets from web crawl data. In *Proceedings of The 12th Language Resources and Evaluation Conference*, pages 4003–4012, 2020.

Dominic Widdows and Chris Brew. Language identification with a reciprocal rank classifier. *CoRR*, abs/2109.09862, 2021. URL `https://arxiv.org/abs/2109.09862`.

Carole-Jean Wu, Ramya Raghavendra, Udit Gupta, Bilge Acun, Newsha Ardalani, Kiwan Maeng, Gloria Chang, Fiona Aga, Jinshi Huang, Charles Bai, et al. Sustainable ai: Environmental implications, challenges and opportunities. *Proceedings of Machine Learning and Systems*, 4:795–813, 2022.

Shijie Wu, Ryan Cotterell, and Mans Hulden. Applying the transformer to character-level transduction. In *Proceedings of the 16th Conference of the European Chapter of the Association for Computational Linguistics: Main Volume*, pages 1901–1907, Online, April 2021. Association for Computational Linguistics. doi: 10.18653/v1/2021.eacl-main.163. URL `https://aclanthology.org/2021.eacl-main.163`.

Yonghui Wu, Mike Schuster, Zhifeng Chen, Quoc V. Le, Mohammad Norouzi, Wolfgang Macherey, Maxim Krikun, Yuan Cao, Qin Gao, Klaus Macherey, Jeff Klingner, Apurva Shah, Melvin Johnson, Xiaobing Liu, Lukasz Kaiser, Stephan Gouws, Yoshikiyo Kato, Taku Kudo, Hideto Kazawa, Keith Stevens, George Kurian, Nishant Patil, Wei Wang, Cliff Young, Jason Smith, Jason Riesa, Alex Rudnick, Oriol Vinyals, Greg Corrado, Macduff Hughes, and Jeffrey Dean. Google's neural machine translation system: Bridging the gap between human and machine translation. *CoRR*, abs/1609.08144, 2016. URL `http://arxiv.org/abs/1609.08144`.

Ruibin Xiong, Yunchang Yang, Di He, Kai Zheng, Shuxin Zheng, Chen Xing, Huishuai Zhang, Yanyan Lan, Liwei Wang, and Tieyan Liu. On layer normalization in the transformer architecture. In *International Conference on Machine Learning*, pages 10524–10533. PMLR, 2020.

Albert Xu, Eshaan Pathak, Eric Wallace, Suchin Gururangan, Maarten Sap, and Dan Klein. Detoxifying language models risks marginalizing minority voices. In *Proceedings of the 2021 Conference of the North American Chapter of the Association for Computational Linguistics: Human Language Technologies*, pages 2390–2397, Online, June 2021a. Association for Computational Linguistics. doi: 10.18653/v1/2021.naacl-main.190. URL `https://aclanthology.org/2021.naacl-main.190`.

Jing Xu, Da Ju, Margaret Li, Y-Lan Boureau, Jason Weston, and Emily Dinan. Recipes for safety in open-domain chatbots. *CoRR*, abs/2010.07079, 2020. URL `https://arxiv.org/abs/2010.07079`.

Jing Xu, Arthur Szlam, and Jason Weston. Beyond goldfish memory: Long-term open-domain conversation. In *Proceedings of the 60th Annual Meeting of the Association for Computational Linguistics (Volume 1: Long Papers)*, pages 5180–5197, Dublin, Ireland, May 2022. Association for Computational Linguistics. doi: 10.18653/v1/2022.acl-long.356. URL `https://aclanthology.org/2022.acl-long.356`.

Weijia Xu, Shuming Ma, Dongdong Zhang, and Marine Carpuat. How does distilled data complexity impact the quality and confidence of non-autoregressive machine translation? In *Findings of the Association for Computational Linguistics: ACL-IJCNLP 2021*, pages 4392–4400, 2021b.

Yinfei Yang, Gustavo Hernandez Abrego, Steve Yuan, Mandy Guo, Qinlan Shen, Daniel Cer, Yun-hsuan Sung, Brian Strope, and Ray Kurzweil. Improving multilingual sentence embedding using bi-directional dual encoder with additive margin softmax. In *IJCAI*, pages 5370–5378, 2019.

Zhen Yang, Wei Chen, Feng Wang, and Bo Xu. Unsupervised neural machine translation with weight sharing. *CoRR*, abs/1804.09057, 2018. URL http://arxiv.org/abs/1804.09057.

Marcos Zampieri, Binyam Gebrekidan Gebre, Hernani Costa, and Josef Van Genabith. Comparing approaches to the identification of similar languages. In *Proceedings of the Joint Workshop on Language Technology for Closely Related Languages, Varieties and Dialects*, pages 66–72, 2015a.

Marcos Zampieri, Liling Tan, Nikola Ljubešić, Jörg Tiedemann, and Preslav Nakov. Overview of the DSL shared task 2015. In *Proceedings of the Joint Workshop on Language Technology for Closely Related Languages, Varieties and Dialects*, pages 1–9, 2015b.

Marcos Zampieri, Shervin Malmasi, Preslav Nakov, Sara Rosenthal, Noura Farra, and Ritesh Kumar. SemEval-2019 task 6: Identifying and categorizing offensive language in social media (OffensEval). In *Proceedings of the 13th International Workshop on Semantic Evaluation*, pages 75–86, Minneapolis, Minnesota, USA, June 2019. Association for Computational Linguistics. doi: 10.18653/v1/S19-2010. URL https://aclanthology.org/S19-2010.

Biao Zhang, Philip Williams, Ivan Titov, and Rico Sennrich. Improving massively multilingual neural machine translation and zero-shot translation. In *Proceedings of the 58th Annual Meeting of the Association for Computational Linguistics*, pages 1628–1639, 2020.

Biao Zhang, Ankur Bapna, Rico Sennrich, and Orhan Firat. Share or not? learning to schedule language-specific capacity for multilingual translation. In *International Conference on Learning Representations*, 2021. URL https://openreview.net/forum?id=Wj4ODo0uyCF.

Dakun Zhang, Josep M Crego, and Jean Senellart. Analyzing knowledge distillation in neural machine translation. In *Proceedings of the 15th International Conference on Spoken Language Translation*, pages 23–30, 2018.

Mike Zhang and Antonio Toral. The effect of translationese in machine translation test sets. *arXiv preprint arXiv:1906.08069*, 2019.

Chunting Zhou, Graham Neubig, and Jiatao Gu. Understanding knowledge distillation in non-autoregressive machine translation. *CoRR*, abs/1911.02727, 2020. URL http://arxiv.org/abs/1911.02727.

Barret Zoph, Deniz Yuret, Jonathan May, and Kevin Knight. Transfer learning for low-resource neural machine translation. In *Proceedings of the 2016 Conference on Empirical Methods in Natural Language Processing*, pages 1568–1575, Austin, Texas, November 2016. Association for Computational Linguistics. doi: 10.18653/v1/D16-1163. URL https://aclanthology.org/D16-1163.

Barret Zoph, Irwan Bello, Sameer Kumar, Nan Du, Yanping Huang, Jeff Dean, Noam Shazeer, and William Fedus. Designing effective sparse expert models. *arXiv preprint arXiv:2202.08906*, 2022.

Shoshana Zuboff. *The age of surveillance capitalism: The fight for a human future at the new frontier of power: Barack Obama's books of 2019.* Profile books, 2019.

Ethan Zuckerman. The polyglot internet, October 2008. URL https://ethanzuckerman.com/the-polyglot-internet/.

Alp Öktem, Muhannad Albayk Jaam, Eric DeLuca, and Grace Tang. Gamayun – language technology for humanitarian response. In *2020 IEEE Global Humanitarian Technology Conference (GHTC)*, pages 1–4, 2020. doi: 10.1109/GHTC46280.2020.9342939.

# Appendix A. Languages

## A.1 Ethical Considerations around Language Standardization

Our work contains many languages that have varying levels of standardization. What may appear as a single language might in fact be hiding several competing standards for script, spelling, word formation, the acceptance of neologisms and borrowed terms, and more generally grammatical and style guidelines. Examples include languages such as Fulah, which includes several distinct languages and languages such as Kashmiri and Central Kanuri which represent languages where multiple scripts are in common use.

The history of standardization of languages is complex, and has often historically been related to power and identity. Certain governments or institutions might push for a certain standard, or more subtle programs by specific circles (cultural elites, religious groups, economically powerful regions, major publishing houses) might exist with the aim of "civilizing" others or establish distinction (Bourdieu, 1987; De Mauro, 2014; Haugen, 1959; Rickford, 2012). At the same time, we may observe anti-standardization, groups of people that reject and contest the separate status of languages of the elite as 'the standard' (Armstrong and Mackenzie, 2013).

Work on language technologies has the potential to affect the way people use language and how it evolves, which opens up questions about responsibility. In order to consider the effect of our models on language communities, we examine two diametrically opposed cases: centrally and distributedly standardized languages.

**Centralized standardization.** A large group of languages in our scope have generally recognized central regulatory bodies, in the form of either governmental, academic or community-run structures that possess deep organizational resources and powers. For these languages, grammars and dictionaries will typically be available, as well as formal ways to decide on the adoption of neologisms. While such resources are usually not directly used by translation technologies such as neural machine translation or language identification, they can be used as references for data collection, provided to annotators as guidelines, or used by scientists to decide how training datasets should be filtered. In training our models, however, we also make use of large amounts of linguistic data mined from the web. This data has the potential of capturing the more fluid state of language development and of including neologisms and language constructs that may not be officially sanctioned (Plank, 2016).

**Distributed standardization.** In contrast to centrally institutionalized languages, many other languages might have an existing standardizing body whose work is however not recognized by large parts of the community; or multiple standardizing bodies might coexist, providing conflicting guidance; or multiple weakly-standardized variants might have arisen without any regulatory bodies. Many of the languages in this category are considered low resource due to the absence or extreme scarcity of annotated data. In order to develop models, it was necessary to annotate some datasets, which in turn required the development of guidelines for the annotators. This may steer the development of languages in potentially undesirable ways, such as favoring one variant over others. As a result, certain groups might end up feeling left out, or forced to comply with standards that feel alien to them. While only specific variants of certain languages might be supported out of the box, it should be possible to extend support to other variants with only limited data annotation efforts. Indeed, it's been shown that closely related languages strongly benefit from transfer learning (Fan et al., 2020; Gu et al., 2018; Sajjad et al., 2020; Zoph et al., 2016), in such a way that "competing" variants might end up benefiting from each other's presence in the dataset. Furthermore, simple adaptation between certain variants can be performed in an automatic manner, as is the case of transliteration (see Section 8.7) or mapping between different spelling standards (Bollmann, 2019; Lusito et al., 2022).

## Appendix B. Evaluation

### B.1 Toxicity Lists

The toxicity lists for languages other than English were created through human translation using an English-language prompt, except the Luxembourgish and Asturian lists, the human translation of which was pivoted via German and Spanish translations of the English prompt, respectively. However, the process was iterative, and the prompt had to be modified in order to reduce translation errors caused by ambiguities and misunderstandings.

For most of the languages that were part of FLORES 101, we used existing lists produced over the course of a previous project. The English prompt for this iteration did not provide any definitions for the words it included, or part-of-speech and register information. Apart from translations directly related to the prompt, the lists could be expanded without any restrictions to allow for the inclusion of cultural specificities and spelling variants (including inflection, leetspeak, and nonstandard spelling). The lists resulting from this first method comprise numerous items that cannot be traced back to any known dictionary form, which makes checking for quality and anomalies a particularly daunting task. It is already not uncommon for toxic words to be excluded from widely available dictionaries, even in their more standard form; nonstandard forms further complicate the matter, especially when any number of suggestions can be freely added. It is unclear how important the inclusion of variants such as leetspeak and nonstandard spelling is. Nonstandard variants can, by definition, be nonstandard in a number of ways, which string-matching methods cannot cover exhaustively.

For languages that were added to those mentioned above, the English prompt was supplemented with part-of-speech and register information, as well as links to definitions, in an effort to reduce errors associated with polysemy and other ambiguities. Translators were encouraged to indicate whether a prompt item had an equivalent in their languages, and to suggest items that an English-centric prompt would miss. However, suggestions were limited to around forty items without specific restrictions as to the number of derived word forms per item apart from the general guidance of keeping within the boundaries of frequently used word forms (i.e. steering clear of infrequent and archaic word forms). In this method, translators were not asked to produce leetspeak or nonstandard spelling variants, yet not discouraged from including them where they saw fit.

Both methods are met with at least one common challenge. Despite the additional grammatical and semantic information provided by the second type of prompt, as well as the complementary training, translators seem to face similar difficulties when deciding whether certain items qualify as slurs or as nontoxic language. This is particularly salient in the case of some racial slurs and of slurs against the LGBTQ+ community. One important factor may be, as was previously noted, that it is sometimes difficult to find translators with diverse backgrounds, who will accept this sort of assignment. Another factor may be that some of these items may have been reclaimed by the community of reference but even allies of the community who are not themselves members thereof would hesitate to use them.

## Appendix C. Data

To assess the performance of our language identification (LID) system, we leverage the high-quality annotations from the Flores-200 dataset. Precision, recall and F1 scores across all languages on Flores-200 `devtest` are shown in Table 49.

As discussed in Section 5.1.4, there is a significant domain mismatch between the data used to train our LID system and the web corpus that system is used on in our pipeline. To assess the impact, we conducted an extensive human evaluation. We selected 74 low-resource languages on which a preliminary LID model yielded low F1 scores. We randomly picked several thousand sentences predicted to be among those languages and asked annotators to assess whether each prediction was correct. We built a challenge set based on these annotations to benchmark our final LID model. Table 50 compares the performance of our model against the readily available `CLD3` system.

| Language | F1 | Precision | Recall | FPR | Language | F1 | Precision | Recall | FPR |
|---|---|---|---|---|---|---|---|---|---|
| ace_Arab | 97.0 | 98.5 | 95.6 | 0.0074 | lit_Latn | 99.9 | 99.8 | 99.9 | 0.0010 |
| ace_Latn | 99.5 | 99.3 | 99.7 | 0.0035 | lmo_Latn | 97.8 | 98.0 | 97.6 | 0.0099 |
| afr_Latn | 99.9 | 99.8 | 100.0 | 0.0010 | ltg_Latn | 99.1 | 100.0 | 98.2 | 0.0000 |
| aka_Latn | 99.9 | 100.0 | 99.8 | 0.0000 | ltz_Latn | 100.0 | 100.0 | 99.9 | 0.0000 |
| amh_Ethi | 99.9 | 99.8 | 100.0 | 0.0010 | lua_Latn | 99.6 | 99.7 | 99.5 | 0.0015 |
| arb_Arab | 96.9 | 94.0 | 100.0 | 0.2667 | lug_Latn | 99.3 | 98.7 | 99.9 | 0.0064 |
| arb_Latn | 99.7 | 99.9 | 99.5 | 0.0005 | luo_Latn | 99.9 | 99.9 | 99.9 | 0.0005 |
| asm_Beng | 100.0 | 100.0 | 99.9 | 0.0000 | lus_Latn | 99.6 | 100.0 | 99.1 | 0.0000 |
| ast_Latn | 99.1 | 98.8 | 99.4 | 0.0059 | mag_Deva | 97.1 | 97.8 | 96.3 | 0.0109 |
| awa_Deva | 96.5 | 98.6 | 94.5 | 0.0069 | mai_Deva | 99.1 | 99.9 | 98.3 | 0.0005 |
| ayr_Latn | 100.0 | 100.0 | 99.9 | 0.0000 | mal_Mlym | 100.0 | 100.0 | 100.0 | 0.0000 |
| azb_Arab | 88.1 | 98.5 | 79.7 | 0.0059 | mar_Deva | 99.8 | 99.6 | 100.0 | 0.0020 |
| azj_Latn | 99.8 | 99.5 | 100.0 | 0.0025 | min_Latn | 54.1 | 100.0 | 37.1 | 0.0000 |
| bak_Cyrl | 99.9 | 99.9 | 99.9 | 0.0005 | mkd_Cyrl | 100.0 | 100.0 | 100.0 | 0.0000 |
| bam_Latn | 61.3 | 47.4 | 86.9 | 0.4817 | plt_Latn | 100.0 | 100.0 | 100.0 | 0.0000 |
| ban_Latn | 97.0 | 99.5 | 94.7 | 0.0025 | mlt_Latn | 100.0 | 99.9 | 100.0 | 0.0005 |
| bel_Cyrl | 100.0 | 100.0 | 100.0 | 0.0000 | mni_Beng | 100.0 | 100.0 | 99.9 | 0.0000 |
| bem_Latn | 97.5 | 95.4 | 99.7 | 0.0242 | khk_Cyrl | 100.0 | 100.0 | 100.0 | 0.0000 |
| ben_Beng | 99.9 | 99.8 | 100.0 | 0.0010 | mos_Latn | 96.9 | 100.0 | 94.0 | 0.0000 |
| bho_Deva | 95.5 | 98.4 | 92.7 | 0.0074 | mri_Latn | 99.8 | 99.8 | 99.8 | 0.0010 |
| bjn_Arab | 95.1 | 96.9 | 93.4 | 0.0148 | zsm_Latn | 93.9 | 93.8 | 94.1 | 0.0311 |
| bjn_Latn | 83.8 | 74.7 | 95.6 | 0.1621 | mya_Mymr | 100.0 | 100.0 | 100.0 | 0.0000 |
| bod_Tibt | 99.1 | 98.3 | 100.0 | 0.0089 | nld_Latn | 98.8 | 97.7 | 100.0 | 0.0119 |
| bos_Latn | 59.0 | 80.5 | 46.5 | 0.0563 | nno_Latn | 98.1 | 98.0 | 98.2 | 0.0099 |
| bug_Latn | 97.4 | 98.9 | 95.9 | 0.0054 | nob_Latn | 98.5 | 97.8 | 99.3 | 0.0114 |
| bul_Cyrl | 100.0 | 100.0 | 99.9 | 0.0000 | npi_Deva | 99.8 | 99.5 | 100.0 | 0.0025 |
| cat_Latn | 99.3 | 98.6 | 100.0 | 0.0069 | nso_Latn | 98.3 | 97.2 | 99.5 | 0.0143 |
| ceb_Latn | 99.9 | 99.9 | 99.8 | 0.0005 | nus_Latn | 99.8 | 99.7 | 99.9 | 0.0015 |
| ces_Latn | 99.9 | 99.8 | 100.0 | 0.0010 | nya_Latn | 94.9 | 97.6 | 92.3 | 0.0114 |
| cjk_Latn | 87.4 | 97.3 | 79.3 | 0.0109 | oci_Latn | 98.6 | 97.3 | 100.0 | 0.0138 |
| ckb_Arab | 100.0 | 100.0 | 100.0 | 0.0000 | gaz_Latn | 100.0 | 99.9 | 100.0 | 0.0005 |
| crh_Latn | 98.3 | 100.0 | 96.6 | 0.0000 | ory_Orya | 100.0 | 100.0 | 100.0 | 0.0000 |
| cym_Latn | 100.0 | 100.0 | 100.0 | 0.0000 | pag_Latn | 99.5 | 99.8 | 99.2 | 0.0010 |
| dan_Latn | 99.7 | 99.8 | 99.5 | 0.0010 | pan_Guru | 100.0 | 100.0 | 100.0 | 0.0000 |
| deu_Latn | 99.1 | 98.2 | 100.0 | 0.0094 | pap_Latn | 98.8 | 97.8 | 99.8 | 0.0114 |
| dik_Latn | 99.4 | 100.0 | 98.8 | 0.0000 | pol_Latn | 98.8 | 97.6 | 100.0 | 0.0124 |
| diq_Latn | —- | 0.0 | —- | 0.0010 | por_Latn | 99.1 | 98.3 | 99.9 | 0.0089 |
| dyu_Latn | 5.6 | 33.7 | 3.1 | 0.0301 | prs_Arab | 54.4 | 94.8 | 38.1 | 0.0104 |
| dzo_Tibt | 99.8 | 100.0 | 99.6 | 0.0000 | pbt_Arab | 99.8 | 99.7 | 99.8 | 0.0015 |
| ell_Grek | 100.0 | 100.0 | 100.0 | 0.0000 | quy_Latn | 100.0 | 100.0 | 100.0 | 0.0000 |
| eng_Latn | 97.0 | 94.2 | 100.0 | 0.0306 | ron_Latn | 99.8 | 99.6 | 100.0 | 0.0020 |
| epo_Latn | 99.7 | 99.4 | 100.0 | 0.0030 | run_Latn | 97.9 | 98.1 | 97.6 | 0.0094 |
| est_Latn | 99.9 | 99.8 | 100.0 | 0.0010 | rus_Cyrl | 100.0 | 99.9 | 100.0 | 0.0005 |
| eus_Latn | 99.9 | 99.8 | 100.0 | 0.0010 | sag_Latn | 99.7 | 99.9 | 99.5 | 0.0005 |
| ewe_Latn | 99.8 | 99.6 | 100.0 | 0.0020 | san_Deva | 99.6 | 99.9 | 99.2 | 0.0005 |
| fao_Latn | 49.1 | 100.0 | 32.5 | 0.0000 | sat_Olck | 100.0 | 100.0 | 100.0 | 0.0000 |
| pes_Arab | 69.8 | 54.2 | 98.0 | 0.4140 | scn_Latn | 99.4 | 98.9 | 99.8 | 0.0054 |
| fij_Latn | 100.0 | 99.9 | 100.0 | 0.0005 | shn_Mymr | 100.0 | 100.0 | 100.0 | 0.0000 |
| fin_Latn | 100.0 | 100.0 | 100.0 | 0.0000 | sin_Sinh | 100.0 | 100.0 | 100.0 | 0.0000 |
| fon_Latn | 99.8 | 100.0 | 99.6 | 0.0000 | slk_Latn | 100.0 | 100.0 | 100.0 | 0.0000 |
| fra_Latn | 99.8 | 99.7 | 99.9 | 0.0015 | slv_Latn | 99.9 | 99.7 | 100.0 | 0.0015 |
| fur_Latn | 99.8 | 99.9 | 99.7 | 0.0005 | smo_Latn | 100.0 | 100.0 | 99.9 | 0.0000 |
| fuv_Latn | 98.4 | 99.4 | 97.4 | 0.0030 | sna_Latn | 99.3 | 98.7 | 99.9 | 0.0064 |
| gla_Latn | 99.9 | 99.7 | 100.0 | 0.0015 | snd_Arab | 99.8 | 99.6 | 100.0 | 0.0020 |
| gle_Latn | 100.0 | 99.9 | 100.0 | 0.0005 | som_Latn | 100.0 | 99.9 | 100.0 | 0.0005 |
| glg_Latn | 99.6 | 99.6 | 99.6 | 0.0020 | sot_Latn | 75.4 | 100.0 | 60.6 | 0.0000 |
| grn_Latn | 100.0 | 100.0 | 99.9 | 0.0000 | spa_Latn | 99.5 | 98.9 | 100.0 | 0.0054 |
| guj_Gujr | 100.0 | 100.0 | 100.0 | 0.0000 | als_Latn | 99.9 | 99.7 | 100.0 | 0.0015 |
| hat_Latn | 99.9 | 99.9 | 99.9 | 0.0005 | srd_Latn | 98.2 | 100.0 | 96.4 | 0.0000 |
| hau_Latn | 99.8 | 99.6 | 100.0 | 0.0020 | srp_Cyrl | 100.0 | 99.9 | 100.0 | 0.0005 |
| heb_Hebr | 100.0 | 100.0 | 100.0 | 0.0000 | ssw_Latn | 99.2 | 99.8 | 98.5 | 0.0010 |
| hin_Deva | 89.2 | 80.6 | 99.9 | 0.1206 | sun_Latn | 96.4 | 95.9 | 96.9 | 0.0208 |
| hne_Deva | 94.4 | 98.6 | 90.6 | 0.0064 | swe_Latn | 100.0 | 100.0 | 100.0 | 0.0000 |
| hrv_Latn | 73.3 | 62.3 | 88.9 | 0.2688 | swh_Latn | 88.4 | 79.2 | 100.0 | 0.1309 |
| hun_Latn | 99.1 | 98.3 | 100.0 | 0.0089 | szl_Latn | 99.0 | 99.9 | 98.1 | 0.0005 |
| hye_Armn | 100.0 | 100.0 | 100.0 | 0.0000 | tah_Latn | —- | 0.0 | —- | 0.0005 |
| ibo_Latn | 100.0 | 99.9 | 100.0 | 0.0005 | tam_Taml | 100.0 | 100.0 | 100.0 | 0.0000 |

| Language | F1 | Precision | Recall | FPR | Language | F1 | Precision | Recall | FPR |
|---|---|---|---|---|---|---|---|---|---|
| ilo_Latn | 99.9 | 99.8 | 100.0 | 0.0010 | tat_Cyrl | 100.0 | 100.0 | 99.9 | 0.0000 |
| ind_Latn | 82.0 | 70.2 | 98.5 | 0.2095 | tel_Telu | 100.0 | 100.0 | 100.0 | 0.0000 |
| isl_Latn | 75.2 | 60.2 | 100.0 | 0.3305 | tgk_Cyrl | 100.0 | 100.0 | 100.0 | 0.0000 |
| ita_Latn | 98.3 | 96.7 | 99.9 | 0.0168 | tgl_Latn | 99.8 | 99.5 | 100.0 | 0.0025 |
| jav_Latn | 97.6 | 95.5 | 99.7 | 0.0232 | tha_Thai | 100.0 | 100.0 | 100.0 | 0.0000 |
| jpn_Jpan | 98.6 | 98.0 | 99.3 | 0.0104 | tir_Ethi | 99.9 | 100.0 | 99.8 | 0.0000 |
| kab_Latn | 86.1 | 75.7 | 99.9 | 0.1606 | taq_Latn | 79.7 | 100.0 | 66.3 | 0.0000 |
| kac_Latn | 100.0 | 100.0 | 100.0 | 0.0000 | ton_Latn | —- | 0.0 | —- | 0.0231 |
| kam_Latn | 77.5 | 99.7 | 63.3 | 0.0010 | tpi_Latn | 99.8 | 100.0 | 99.6 | 0.0000 |
| kan_Knda | 100.0 | 100.0 | 100.0 | 0.0000 | tsn_Latn | 84.4 | 73.1 | 99.8 | 0.1838 |
| kas_Arab | 97.2 | 100.0 | 94.5 | 0.0000 | tso_Latn | 99.0 | 98.2 | 99.9 | 0.0094 |
| kas_Deva | 98.5 | 99.9 | 97.1 | 0.0005 | tuk_Latn | 100.0 | 100.0 | 100.0 | 0.0000 |
| kat_Geor | 100.0 | 100.0 | 100.0 | 0.0000 | tum_Latn | 98.3 | 96.8 | 99.9 | 0.0163 |
| knc_Arab | 70.3 | 100.0 | 54.2 | 0.0000 | tur_Latn | 98.4 | 96.8 | 100.0 | 0.0163 |
| knc_Latn | 99.8 | 99.7 | 99.8 | 0.0015 | tzm_Tfng | 100.0 | 100.0 | 99.9 | 0.0000 |
| kaz_Cyrl | 100.0 | 100.0 | 99.9 | 0.0000 | uig_Arab | 100.0 | 99.9 | 100.0 | 0.0005 |
| kbp_Latn | 100.0 | 100.0 | 100.0 | 0.0000 | ukr_Cyrl | 100.0 | 100.0 | 100.0 | 0.0000 |
| kea_Latn | 97.0 | 100.0 | 94.2 | 0.0000 | umb_Latn | 96.8 | 95.7 | 97.9 | 0.0217 |
| khm_Khmr | 100.0 | 100.0 | 99.9 | 0.0000 | urd_Arab | 97.4 | 95.0 | 100.0 | 0.0262 |
| kik_Latn | 96.3 | 93.4 | 99.4 | 0.0351 | uzn_Latn | 99.9 | 99.8 | 100.0 | 0.0010 |
| kin_Latn | 98.0 | 97.6 | 98.3 | 0.0119 | vec_Latn | 99.2 | 99.1 | 99.3 | 0.0044 |
| kir_Cyrl | 100.0 | 100.0 | 100.0 | 0.0000 | vie_Latn | 99.1 | 98.3 | 100.0 | 0.0089 |
| kmb_Latn | 93.7 | 90.7 | 96.9 | 0.0499 | war_Latn | 100.0 | 100.0 | 100.0 | 0.0000 |
| kon_Latn | 99.3 | 98.7 | 99.9 | 0.0064 | wes_Arab | —- | 0.0 | —- | 0.0005 |
| kor_Hang | 99.4 | 98.7 | 100.0 | 0.0064 | wol_Latn | 99.6 | 99.8 | 99.3 | 0.0010 |
| krc_Cyrl | —- | 0.0 | —- | 0.0005 | xho_Latn | 97.9 | 97.1 | 98.7 | 0.0148 |
| kmr_Latn | 99.7 | 99.3 | 100.0 | 0.0035 | ydd_Hebr | 100.0 | 100.0 | 100.0 | 0.0000 |
| lao_Laoo | 100.0 | 100.0 | 99.9 | 0.0000 | yor_Latn | 99.7 | 99.6 | 99.8 | 0.0020 |
| lvs_Latn | 99.3 | 98.5 | 100.0 | 0.0074 | yue_Hant | 49.0 | 45.4 | 53.2 | 0.3192 |
| lij_Latn | 98.4 | 99.4 | 97.4 | 0.0030 | zho_Hans | 85.4 | 94.6 | 77.8 | 0.0222 |
| lim_Latn | 99.0 | 99.8 | 98.2 | 0.0010 | zho_Hant | 46.7 | 49.6 | 44.2 | 0.2248 |
| lin_Latn | 99.8 | 99.8 | 99.8 | 0.0010 | zul_Latn | 97.3 | 95.4 | 99.4 | 0.0242 |

Table 49: **LID results on all Flores-200 languages.**

| | F1 | | FPR | | | F1 | | FPR | |
|---|---|---|---|---|---|---|---|---|---|
| | Ours | CLD3 | Ours | CLD3 | | Ours | CLD3 | Ours | CLD3 |
| umb_Latn | 0.80 | —- | 1.59 | —- | diq_Latn | 65.07 | —- | 0.47 | —- |
| xho_Latn | 10.64 | 6.40 | 2.27 | 1.26 | kab_Latn | 65.29 | —- | 1.70 | —- |
| kam_Latn | 10.88 | —- | 1.18 | —- | run_Latn | 66.33 | —- | 0.62 | —- |
| sna_Latn | 14.13 | 11.73 | 2.47 | 2.60 | smo_Latn | 72.05 | 78.25 | 1.63 | 0.94 |
| pag_Latn | 15.72 | —- | 2.43 | —- | bho_Deva | 73.06 | —- | 0.03 | —- |
| tso_Latn | 19.75 | —- | 2.60 | —- | scn_Latn | 73.59 | —- | 1.57 | —- |
| war_Latn | 20.72 | —- | 2.56 | —- | bak_Cyrl | 73.95 | —- | 0.44 | —- |
| ast_Latn | 21.24 | —- | 2.70 | —- | nno_Latn | 74.08 | —- | 1.15 | —- |
| lmo_Latn | 24.79 | —- | 3.73 | —- | srd_Latn | 76.64 | —- | 0.05 | —- |
| zul_Latn | 27.22 | 19.32 | 1.14 | 1.45 | ceb_Latn | 78.26 | 66.08 | 0.49 | 0.73 |
| lim_Latn | 28.71 | —- | 3.22 | —- | oci_Latn | 79.59 | —- | 0.37 | —- |
| kon_Latn | 29.03 | —- | 2.18 | —- | bul_Cyrl | 81.86 | 68.81 | 0.93 | 0.92 |
| bem_Latn | 29.42 | —- | 2.60 | —- | lus_Latn | 84.60 | —- | 0.34 | —- |
| lug_Latn | 29.89 | —- | 1.69 | —- | kaz_Cyrl | 85.81 | 72.79 | 0.40 | 0.66 |
| mos_Latn | 31.24 | —- | 2.37 | —- | hat_Latn | 86.51 | 65.02 | 0.31 | 1.31 |
| twi_Latn | 31.97 | —- | 2.52 | —- | azb_Arab | 86.63 | —- | 0.22 | —- |
| sot_Latn | 33.97 | 38.31 | 2.87 | 1.93 | glg_Latn | 86.70 | 68.79 | 0.28 | 0.94 |
| dyu_Latn | 36.32 | —- | 0.52 | —- | hrv_Latn | 87.54 | 23.72 | 0.24 | 2.19 |
| min_Latn | 37.28 | —- | 2.55 | —- | san_Deva | 88.78 | —- | 0.21 | —- |
| hau_Latn | 43.27 | 39.63 | 2.12 | 2.16 | bod_Tibt | 90.67 | —- | 0.11 | —- |
| kmb_Latn | 44.28 | —- | 2.09 | —- | quy_Latn | 90.97 | —- | 0.09 | —- |
| kin_Latn | 46.26 | —- | 2.00 | —- | bos_Latn | 92.68 | 63.60 | 0.74 | 1.43 |
| cjk_Latn | 47.08 | —- | 1.91 | —- | mai_Deva | 92.73 | —- | 0.15 | —- |
| ewe_Latn | 49.48 | —- | 2.26 | —- | jav_Latn | 92.82 | 68.95 | 0.08 | 0.67 |
| lua_Latn | 50.19 | —- | 0.88 | —- | kat_Geor | 93.19 | 76.18 | 0.08 | 0.12 |
| fuv_Latn | 51.06 | —- | 0.89 | —- | lvs_Latn | 93.64 | 84.49 | 0.17 | 0.28 |
| ilo_Latn | 51.58 | —- | 2.12 | —- | slv_Latn | 93.95 | 76.94 | 0.10 | 0.32 |
| tum_Latn | 51.59 | —- | 2.00 | —- | kbp_Latn | 94.05 | —- | 0.20 | —- |
| vec_Latn | 51.85 | —- | 1.70 | —- | kik_Latn | 94.81 | —- | 0.10 | —- |
| tsn_Latn | 52.97 | —- | 2.48 | —- | sun_Latn | 95.90 | 74.26 | 0.07 | 0.79 |
| pap_Latn | 57.64 | —- | 2.05 | —- | kmr_Latn | 96.10 | 63.98 | 0.10 | 1.59 |
| tgk_Cyrl | 59.66 | 51.40 | 0.80 | 0.79 | kir_Cyrl | 96.98 | 87.37 | 0.06 | 0.37 |
| eus_Latn | 61.32 | 64.25 | 0.91 | 0.60 | npi_Deva | 97.17 | 76.76 | 0.16 | 2.61 |
| fij_Latn | 61.54 | —- | 1.66 | —- | khk_Cyrl | 98.47 | 96.05 | 0.04 | 0.09 |
| fon_Latn | 62.85 | —- | 1.81 | —- | tuk_Latn | 99.19 | —- | 0.03 | —- |
| sag_Latn | 63.30 | —- | 1.33 | —- | sat_Olck | 99.60 | —- | —- | —- |

Table 50: **Comparison of CLD3 and our model on a challenge set built from human annotations.** False Positive Rates (FPR) are reported on top of F1 scores, which can be misleading when an LID system is eventually to be applied on web data with different class balances than the development set. Cells corresponding to languages unsupported by CLD3 are left blank.

## Appendix D. Modeling

### D.1 Ablation Dataset

In Table 51 we list all the languages in our *ablation* dataset used for experimentation in Section 6.

Table 51 contains the exact set of 110 language pairs used in the ablation dataset for experiments in Section 6.

### D.2 SMT vs MMT

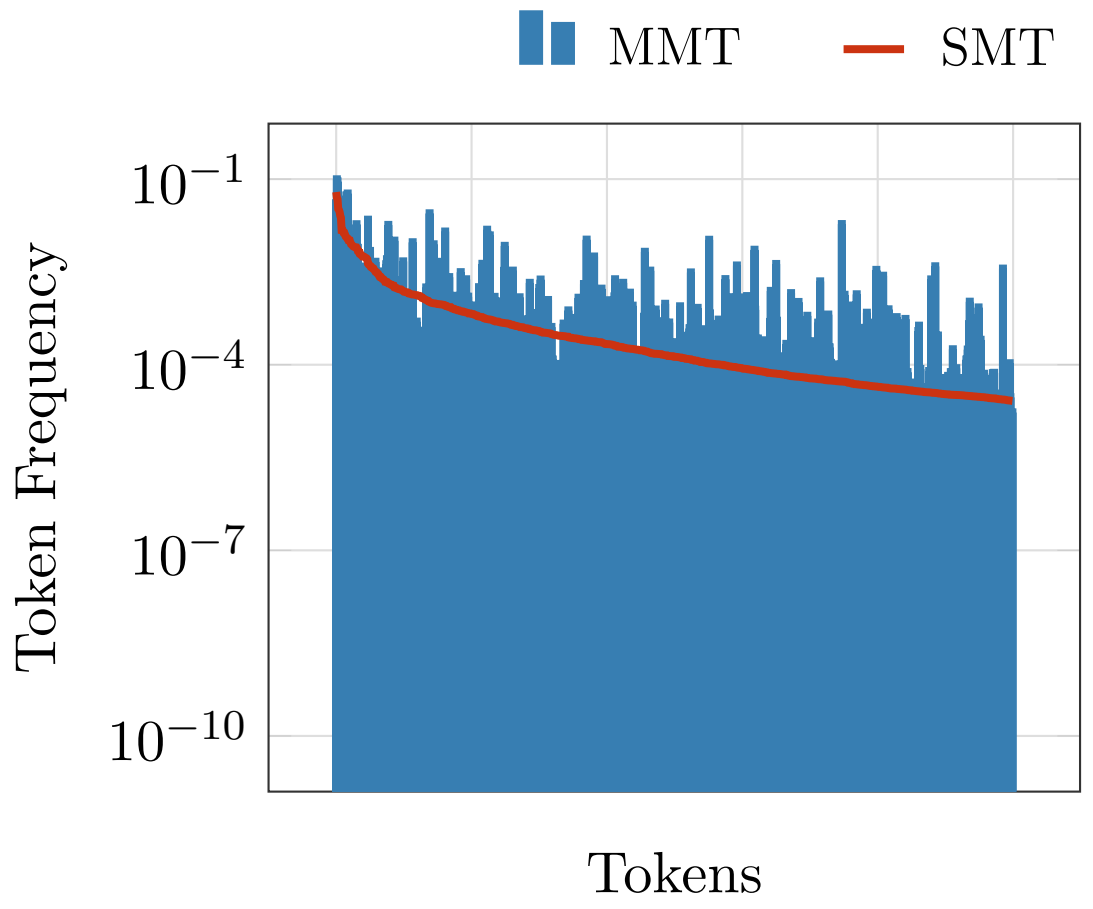

(a) `kau_Arab-eng_Latn`, $\rho = 0.26, \tau = 0.19$

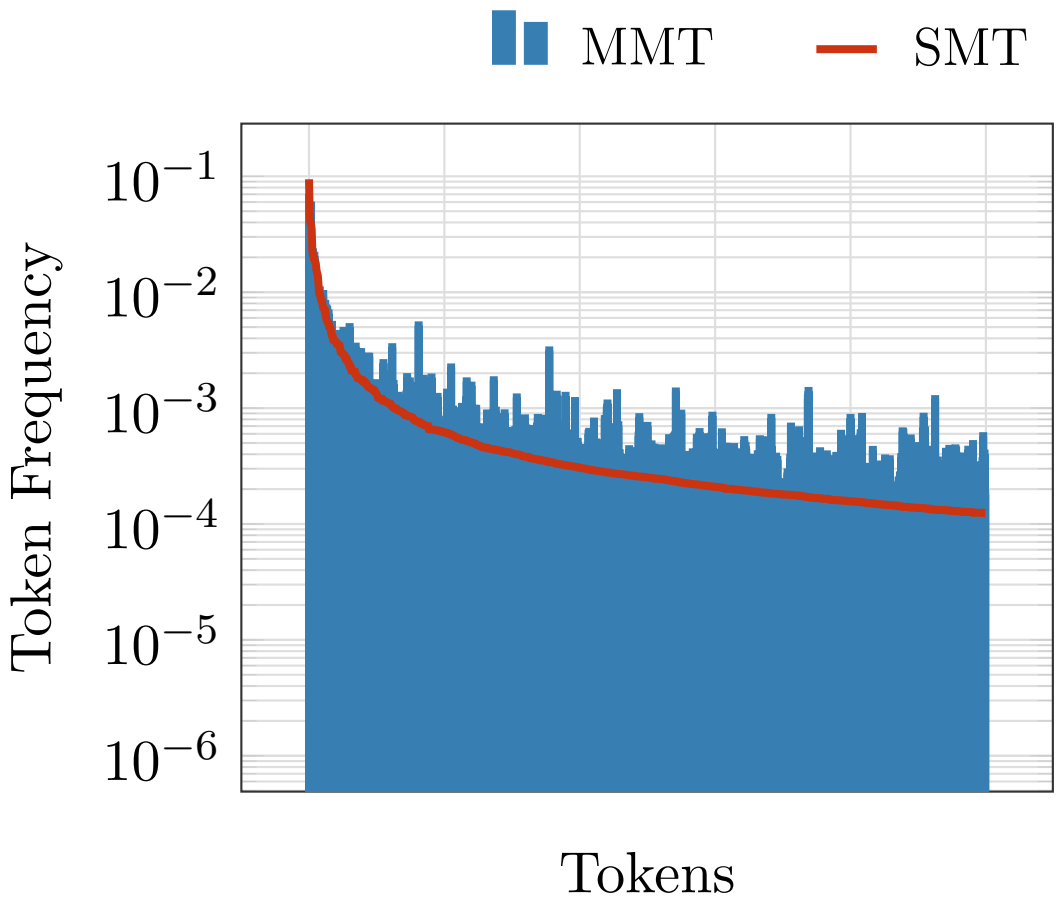

(b) `awa_Deva-eng_Latn`, $\rho = 0.52, \tau = 0.45$

Figure 42: **Comparison of token counts generated by MMT and SMT** for the same set of source sentences. Tokens on the $x$-axis are sorted by decreasing SMT generation counts. We also report Spearman's $\rho$ and Kendall's $\tau$ rank correlation coefficients.

In Section 6.4.1 we observed that combining backtranslations from Statistical Machine Translation (SMT) and Multilingual Neural Machine Translation (MMT) is beneficial. We hypothesized that this is because the two models provide different, complementary sources of noise.

One way to visualize these differences is to plot the token frequencies of the translations produced by the two types of model. We show this in Figure 42 for two directions. Namely, Figure 42a shows the same set of `kau_Arab` sentences translated into `eng_Latn` by the two models; and Figure 42b does this for `awa_Deva`.

| Language Pairs | | | | |
|---|---|---|---|---|
| arb_Arab-sin_Sinh | eng_Latn-pes_Arab | eng_Latn-kin_Latn | eng_Latn-sin_Sinh | eng_Latn-yue_Hant |
| sin_Sinh-arb_Arab | pes_Arab-eng_Latn | kin_Latn-eng_Latn | sin_Sinh-eng_Latn | yue_Hant-eng_Latn |
| eng_Latn-ace_Latn | eng_Latn-fin_Latn | eng_Latn-kon_Latn | eng_Latn-snd_Arab | eng_Latn-zho_Hans |
| ace_Latn-eng_Latn | fin_Latn-eng_Latn | kon_Latn-eng_Latn | snd_Arab-eng_Latn | zho_Hans-eng_Latn |
| eng_Latn-afr_Latn | eng_Latn-fon_Latn | eng_Latn-lvs_Latn | eng_Latn-tam_Taml | eus_Latn-por_Latn |
| afr_Latn-eng_Latn | fon_Latn-eng_Latn | lvs_Latn-eng_Latn | tam_Taml-eng_Latn | por_Latn-eus_Latn |
| eng_Latn-arb_Arab | eng_Latn-fra_Latn | eng_Latn-lin_Latn | eng_Latn-tel_Telu | fra_Latn-hau_Latn |
| arb_Arab-eng_Latn | fra_Latn-eng_Latn | lin_Latn-eng_Latn | tel_Telu-eng_Latn | hau_Latn-fra_Latn |
| eng_Latn-ast_Latn | eng_Latn-fuv_Latn | eng_Latn-luo_Latn | eng_Latn-tir_Ethi | fra_Latn-kon_Latn |
| ast_Latn-eng_Latn | fuv_Latn-eng_Latn | luo_Latn-eng_Latn | tir_Ethi-eng_Latn | kon_Latn-fra_Latn |
| eng_Latn-ayr_Latn | eng_Latn-hau_Latn | eng_Latn-mal_Mlym | eng_Latn-tso_Latn | fra_Latn-lin_Latn |
| ayr_Latn-eng_Latn | hau_Latn-eng_Latn | mal_Mlym-eng_Latn | tso_Latn-eng_Latn | lin_Latn-fra_Latn |
| eng_Latn-bel_Cyrl | eng_Latn-hin_Deva | eng_Latn-mar_Deva | eng_Latn-twi_Latn | fra_Latn-swh_Latn |
| bel_Cyrl-eng_Latn | hin_Deva-eng_Latn | mar_Deva-eng_Latn | twi_Latn-eng_Latn | swh_Latn-fra_Latn |
| eng_Latn-bul_Cyrl | eng_Latn-isl_Latn | eng_Latn-nso_Latn | eng_Latn-urd_Arab | hin_Deva-tam_Taml |
| bul_Cyrl-eng_Latn | isl_Latn-eng_Latn | nso_Latn-eng_Latn | urd_Arab-eng_Latn | tam_Taml-hin_Deva |
| eng_Latn-cjk_Latn | eng_Latn-ita_Latn | eng_Latn-oci_Latn | eng_Latn-vie_Latn | jpn_Jpan-kor_Hang |
| cjk_Latn-eng_Latn | ita_Latn-eng_Latn | oci_Latn-eng_Latn | vie_Latn-eng_Latn | kor_Hang-jpn_Jpan |
| eng_Latn-cym_Latn | eng_Latn-kea_Latn | eng_Latn-run_Latn | eng_Latn-wol_Latn | rus_Cyrl-tat_Cyrl |
| cym_Latn-eng_Latn | kea_Latn-eng_Latn | run_Latn-eng_Latn | wol_Latn-eng_Latn | tat_Cyrl-rus_Cyrl |
| eng_Latn-ewe_Latn | eng_Latn-kik_Latn | eng_Latn-rus_Cyrl | eng_Latn-yor_Latn | swh_Latn-tsn_Latn |
| ewe_Latn-eng_Latn | kik_Latn-eng_Latn | rus_Cyrl-eng_Latn | yor_Latn-eng_Latn | tsn_Latn-swh_Latn |

Table 51: Language Pairs in the Ablation Dataset used in Section 6

## Appendix E. Bringing it All Together

### E.1 Preparing the Data

#### E.1.1 Primary Dataset Composition

For reference, we summarize in Table 52 some of the main datasets used in training our model NLLB-200. Our data was largely downloaded via OPUS (Tiedemann, 2012) and with the help of the `mtdata` tool (Gowda et al., 2021). Direction counts refer to the number of directions used in this work, which may differ from to the total number of directions made available by the corpus. They also do not include reverse directions, such that e.g. `eng_Latn-fra_Latn` does not also contribute to the count as `fra_Latn-eng_Latn`. Complete training configuration files are available in our repository[49].

#### E.1.2 Importance of Backtranslation Quality on Model Scaling

We study the importance of backtranslation quality on model scaling. Data augmentation strategies such as backtranslation, self-training, and even large-scale mining form a significant portion of training data for modern translation systems. However, they are not as high quality as human translated data and data augmentation quality may limit a translation model's overall quality.

We train a multilingual model on 60 African language translation directions, to and from French and English. Subsequently, we investigate the importance of BT quality on model performance for 8 languages: `fuv_Latn`, `kmb_Latn`, `lug_Latn`, `nya_Latn`, `swh_Latn`,

49. `https://github.com/facebookresearch/fairseq/tree/nllb/data`

| Corpus Name | Citation | # Directions | # Languages |
|---|---|---|---|
| AAU Ethiopian Languages | Abate et al. (2018) | 3 | 4 |
| AI4D | Degila et al. (2020); Siminyu et al. (2021) | 3 | 5 |
| DGT | Tiedemann (2012) | 94 | 24 |
| ECB | Tiedemann (2012) | 74 | 19 |
| EMEA | Tiedemann (2012) | 86 | 22 |
| English-Twi | Azunre et al. (2021a,b) | 2 | 1 |
| EU Bookshop | Skadiņš et al. (2014b) | 160 | 38 |
| GlobalVoices | Tiedemann (2012) | 235 | 41 |
| HornMT | Hadgu et al. (2021) | 10 | . 5 |
| InfoPankki v1 | Tiedemann (2012) | 30 | 12 |
| QCRI Educational Domain | Abdelali et al. (2014) | 866 | 135 |
| JHU Bible | McCarthy et al. (2020) | 300 | 155 |
| MADAR | Bouamor et al. (2019) | 5 | 6 |
| Mburisano | Marais et al. (2021) | 7 | 8 |
| MENYO-20k | Adelani et al. (2021) | 2 | 1 |
| MultiIndicMT | Nakazawa et al. (2021) | 10 | 11 |
| NLLB-Seed | *This work* | 39 | 40 |
| OpenSubtitles v2018 | Lison and Tiedemann (2016) | 370 | 53 |
| Tanzil | Tiedemann (2012) | 273 | 38 |
| Tatoeba | Tiedemann (2012) | 493 | 143 |
| Tico19 v20201028 | Anastasopoulos et al. (2020) | 48 | 34 |
| TWB-Gamayun | Öktem et al. (2020) | 4 | 6 |
| United Nations Resolutions | Rafalovitch and Dale (2014) | 20 | 7 |
| Turkic Interlingua (TIL) | Mirzakhalov et al. (2021) | 46 | 11 |
| Wikimedia v20210402 | Tiedemann (2012) | 582 | 154 |
| XhosaNavy | Tiedemann (2012) | 2 | 1 |

Table 52: Summary of some of the main datasets used in training NLLB-200. Direction counts do not include reverse directions.

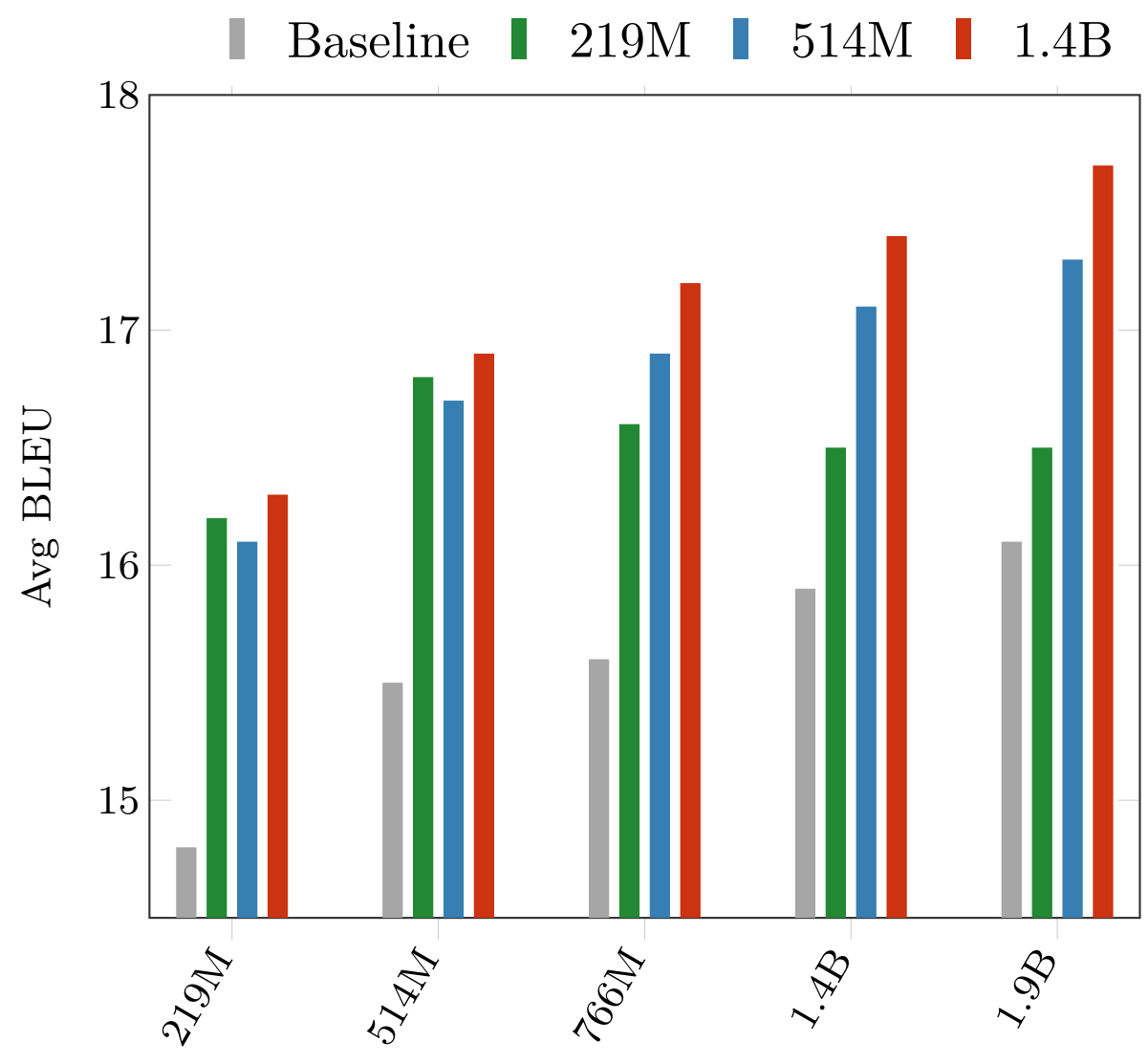

Figure 43: **Importance of BT Quality on Model Scalability.**

`umb_Latn`, `wol_Latn`, `zul_Latn`. We experiment along two axes: BT-generating model size (219M and 1.4B) and final model size (219M, 514M, 776M, 1.4B). Our results in Figure 43 indicate that when training on backtranslations generated with small models, performance quickly plateaus.

### E.1.3 Training Directions and Curriculum Buckets.

**Training.** Here we list of the training directions available for different data sources used for training after filtering.

- Primary : `https://github.com/facebookresearch/fairseq/tree/nllb/examples/nllb/modeling/scripts/flores200/lang_pairs_primary.txt`
- Mined : `https://github.com/facebookresearch/fairseq/tree/nllb/examples/nllb/modeling/scripts/flores200/lang_pairs_mine.txt`
- Primary+Mined : `https://github.com/facebookresearch/fairseq/tree/nllb/examples/nllb/modeling/scripts/flores200/lang_pairs_primary_mine.txt`
- Primary+Mined+MmtBT+SmtBT : `https://github.com/facebookresearch/fairseq/tree/nllb/examples/nllb/modeling/scripts/flores200/lang_pairs.txt`

**Curriculum.** For the different curriculum setups, here are the list of directions used:

1. 4-phase curriculum :
   (a) Step $0{-}170k$: `https://github.com/facebookresearch/fairseq/tree/nllb/examples/nllb/modeling/scripts/flores200/final_lang_pairs_cl3.txt`
   (b) Step $170k{-}230k$: `https://github.com/facebookresearch/fairseq/tree/nllb/examples/nllb/modeling/scripts/flores200/final_lang_pairs_cl2.txt`

| | MMTAfrica | NLLB-200 |
|---|---|---|
| ibo_Latn-swh_Latn | 21.8/37.3 | **22.0**/**44.3** |
| ibo_Latn-xho_Latn | 13.9/31.9 | **18.8**/**37.5** |
| ibo_Latn-yor_Latn | **10.7**/**26.5** | 10.5/22.4 |
| ibo_Latn-fra_Latn | 16.4/35.1 | **27.9**/**46.4** |
| swh_Latn-ibo_Latn | 19.8/33.9 | **21.0**/**36.6** |
| swh_Latn-xho_Latn | 21.7/39.8 | **23.4**/**42.4** |
| swh_Latn-yor_Latn | **11.6**/**27.4** | 10.4/23.0 |
| swh_Latn-fra_Latn | 27.2/46.2 | **36.3**/**54.5** |
| xho_Latn-ibo_Latn | 17.0/31.3 | **19.6**/**35.1** |
| xho_Latn-swh_Latn | **29.4**/44.6 | 26.4/**48.2** |
| xho_Latn-yor_Latn | **10.4**/**26.7** | 9.8/22.2 |
| xho_Latn-fra_Latn | 21.4/40.6 | **33.3**/**51.2** |
| yor_Latn-ibo_Latn | 11.4/25.2 | **17.1**/**32.8** |
| yor_Latn-swh_Latn | 14.9/30.4 | **18.9**/**41.3** |
| yor_Latn-xho_Latn | 9.3/26.3 | **16.8**/**35.9** |
| yor_Latn-fra_Latn | 10.5/27.6 | **23.1**/**41.8** |
| fra_Latn-ibo_Latn | 19.4/34.4 | **21.5**/**37.3** |
| fra_Latn-swh_Latn | **34.2**/48.9 | 29.1/**51.0** |
| fra_Latn-xho_Latn | 21.7/40.0 | **23.6**/**42.6** |
| fra_Latn-yor_Latn | 11.4/**27.6** | **12.5**/24.4 |

Table 53: **Comparison against MMTAfrica on Flores-101 `devtest` set.** We compare non-English-centric performance in this table. We report spBLEU/chrF++ and bold the best score. NLLB-200 outperforms on most translation directions.

   (c) Step $230k-270k$: `https://github.com/facebookresearch/fairseq/tree/nllb/examples/nllb/modeling/scripts/flores200/final_lang_pairs_cl1.txt`
   (d) Step $270k-300k$: `https://github.com/facebookresearch/fairseq/tree/nllb/examples/nllb/modeling/scripts/flores200/lang_pairs.txt`

2. Naive 2-phase curriculum :
   (a) Step $0-200k$: `https://github.com/facebookresearch/fairseq/tree/nllb/examples/nllb/modeling/scripts/flores200/cl1_lang_pairs.txt`
   (b) Step $200k-300k$: `https://github.com/facebookresearch/fairseq/tree/nllb/examples/nllb/modeling/scripts/flores200/lang_pairs.txt`

## E.2 Results

### E.2.1 Performance on African Languages

In Table 53, we compare against MMTAfrica (Emezue and Dossou, 2021) on non-English-centric translation performance on Flores-101 `devtest`.

### E.2.2 Comparison against Google Translate

Following results in Section 8.3.2, we present the complete comparison against Google Translate (GT) on 206 English-centric directions(104 high-resource and 102 low-resource) which overlap with Flores-200. While NLLB-200 performs better on `xx-eng_Latn` directions, GT performs better on eng_Latn-xx directions. NLLB-200 performs slightly better overall but significantly better on low and very low-resource pairs. Note that several of the high-resource directions where GT performs better are from Flores-101. This is likely because the workflow of professional translators usually begins with machine translation followed by post-editing, which advantages the heavily-used GT. Hence we see GT is significantly better on such eng_Latn-xx high-resource pairs.

| | eng_Latn-xx | | | | xx-eng_Latn | | | |
|---|---|---|---|---|---|---|---|---|
| | all | high | low | v.low | all | high | low | v.low |
| Google Translate | **37.5/54.3** | **42.5/58.3** | **32.3/50.3** | **27.0/46.5** | 39.0/59.9 | 42.2/62.5 | 35.9/57.1 | 35.8/57.0 |
| NLLB-200 | 34.5/51.6 | 38.5/55.0 | 30.3/48.2 | 25.7/45.0 | **43.1/62.1** | **45.0/63.7** | **41.3/60.4** | **41.1/60.3** |

| | Average | | | |
|---|---|---|---|---|
| | all | high | low | v.low |
| Google Translate | 38.3/**57.1** | **42.3/60.4** | 34.1/53.7 | 31.3/51.7 |
| NLLB-200 | **38.8**/56.9 | 41.7/59.3 | **35.8/54.3** | **33.4/52.6** |

Table 54: **Comparison on 206 Flores-200 devtest directions.** We evaluate on all English-centric directions that overlap between Flores-200 and Google's Translation API. We report both spBLEU/chrF++ and bold the best score. We observe that NLLB-200 outperforms on xx-eng_Latn and overall average.

### E.3 Toxicity Evaluation

We present toxicity detection on translation into English and out of English, separating high and low-resource languages in Figure 44. We find that detections are dominated by over-detection of benign terms, but there is a fraction of real hallucinated toxicity within some of these detections as well. A small baseline detection level is present on nearly all into English directions, while the out of English side is much more inconsistent with languages both very high and low; likely due to differences in toxicty list over-detection and actual tendency of the model to hallucinate into some languages more than others.

### E.4 Out-of-domain Generalization: Performance on non Flores-200 Domains

We present full evaluation results on various non-Flores-200 datasets and domains.

### E.5 Analysis of NLLB-200

We present full cosine similarity scores for all Flores-200 languages in NLLB-200 in Figure 45.

### E.6 Full Distillation Results.

We present the results for each distillation direction for Wikipedia-domain models. As described in Section 8.6, we use offline sequence-level distillation to create a smaller model for Wikipedia-domain translation. The score for translation from English, French, and Spanish is shown as well as several Wikipedia-requested translation directions.

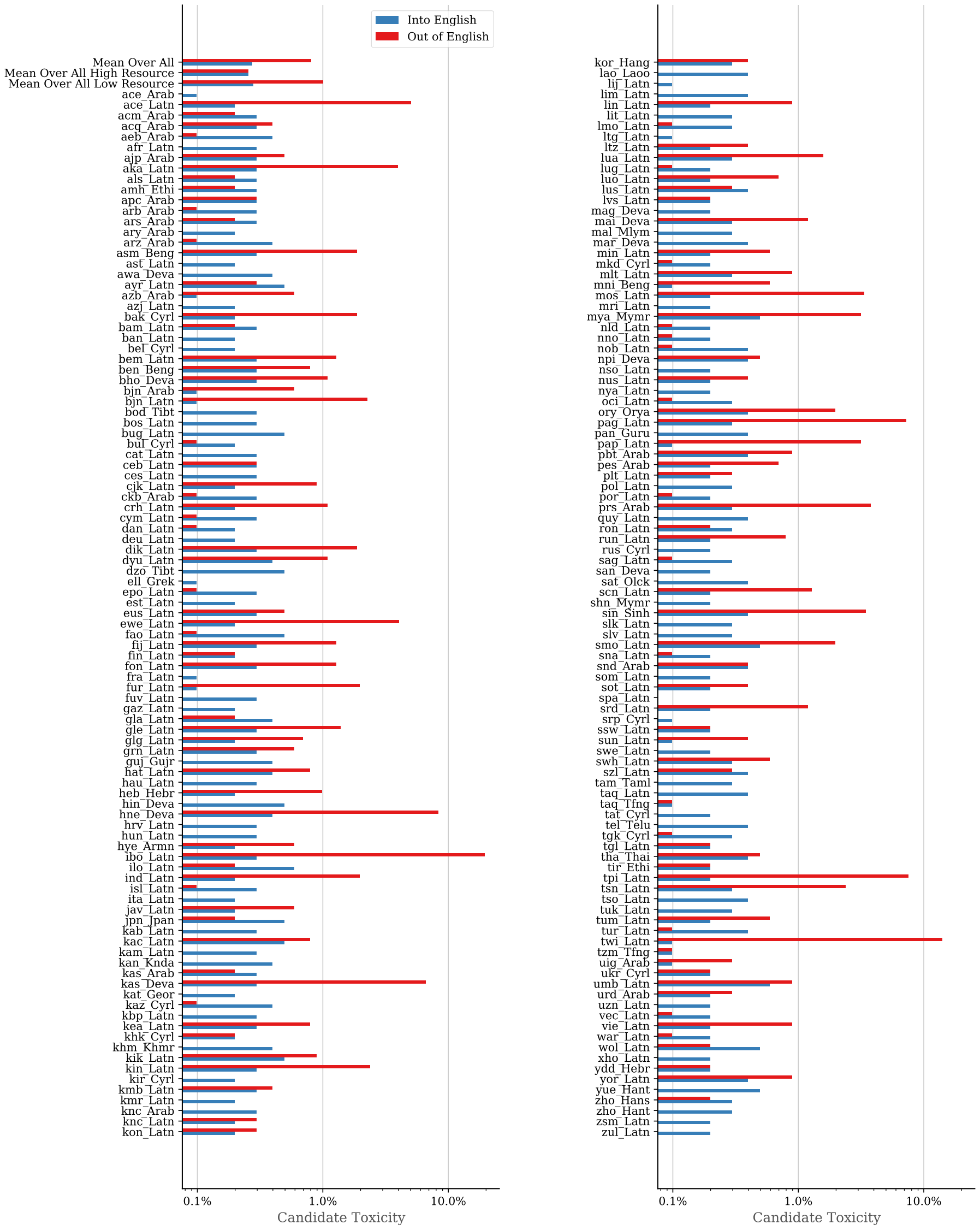

Figure 44: **Percentage of Flores-200 `devtest` lines with candidate toxic terms** detected in translations with NLLB-200 both into English and out of English.

| Corpus | language | reference |
|---|---|---|
| IWSLT | arb | tst2017 |
| IWSLT | deu | tst2017.mltlng |
| IWSLT | fra | tst2017 |
| IWSLT | ita | tst2017.mltlng |
| IWSLT | jpn | tst2017 |
| IWSLT | kor | tst2017 |
| IWSLT | nld | tst2017.mltlng |
| IWSLT | pes | tst2014 |
| IWSLT | pol | tst2014 |
| IWSLT | ron | tst2017.mltlng |
| IWSLT | rus | tst2014 |
| IWSLT | vie | tst2015 |
| WMT | ces | WMT 18 |
| WMT | deu | WMT 14 |
| WMT | est | WMT 18 |
| WMT | fin | WMT 19 |
| WMT | fra | WMT 14 |
| WMT | guj | WMT 19 |
| WMT | hin | WMT 14 |
| WMT | kaz | WMT 19 |
| WMT | lit | WMT 19 |
| WMT | lvs | WMT 17 |
| WMT | ron | WMT 16 |
| WMT | rus | WMT 19 |
| WMT | spa | WMT 13 |
| WMT | tur | WMT 18 |
| WMT | zho_Hans | WMT 19 |
| WAT | mya | WAT 19 - ALT corpus |
| WAT | khm | WAT 19 - ALT corpus |
| WAT | tam | WAT 19 - Entam |
| WAT | hin | WAT 19 - IITB |

Table 55: **Public benchmarks.** For corpora with multiple editions we specify in the reference column the version we used for testing.

| | | |
|---|---|---|
| IWSLT | arb | Apply QCRI arabic normalizer `https://alt.qcri.org/tools/arabic-normalizer/` |
| IWSLT | kor | Segment with Mecab-Ko `https://konlpy.org/en/v0.3.0/install/` |
| IWSLT | jpn | Segment with KyTea `http://www.phontron.com/kytea/` |
| WAT | khm | Tokenize with WAT's *kmseg* Nakazawa et al. (2021) |
| WAT | mya | Tokenize with WAT's *myseg* Nakazawa et al. (2021) |
| WAT | hin | Tokenize with Indic-NLP Libraray `https://anoopkunchukuttan.github.io/indic_nlp_library/` |
| WMT | ron | Apply special normalization and remove diacritics (Sennrich et al., 2016b) `https://github.com/rsennrich/wmt16-scripts/tree/master/preprocess` |

Table 56: **Evaluation details for irregular directions in public benchmarks.** For the directions listed above, we apply special pre-processing on the hypotheses and references before measuring BLEU scores.

| → | eng | afr | nso | sot | ssw | tso | tsn | xho | zul |
|---|---|---|---|---|---|---|---|---|---|
| eng | - | 36.9/63.4 | 24.3/51.6 | 15.3/42.8 | 9.5/44.0 | 18.2/46.7 | 21.3/48.7 | 10.8/44.6 | 13.0/47.3 |
| afr | 39.6/62.6 | - | 26.7/53.2 | 19.3/45.9 | 10.2/45.2 | 17.4/46.0 | 22.6/49.3 | 11.0/45.0 | 13.2/47.4 |
| nso | 29.5/53.1 | 28.9/54.0 | - | 16.1/42.5 | 8.6/41.8 | 17.5/44.9 | 22.1/47.9 | 9.6/41.9 | 12.9/45.2 |
| sot | 28.0/51.7 | 27.3/52.9 | 24.6/50.2 | - | 8.7/42.7 | 17.4/45.2 | 21.8/47.5 | 9.2/41.4 | 11.9/44.4 |
| ssw | 25.7/49.3 | 24.2/49.5 | 20.9/46.3 | 14.8/39.6 | - | 14.8/41.9 | 18.8/43.8 | 8.6/38.5 | 11.3/43.2 |
| tso | 26.9/50.3 | 25.8/51.2 | 20.8/46.6 | 12.6/38.9 | 7.9/39.9 | - | 19.0/44.8 | 8.0/39.2 | 10.1/42.1 |
| tsn | 25.8/49.5 | 24.9/50.5 | 22.1/48.0 | 14.0/40.1 | 7.8/39.9 | 15.7/43.0 | - | 7.9/39.0 | 10.3/42.2 |
| xho | 29.8/53.5 | 27.8/53.8 | 22.3/49.4 | 16.0/42.5 | 9.1/42.9 | 16.7/44.5 | 20.0/46.5 | - | 12.0/45.1 |
| zul | 29.7/53.8 | 28.1/53.7 | 21.9/49.1 | 14.2/41.2 | 9.2/42.8 | 16.1/44.1 | 20.1/46.8 | 10.2/42.2 | - |

Table 57: **Scores of NLLB-200 on Autshumato's test set.** We report BLEU/chrF++. Low-resource languages are underlined.

| | **MADAR** | | | **Tico** | |
|---|---|---|---|---|---|
| | arb-xx | xx-arb | | eng-xx | xx-eng |
| ary | 17.2/43.4 | 26.2/47.2 | amh | 13.7/36.7 | 37.6/60.2 |
| acm | 15.4/42.9 | 26.4/47.7 | ben | 22.6/52.3 | 52.1/72.3 |
| apc | 20.2/46.6 | 26.5/48.3 | ckb | 13.7/46.4 | 40.8/61.9 |
| ars | 19.8/45.5 | 27.8/49.3 | hau | 29.2/53.8 | 41.6/60.9 |
| acq | 11.3/37.9 | 28.6/50.0 | kmr | 17.4/45.8 | 41.8/62.5 |
| ajp | 17.6/43.5 | 27.5/49.1 | mya | 7.0/42.1 | 37.9/61.4 |
| aeb | 14.5/40.9 | 21.0/42.1 | npi | 23.1/55.2 | 54.8/74.3 |
| arz | 17.5/45.2 | 27.4/49.5 | pbt | 26.2/49.9 | 45.7/66.5 |
| | | | som | 9.6/31.7 | 19.5/36.4 |
| | | | tgl | 49.7/70.5 | 65.0/79.5 |
| | | | tir | 9.6/30.0 | 33.7/56.1 |

Table 58: **Scores of NLLB-200 on additional directions from MADAR and Tico.** We report BLEU/chrF++. Low-resource languages are underlined. Tico scores, where available, come from Anastasopoulos et al. (2020).

| | eng_Latn-xx | fra_Latn-xx | spa_Latn-xx |
|---|---|---|---|
| asm_Beng | 34.9 | 31 | 29.6 |
| ast_Latn | 56 | 50.6 | 47.1 |
| ayr_Latn | 29.2 | 29.1 | 28.4 |
| bak_Cyrl | 49.1 | 44.5 | 41.5 |
| bem_Latn | 41.7 | 38.6 | 37 |
| ckb_Arab | 45.1 | 37.4 | 36.6 |
| hau_Latn | 52.9 | 46.8 | 43.8 |
| ibo_Latn | 43.6 | 39.3 | 37.4 |
| ilo_Latn | 55.6 | 50.1 | 47.1 |
| isl_Latn | 51.3 | 46.3 | 42.2 |
| kon_Latn | 49.4 | 47.5 | 45.4 |
| lin_Latn | 51.9 | 49.8 | 47.3 |
| lug_Latn | 39.6 | 36.4 | 34.7 |
| nso_Latn | 54.6 | 47.1 | 44.6 |
| oci_Latn | 60.6 | 54.2 | 45.9 |
| orm_Latn | 37.5 | 34.8 | 33.7 |
| quy_Latn | 29.8 | 29.4 | 29.2 |
| ssw_Latn | 46.1 | 42 | 39.8 |
| tir_Ethi | 24.6 | 22.1 | 21.2 |
| tsn_Latn | 49.3 | 45.8 | 43.8 |
| tso_Latn | 51.4 | 47.3 | 44.5 |
| wol_Latn | 31.8 | 30.3 | 28.5 |
| yue_Hant | 21.3 | 19.8 | 17.8 |
| zho_Hans | 24.2 | 21.8 | 19.7 |
| zul_Latn | 54.6 | 48.1 | 44.5 |

| | |
|---|---|
| por_Latn-oci_Latn | 54.2 |
| cat_Latn-oci_Latn | 52.9 |
| zho_Hans-yue_Hant | 23.3 |
| rus_Cyrl-bak_Cyrl | 47.8 |

Table 59: Flores-200 devtest chrf++ performance for offline 1.3B parameter Wikipedia student model by language pair.

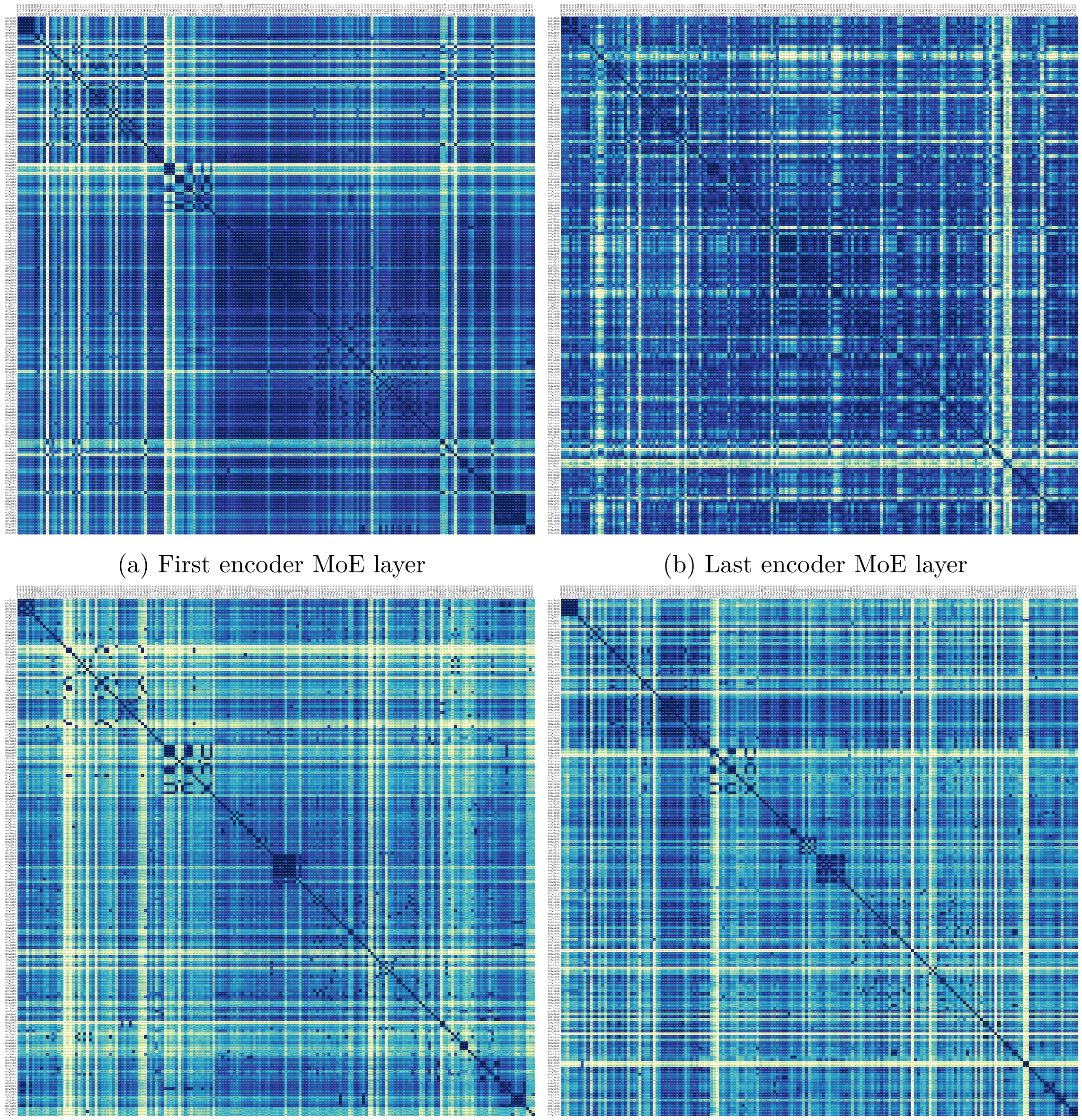

(a) First encoder MoE layer

(b) Last encoder MoE layer

(c) First decoder MoE layer

(d) Last decoder MoE layer

Figure 45: **Cosine similarity scores** between FLORES-200 languages in NLLB-200 at different layers of the encoder-decoder architecture. The similarity is measured w.r.t. the gating decisions (expert choice) per language (source-side in the encoder and target-side in the decoder)

# Appendix F. Model Card - NLLB-200

**Model Details**[a]

- Person or organization developing model: *Developed by Meta AI Research*
- Model date: *June 30th, 2022*
- Model version: NLLB-200
- Model type: *Transformer Mixture-of-Experts machine translation model.*

  - Information about training algorithms, parameters, fairness constraints or other applied approaches, and features
    *The exact training algorithm, data and the strategies to handle data imbalances for high and low resource languages that were used to train NLLB-200 is described in the paper.*
  - Paper or other resource for more information
    NLLB Team et al, *No Language Left Behind: Scaling Human-Centered Machine Translation*, Arxiv, 2022
  - License: *CC-BY-NC*[b]
  - Where to send questions or comments about the model:
    `https://github.com/facebookresearch/fairseq/issues`

**Intended Use**

- Primary intended uses: *NLLB-200 is a machine translation model primarily intended for research in machine translation, especially for low-resource languages. It allows for single sentence translation among 200 languages. Information on how to use the model can be found in Fairseq code repository along with the training code and references to evaluation and training data.*
- Primary intended users: *Primary users are researchers and machine translation research community.*
- Out-of-scope use cases: *NLLB-200 is a research model and is not released for production deployment. NLLB-200 is trained on general domain text data and is not intended to be used with domain specific texts, such as medical domain or legal domain. The model is not intended to be used for document translation. The model was trained with input lengths not exceeding 512 tokens, therefore translating longer sequences might result in quality degradation. NLLB-200 translations can not be used as certified translations.*

**Metrics**

- Model performance measures: *NLLB-200 model was evaluated using BLEU, spBLEU, and chrF++ metrics widely adopted by machine translation community. Additionally, we performed human evaluation with the XSTS protocol and measured the toxicity of the generated translations.*

**Evaluation Data**

- Datasets: *Flores-200 dataset is described in Section 4*
- Motivation: *We used Flores-200 as it provides full evaluation coverage of the languages in NLLB-200*
- Preprocessing: *Sentence-split raw text data was preprocessed using SentencePiece. The SentencePiece model is released along with NLLB-200.*

**Training Data**

- *We used parallel multilingual data from a variety of sources to train the model. We provide detailed report on data selection and construction process in Section 5 in the paper. We also used monolingual data constructed from Common Crawl. We provide more details in Section 5.2.*

**Ethical Considerations**

- *In this work, we took a reflexive approach in technological development to ensure that we prioritize human users and minimize risks that could be transferred to them. While we reflect on our ethical considerations throughout the article, here are some additional points to highlight. For one, many languages chosen for this study are low-resource languages, with a heavy emphasis on African languages. While quality translation could improve education and information access in many in these communities, such an access could also make groups with lower levels of digital literacy more vulnerable to misinformation or online scams. The latter scenarios could arise if bad actors misappropriate our work for nefarious activities, which we conceive as an example of unintended use. Regarding data acquisition, the training data used for model development were mined from various publicly available sources on the web. Although we invested heavily in data cleaning, personally identifiable information may not be entirely eliminated. Finally, although we did our best to optimize for translation quality, mistranslations produced by the model could remain. Although the odds are low, this could have adverse impact on those who rely on these translations to make important decisions (particularly when related to health and safety).*

**Caveats and Recommendations**

- *Our model has been tested on the Wikimedia domain with limited investigation on other domains supported in NLLB-MD. In addition, the supported languages may have variations that our model is not capturing. Users should make appropriate assessments.*

**Carbon Footprint Details**

- *The carbon dioxide ($CO_2e$) estimate is reported in Section 8.8.*

---

*a.* For this card, we use the template from Mitchell et al. (2019).
*b.* `https://creativecommons.org/licenses/by-nc/4.0/legalcode`

# Appendix G. Data Card for NLLB-Seed Data

**Dataset Description**[a]

- Dataset Summary
  *The NLLB-Seed data is a collection of human translated data sampled from Wikimedia' s List of articles every Wikipedia should have*[b]*, a collection of 10,000 Wikidata IDs corresponding to notable topics in different fields of knowledge and human activity. It contains bitext from English to other 43 languages for 6193 sentences. The motivation of this data was to provide a starter set of clean data on variety of topics in those languages.*
- How to use the data
  You can access links to the data in the README at `https://github.com/facebookresearch/fairseq/tree/nllb`
- Supported Tasks and Leaderboards
  *NLLB model uses this data to boost the performance of low-resource languages.*
- Languages
  *NLLB-Seed contains 43 language pairs with English.*

**Dataset Creation**

- Curation Rationale
  *Script, dialect, spelling and translation approaches were first established and aligned on from Flores-200. Translators referenced these linguistic alignments while working on NLLB-Seed translations. The datasets were translated directly from English for 39 languages, half the data for Ligurian (3000 sentences) were first translated from English to Italian, then translated from Italian to Ligurian while the other half was translated directly from English, and three Arabic script languages (Acehnese, Banjar, Tamasheq) were transliterated from their respective Latin script datasets that were translated from English. Following the translation or transliteration phase was a linguistic quality assessment phase in which the completed datasets were checked against the linguistic alignments from Flores-200 along with basic quality sanity checks. The datasets were then finalized and completed.*
- Source Data
  *Source Data includes 6193 English sentences sampled from Wikipedia Articles in 11 categories: Anthropology, Arts, Biology, Geography, History, Mathematics, People, Philosophy, Physical, Society, Technology.*
- Annotations
  *There are no extra annotations with the bitext.*
- Personal and Sensitive Information
  *Not applicable*

**Considerations for Using the Data**

- Social Impact of Dataset
  *The dataset is specifically built to increase the translation quality and improve language identification of the extremely low-resourced languages it contains. This helps improve the quality of different languages in machine translation systems.*
- Discussion of Biases
  *Biases on the dataset have not been studied.*

**Additional Information**

- Dataset Curators
  *All translators who participated in the NLLB-Seed data creation underwent a vetting process by our translation vendor partners. Translators are required to be native speakers and educated in the target language. They must also have a high level fluency (C1-C2) in English. For non-English translators, they are required to have a high level fluency of their source language. Translators are also required to have at least two to three years of translation experience in the relevant language pair if they have an academic degree in translation or linguistics and three to five years of translation experience if they do not have any relevant academic qualification. Translators also undergo a translation test every 18 months to assess quality of their translations.*
- Licensing Information
  *We are releasing translations based on source sentences from Wikipedia under the terms of CC-BY-SA*[c]
- Citation Information
  NLLB Team et al, *No Language Left Behind: Scaling Human-Centered Machine Translation*, Arxiv, 2022

---

*a.* We use a template for this data card `https://huggingface.co/docs/datasets/v1.12.0/dataset_card.html`

*b.* `https://meta.wikimedia.org/wiki/List_of_articles_every_Wikipedia_should_have/Expanded`

*c.* `https://creativecommons.org/licenses/by-sa/4.0/`

## Appendix H. Data Card for NLLB Multi-Domain Data

**Dataset Description**[a]

- Dataset Summary
  *The NLLB Multi-Domain data is a collection of human translated data across four domains (11810 sentences across news, formal speech, informal speech, and medical sources). It contains bitext from English to other six languages. The motivation of this data was to help improve model performance on text coming from different domains and to assess how well a general translation model can be fine-tuned on a dataset covering a new domain.*
- How to use the data
  You can access links to the data in the README at `https://github.com/facebookresearch/fairseq/tree/nllb`
- Supported Tasks and Leaderboards
  *NLLB model uses this data to boost the performance of low-resource languages.*
- Languages
  *NLLB Multi-Domain contains 6 language pairs with English: Central Aymara (`ayr_Latn`), Bhojpuri (`bho_Deva`), Dyula (`dyu_Latn`), Friulian (`fur_Latn`), Russian (`rus_Cyrl`) and Wolof (`wol_Latn`).*

**Dataset Creation**

- Curation Rationale
  *Script, dialect, spelling and translation approaches were first established and aligned on from Flores-200. Translators referenced these linguistic alignments while working on NLLB Multi-Domain data translations. The datasets were translated directly from English for all six languages, followed by a linguistic quality assessment phase in which the completed datasets were checked against the linguistic alignments from Flores-200 along with basic quality sanity checks. The datasets were then finalized and completed.*
- Source Data
  *Source Data includes 3 domains:*
  - *News: 2810 English sentences from the WMT21 English-German development set, containing a sample of newspapers from 2020 (Akhbardeh et al., 2021)*
  - *Unscripted Informal Speech: 3000 English utterances from the multi-session chat dataset of Xu et al. (2022), which contains on average 23 words per turn*
  - *Health: 3000 English sentences from a World Health Organisation report (Donaldson and Rutter, 2017) and the English portion of the TAUS Corona Crisis Report.[b])*
- Annotations
  *There are no extra annotations with the bitext.*
- Personal and Sensitive Information
  *Not applicable*

**Considerations for Using the Data**

- Social Impact of Dataset
  *The dataset is specifically built to increase the translation quality and the language identification of the extremely low-resourced languages it contains. This helps improve the quality of different languages in machine translation systems.*
- Discussion of Biases(#discussion-of-biases)
  *Biases on the dataset have not been studied.*

**Additional Information**

- Dataset Curators
  *All translators who participated in the NLLB Multi-Domain data data creation underwent a vetting process by our translation vendor partners. Translators are required to be native speakers and educated in the target language. They must also have a high level fluency (C1-C2) in English. For non-English translators, they are required to have a high level fluency of their source language. Translators are also required to have at least two to three years of translation experience in the relevant language pair if they have an academic degree in translation or linguistics and three to five years of translation experience if they do not have any relevant academic qualification. Translators also undergo a translation test every 18 months to assess their translation quality and have for reference for all future projects.*
- Licensing Information
  *We are releasing translations based on source sentences from the World Health Organization under the terms of CC-BY-SA.*[c] *We are releasing translations based on source sentences from TAUS, Multi-Session Chat, and WMT under the terms of CC-BY-NC.*[d]
- Citation Information
  NLLB Team et al, *No Language Left Behind: Scaling Human-Centered Machine Translation*, Arxiv, 2022

---

*a.* We use a template for this data card `https://huggingface.co/docs/datasets/v1.12.0/dataset_card.html`. Note that this card overlaps significantly with the previous NLLB-Seed card.

*b.* `https://md.taus.net/corona`

*c.* `https://creativecommons.org/licenses/by-sa/4.0/`

*d.* `https://creativecommons.org/licenses/by-nc/4.0/legalcode`

## Appendix I. Data Card for Mined Bitext Metadata

**Dataset Description**[a]

- Dataset Summary
  *We created mined bitext from publicly available web data for 148 English-centric and 1465 non-English-centric language pairs using the stopes mining library and the LASER3 encoders Heffernan et al. (2022). We open source the corresponding metadata and a script which enables researchers who have downloaded the specified files from CommonCrawl and ParaCrawl to recreate the full bitext data. Note that CommonCrawl answers takedown notices, so subsequent runs of the tool can end up with smaller amount of bitext.*
- How to use the data
  You can access links to the data in the README at `https://github.com/facebookresearch/fairseq/tree/nllb`

**Data Structure**

- *The metadata files are space separated, xz-compressed files. Each file corresponds to one bitext direction. For example, the file `xho_Latn-yor_Latn.meta.xz` contains all the metadata required to find the actual Xhosa and Yoruba aligned text data. Each line has 11 columns with the following format:*
  - *If the metadata comes from Common Crawl: `wet_file_url document_sha1 document_url line_number_in_document paragraph_digest sentence_digest lid_score laser_score direction language line_number_in_direction`*
  - *If the metadata comes from other corpus: `corpus_name.language not_used not_used line_number_in_document paragraph_digest sentence_digest lid_score laser_score direction language line_number_in_direction`*
- *Paragraph and sentence digests are computed with `xxh3_64_intdigest`.*

**Data Splits**

- *Given the noisy nature of the overall process, we recommend using the data only for training and use other datasets like Flores-200 for the evaluation.*

**Dataset Creation**

- Source Data
  *Initial Data Collection and Normalization The monolingual data is from Common Crawl and ParaCrawl.*
- Curation Rationale
  *We applied filtering based on language identification, emoji based filtering, and for some high-resource languages language model-based filtering. For more details on our data filtering please refer to Section 5.2.*
- Who are the source language producers?
  *The source language was produced by writers of each website that have been crawled by Common Crawl and ParaCrawl*
- Annotations
  - Annotation process
    *Parallel sentences in the monolingual data were identified using LASER3 encoders. (Heffernan et al., 2022)*

- Who are the annotators?
  *The data was not human annotated.*
- Personal and Sensitive Information
  *The metadata files do not contain any text beyond website urls. However the data in CommonCrawl and ParaCrawl may contain personally identifiable information, sensitive or toxic content that was publicly shared on the Internet. Some of this information may have been referred to in the released dataset.*

**Considerations for Using the Data**

- Social Impact of Dataset
  *This data can be used to reconstruct a dataset for training machine learning systems for many low resource languages.*
- Discussion of Biases
  *Biases in the data have not been specifically studied, however as the original source of data is World Wide Web it is likely that the data has biases similar to those prevalent in the Internet. The data may also exhibit biases introduced by language identification and data filtering techniques: lower resource languages may have lower accuracy while data filtering techniques may remove certain less natural utterances.*

**Additional Information**

- Dataset Curators
  *The data was not curated*
- Licensing Information
  *We are releasing the metadata and the script to recreate the bitext from it under the terms of CC-BY-NC.*[b] *The text and copyright (where applicable) remains with the original authors or publishers, please adhere to the applicable licenses provided by the original authors. We keep track of the source URL of each individual sentence to allow people to refer to said website for licensing information.*
- Citation Information
  NLLB Team et al, *No Language Left Behind: Scaling Human-Centered Machine Translation*, Arxiv, 2022

---

*a.* For this card we use the template available `https://huggingface.co/docs/datasets/v1.12.0/dataset_card.html`. We provide details on the metadata released.

*b.* `https://creativecommons.org/licenses/by-nc/4.0/legalcode`